\documentclass[acmsmall]{acmart}

\usepackage{tcolorbox}
\usepackage{xcolor}
\usepackage{float}
\usepackage{geometry}
\usepackage{multirow,bigstrut}
\usepackage{colortbl}
\usepackage{longtable}
\usepackage{changepage}
\usepackage[printonlyused]{acronym} 
\usepackage{natbib}
\usepackage{placeins}
\usepackage{subcaption}

\newcolumntype{P}[1]{>{\centering\arraybackslash}p{#1}}

\definecolor{Gray1}{gray}{0.95}
\definecolor{Gray2}{gray}{0.89}
\definecolor{ILCblue}{RGB}{68,119,170}
\definecolor{ISpink}{RGB}{170,51,119}
\definecolor{SSyellow}{RGB}{235, 192, 52}
\definecolor{ODgreen}{RGB}{34,136,51}
\definecolor{lightgray}{rgb}{0.95,0.95,0.95}
\usepackage{fontawesome}

\usepackage{calc}

\hypersetup{
  pdfinfo={
  pdfproducer={},
  Title={},
  Subject={},
  Author={},
  }
}
\AtBeginDocument{%
  \providecommand\BibTeX{{%
    \normalfont B\kern-0.5em{\scshape i\kern-0.25em b}\kern-0.8em\TeX}}}

\begin{document}

\DeclareUrlCommand\ULurl@{%
  \def\UrlFont{\ttfamily\color{blue}}%
  \def\UrlLeft{\uline\bgroup}%
  \def\UrlRight{\egroup}}
\def\ULurl@#1{\hyper@linkurl{\ULurl@@{#1}}{#1}}
\DeclareRobustCommand*\ULurl{\hyper@normalise\ULurl@}
\title[Understanding metric-related pitfalls]{Understanding metric-related pitfalls in image analysis validation}


\author{Annika Reinke}
\authornote{\textbf{Corresponding authors:} Annika Reinke: a.reinke@dkfz-heidelberg.de; Minu D. Tizabi: Lena Maier-Hein, l.maier-hein@dkfz-heidelberg.de; Paul F. Jäger: p.jaeger@dkfz-heidelberg.de; Lena Maier-Hein, l.maier-hein@dkfz-heidelberg.de.}
\authornote{Shared first authors: Annika Reinke and Minu D. Tizabi}
\email{a.reinke@dkfz.de}
\affiliation{%
  \institution{German Cancer Research Center (DKFZ) Heidelberg, Division of Intelligent Medical Systems and HI Helmholtz Imaging}
  \country{Germany}
}
\affiliation{%
  \institution{Faculty of Mathematics and Computer Science, Heidelberg University}
  \country{Heidelberg, Germany}
}

\author{Minu D. Tizabi}
\authornotemark[2]
\affiliation{%
  \institution{German Cancer Research Center (DKFZ) Heidelberg, Division of Intelligent Medical Systems}
  \country{Germany}
}
\affiliation{%
  \institution{National Center for Tumor Diseases (NCT), NCT Heidelberg, a partnership between DKFZ and University Medical Center Heidelberg}
  \country{Germany}
}

\author{Michael Baumgartner}
\affiliation{%
  \institution{German Cancer Research Center (DKFZ) Heidelberg, Division of Medical Image Computing}
  \country{Germany}
}

\author{Matthias Eisenmann}
\affiliation{%
  \institution{German Cancer Research Center (DKFZ) Heidelberg, Division of Intelligent Medical Systems}
  \country{Germany}
}

\author{Doreen Heckmann-Nötzel}
\affiliation{
  \institution{German Cancer Research Center (DKFZ) Heidelberg, Division of Intelligent Medical Systems}
  \country{Germany}
}
\affiliation{%
  \institution{National Center for Tumor Diseases (NCT), NCT Heidelberg, a partnership between DKFZ and University Medical Center Heidelberg}
  \country{Germany}
}

\author{A. Emre Kavur}
\affiliation{%
  \institution{HI Applied Computer Vision Lab, Division of Medical Image Computing; German Cancer Research Center (DKFZ) Heidelberg, Division of Intelligent Medical Systems}
  \country{Germany}
}

\author{Tim Rädsch}
\affiliation{%
  \institution{German Cancer Research Center (DKFZ) Heidelberg, Division of Intelligent Medical Systems and HI Helmholtz Imaging}
  \country{Germany}
}

\author{Carole H. Sudre}
\affiliation{%
 \institution{MRC Unit for Lifelong Health and Ageing at UCL and Centre for Medical Image Computing, Department of Computer Science, University College London}
 \country{London, UK}
 }
\affiliation{%
 \institution{School of Biomedical Engineering and Imaging Science, King's College London}
 \country{London, UK}
 }

\author{Laura Acion}
\affiliation{%
  \institution{Instituto de Cálculo, CONICET -- Universidad de Buenos Aires}
  \country{Buenos Aires, Argentina}
}

\author{Michela Antonelli}
\affiliation{%
 \institution{School of Biomedical Engineering and Imaging Science, King's College London}
 \country{London, UK}}
\affiliation{%
 \institution{Centre for Medical Image Computing, University College London}
 \country{London, UK}}

\author{Tal Arbel}
\affiliation{%
  \institution{Centre for Intelligent Machines and MILA (Quebec Artificial Intelligence Institute), McGill University}
  \country{Montreal, Canada}}

\author{Spyridon Bakas}
\affiliation{%
  \institution{Division of Computational Pathology, Dept of Pathology \& Laboratory Medicine, Indiana University School of Medicine, IU Health Information and Translational Sciences Building}
  \country{Indianapolis, USA}}
\affiliation{%
  \institution{Center for Biomedical Image Computing and Analytics (CBICA), University of Pennsylvania, Richards Medical Research Laboratories FL7}
  \country{Philadelphia, PA, USA}}

\author{Arriel Benis}
\affiliation{%
  \institution{Department of Digital Medical Technologies, Holon Institute of Technology}
  \country{Holon, Israel}
}
\affiliation{%
  \institution{European Federation for Medical Informatics}
  \country{Le Mont-sur-Lausanne, Switzerland}
}

\author{Matthew B. Blaschko}
\affiliation{%
  \institution{Center for Processing Speech and Images, Department of Electrical Engineering, KU Leuven}
  \country{Leuven, Belgium}
}

\author{Florian Buettner}
\affiliation{%
  \institution{German Cancer Consortium  (DKTK), partner site Frankfurt/Mainz, a partnership between DKFZ and UCT Frankfurt-Marburg}
  \country{Germany}
}
\affiliation{%
  \institution{German Cancer Research Center (DKFZ) Heidelberg}
  \country{Germany}
}
\affiliation{%
 \institution{Goethe University Frankfurt, Department of Medicine}
  \country{Germany}
}
\affiliation{%
 \institution{Goethe University Frankfurt, Department of Informatics}
  \country{Germany}
}
\affiliation{%
  \institution{Frankfurt Cancer Insititute}
  \country{Germany}
}

\author{M. Jorge Cardoso}
\affiliation{%
 \institution{School of Biomedical Engineering and Imaging Science, King's College London}
 \country{London, UK}}
  
\author{Veronika Cheplygina}
\affiliation{%
  \institution{Department of Computer Science, IT University of Copenhagen}
  \country{Copenhagen, Denmark}}

\author{Jianxu Chen}
\affiliation{%
  \institution{Leibniz-Institut für Analytische Wissenschaften – ISAS – e.V.}
  \country{Dortmund, Germany}}

\author{Evangelia Christodoulou}
\affiliation{%
  \institution{German Cancer Research Center (DKFZ) Heidelberg, Division of Intelligent Medical Systems}
  \country{Germany}
}

\author{Beth A. Cimini}
\affiliation{%
  \institution{Imaging Platform, Broad Institute of MIT and Harvard}
  \country{Cambridge, Massachusetts, USA}}

  \author{Gary S. Collins}
\affiliation{%
  \institution{Centre for Statistics in Medicine, University of Oxford}
  \country{Oxford, UK}}

\author{Keyvan Farahani}
\affiliation{%
  \institution{Center for Biomedical Informatics and Information Technology, National Cancer Institute}
  \country{Bethesda, MD, USA}}
  
\author{Luciana Ferrer}
\affiliation{%
\institution{Instituto de Investigación en Ciencias de la Computación (ICC), CONICET-UBA}
  \country{Ciudad Universitaria, Ciudad Autónoma de Buenos Aires, Argentina}}
  
\author{Adrian Galdran}
\affiliation{%
  \institution{Universitat Pompeu Fabra}
  \country{Barcelona, Spain}}
\affiliation{%
  \institution{University of Adelaide}
  \country{Adelaide, Australia}}  

  \author{Bram van Ginneken}
\affiliation{%
  \institution{Fraunhofer MEVIS}
  \country{Bremen, Germany}
}
\affiliation{%
  \institution{Radboud Institute for Health Sciences, Radboud University Medical Center}
  \country{Nijmegen, The Netherlands}
}

\author{Ben Glocker}
\affiliation{%
  \institution{Department of Computing, Imperial College London}
  \country{London, UK}}
  
\author{Patrick Godau}
\affiliation{%
  \institution{German Cancer Research Center (DKFZ) Heidelberg, Division of Intelligent Medical Systems}
  \country{Germany}
}
\affiliation{%
  \institution{Faculty of Mathematics and Computer Science, Heidelberg University}
  \country{Heidelberg, Germany}
}
\affiliation{%
  \institution{National Center for Tumor Diseases (NCT), NCT Heidelberg, a partnership between DKFZ and University Medical Center Heidelberg}
  \country{Germany}
}

\author{Robert Haase}
\affiliation{%
  \institution{Now with: Center for Scalable Data Analytics and Artificial Intelligence (ScaDS.AI), Leipzig University}
  \country{Leipzig, Germany}
}
\affiliation{%
  \institution{DFG Cluster of Excellence "Physics of Life", Technische Universität (TU) Dresden}
  \country{Dresden, Germany}
}
\affiliation{%
  \institution{Center for Systems Biology }
  \country{Dresden, Germany}
}
  
\author{Daniel A. Hashimoto}
\affiliation{%
  \institution{Department of Surgery, Perelman School of Medicine}
  \country{Philadelphia, PA, USA}}
\affiliation{%
  \institution{General Robotics Automation Sensing and Perception Laboratory, School of Engineering and Applied Science, University of Pennsylvania}
  \country{Philadelphia, PA, USA}}
  
\author{Michael M. Hoffman}
\affiliation{%
  \institution{Princess Margaret Cancer Centre, University Health Network}
  \country{Toronto, Canada}}
\affiliation{%
  \institution{Department of Medical Biophysics, University of Toronto}
  \country{Toronto, Canada}}
\affiliation{%
  \institution{Department of Computer Science, University of Toronto}
  \country{Toronto, Canada}}
\affiliation{%
  \institution{Vector Institute for Artificial Intelligence}
  \country{Toronto, Canada}}

\author{Merel Huisman}
\affiliation{%
  \institution{Department of Radiology and Nuclear Medicine, Radboud University Medical Center}
  \country{Nijmegen, The Netherlands}
}

\author{Fabian Isensee}
\affiliation{%
  \institution{German Cancer Research Center (DKFZ) Heidelberg, Division of Medical Image Computing and HI Applied Computer Vision Lab}
  \country{Germany}
}

\author{Pierre Jannin}
\affiliation{%
  \institution{Laboratoire Traitement du Signal et de l’Image – UMR\_S 1099, Université de Rennes 1}
  \country{Rennes, France}
}
\affiliation{%
  \institution{INSERM}
  \country{Paris Cedex, France}
}

\author{Charles E. Kahn}
\affiliation{%
  \institution{Department of Radiology and Institute for Biomedical Informatics, University of Pennsylvania}
  \country{Philadelphia, PA, USA}
}

\author{Dagmar Kainmueller}
\affiliation{%
  \institution{Max-Delbrück Center for Molecular Medicine in the Helmholtz Association (MDC), Biomedical Image Analysis and HI Helmholtz Imaging}
  \country{Berlin, Germany}
}
\affiliation{%
  \institution{University of Potsdam, Digital Engineering Faculty}
  \country{Potsdam, Germany}
}

\author{Bernhard Kainz}
\affiliation{%
  \institution{Department of Computing, Faculty of Engineering, Imperial College London}
  \country{London, UK}
}
\affiliation{%
  \institution{Department AIBE, Friedrich-Alexander-Universität (FAU)}
  \country{Erlangen-Nürnberg, Germany}
}

\author{Alexandros Karargyris}
\affiliation{%
  \institution{IHU Strasbourg}
  \country{Strasbourg, France}
}

\author{Alan Karthikesalingam}
\affiliation{%
  \institution{Google Health DeepMind}
  \country{London, UK}
}

\author{Hannes Kenngott}
\affiliation{%
  \institution{Department of General, Visceral and Transplantation Surgery, Heidelberg University Hospital}
  \country{Heidelberg, Germany}
}

\author{Jens Kleesiek}
\affiliation{%
  \institution{Translational Image-guided Oncology (TIO), Institute for AI in Medicine (IKIM), University Medicine Essen}
  \country{Essen, Germany}
}

\author{Florian Kofler}
\affiliation{%
  \institution{Helmholtz AI}
  \country{München, Germany}
}

\author{Thijs Kooi}
\affiliation{%
  \institution{Lunit}
  \country{Seoul, South Korea}
}

\author{Annette Kopp-Schneider}
\affiliation{%
  \institution{German Cancer Research Center (DKFZ) Heidelberg, Division of Biostatistics}
  \country{Germany}
}

\author{Michal Kozubek}
\affiliation{%
  \institution{Centre for Biomedical Image Analysis and Faculty of Informatics, Masaryk University}
  \country{Brno, Czech Republic}
}

\author{Anna Kreshuk}
\affiliation{%
  \institution{Cell Biology and Biophysics Unit, European Molecular Biology Laboratory (EMBL)}
  \country{Heidelberg, Germany}
}

\author{Tahsin Kurc}
\affiliation{%
  \institution{Department of Biomedical Informatics, Stony Brook University}
  \country{Stony Brook, NY, USA}
}

\author{Bennett A. Landman}
\affiliation{%
  \institution{Electrical Engineering, Vanderbilt University}
  \country{Nashville, TN, USA}
}

\author{Geert Litjens}
\affiliation{%
  \institution{Department of Pathology, Radboud University Medical Center}
  \country{Nijmegen, The Netherlands}
}

\author{Amin Madani}
\affiliation{%
  \institution{Department of Surgery, University Health Network}
  \country{Philadelphia, PA, Canada}
}

\author{Klaus Maier-Hein}
\affiliation{%
  \institution{German Cancer Research Center (DKFZ) Heidelberg, Division of Medical Image Computing and HI Helmholtz Imaging}
  \country{Germany}
}
\affiliation{%
  \institution{Pattern Analysis and Learning Group, Department of Radiation Oncology, Heidelberg University Hospital}
  \country{Heidelberg, Germany}
}
\author{Anne L. Martel}
\affiliation{%
  \institution{Physical Sciences, Sunnybrook Research Institute}
    \country{Toronto, Canada}
}
\affiliation{%
  \institution{Department of Medical Biophysics, University of Toronto}
  \country{Toronto, ON, Canada}
}

\author{Peter Mattson}
\affiliation{%
  \institution{Google}
  \country{Mountain View, USA}
}

\author{Erik Meijering}
\affiliation{%
  \institution{School of Computer Science and Engineering, University of New South Wales}
  \country{Sydney, Australia}
}

\author{Bjoern Menze}
\affiliation{%
  \institution{Department of Quantitative Biomedicine, University of Zurich}
  \country{Zurich, Switzerland}
}

\author{Karel G.M. Moons}
\affiliation{%
  \institution{Julius Center for Health Sciences and Primary Care, UMC Utrecht, Utrecht University}
  \country{Utrecht, The Netherlands}
}

\author{Henning Müller}
\affiliation{%
  \institution{Information Systems Institute, University of Applied Sciences Western Switzerland (HES-SO)}
  \country{Sierre, Switzerland}
}
\affiliation{%
  \institution{Medical Faculty, University of Geneva}
  \country{Geneva, Switzerland}
}

\author{Brennan Nichyporuk}
\affiliation{%
  \institution{MILA (Quebec Artificial Intelligence Institute)}
  \country{Montréal, Canada}}
  
\author{Felix Nickel}
\affiliation{%
  \institution{Department of General, Visceral and Thoracic Surgery, University Medical Center Hamburg-Eppendorf}
  \country{Hamburg, Germany}
}

\author{Jens Petersen}
\affiliation{%
  \institution{German Cancer Research Center (DKFZ) Heidelberg, Division of Medical Image Computing}
  \country{Germany}
}

\author{Susanne M. Rafelski}
\affiliation{%
  \institution{Allen Institute for Cell Science}
  \country{Seattle, WA, USA}}

\author{Nasir Rajpoot}
\affiliation{%
  \institution{Tissue Image Analytics Laboratory, Department of Computer Science, University of Warwick}
  \department{Tissue Image Analytics Laboratory, Department of Computer Science}
  \country{Coventry, UK}
}

\author{Mauricio Reyes}
\affiliation{%
  \institution{ARTORG Center for Biomedical Engineering Research, University of Bern}
  \country{Bern, Switzerland}
}
\affiliation{%
  \institution{Department of Radiation Oncology, University Hospital Bern, University of Bern}
  \country{Bern, Switzerland}
}

\author{Michael A. Riegler}
\affiliation{%
  \institution{Simula Metropolitan Center for Digital Engineering}
  \country{Oslo, Norway}
}
\affiliation{%
  \institution{UiT The Arctic University of Norway}
  \country{Tromsø, Norway}
}

\author{Nicola Rieke}
\affiliation{%
  \institution{NVIDIA GmbH}
  \country{München, Germany}
}

\author{Julio Saez-Rodriguez}
\affiliation{%
  \institution{Institute for Computational Biomedicine, Heidelberg University}
  \country{Heidelberg. Germany}
}
\affiliation{%
  \institution{Faculty of Medicine, Heidelberg University Hospital}
  \country{Heidelberg, Germany}
}

\author{Clara I. Sánchez}
\affiliation{%
  \institution{Informatics Institute, Faculty of Science, University of Amsterdam}
  \country{Amsterdam, The Netherlands}
}

\author{Shravya Shetty}
\affiliation{%
  \institution{Google Health, Google}
  \country{CA, USA}
}

\author{Ronald M. Summers}
\affiliation{%
  \institution{National Institutes of Health Clinical Center}
  \country{Bethesda, MD, USA}
}

\author{Abdel A. Taha}
\affiliation{%
\institution{Institute of Information Systems Engineering, TU Wien}
  \country{Vienna, Austria}
}

\author{Aleksei Tiulpin}
\affiliation{%
\institution{Research Unit of Health Sciences and Technology, Faculty of Medicine, University of Oulu}
  \country{Oulu, Finland}
}
\affiliation{%
\institution{Neurocenter Oulu, Oulu University Hospital}
  \country{Oulu, Finland}
}

\author{Sotirios A. Tsaftaris}
\affiliation{%
  \institution{School of Engineering, The University of Edinburgh}
  \country{Edinburgh, Scotland}
}

\author{Ben Van Calster}
\affiliation{%
  \institution{Department of Development and Regeneration and EPI-centre, KU Leuven}
  \country{Leuven, Belgium}
}
\affiliation{%
  \institution{Department of Biomedical Data Sciences, Leiden University Medical Center}
  \country{Leiden, The Netherlands}
}

\author{Gaël Varoquaux}
\affiliation{%
  \institution{Parietal project team, INRIA Saclay-Île de France}
  \country{Palaiseau, France}
}

\author{Ziv R. Yaniv}
\affiliation{%
  \institution{National Institute of Allergy and Infectious Diseases, National Institutes of Health}
  \country{Bethesda, MD, USA}
}

\author{Paul F. Jäger}
\authornote{Shared last authors: Paul F. Jäger and Lena Maier-Hein}
\affiliation{%
  \institution{German Cancer Research Center (DKFZ) Heidelberg, Interactive Machine Learning Group and HI Helmholtz Imaging}
  \country{Germany}
}
\author{Lena Maier-Hein}
\authornotemark[3]
\affiliation{%
  \institution{German Cancer Research Center (DKFZ) Heidelberg, Division of Intelligent Medical Systems and HI Helmholtz Imaging}
  \country{Germany}
}
\affiliation{%
  \institution{Faculty of Mathematics and Computer Science and Medical Faculty, Heidelberg University}
  \country{Heidelberg, Germany}
}
\affiliation{%
  \institution{National Center for Tumor Diseases (NCT), NCT Heidelberg, a partnership between DKFZ and University Medical Center Heidelberg}
  \country{Germany}
}

\renewcommand{\shortauthors}{Reinke/Tizabi et al.}

\begin{abstract}
\newpage
\textbf{Abstract:} 
Validation metrics are key for tracking scientific progress and bridging the current chasm between \ac{AI} research and its translation into practice.
However, increasing evidence shows that particularly in image analysis, metrics are often chosen inadequately. While taking into account the individual strengths, weaknesses, and limitations of validation metrics is a critical prerequisite to making educated choices, the relevant knowledge is currently scattered and poorly accessible to individual researchers. Based on a multi-stage Delphi process conducted by a multidisciplinary expert consortium as well as extensive community feedback, the present work provides the first reliable and comprehensive common point of access to information on pitfalls related to validation metrics in image analysis. While focused on biomedical image analysis, the addressed pitfalls generalize across application domains and are categorized according to a newly created, domain-agnostic taxonomy. The work serves to enhance global comprehension of a key topic in image analysis validation.
\end{abstract}


\maketitle
\setlength{\parskip}{0.5em}

\newpage
\acresetall
\section*{Main}
\begin{figure}
    \centering
    \includegraphics[width=0.8\linewidth]{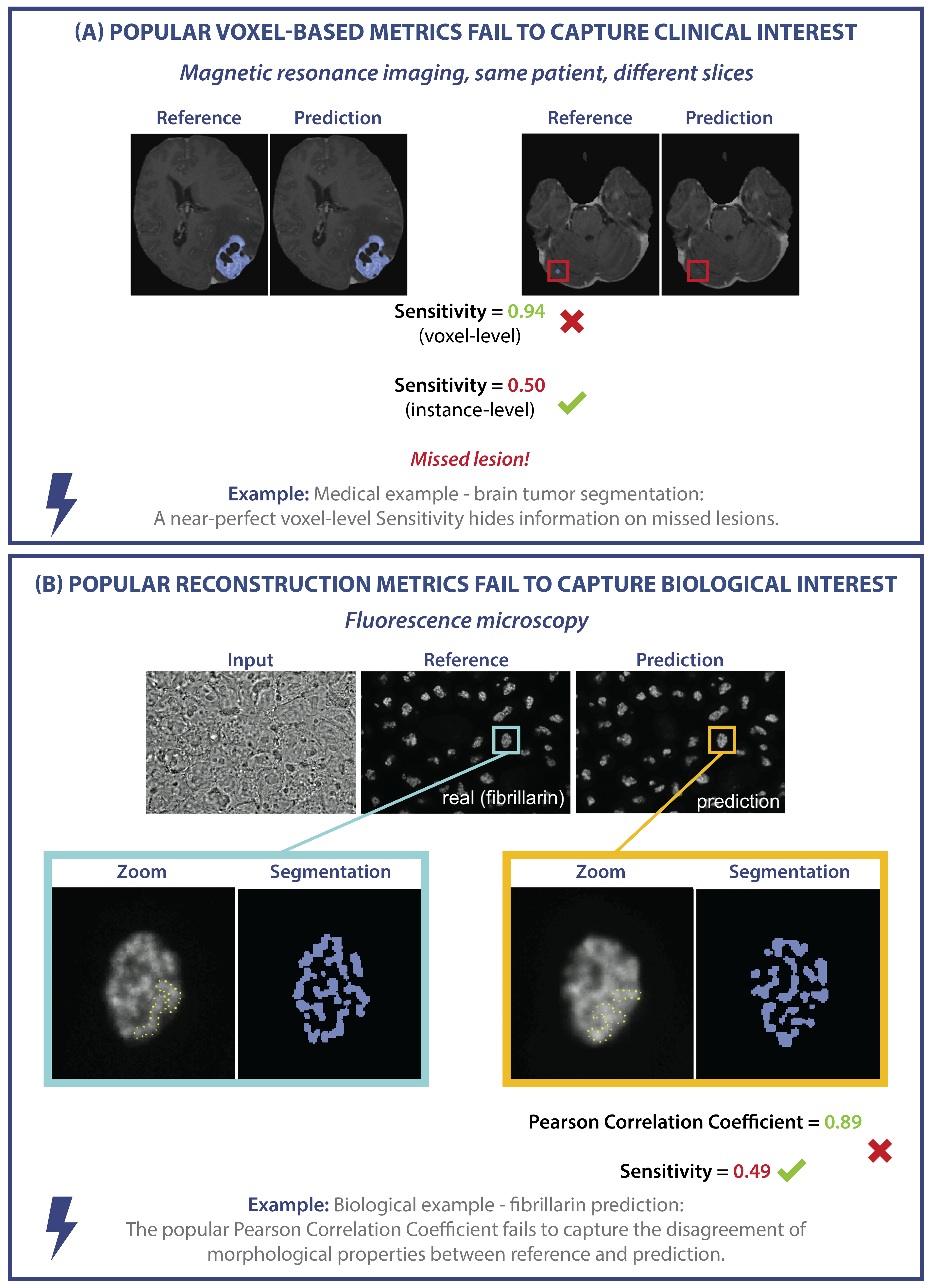}
    \caption{Examples of metric-related pitfalls in image analysis validation. (A) Medical image analysis example: Voxel-based metrics are not appropriate for detection problems. Measuring the voxel-level performance of a prediction yields a near-perfect Sensitivity. However, the Sensitivity at the instance level reveals that lesions are actually missed by the algorithm. (B) Biological image analysis example: The task of predicting fibrillarin in the dense fibrillary component of the nucleolus should be phrased as a segmentation task, for which segmentation metrics reveal the low quality of the prediction. Phrasing the task as image reconstruction instead and validating it using metrics such as the Person Correlation Coefficient yields misleadingly high metric scores  \citep{roberts2017systematic, chen2020allen, viana2023integrated, ounkomol2018label, viana2023integrated}.}
    \label{fig:figure1}
\end{figure}

\addcontentsline{toc}{chapter}{\protect\numberline{}Main}
\label{sec:main}
Measuring performance and progress in any given field critically depends on the availability of meaningful outcome metrics. In a field such as athletics, this process is straightforward because the performance measurements (e.g., the time it takes an athlete to run a given distance) exactly reflect the underlying interest (e.g., which athlete runs a given distance the fastest?). In image analysis, the situation is much more complex. Depending on the underlying research question, vastly different aspects of an algorithm’s performance might be of interest (Fig.~\ref{fig:figure1}) and meaningful in determining its future practical, for example clinical, applicability. If the performance of an image analysis algorithm is not measured according to relevant validation metrics, no reliable statement can be made about the suitability of this algorithm in solving the proposed task, and the algorithm is unlikely to ever reach the stage of real-life application. Moreover, unsuitable algorithms could be wrongly regarded as the best-performing ones, sparking entirely futile resource investment and follow-up research while obscuring true scientific advancements. In determining new state-of-the-art methods and informing future directions, the use of validation metrics actively shapes the evolution of research. In summary, \textit{validation metrics are the key for both measuring and informing scientific progress, as well as bridging the current chasm between image analysis research and its translation into practice}.

In image analysis, while for some applications it might, for instance, be sufficient to draw a box around the structure of interest (e.g., detecting individual mitotic cells or regions with apoptotic cell debrisa) and optionally associate that region with a classification (e.g., a mitotic vs an interphase cell), other applications (e.g., cell tracing for fluorescent signal quantification) could require determining the exact structure boundaries. The suitability of any individual validation metric thus depends crucially on the properties of the driving image analysis problem. 
As a result, numerous metrics have so far been proposed in the field of image processing. In our previous work, we analyzed all biomedical image analysis competitions conducted within a period of about 15 years~\cite{maier2018rankings}. We found a total of 97 different metrics reported in the field of biomedicine alone, each with its own individual strengths, weaknesses, and limitations, and hence varying degrees of suitability for meaningfully measuring algorithm performance on any given research problem. Such a vast range of options makes tracking all related information impossible for any individual researcher and consequently renders the process of metric selection error-prone. Thus, the frequent reliance on flawed, historically grown validation practices in current literature comes as no surprise. To make matters worse, there is currently no comprehensive resource that can provide an overview of the relevant definitions, (mathematical) properties, limitations, and pitfalls pertaining to a metric of interest. \textit{While taking into account the individual properties and limitations of metrics is imperative for choosing adequate validation metrics, the required knowledge is thus largely inaccessible.} 

As a result, numerous flaws and pitfalls are prevalent in image analysis validation, with researchers often being unaware of them due to a lack of knowledge of intricate metric properties and limitations. Accordingly, increasing evidence shows that metrics are often selected inadequately in image analysis (e.g., \citep{kofler2021DICE, gooding2018comparative, vaassen2020evaluation}). In the absence of a central information resource, it is common for researchers to resort to popular validation metrics, which, however, can be entirely unsuitable, for instance due to a mismatch of the metric's inherent mathematical properties with the underlying research question and specifications of the data set at hand (see Fig.~\ref{fig:figure1}). 

The present work addresses this important roadblock in image analysis research with a crowd\-sourcing-based approach that involved both a Delphi process undergone by a multidisciplinary expert consortium as well as a social media campaign. It represents the \textit{first comprehensive collection, visualization, and detailed discussion of pitfalls, drawbacks, and limitations regarding validation metrics commonly used in image analysis}. Our work provides researchers with a \textit{reliable, single point of access} to this critical information. Owing to the enormous complexity of the matter, 
the metric properties and pitfalls are discussed in the specific context of classification problems, i.e., image analysis problems that can be considered classification tasks at either the image, object, or pixel level. Specifically, these encompass the four problem categories of image-level classification, semantic segmentation, object detection, and instance segmentation. Our contribution includes a dedicated profile for each metric (\ref{app:steckbriefe}) as well as the creation of a new common taxonomy that categorizes pitfalls in a domain-agnostic manner (Fig.~\ref{fig:overview-taxonomy}). The taxonomy is depicted for individual metrics in provided tables (see Extended Data Tabs.~\ref{tab:pitfalls-overview-ilc}-\ref{tab:pitfalls-overview-is-2}) and enables researchers to quickly grasp whether using a certain metric comes with pitfalls in a given use case. 
While our work grew out of image analysis research and practice in the field of biomedicine, a field of high complexity and particularly high stakes due to its direct impact on human health, we believe the identified pitfalls to be transferable to other application areas of imaging research. It should be noted that this work focuses on identifying, categorizing, and illustrating metric pitfalls, while the sister publication of this work gives specific recommendations on which metrics to apply under which circumstances \cite{maier2022metrics}.




\section*{Results}
\addcontentsline{toc}{chapter}{\protect\numberline{}Results}
\label{sec:results}

\subsection*{Information on metric pitfalls is largely inaccessible}
Researchers and algorithm developers seeking to validate image analysis algorithms frequently face the problem of choosing adequate validation metrics while at the same time navigating a range of potential pitfalls. Following common practice is often not the best option, as evidenced by a number of recent publications \citep{maier2018rankings, kofler2021DICE, gooding2018comparative, vaassen2020evaluation}. Making an educated choice is notably complicated by the absence of any comprehensive databases or reviews covering the topic and thus the lack of a central resource for reliable information on validation metrics.

This lack of accessibility is considered by experts to be a major bottleneck in image analysis validation~\cite{maier2018rankings}. To illustrate this point, we searched the literature for available information on commonly used validation metrics. The search was conducted on the platform Google Scholar using search strings that combined different notations of the metric name, including synonyms and acronyms, with search terms indicating problems, such as “pitfall” or “limitation”. The mean and median number of hits for the metrics addressed in the present work were 159,329 and 22,100, respectively, and ranged between 49 for \ac{clDice} and 962,000 for Sensitivity. Moreover, despite valuable literature on individual relevant aspects (e.g.,~\cite{taha2014formal, taha2015metrics, grandini2020metrics, kofler2021DICE, vaassen2020evaluation, chicco2021matthews, chicco2020advantages}), we did not find a common point of entry to metric-related pitfalls in image analysis in the form of a review paper or other credible source.
We conclude that 
\textit{the key knowledge required for making educated decisions and avoiding pitfalls related to the use of validation metrics is highly scattered and not accessible by individuals.} 

\subsection*{Historically grown practices are not always justified}
\label{ssec:historic}

To obtain an initial insight into current common practice regarding validation metrics, we prospectively captured the designs of challenges organized by the IEEE Society of the International Symposium of Biomedical Imaging (ISBI), the \ac{MICCAI} Society and the Medical Imaging with Deep Learning (MIDL) foundation. The organizers of the respective competitions were asked to provide a rationale for the choice of metrics in their competition. An analysis of a total of 138 competitions conducted between 2018 and 2022 revealed that metrics are frequently (in $24\%$ of the competitions) based on common practice in the community. We found, however, that common practices are often not well-justified, and poor practices may even be propagated from one generation to the next.

One remarkable example for this issue is the widespread adoption of an incorrect naming and inconsistent mathematical formulation of a metric proposed for cell instance segmentation. The term "\ac{mAP}" usually refers to one of the most common metrics in object detection (object-level classification)~\cite{lin2014microsoft, reinke2021common}. Here, Precision denotes the \ac{PPV}, which is "averaged" over varying thresholds on the predicted class scores of an object detection algorithm. The "mean" \ac{AP} is then obtained by taking the mean over classes \cite{everingham2010pascal, reinke2021common}. Despite the popularity of \ac{mAP}, a widely known challenge on cell instance segmentation\footnote{\url{https://www.kaggle.com/competitions/data-science-bowl-2018/overview/evaluation}} introduced a new "Mean Average Precision" in 2018. Although the task matches the task of the original "mean" \ac{AP}, object detection, all terms in the newly proposed metric (mean, average, and precision) refer to entirely different concepts. For instance, the common definition of Precision from literature $\mathrm{TP} / (\mathrm{TP} + \mathrm{FP})$ was altered to $\mathrm{TP} / (\mathrm{TP} + \mathrm{FP} + \mathrm{FN})$, where TP, FP, and FN refer to the cardinalities of the confusion matrix (i.e., the true/false positives/negatives). The latter formula actually defines the \ac{IoU} metric. Despite these problems, the terminology was adopted by subsequent influential works \cite{schmidt2018cell,stringer2021cellpose,satorius, hirling2023segmentation}, indicating widespread propagation and usage within the community.

\subsection*{A multidisciplinary Delphi process reveals numerous pitfalls in biomedical image analysis validation}
With the aim of creating a comprehensive, reliable collection and future point of access to biomedical image analysis metric definitions and limitations, we formed an international multidisciplinary consortium of 62 experts from various biomedical image analysis-related fields that engaged in a multi-stage Delphi process \cite{brown1968delphi} for consensus building. The Delphi process comprised multiple surveys, developed by a coordinating team and filled out by the remaining members of the consortium. Based on the survey results, the list of pitfalls was iteratively refined by collecting pitfall sources, specific feedback and suggestions on pitfalls, and final agreement on which pitfalls to include and how to illustrate them. Further pitfalls were crowdsourced through the publication of a dynamic preprint of this work~\cite{reinke2021common} as well as a social media campaign, both of which asked the scientific community for contributions. 
This approach allowed us to integrate distributed, cross-domain knowledge on metric-related pitfalls within a single resource.
In total, the process revealed 37 distinct sources of pitfalls (see Fig.~\ref{fig:overview-taxonomy}). Notably, these pitfall sources (e.g., class imbalances, uncertainties in the reference, or poor image resolution) can occur irrespective of a specific imaging modality or application. As a result, many pitfalls generalize across different problem categories in image processing (image-level classification, semantic segmentation, object detection, and instance segmentation), as well as imaging modalities and domains. A detailed discussion of all pitfalls can be found in \ref{app:pitfalls}.

\subsection*{A common taxonomy enables domain-agnostic categorization of pitfalls}
One of our key objectives was to facilitate information retrieval and provide structure within this vast topic. Specifically, we wanted to enable researchers to identify at a glance which metrics are affected by which types of pitfalls. To this end, we created a comprehensive taxonomy that categorizes the different pitfalls in a semantic fashion. The taxonomy was created in a domain-agnostic manner to reflect the generalization of pitfalls across different imaging domains and modalities. An overview of the taxonomy is presented in Fig.~\ref{fig:overview-taxonomy}, and the relations between the pitfall categories and individual metrics can be found in Extended Data Tabs.~\ref{tab:pitfalls-overview-ilc}-\ref{tab:pitfalls-overview-is-2}. We distinguish the following three main categories:

\newpage
\begin{figure}[H]
    \centering
    \includegraphics[width=1\linewidth]{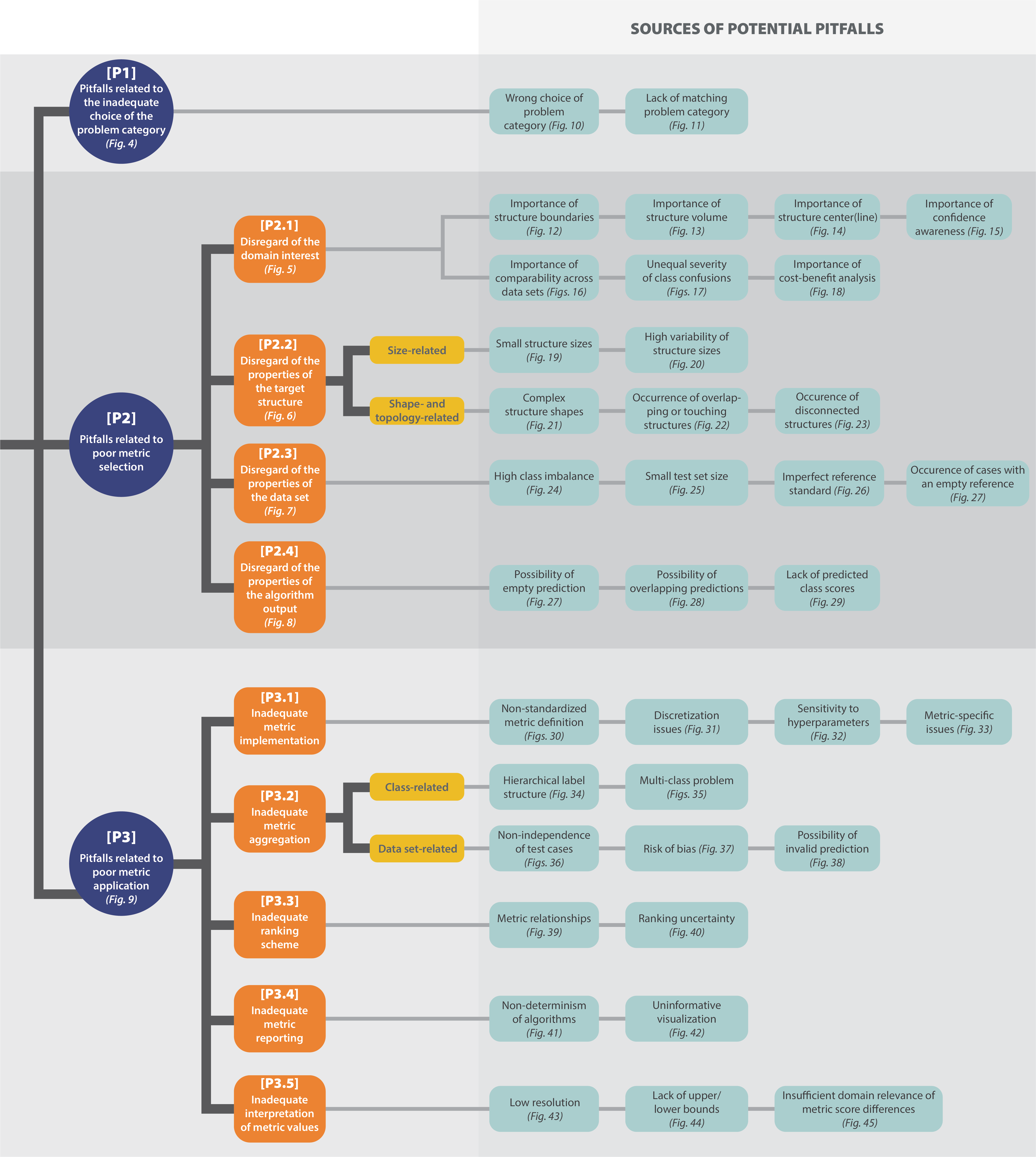}
    \caption{Overview of the taxonomy for metric-related pitfalls. Pitfalls can be grouped into three main categories: [P1] Pitfalls related to the inadequate choice of the problem category, [P2]  pitfalls related to poor metric selection, and [P3] pitfalls related to poor metric application. [P2] and [P3] are further split into subcategories. For all categories, pitfall sources are presented (green), with references to corresponding illustrations of representative examples. Note that the order in which the pitfall sources are presented does not correlate with importance.}
    \label{fig:overview-taxonomy}
\end{figure}

\newpage
\paragraph{\textbf{[P1] Pitfalls related to the inadequate choice of the problem category.}} 
A common pitfall lies in the use of metrics for a problem category they are not suited for because they fail to fulfill crucial requirements of that problem category, and hence do not reflect the domain interest (Fig.~\ref{fig:figure1}).
For instance, popular voxel-based metrics, such as the \ac{DSC} or Sensitivity, are widely used in image analysis problems, although they do not fulfill the critical requirement of detecting all objects in a data set. In a cancer monitoring application they fail to measure instance progress, i.e., the potential increase in number of lesions~(Fig.~\ref{fig:figure1}), which can have serious consequences for the patient. For some problems, there may even be a lack of matching problem category~(Fig.~\ref{fig:context-ratio}), rendering common metrics inadequate. 
We present further examples of pitfalls in this category in Suppl. Note~\ref{sec:pitf:underlying-task}.

\paragraph{\textbf{[P2] Pitfalls related to poor metric selection.}} Pitfalls of this category occur when a validation metric is selected while disregarding specific properties of the given research problem or method used that make this metric unsuitable in the particular context.
[P2] can be further divided into the following four subcategories:

\paragraph{[P2.1] Disregard of the domain interest} Commonly, several requirements arise from the domain interest of the underlying research problem that may clash with particular metric limitations. For example, if there is particular interest in the structure boundaries, it is important to know that overlap-based metrics such as the \ac{DSC} do not take the correctness of an object's boundaries into account, as shown in Fig.~\ref{fig:pitfalls-p2-1}(a). Similar issues may arise if the structure volume~(Fig.~\ref{fig:outline}) or center(line)~(Fig.~\ref{fig:center}) are of particular interest.
Other domain interest-related properties may include an unequal severity of class confusions. This may be important in an ordinal grading use case, in which the severity of a disease is categorized by different scores. Predicting a low severity for a patient that actually suffers from a severe disease should be substantially penalized. Common classification metrics do not fulfill this requirement. An example is provided in Fig.~\ref{fig:pitfalls-p2-1}(b). On pixel level, this property relates to an unequal severity of over- vs. undersegmentation. In applications such as radiotherapy, it may be highly relevant whether an algorithm tends to over- or undersegment the target structure. Common overlap-based metrics, however, do not represent over- and undersegmentation equally \citep{yeghiazaryan2018family}. Further pitfalls may occur if confidence awareness~(Fig.~\ref{fig:calibration}), comparability across data sets~(Fig.~\ref{fig:prevalence-dependency}), or a cost-benefit analysis~(Fig.~\ref{fig:cost-benefit}) are of particular importance, as illustrated in  Suppl. Note~\ref{sec:pitf:poor-selection-domain}.

\paragraph{[P2.2] Disregard of the properties of the target structures} For problems that require capturing local properties (object detection, semantic or instance segmentation), the properties of the target structures to be localized and/or segmented may have important implications for the choice of metrics. Here, we distinguish between \textit{size-related} and \textit{shape- and topology-related} pitfalls. Common metrics, for example, are sensitive to structure sizes, such that single-pixel differences may hugely impact the metric scores, as shown in Extended Data Fig.~\ref{fig:pitfalls-p2-2}(a). Shape- and topology-related pitfalls may relate to the fact that common metrics disregard complex shapes (Extended Data Fig.~\ref{fig:pitfalls-p2-2}(b)) or that bounding boxes do not capture the disconnectedness of structures (Fig.~\ref{fig:disconnected}). A high variability of structure sizes (Fig.~\ref{fig:high-variability}) and overlapping or touching structures (Fig.~\ref{fig:multi-labels}) may also influence metric values. We present further examples of [P2.2] pitfalls in Suppl. Note~\ref{sec:pitf:poor-selection-target}.

\newpage
\begin{figure}[H]
    \centering
    \includegraphics[width=0.9\linewidth]{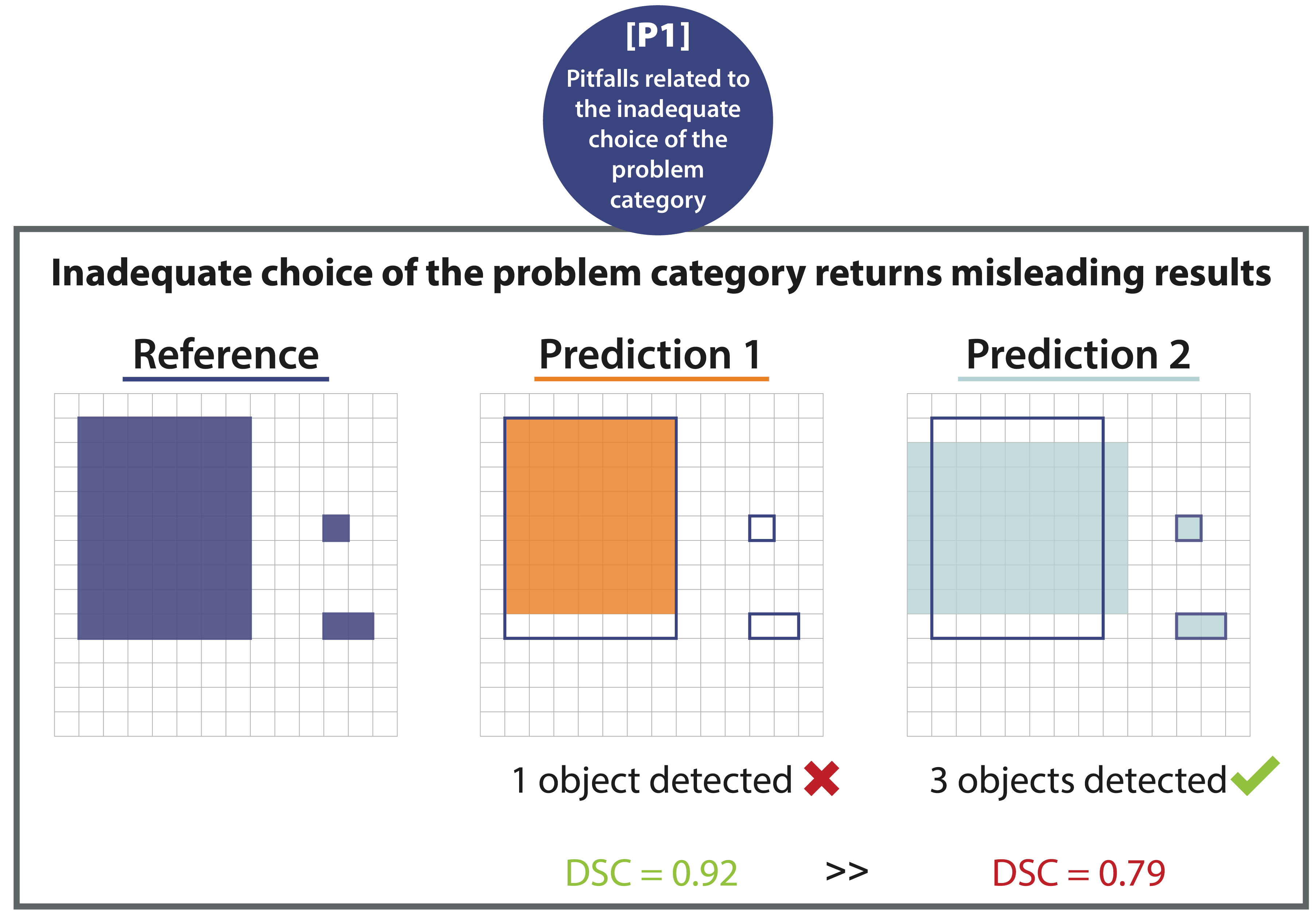}
    \caption{[P1] Pitfalls related to the inadequate choice of the problem category. \textbf{Wrong choice of problem category.} Effect of using segmentation metrics for object detection problems. The pixel-level \acf{DSC} of a prediction recognizing every structure (\textit{Prediction 2}) is lower than that of a prediction that only recognizes one of the three structures (\textit{Prediction 1}).}
    \label{fig:pitfalls-p1}
\end{figure}

\paragraph{[P2.3] Disregard of the properties of the data set} Various properties of the data set such as class imbalances (Fig.~\ref{fig:pitfalls-p2-3}(a)), small sample sizes (Fig.~\ref{fig:pitfalls-p2-3}(b)), or the quality of the reference annotations, may directly affect metric values. Common metrics such as the \ac{BA}, for instance, may yield a very high score for a model that predicts many \ac{FP} samples in an imbalanced setting (see Fig.~\ref{fig:pitfalls-p2-3}(a)). When only small test data sets are used, common calibration metrics (which are typically biased estimators) either underestimate or overestimate the true calibration error of a model (Fig.~\ref{fig:pitfalls-p2-3}(b)) \citep{gruber2022better}. On the other hand, metric values may be impacted by reference annotations (Fig.~\ref{fig:low-quality}). Spatial outliers in the reference may have a huge impact on distance-based metrics such as the \ac{HD} (Fig.~\ref{fig:pitfalls-p2-3}(c)). Additional pitfalls may arise from the occurrence of cases with an empty reference~(Extended Data Fig.~\ref{fig:pitfalls-p2-4}(b)), causing division by zero errors. We present further examples of [P2.3] pitfalls in Suppl. Note~\ref{sec:pitf:poor-selection-data-output}.

\paragraph{[P2.4] Disregard of the properties of the algorithm output} Reference-based metrics compare the algorithm output to a reference annotation to compute a metric score. Thus, the content and format of the prediction are of high importance when considering metric choice. Overlapping predictions in segmentation problems, for instance, may return misleading results. In Extended Data Fig.~\ref{fig:pitfalls-p2-4}(a), the predictions only overlap to a certain extent, not representing that the reference instances actually overlap substantially. This is not detected by common metrics. Another example are empty predictions that may cause division by zero errors in metric calculations, as illustrated in Extended Data Fig.~\ref{fig:pitfalls-p2-4}(b), or the lack of predicted class scores (Fig.~\ref{fig:lack-of-scores}).  
We present further examples of [P2.4] pitfalls in Suppl. Note~\ref{sec:pitf:poor-selection-data-output}.

\begin{figure}[H]
    \centering
    \includegraphics[width=0.75\linewidth]{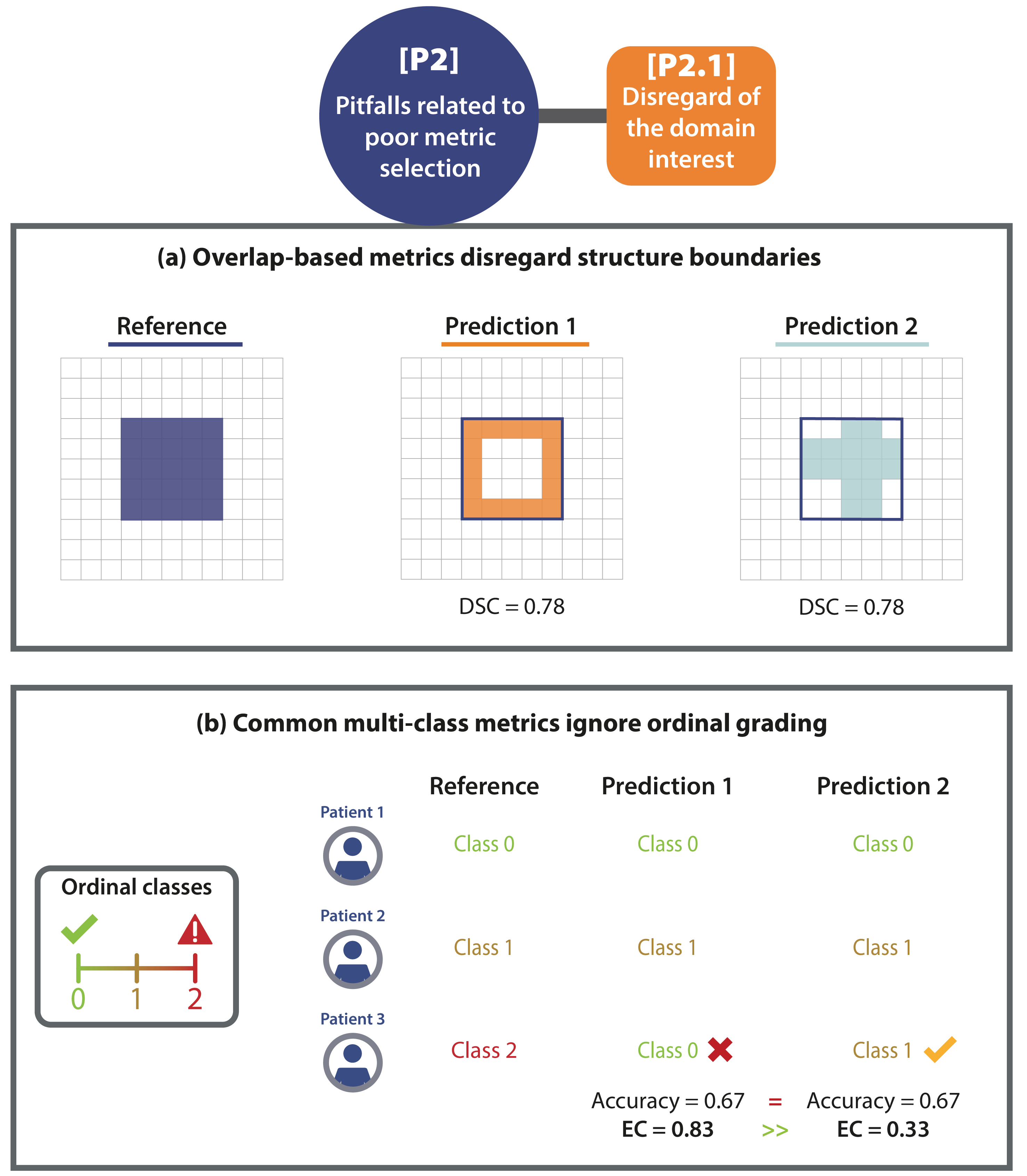}
    \caption{[P2.1] \textbf{Disregard of the domain interest}. \textbf{(a) Importance of structure boundaries.} The predictions of two algorithms (\textit{Prediction 1/2}) capture the boundary of the given structure substantially differently, but lead to the exact same \acf{DSC}, due to its boundary unawareness. This pitfall is also relevant for other overlap-based metrics such as \acf{clDice}, pixel-level F$_\beta$ Score, and \acf{IoU}, as well as localization criteria such as Box/Approx/Mask \ac{IoU}, Center Distance, Mask \ac{IoU} > 0, Point inside Mask/Box/Approx, and \acf{IoR}. \textbf{(b) Unequal severity of class confusions.} When predicting the severity of a disease for three patients in an ordinal classification problem, \textit{Prediction 1} assumes a much lower severity for \textit{Patient 3} than actually observed. This critical issue is overlooked by common metrics (here: Accuracy), which measure no difference to \textit{Prediction 2}, which assesses the severity much better. Metrics with pre-defined weights (here: \acf{EC}) correctly penalize \textit{Prediction 1} much more than \textit{Prediction 2}. This pitfall is also relevant for other counting metrics, such as \acf{BA}, F$_\beta$ Score, \acf{LR+}, \acf{MCC}, \acf{NB}, \acf{NPV}, \acf{PPV}, Sensitivity, and Specificity.}
    \label{fig:pitfalls-p2-1}
\end{figure}


\begin{figure}[H]
    \centering
    \includegraphics[width=1\linewidth]{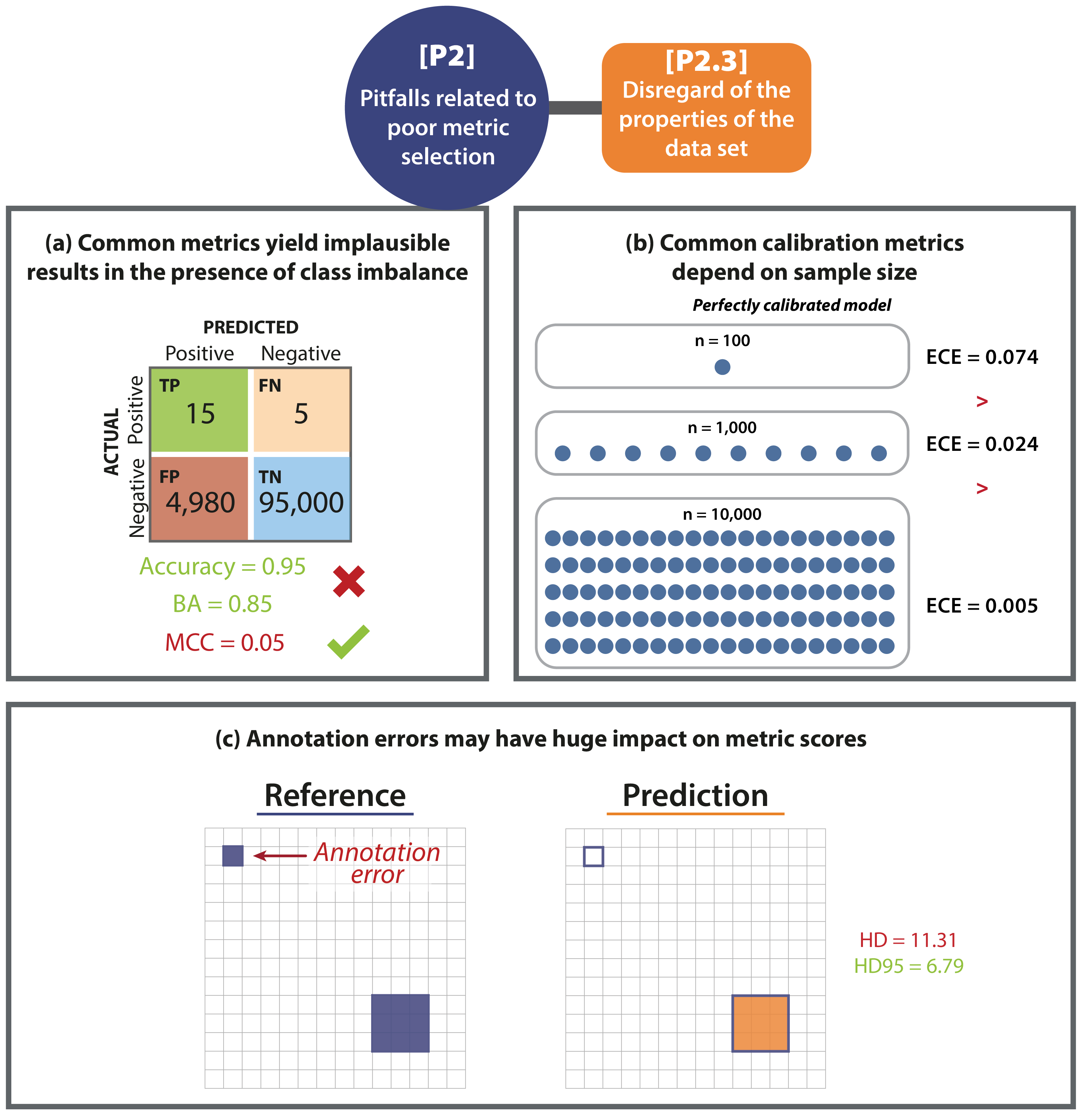}
    \caption{[P2.3] \textbf{Disregard of the properties of the data set}. \textbf{(a) High class imbalance.} In the case of underrepresented classes, common metrics may yield misleading values. In the given example, Accuracy and \acf{BA} have a high score despite the high amount of \acf{FP} samples. The class imbalance is only uncovered by metrics considering predictive values (here: \acf{MCC}). This pitfall is also relevant for other counting and multi-threshold metrics such as \acf{AUROC}, \acf{EC} (depending on the chosen costs), \acf{LR+}, \acf{NB}, Sensitivity, Specificity, and \acf{WCK}. \textbf{(b) Small test set size.} The values of the \acf{ECE} depend on the sample size. Even for a simulated perfectly calibrated model, the \ac{ECE} will be substantially greater than zero for small sample sizes \citep{gruber2022better}. \textbf{(c) Imperfect reference standard.} A single erroneously annotated pixel may lead to a large decrease in performance, especially in the case of the \acf{HD} when applied to small structures. The \acf{HD95}, on the other hand, was designed to deal with spatial outliers. This pitfall is also relevant for localization criteria such as Box/Approx \acf{IoU} and Point inside Box/Approx. Further abbreviations: \acf{TP}, \acf{FN}, \acf{TN}.}
    \label{fig:pitfalls-p2-3}
\end{figure}

\begin{figure}[H]
    \centering
    \includegraphics[width=0.8\linewidth]{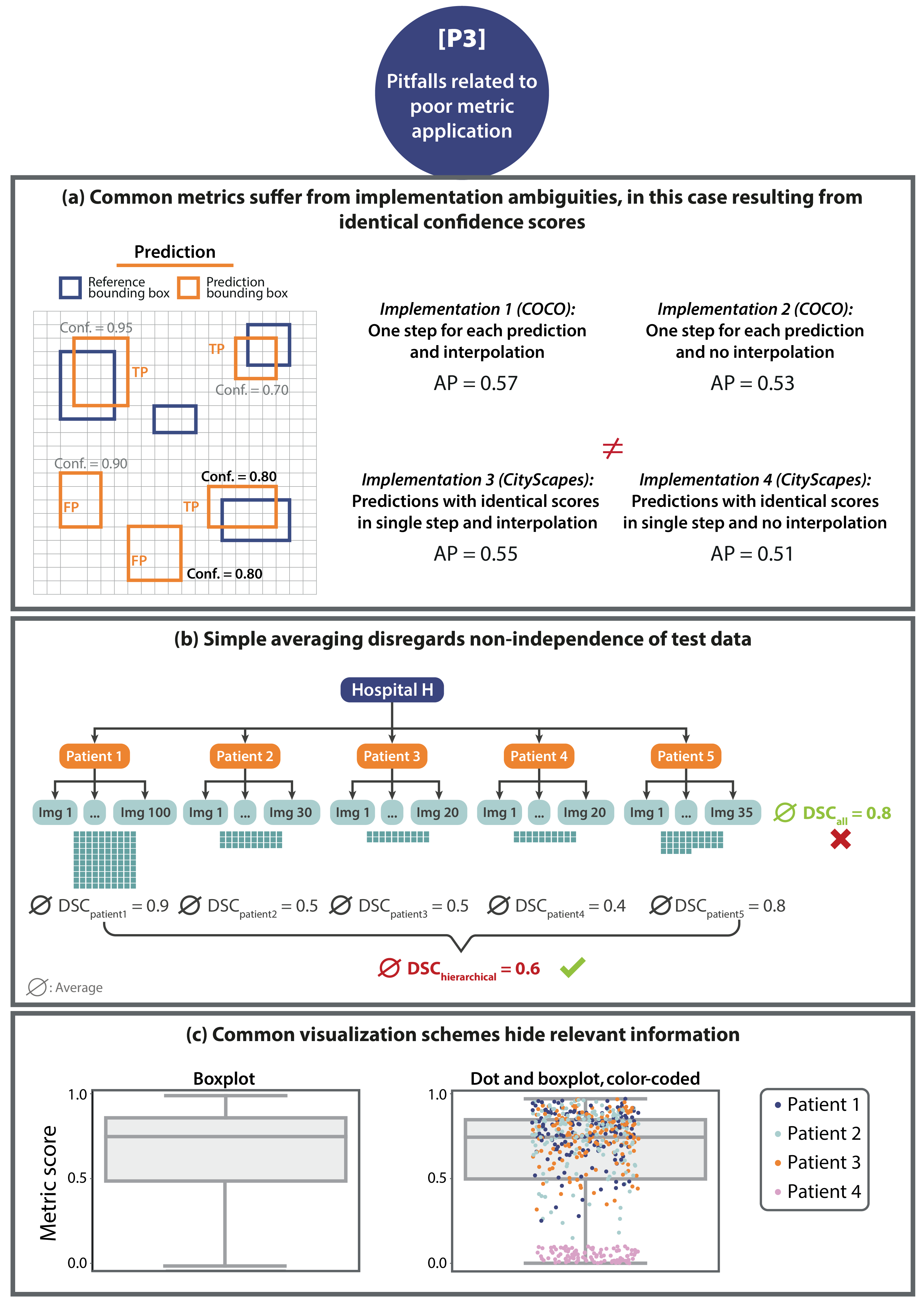}
    \caption{[P3] \textbf{Pitfalls related to poor metric application}.  \textbf{(a) Non-standardized metric implementation.} In the case of the \acf{AP} metric and the construction of the \acf{PR}-curve, the strategy of how identical scores (here: confidence score of 0.80 is present twice) are treated has a substantial impact on the metric scores. Microsoft \acf{COCO} \citep{lin2014microsoft} and CityScapes \citep{Cordts2015Cvprw} are used as examples. \textbf{(b) Non-independence of test cases.} The number of images taken from \textit{Patient 1} is much higher compared to that acquired from \textit{Patients 2-5}. Averaging over all \acf{DSC} values, denoted by $\varnothing$, results in a high averaged score. Aggregating metric values per patient reveals much higher scores for \textit{Patient 1} compared to the others, which would have been hidden by simple aggregation. \textbf{(c) Uninformative visualization.} A single box plot (left) does not give sufficient information about the raw metric value distribution. Adding the raw metric values as jittered dots on top (right) adds important information (here: on clusters). In the case of non-independent validation data, color/shape-coding helps reveal data clusters.}
    \label{fig:pitfalls-p3}
\end{figure}

\paragraph{\textbf{[P3] Pitfalls related to poor metric application.}} Once selected, the metrics need to be applied to an image or an entire data set. This step is not straightforward and comes with several pitfalls. For instance, when aggregating metric values over multiple images or patients, a common mistake is to ignore the hierarchical data structure, such as data from several hospitals or a varied number of images per patient. We present three examples of [P3] pitfalls in Fig.~\ref{fig:pitfalls-p3}; for more pitfalls in this category, please refer to Suppl. Note~\ref{sec:pitf:poor-application}. [P3] can further be divided into five subcategories that are presented in the following paragraphs.

\paragraph{[P3.1] Inadequate metric implementation} Metric implementation is, unfortunately, not standardized. As shown by \citep{gooding2022multicenter}, different researchers typically employ various different implementations for the same metric, which may yield a substantial variation in the metric scores. While some metrics are straightforward to implement, others require more advanced techniques and offer different possibilities. In the following, we provide some examples for inadequate metric implementation:
\begin{itemize}
    \item The method of how identical confidence scores are handled in the computation of the \ac{AP} metric may lead to substantial differences in the metric scores. Microsoft \ac{COCO} \citep{lin2014microsoft}, for instance, processes each prediction individually, while CityScapes \citep{Cordts2015Cvprw} processes all predictions with the same score in one joint step. Fig.~\ref{fig:pitfalls-p3}(a) provides an example with two predictions having the same confidence score, in which the final metric scores differ depending on the chosen handling strategy for identical confidence scores. Similar issues may arise with other curve-based metrics, such as \ac{AUROC}, \ac{AP}, or \ac{FROC} scores (see e.g.,~\citep{muschelli2020roc}).
    \item Metric implementation may be subject to discretization issues such as the chosen discretization of continuous variables, which may cause differences in the metric scores, as exemplary illustrated in Fig.~\ref{fig:ece-mce-bins}.
    \item For metrics assessing structure boundaries, such as the \ac{ASSD}, the exact boundary extraction method is not standardized. Thus, for example, the boundary extraction method implemented by the Liver Tumor Segmentation (LiTS) challenge \cite{bilic2023liver} and that implemented by Google DeepMind\footnote{\url{https://github.com/deepmind/surface-distance}} may produce different metric scores for the \ac{ASSD}. This is especially critical for metrics that are sensitive to small contour changes, such as the \ac{HD}.
    \item Suboptimal choices of hyperparameters may also lead to metric scores that do not reflect the domain interest. For example, the choice of a threshold on a localization criterion (see Fig.~\ref{fig:sensitivity-hyperparam}) or the chosen hyperparameter for the F$_\beta$ Score will heavily influence the subsequent metric scores~\cite{tran2022sources}.
\end{itemize}
More [P3.1] pitfalls can be found in Suppl. Note~\ref{sec:pitf:poor-application-impl}.

\paragraph{[P3.2] Inadequate metric aggregation} A common pitfall with respect to metric application is to simply aggregate metric values over the entire data set and/or all classes. As detailed in Fig.~\ref{fig:pitfalls-p3}(b) and Suppl. Note~\ref{sec:pitf:poor-application-aggregation}, important information may get lost in this process, and metric results can be misleading. For example, the popular \texttt{TorchMetrics} framework calculates the \ac{DSC} metric by default as a global average over all pixels in the data set without considering their image or class of origin\footnote{\url{https://torchmetrics.readthedocs.io/en/stable/classification/dice.html?highlight=dice}}. Such a calculation eliminates the possibility of interpreting the final metric score with respect to individual images and classes. For example, errors in small structures may be suppressed by correctly segmented larger structures in other images (see e.g., Fig.~\ref{fig:aggr-per-class}). An adequate aggregation scheme is also crucial for handling hierarchical class structure (Fig.~\ref{fig:multi-class-interdependencies}), missing values~(Fig.~\ref{fig:missings}), and potential biases~(Fig.~\ref{fig:stratification-gender}) of the algorithm. Further [P3.2] pitfalls are shown in Suppl. Note~\ref{sec:pitf:poor-application-aggregation}.

\paragraph{[P3.3] Inadequate ranking scheme} Rankings are often created to compare algorithm performances. In this context, several pitfalls pertain to either metric relationships or ranking uncertainty. For example, to assess different properties of an algorithm, it is advisable to select multiple metrics and determine their values. However, the chosen metrics should assess complementary properties and should not be mathematically related. For example, the \ac{DSC} and \ac{IoU} are closely related, so using both in combination would not provide any additional information over using either of them individually~(Fig.~\ref{fig:combination}). Note in this context that unawareness of metric synonyms can equally mislead. Metrics can be known under different names; for instance, Sensitivity and Recall refer to the same mathematical formula. Despite this fact potentially appearing trivial, an analysis of 138 biomedical image analysis challenges \citep{maier2022metrics} found three challenges that unknowingly used two versions of the same metric to calculate their rankings. Moreover, rankings themselves may be unstable~(Fig.~\ref{fig:ranking-uncertainty}). \citep{maier2018rankings} and \citep{wiesenfarth2021methods} demonstrated that rankings are highly sensitive to altering the metric aggregation operators, the underlying data set, or the general ranking method. Thus, if the robustness of rankings is disregarded, the winning algorithm may be identified by chance rather than true superiority. 

\paragraph{[P3.4] Inadequate metric reporting} A thorough reporting of metric values and aggregates is important both in terms of transparency and interpretability. However, several pitfalls are to be avoided in this regard. Notably, different types of visualization may vary substantially in terms of interpretability, as shown in Figs~\ref{fig:pitfalls-p3}(c). For example, while a box plot provides basic information, it does not depict the distribution of metric values. This may conceal important information, such as specific images on which an algorithm performed poorly. Other pitfalls in this category relate to the non-determinism of algorithms, which introduces a natural variability to the results of a neural network, even with fixed seeds~(Fig.~\ref{fig:non-determinism}). This issue is aggravated by inadequate reporting, for instance, reporting solely the results from the best run instead of proper cross-validation and reporting of the variability across different runs. Generally, shortcomings in reporting, such as providing no standard deviation or confidence intervals in the presented results, are common. Concrete examples of [P3.4] pitfalls can be found in Suppl. Note~\ref{sec:pitf:poor-application-reporting}.

\paragraph{[P3.5] Inadequate interpretation of metric values} Interpreting metric scores and aggregates is an important step for the analysis of algorithm performances. However, several pitfalls can arise from the interpretation. In rankings, for example, minor differences in metric scores may not be relevant from an application perspective but may still yield better ranks~(Fig.~\ref{fig:ranking-relevance}). Furthermore, some metrics do not have upper or lower bounds, or the theoretical bounds may not be achievable in practice, rendering interpretation difficult~(Fig.~\ref{fig:lack-bounds}). More information on interpretation-based pitfalls can be found in Suppl. Note~\ref{sec:pitf:poor-application-interpretation}.

\newpage



\newpage
\subsection*{The first illustrated common access point to metric definitions and pitfalls}
 To underline the importance of a common access point to metric pitfalls, we conducted a search for individual metric-related pitfalls on the platforms Google Scholar and Google, with the purpose of determining how many of the pitfalls identified through our work could be located in existing resources. We were only able to locate a portion of the pitfalls identified by our approach in existing research literature (68\%) or online resources such as blog posts (11\%; 8\% were found in both). Only 27\% of the located pitfalls were presented visually. \\

Our work now provides this key resource in a highly structured and easily understandable form. \ref{app:pitfalls}, contains a dedicated illustration for each of the pitfalls discussed, thus facilitating reader comprehension and making the information accessible to everyone regardless of their level of expertise. A further core contribution of our work are the metric profiles presented in \ref{app:pitfalls}, which, for each metric, summarize the most important information deemed of particular relevance by the \textit{Metrics Reloaded} consortium of the sister work to this publication ~\cite{maier2022metrics}. The profiles provide the reader with a compact, at-a-glance overview of each metric and an enumeration of the limitations and pitfalls identified in the Delphi process conducted for this work. 


\section*{Discussion}
\addcontentsline{toc}{chapter}{\protect\numberline{}Discussion}
\label{sec:discussion}

Flaws in the validation of biomedical image analysis algorithms significantly impede the translation of methods into (clinical) practice and undermine the assessment of scientific progress in the field \cite{lennerz2022unifying}. They are frequently caused by poor choices due to disregarding the specific properties and limitations of individual validation metrics. The present work represents the first comprehensive collection of pitfalls and limitations to be considered when using validation metrics in image-level classification, semantic segmentation, instance segmentation, and object detection tasks. Our work enables researchers to gain a deep understanding of and familiarity with both the overall topic and individual metrics by providing a common access point to previously largely scattered and inaccessible information — key knowledge they can resort to when conducting validation of image analysis algorithms. This way, our work aims to disrupt the current common practice of choosing metrics based on their popularity rather than their suitability to the underlying research problem. This practice, which, for instance, often manifests itself in the unreflected and inadequate use of the \ac{DSC}, is concerningly prevalent even among prestigious, high-quality biomedical image analysis competitions (challenges) \cite{honauer2015hci, correia2006video, kofler2021DICE, gooding2018comparative, vaassen2020evaluation, konukoglu2012discriminative, margolin2014evaluate, maier2018rankings}. The educational aspect of our work is complemented by dedicated 'metric profiles' which detail the definitions and properties of all metrics discussed. Notably, our work pioneers the examination of \ac{AI} validation pitfalls in the biomedical domain, a domain in which they are arguably more critical than in many others as flaws in biomedical algorithm validation can directly affect patient wellbeing and safety.

We posited that shortcomings in current common practice are marked by the low accessibility of information on the pitfalls and limitations of commonly used validation metrics. A literature search conducted from the point of view of a researcher seeking information on individual metrics confirmed that the number of search results far exceeds any amount that could be overseen within reasonable time and effort, as well as the lack of a common point of entry to reliable metric information. Even when knowing the specific pitfalls and related keywords uncovered by our consortium, only a fraction of those pitfalls could be found in existing literature, indicating the novelty and added value of our work. 

For transparency, several constraints regarding our literature search must be noted. First, it must be acknowledged that the remarkably high search result numbers inevitably include duplicates of papers (e.g., the same work in a conference paper and on arXiv) as well as results that are out of scope (e.g., \cite{carbonell2006effects}, \citep{di2012nonceliac}), in the cited examples for instance due to a metric acronym (AUC) simultaneously being an acronym for another entity (a trinucleotide) in a different domain, or the word "sensitivity" being used in its common, non-metric meaning. Moreover, common words used to describe pitfalls such as “problem” or “issue” are by nature present in many publications discussing any kind of research, rendering them unusable for a dedicated search, which could, in turn, account for missing publications that do discuss pitfalls in these terms. Similarly, when searching for specific pitfalls, many of the returned results containing the appropriate keywords did not actually refer to metrics or algorithm validation but to other parts of a model or biomedical problem (e.g., the need for stratification is commonly discussed with regard to the design of clinical studies but not with regard to their validation). Character limits in the Google Scholar search bar further complicate or prevent the use of comprehensive search strings. Finally, it is both possible and probable that our literature search did not retrieve all publications or non-peer-reviewed online resources that mention a particular pitfall, since even extensive search strings might not cover the particular words used for a pitfall description. 

None of these observations, however, detracts from our hypothesis. In fact, all of the above observations reinforce our finding that, for any individual researcher, retrieving information on metrics of interest is difficult to impossible. In many cases, finding information on pitfalls only appears feasible if the specific pitfall and its related keywords are exactly known, which, of course, is not the situation most researchers realistically find themselves in. Overall accessibility of such vital information, therefore, currently leaves much to be desired. 

Compiling this information through a multi-stage Delphi process allowed us to leverage distributed knowledge from experts across different biomedical imaging domains and thus ensure that the resulting illustrated collection of metric pitfalls and limitations is both comprehensive and of maximum practical relevance. Continued proximity of our work to issues occurring in practical application was achieved through sharing the first results of this process as a dynamic preprint \citep{reinke2021commonarxiv} with dedicated calls for feedback, as well as crowdsourcing further suggestions on social media.

Although their severity and practical consequences might differ between applications, we found that the pitfalls generalize across different imaging modalities and application domains. By categorizing them solely according to their underlying sources, we were able to create an overarching taxonomy that goes beyond domain-specific concerns and thus enjoys broad applicability. Given the large number of identified pitfalls, our taxonomy crucially establishes structure in the topic. Moreover, by relating types of pitfalls to the respective metrics they apply to and illustrating them, it enables researchers to gain a deeper, systemic understanding of the causes of metric failure.

Our complementary \textit{Metrics Reloaded} recommendation framework, which guides researchers towards the selection of appropriate validation metrics for their specific tasks and is introduced in a sister publication to this work \citep{maier2022metrics}, shares the same principle of domain independence. Its recommendations are based on the creation of a 'problem fingerprint' that abstracts from specific domain knowledge and, informed by the pitfalls discussed here, captures all properties relevant to metric selection for a specific biomedical problem. In this sister publication, we present recommendations to avoid the pitfalls presented in this work. Importantly, the finding that pitfalls generalize and can be categorized in a domain-independent manner opens up avenues for future expansion of our work to other fields of ML-based imaging, such as general computer vision (see below), thus freeing it from its major constraint of exclusively focusing on biomedical problems. 

It is worth mentioning that we only examined pitfalls related to the tasks of image-level classification, semantic segmentation, instance segmentation, and object detection, as these can all be considered classification tasks at different levels (image/object/pixel) and hence share similarities in their validation. While including a wider range of biomedical problems not considered classification tasks, such as regression or registration, would have gone beyond the scope of the present work, we envision this expansion in future work. Moreover, our work focused on pitfalls related to reference-based metrics. Including pitfalls pertaining to non-reference-based metrics, such as metrics that assess speed, memory consumption, or carbon footprint, could be a future direction to take. Finally, while we aspired to be as comprehensive as possible in our compilation, we cannot exclude that there are further pitfalls to be taken into account that the consortium and the participating community have so far failed to recognize. Should this be the case, our dynamic \textit{Metrics Reloaded} online platform, which is currently under development and will continuously be updated after release, will allow us to easily and transparently append missed pitfalls. This way, our work can remain a reliable point of access, reflecting the state of the art at any given moment in the future. In this context, we note that we explicitly welcome feedback and further suggestions from the readership of \textit{Nature Methods}. 

The expert consortium was primarily compiled in a way to cover the required expertise from various fields but also consisted of researchers of different countries, (academic) ages, roles, and backgrounds (details can be found in the Suppl. Methods).  It mainly focused on biomedical applications. The pitfalls presented here are therefore of the highest relevance for biological and clinical use cases. Their clear generalization across different biomedical imaging domains, however, indicates broader generalizability to fields such as general computer vision. Future work could thus see a major expansion of our scope to \ac{AI} validation well beyond biomedical research. Regardless of this possibility, we strongly believe that by raising awareness of metric-related pitfalls, our work will kick off a necessary scientific debate. Specifically, we see its potential in inducing the scientific communities in other areas of \ac{AI} research to follow suit and investigate pitfalls and common practices impairing progress in their specific domains. 

In conclusion, our work presents the first comprehensive and illustrated access point to information on validation metric properties and their pitfalls. We envision it to not only impact the quality of algorithm validation in biomedical imaging and ultimately catalyze faster translation into practice, but to raise awareness on common issues and call into question flawed \ac{AI} validation practice far beyond the boundaries of the field.

\newpage
\section*{Extended Data}
\addcontentsline{toc}{chapter}{\protect\numberline{}Extended Data}

\setcounter{figure}{0}
\setcounter{table}{0}

\makeatletter
\renewcommand{\fnum@figure}{Extended Data Fig. \thefigure}
\renewcommand{\fnum@table}{Extended Data Tab. \thetable}
\makeatother

\begin{figure}[H]
    \centering
    \includegraphics[width=1\linewidth]{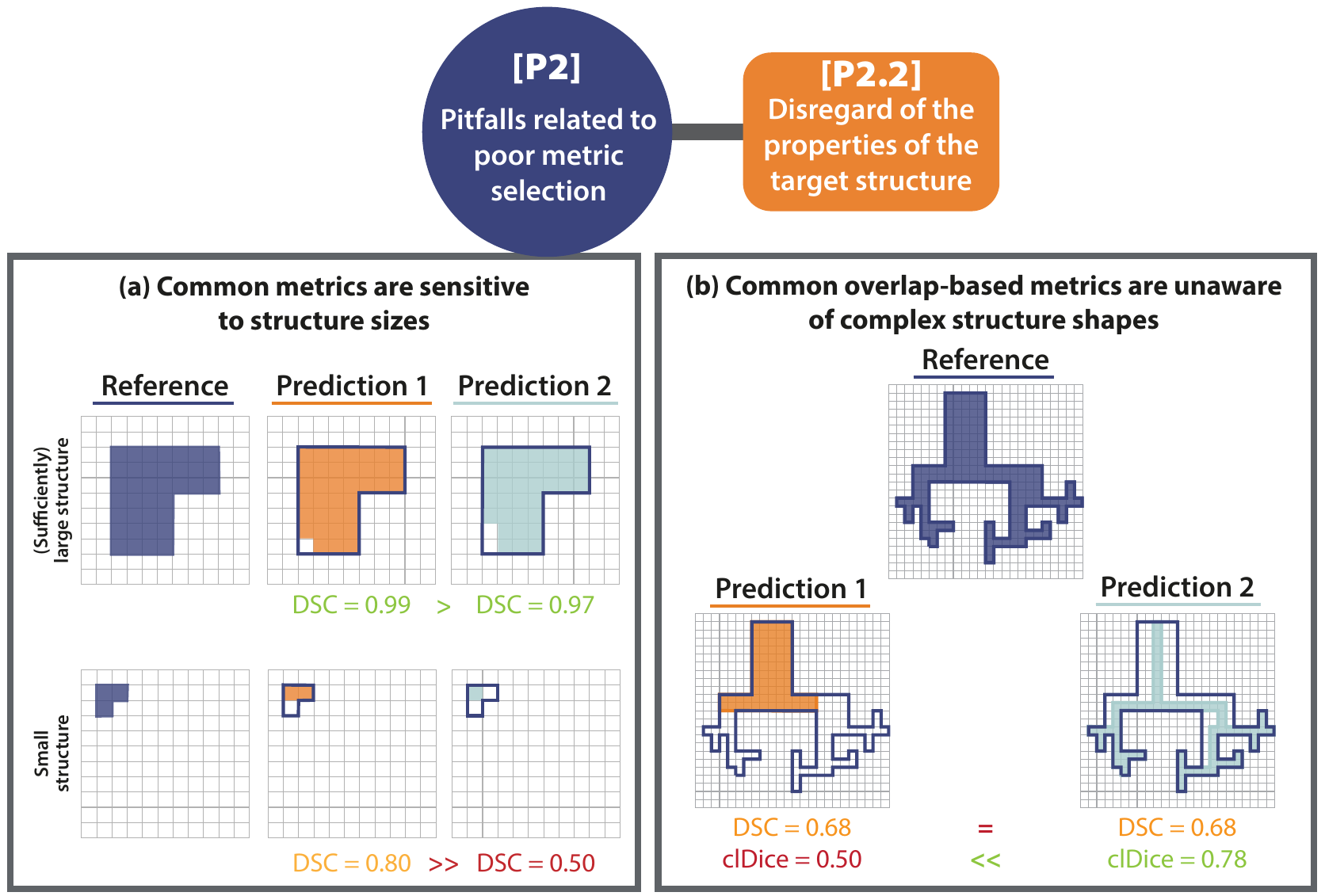}
    \caption{[P2.2] \textbf{Disregard of the properties of the target structures}. \textbf{(a) Small structure sizes.} The predictions of two algorithms (\textit{Prediction 1/2}) differ in only a single pixel. In the case of the small structure (bottom row), this has a substantial effect on the corresponding Dice Similarity Coefficient (DSC) metric value (similar for the Intersection over Union (IoU)). This pitfall is also relevant for other overlap-based metrics such as the centerline Dice Similarity Coefficient (clDice), and localization criteria such as Box/Approx/Mask IoU and Intersection over Reference (IoR). \textbf{(b) Complex structure shapes.} Common overlap-based metrics (here: DSC) are unaware of complex structure shapes and treat \textit{Predictions 1} and \textit{2} equally. The clDice uncovers the fact that \textit{Prediction 1} misses the fine-granular branches of the reference and favors \textit{Prediction 2}, which focuses on the center line of the object. This pitfall is also relevant for other overlap-based such as metrics IoU and pixel-level F$_\beta$ Score as well as localization criteria such as Box/Approx/Mask IoU, Center Distance, Mask IoU > 0, Point inside Mask/Box/Approx, and IoR.}
    \label{fig:pitfalls-p2-2}
\end{figure}

\newpage
\begin{figure}[H]
    \centering
    \includegraphics[width=0.9\linewidth]{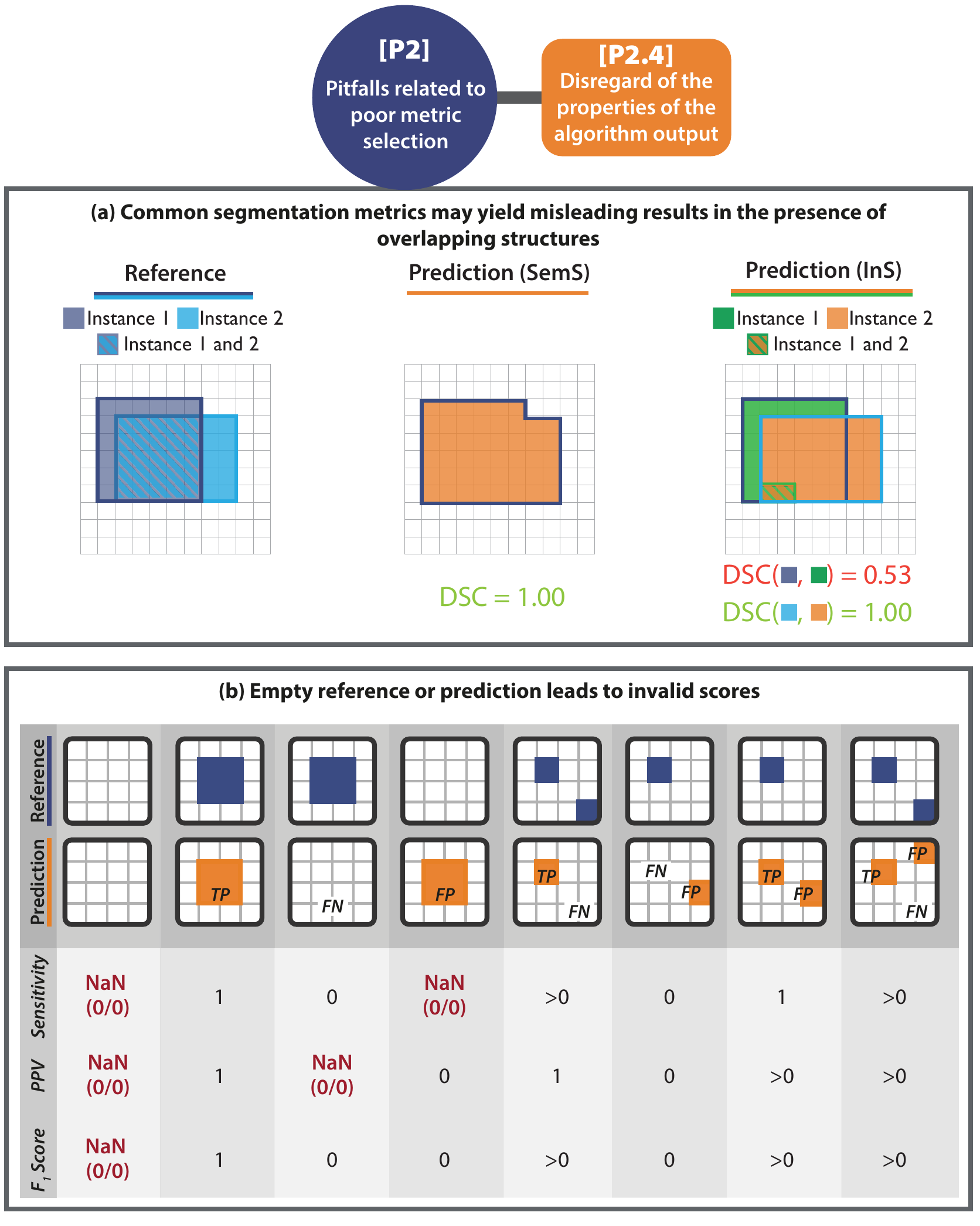}
    \caption{[P2.4] \textbf{Disregard of the properties of the algorithm output}. \textbf{(a) Possibility of overlapping predictions.} If multiple structures of the same type can be seen within the same image (here: reference objects \textit{R1} and \textit{R2}), it is generally advisable to phrase the problem as  instance segmentation (InS; right) rather than semantic segmentation (SemS; left). This way, issues with boundary-based metrics resulting from comparing a given structure boundary to the boundary of the wrong instance in the reference can be avoided. In the provided example, the distance of the red boundary pixel to the reference, as measured by a boundary-based metric in SemS problems, would be zero, because different instances of the same structure cannot be distinguished. This problem is overcome by phrasing the problem as InS. In this case, (only) the boundary of the matched instance (here: R2) is considered for distance computation. \textbf{(b) Possibility of empty prediction or reference.} Each column represents a potential scenario for per-image validation of objects, categorized by whether True Positives (TPs), False Negatives (FNs), and False Positives (FPs) are present (n > 0) or not (n = 0) after matching/assignment. The sketches on the top showcase each scenario when setting "n > 0" to "n = 1". For each scenario, Sensitivity, Positive Predictive Value (PPV), and the F$_\text{1}$ Score are calculated. Some scenarios yield undefined values (Not a Number (NaN)).}
    \label{fig:pitfalls-p2-4}
\end{figure}

\newpage
\begin{table}[H]
\caption{\textbf{Overview of pitfall sources for \textit{\textcolor{ILCblue}{image-level classification metrics}}} ((a): counting metrics, (b): multi-threshold metrics) related to poor metric selection [P2]. A warning sign indicates a potential pitfall for the metric in the corresponding column, in case the property represented by the respective row holds true. Comprehensive illustrations of pitfalls are available in \ref{app:pitfalls}. A comprehensive list of pitfalls is provided separately for each metrics in the metrics cheat sheets (\ref{app:steckbriefe}). Note that we only list sources of pitfalls relevant to the considered metrics. Other sources of pitfalls are neglected for this table.}
\label{tab:pitfalls-overview-ilc}
\hspace*{-1em}
\centering
\tiny

\begin{subtable}{1\textwidth}
    \subcaption{\textbf{Counting metrics.} Considered metrics: Accuracy (Fig.~\ref{fig:cheat-sheet-accuracy}), \acf{BA} (Fig.~\ref{fig:cheat-sheet-ba}), \acf{EC} (Fig.~\ref{fig:cheat-sheet-ec}), F$_\beta$ Score (Fig.\ref{fig:cheat-sheet-fbeta}), \acf{MCC} (Fig.~\ref{fig:cheat-sheet-mcc}), \acf{NB} (Fig.~\ref{fig:cheat-sheet-nb}), \acf{NPV} (Fig.~\ref{fig:cheat-sheet-npv}), \acf{LR+} (Fig.~\ref{fig:cheat-sheet-lr+}), \acf{PPV} (Fig.~\ref{fig:cheat-sheet-ppv}), Sensitivity (Sens) (Fig.~\ref{fig:cheat-sheet-sensitivity}), Specificity (Spec) (Fig.~\ref{fig:cheat-sheet-specificity}), \acf{WCK} (Fig.~\ref{fig:cheat-sheet-wck}).}

    \begin{tabular}{p{1.4cm}P{1cm}P{0.9cm}P{0.9cm}P{1cm}P{0.9cm}P{0.9cm}P{0.9cm}P{0.9cm}P{0.9cm}P{0.9cm}}
    \toprule
    \textbf{Source of potential pitfall} & \textbf{Accuracy} & \textbf{BA} & \textbf{EC} & \textbf{F$_\beta$ Score} & \textbf{LR+} & \textbf{MCC} & \textbf{NB} & \textbf{PPV/ NPV} & \textbf{Sens/ Spec} & \textbf{WCK} \\\midrule
    
    Importance of confidence awareness & \textcolor{ILCblue}{\faWarning}* & \textcolor{ILCblue}{\faWarning}* & \textcolor{ILCblue}{\faWarning}* & \textcolor{ILCblue}{\faWarning}* & \textcolor{ILCblue}{\faWarning}* & \textcolor{ILCblue}{\faWarning}* & \textcolor{ILCblue}{\faWarning}* & \textcolor{ILCblue}{\faWarning}* & \textcolor{ILCblue}{\faWarning}* & \textcolor{ILCblue}{\faWarning}* \\
    
    \rowcolor{lightgray}Importance of comparability across data sets & \textcolor{ILCblue}{\faWarning} (Fig.~\ref{fig:prevalence-dependency})&  & \textcolor{ILCblue}{\faWarning}** (Fig.~\ref{fig:prevalence-dependency})& \textcolor{ILCblue}{\faWarning} (Fig.~\ref{fig:prevalence-dependency})&  & \textcolor{ILCblue}{\faWarning} (Fig.~\ref{fig:prevalence-dependency})& \textcolor{ILCblue}{\faWarning} (Fig.~\ref{fig:prevalence-dependency})& \textcolor{ILCblue}{\faWarning} (Fig.~\ref{fig:prevalence-dependency})&  & \textcolor{ILCblue}{\faWarning} (Fig.~\ref{fig:prevalence-dependency})\\
    
    Unequal severity of class confusions  & \textcolor{ILCblue}{\faWarning} (Fig.~\ref{fig:pitfalls-p2-1}b)  & \textcolor{ILCblue}{\faWarning} (Fig.~\ref{fig:pitfalls-p2-1}b)  &  & \textcolor{ILCblue}{\faWarning}  *** (Fig.~\ref{fig:pitfalls-p2-1}b)& \textcolor{ILCblue}{\faWarning} (Fig.~\ref{fig:pitfalls-p2-1}b)  & \textcolor{ILCblue}{\faWarning} (Fig.~\ref{fig:pitfalls-p2-1}b)  &  & \textcolor{ILCblue}{\faWarning} (Fig.~\ref{fig:pitfalls-p2-1}b)  & \textcolor{ILCblue}{\faWarning} (Fig.~\ref{fig:pitfalls-p2-1}b)  &  \\
    
    \rowcolor{lightgray}Importance of cost-benefit analysis  & \textcolor{ILCblue}{\faWarning} (Fig.~\ref{fig:cost-benefit})& \textcolor{ILCblue}{\faWarning} (Fig.~\ref{fig:cost-benefit})&  & \textcolor{ILCblue}{\faWarning}*** (Fig.~\ref{fig:cost-benefit})& \textcolor{ILCblue}{\faWarning} (Fig.~\ref{fig:cost-benefit})& \textcolor{ILCblue}{\faWarning} (Fig.~\ref{fig:cost-benefit})&  & \textcolor{ILCblue}{\faWarning} (Fig.~\ref{fig:cost-benefit})& \textcolor{ILCblue}{\faWarning} (Fig.~\ref{fig:cost-benefit})&  \\ 
    
    High class imbalance & \textcolor{ILCblue}{\faWarning} (Figs.~\ref{fig:pitfalls-p2-3}a, \ref{fig:class-imbalance}) & \textcolor{ILCblue}{\faWarning} (Fig.~\ref{fig:pitfalls-p2-3}a) & \textcolor{ILCblue}{\faWarning}** (Fig.~\ref{fig:pitfalls-p2-3}a) &  & \textcolor{ILCblue}{\faWarning} (Fig.~\ref{fig:pitfalls-p2-3}a) &  & \textcolor{ILCblue}{\faWarning} (Figs.~\ref{fig:pitfalls-p2-3}a, \ref{fig:class-imbalance}) & \textbf{NPV:} \textcolor{ILCblue}{\faWarning} (Figs.~\ref{fig:pitfalls-p2-3}a, \ref{fig:class-imbalance}) & \textcolor{ILCblue}{\faWarning} (\textbf{Sens:} Fig.~\ref{fig:pitfalls-p2-3}a, \textbf{Spec:} Figs.~\ref{fig:pitfalls-p2-3}a, \ref{fig:class-imbalance}) & \textcolor{ILCblue}{\faWarning} (Figs.~\ref{fig:pitfalls-p2-3}a, \ref{fig:class-imbalance})\\
    
    \rowcolor{lightgray}Small test set size  & \textcolor{ILCblue}{\faWarning} (Fig.~\ref{fig:auroc-small-sample-sizes})  & \textcolor{ILCblue}{\faWarning} (Fig.~\ref{fig:auroc-small-sample-sizes})  & \textcolor{ILCblue}{\faWarning} (Fig.~\ref{fig:auroc-small-sample-sizes})  & \textcolor{ILCblue}{\faWarning} (Fig.~\ref{fig:auroc-small-sample-sizes})  & \textcolor{ILCblue}{\faWarning} (Fig.~\ref{fig:auroc-small-sample-sizes})  & \textcolor{ILCblue}{\faWarning} (Fig.~\ref{fig:auroc-small-sample-sizes})  & \textcolor{ILCblue}{\faWarning} (Fig.~\ref{fig:auroc-small-sample-sizes})  & \textcolor{ILCblue}{\faWarning} (Fig.~\ref{fig:auroc-small-sample-sizes})  & \textcolor{ILCblue}{\faWarning} (Fig.~\ref{fig:auroc-small-sample-sizes})  & \textcolor{ILCblue}{\faWarning} (Fig.~\ref{fig:auroc-small-sample-sizes})  \\
    
    \bottomrule
     \multicolumn{11}{l}{* \textit{Discrimination metrics do not assess whether the predicted class scores reflect the confidence of the classifier. This is typically achieved with}} \\
      \multicolumn{11}{l}{\textit{additional calibration metrics, which come with their own pitfalls (see Figs.~\ref{fig:calibration} and~\ref{fig:ece-mce-bins}, Extended Data Fig.~\ref{fig:pitfalls-p2-2}b and the metric profiles in Suppl. Note~\ref{app:steckbriefe:calibration}).}} \\
        \multicolumn{11}{l}{** \textit{The weights in EC can be adjusted to avoid this pitfall.}} \\
    \multicolumn{11}{l}{*** \textit{The hyperparameter $\beta$ can be used as a penalty for class confusions in the binary case. 
    This property is not applicable to multi-class problems.}}\\
    &  &  &  &   &  &  &  &   &   & \\ 
    \end{tabular}
\end{subtable}

\newpage

\begin{subtable}{1\textwidth}
    \subcaption{\textbf{Multi-threshold metrics.} Considered metrics: \acf{AUROC} (Fig.~\ref{fig:cheat-sheet-auroc}) and \acf{AP} (Fig.~\ref{fig:cheat-sheet-ap}).}

    \begin{tabular}{p{6cm}P{3.5cm}P{3.5cm}}
\toprule
\textbf{Source of potential pitfall} & \textbf{AP} & \textbf{AUROC} \\\midrule

    Importance of confidence awareness & \textcolor{ILCblue}{\faWarning}* & \textcolor{ILCblue}{\faWarning}*  \\
    \rowcolor{lightgray}Importance of comparability across data sets & \textcolor{ILCblue}{\faWarning} (Fig.~\ref{fig:prevalence-dependency})& \\
    
    High class imbalance & & \textcolor{ILCblue}{\faWarning}  (Fig.~\ref{fig:pitfalls-p2-3}a) \\
    \rowcolor{lightgray}Small test set size  & \textcolor{ILCblue}{\faWarning} (Fig.~\ref{fig:auroc-small-sample-sizes})  & \textcolor{ILCblue}{\faWarning} (Fig.~\ref{fig:auroc-small-sample-sizes})   \\
    Lack of predicted class scores  & \textcolor{ILCblue}{\faWarning} (Fig.~\ref{fig:lack-of-scores})  & \textcolor{ILCblue}{\faWarning} (Fig.~\ref{fig:lack-of-scores})\\ \bottomrule
    \multicolumn{3}{l}{* \textit{Discrimination metrics do not assess whether the predicted class scores reflect the confidence of the classifier. This is typically achieved with}} \\
    \multicolumn{3}{l}{\textit{additional calibration metrics, which come with their own pitfalls (see Figs.~\ref{fig:calibration} and~\ref{fig:ece-mce-bins}, Extended Data Fig.~\ref{fig:pitfalls-p2-2}b and the metric profiles in Suppl. Note~\ref{app:steckbriefe:calibration}).}} \\
    \end{tabular}
\end{subtable}
\end{table}
\newpage
\begin{table}[H]
\caption{\textbf{Overview of pitfall sources for \textit{\textcolor{SSyellow}{semantic segmentation metrics}}} ((a): overlap-based metrics, (b): boundary-based metrics) related to poor metric selection [P2]. A warning sign indicates a potential pitfall for the metric in the corresponding column, in case the property represented by the respective row holds true. Comprehensive illustrations of pitfalls are available in \ref{app:pitfalls}. A comprehensive list of pitfalls is provided separately for each metrics in the metrics cheat sheets (\ref{app:steckbriefe}). Note that we only list sources of pitfalls relevant to the considered metrics. Other sources of pitfalls are neglected for this table.}

\label{tab:pitfalls-overview-ss}
\hspace*{-1em}
\centering
\tiny
\begin{subtable}{1\textwidth}
    \subcaption{\textbf{Overlap-based metrics.} Considered metrics: Considered metrics: \acf{clDice} (Fig.~\ref{fig:cheat-sheet-cldice}), \acf{DSC} (Fig.~\ref{fig:cheat-sheet-dsc}), F$_\beta$ Score (Fig.~\ref{fig:cheat-sheet-fbeta}), \acf{IoU} (Fig.~\ref{fig:cheat-sheet-iou}).}
    \begin{tabular}{p{5.1cm}P{2cm}P{3.5cm}P{2cm}}
    \toprule
    \textbf{Source of potential pitfall} & \textbf{clDice} & \textbf{DSC/IoU} & \textbf{F$_\beta$ Score} \\ \midrule
    
    Importance of structure boundaries & \textcolor{SSyellow}{\faWarning} (Fig.~\ref{fig:pitfalls-p2-1}a) & \textcolor{SSyellow}{\faWarning} (Fig.~\ref{fig:pitfalls-p2-1}a) & \textcolor{SSyellow} {\faWarning} (Fig.~\ref{fig:pitfalls-p2-1}a) \\
    
    \rowcolor{lightgray}Importance of structure center(line)  &  & \textcolor{SSyellow}{\faWarning} (Fig.~\ref{fig:center}, Extended Data Fig.~\ref{fig:pitfalls-p2-2}b) & \textcolor{SSyellow}{\faWarning} (Fig.~\ref{fig:center}, Extended Data Fig.~\ref{fig:pitfalls-p2-2}b) \\
    
    Unequal severity of class confusions  & \textcolor{SSyellow}{\faWarning} (Fig.~\ref{fig:DSC-overunder}) &  \textcolor{SSyellow}{\faWarning} (Fig.~\ref{fig:DSC-overunder})&  \\%
    
    \rowcolor{lightgray}Small structure sizes  & \textcolor{SSyellow}{\faWarning} (Fig.~\ref{fig:boundary-mask-iou} , Extended Data Fig.~\ref{fig:pitfalls-p2-2}a)&  \textcolor{SSyellow}{\faWarning} (Fig.~\ref{fig:boundary-mask-iou} , Extended Data Fig.~\ref{fig:pitfalls-p2-2}a)& \textcolor{SSyellow}{\faWarning} (Fig.~\ref{fig:boundary-mask-iou} , Extended Data Fig.~\ref{fig:pitfalls-p2-2}a)\\
    
    High variability of structure sizes  & \textcolor{SSyellow}{\faWarning} (Fig.~\ref{fig:high-variability})& \textcolor{SSyellow}{\faWarning} (Fig.~\ref{fig:high-variability})& \textcolor{SSyellow}{\faWarning} (Fig.~\ref{fig:high-variability})\\
    \rowcolor{lightgray}Complex structure shapes  &  &  \textcolor{SSyellow}{\faWarning} (Fig.~\ref{fig:complex-shapes})&  \textcolor{SSyellow}{\faWarning} (Fig.~\ref{fig:complex-shapes})\\
    
    Occurrence of overlapping or touching structures  & \textcolor{SSyellow}{\faWarning} (Fig.~\ref{fig:multi-labels})& \textcolor{SSyellow}{\faWarning} (Fig.~\ref{fig:multi-labels})& \textcolor{SSyellow}{\faWarning} (Fig.~\ref{fig:multi-labels})\\
    
    \rowcolor{lightgray}Imperfect reference standard  & & \textcolor{SSyellow}{\faWarning} (Fig.~\ref{fig:low-quality}) & \textcolor{SSyellow}{\faWarning} (Fig.~\ref{fig:low-quality})\\
    Occurrence of cases with an empty reference  & \textcolor{SSyellow}{\faWarning} (Fig.~\ref{fig:empty}))& \textcolor{SSyellow}{\faWarning} (Fig.~\ref{fig:empty}))& \textcolor{SSyellow}{\faWarning} (Fig.~\ref{fig:empty}))\\
    
    \rowcolor{lightgray}Possibility of empty prediction & \textcolor{SSyellow}{\faWarning} (Fig.~\ref{fig:empty}))& \textcolor{SSyellow}{\faWarning} (Fig.~\ref{fig:empty}))& \textcolor{SSyellow}{\faWarning} (Fig.~\ref{fig:empty}))\\
    
    Possibility of overlapping predictions  & \textcolor{SSyellow}{\faWarning} (Fig.~\ref{fig:overlapping-pred}, Extended Data Fig.~\ref{fig:pitfalls-p2-4}a)& \textcolor{SSyellow}{\faWarning} (Fig.~\ref{fig:overlapping-pred}, Extended Data Fig.~\ref{fig:pitfalls-p2-4}a)& \textcolor{SSyellow}{\faWarning} (Fig.~\ref{fig:overlapping-pred}, Extended Data Fig.~\ref{fig:pitfalls-p2-4}a)\\ \bottomrule
    & &&\\
    \end{tabular}
\end{subtable}
\begin{subtable}{1\textwidth}
    \subcaption{\textbf{Boundary-based metrics.} Considered metrics: \acf{ASSD} (Fig.~\ref{fig:cheat-sheet-assd}), \acf{Boundary IoU} (Fig.~\ref{fig:cheat-sheet-boundary-iou}), \acf{HD} (Fig.~\ref{fig:cheat-sheet-hd}), \acf{HD95} (Fig.~\ref{fig:cheat-sheet-hd95}), \acf{MASD} (Fig.~\ref{fig:cheat-sheet-masd}), \acf{NSD} (Fig.~\ref{fig:cheat-sheet-nsd}).}
    \begin{tabular}{p{3.5cm}P{1.2cm}P{1.8cm}P{1.2cm}P{1.5cm}P{1.2cm}P{1.2cm}P{1.2cm}}
    \toprule
    \textbf{Source of potential pitfall} & \textbf{ASSD} & \textbf{Boundary IoU} & \textbf{HD} & \textbf{HD95} & \textbf{MASD} & \textbf{NSD} \\ \midrule
    
    Importance of structure volume & \textcolor{SSyellow}{\faWarning} (Fig.~\ref{fig:outline}) & \textcolor{SSyellow}{\faWarning} (Fig.~\ref{fig:outline}) & \textcolor{SSyellow}{\faWarning} (Fig.~\ref{fig:outline})  & \textcolor{SSyellow}{\faWarning} (Fig.~\ref{fig:outline}) & \textcolor{SSyellow}{\faWarning} (Fig.~\ref{fig:outline}) & \textcolor{SSyellow}{\faWarning} (Fig.~\ref{fig:outline}) \\
    \rowcolor{lightgray}Importance of structure center(line)  & \textcolor{SSyellow}{\faWarning} (Fig.~\ref{fig:center}, Extended Data Fig.~\ref{fig:pitfalls-p2-2}b)& \textcolor{SSyellow}{\faWarning} (Fig.~\ref{fig:center}, Extended Data Fig.~\ref{fig:pitfalls-p2-2}b)& \textcolor{SSyellow}{\faWarning} (Fig.~\ref{fig:center}, Extended Data Fig.~\ref{fig:pitfalls-p2-2}b)& \textcolor{SSyellow}{\faWarning} (Fig.~\ref{fig:center}, Extended Data Fig.~\ref{fig:pitfalls-p2-2}b)& \textcolor{SSyellow}{\faWarning} (Fig.~\ref{fig:center}, Extended Data Fig.~\ref{fig:pitfalls-p2-2}b)& \textcolor{SSyellow}{\faWarning} (Fig.~\ref{fig:center}, Extended Data Fig.~\ref{fig:pitfalls-p2-2}b)\\

    Occurrence of overlapping or touching structures  & \textcolor{SSyellow}{\faWarning} (Fig.~\ref{fig:multi-labels})& \textcolor{SSyellow}{\faWarning} (Fig.~\ref{fig:multi-labels})& \textcolor{SSyellow}{\faWarning} (Fig.~\ref{fig:multi-labels})& \textcolor{SSyellow}{\faWarning} (Fig.~\ref{fig:multi-labels})& \textcolor{SSyellow}{\faWarning} (Fig.~\ref{fig:multi-labels})& \textcolor{SSyellow}{\faWarning} (Fig.~\ref{fig:multi-labels})\\

    \rowcolor{lightgray}Imperfect reference standard  & \textcolor{SSyellow}{\faWarning} (Figs. 5c, SN 2.17)& \textcolor{SSyellow}{\faWarning} (Figs. 5c, SN 2.17) & \textcolor{SSyellow}{\faWarning} (Figs. 5c, SN 2.17)& \textcolor{SSyellow}{\faWarning} (Figs.~5c*, SN 2.17)& \textcolor{SSyellow}{\faWarning} (Figs. 5c, SN 2.17)&  \\
    Occurrence of cases with an empty reference  & \textcolor{SSyellow}{\faWarning} (Fig.~\ref{fig:empty}))& \textcolor{SSyellow}{\faWarning} (Fig.~\ref{fig:empty}))& \textcolor{SSyellow}{\faWarning} (Fig.~\ref{fig:empty}))& \textcolor{SSyellow}{\faWarning} (Fig.~\ref{fig:empty}))& \textcolor{SSyellow}{\faWarning} (Fig.~\ref{fig:empty}))& \textcolor{SSyellow}{\faWarning} (Fig.~\ref{fig:empty}))\\

   \rowcolor{lightgray} Possibility of empty prediction& \textcolor{SSyellow}{\faWarning} (Fig.~\ref{fig:empty}))& \textcolor{SSyellow}{\faWarning} (Fig.~\ref{fig:empty}))& \textcolor{SSyellow}{\faWarning} (Fig.~\ref{fig:empty}))& \textcolor{SSyellow}{\faWarning} (Fig.~\ref{fig:empty}))& \textcolor{SSyellow}{\faWarning} (Fig.~\ref{fig:empty}))& \textcolor{SSyellow}{\faWarning} (Fig.~\ref{fig:empty}))\\
    Possibility of overlapping predictions  & \textcolor{SSyellow}{\faWarning} (Fig.~\ref{fig:overlapping-pred}, Extended Data Fig.~\ref{fig:pitfalls-p2-4}a)& \textcolor{SSyellow}{\faWarning} (Fig.~\ref{fig:overlapping-pred}, Extended Data Fig.~\ref{fig:pitfalls-p2-4}a)& \textcolor{SSyellow}{\faWarning} (Fig.~\ref{fig:overlapping-pred}, Extended Data Fig.~\ref{fig:pitfalls-p2-4}a)& \textcolor{SSyellow}{\faWarning} (Fig.~\ref{fig:overlapping-pred}, Extended Data Fig.~\ref{fig:pitfalls-p2-4}a)& \textcolor{SSyellow}{\faWarning} (Fig.~\ref{fig:overlapping-pred}, Extended Data Fig.~\ref{fig:pitfalls-p2-4}a)& \textcolor{SSyellow}{\faWarning} (Fig.~\ref{fig:overlapping-pred}, Extended Data Fig.~\ref{fig:pitfalls-p2-4}a)\\\bottomrule
    \multicolumn{7}{l}{* \textit{Can be mitigated by the choice of the percentile.}} 
    \end{tabular}
\end{subtable}
\end{table}

\newpage
\begin{table}[H]
\caption{\textbf{Overview of sources of pitfalls for \textit{\textcolor{ODgreen}{object detection metrics}}} ((a): detection metrics, (b): localization criteria) related to poor metric selection [P2]. A warning sign indicates a potential pitfall for the metric in the corresponding column, in case the property represented by the respective row holds true. Comprehensive illustrations of pitfalls are available in \ref{app:pitfalls}. A comprehensive list of pitfalls is provided separately for each metrics in the metrics cheat sheets (\ref{app:steckbriefe}). Note that we only list sources of pitfalls relevant to the considered metrics. Other sources of pitfalls are neglected for this table.}

\label{tab:pitfalls-overview-od}
\hspace*{-1em}
\centering
\tiny

\begin{subtable}{1\textwidth}
\subcaption{\textbf{Detection metrics.} Considered counting metrics: F$_\beta$ Score (Fig.~\ref{fig:cheat-sheet-fbeta}), \acf{PPV} (Fig.~\ref{fig:cheat-sheet-ppv}), Sensitivity (Sens) (Fig.~\ref{fig:cheat-sheet-sensitivity}). Considered multi-threshold metrics: \acf{AP} (Fig.~\ref{fig:cheat-sheet-ap}) and \acf{FROC} (Fig.~\ref{fig:cheat-sheet-froc}).}
    \begin{tabular}{p{5.5cm}P{1.5cm}P{1cm}P{1cm}P{1cm}P{2cm}}
    \toprule
    \textbf{Source of potential pitfall} & \textbf{F$_\beta$ Score} & \textbf{PPV} & \textbf{Sens} & \textbf{AP} & \textbf{FROC Score} \\\midrule
    Unequal severity of class confusions & \textcolor{ODgreen}{\faWarning}* (Fig.~\ref{fig:pitfalls-p2-1}b)& \textcolor{ODgreen}{\faWarning} (Fig.~\ref{fig:pitfalls-p2-1}b)& \textcolor{ODgreen}{\faWarning} (Fig.~\ref{fig:pitfalls-p2-1}b)&\textcolor{ODgreen}{\faWarning} (Fig.~\ref{fig:pitfalls-p2-1}b) & \textcolor{ODgreen}{\faWarning} (Fig.~\ref{fig:pitfalls-p2-1}b)\\

    
    
    \rowcolor{lightgray}High class imbalance  & & & \textcolor{ODgreen}{\faWarning} (Fig.~\ref{fig:pitfalls-p2-3}a)& & \\
    Small test set size  & \textcolor{ODgreen}{\faWarning} (Fig.~\ref{fig:auroc-small-sample-sizes})& \textcolor{ODgreen}{\faWarning} (Fig.~\ref{fig:auroc-small-sample-sizes})& \textcolor{ODgreen}{\faWarning} (Fig.~\ref{fig:auroc-small-sample-sizes})& \textcolor{ODgreen}{\faWarning} (Fig.~\ref{fig:auroc-small-sample-sizes})& \textcolor{ODgreen}{\faWarning} (Fig.~\ref{fig:auroc-small-sample-sizes})\\
    \rowcolor{lightgray}Occurrence of cases with an empty reference & \textcolor{ODgreen}{\faWarning} (Fig.~\ref{fig:empty}, Extended Data Fig.~\ref{fig:pitfalls-p2-4}b)& \textcolor{ODgreen}{\faWarning} (Fig.~\ref{fig:empty}, Extended Data Fig.~\ref{fig:pitfalls-p2-4}b)& \textcolor{ODgreen}{\faWarning} (Fig.~\ref{fig:empty}, Extended Data Fig.~\ref{fig:pitfalls-p2-4}b)& \textcolor{ODgreen}{\faWarning} (Fig.~\ref{fig:empty}, Extended Data Fig.~\ref{fig:pitfalls-p2-4}b)& \textcolor{ODgreen}{\faWarning} (Fig.~\ref{fig:empty}, Extended Data Fig.~\ref{fig:pitfalls-p2-4}b)\\
    Possibility of empty prediction & \textcolor{ODgreen}{\faWarning} (Fig.~\ref{fig:empty}, Extended Data Fig.~\ref{fig:pitfalls-p2-4}b)& \textcolor{ODgreen}{\faWarning} (Fig.~\ref{fig:empty}, Extended Data Fig.~\ref{fig:pitfalls-p2-4}b)& \textcolor{ODgreen}{\faWarning} (Fig.~\ref{fig:empty}, Extended Data Fig.~\ref{fig:pitfalls-p2-4}b)& \textcolor{ODgreen}{\faWarning} (Fig.~\ref{fig:empty}, Extended Data Fig.~\ref{fig:pitfalls-p2-4}b)& \textcolor{ODgreen}{\faWarning} (Fig.~\ref{fig:empty}, Extended Data Fig.~\ref{fig:pitfalls-p2-4}b)\\
    \rowcolor{lightgray}Lack of predicted class scores  & & & & \textcolor{ODgreen}{\faWarning} (Fig.~\ref{fig:lack-of-scores})& \textcolor{ODgreen}{\faWarning} (Fig.~\ref{fig:lack-of-scores})\\ \bottomrule
    \multicolumn{6}{l}{* \textit{The hyperparameter $\beta$ can be used as a penalty for class confusions in the binary case.}}\\
    \multicolumn{6}{l}{  \textit{This property is not applicable to multi-class problems.}} \\
    & &&&&\\
    \end{tabular}
\end{subtable}

\begin{subtable}{1\textwidth}
    \subcaption{\textbf{Localization criteria.} Considered localization criteria: Box/Approx \acf{IoU} (Fig.~\ref{fig:cheat-sheet-iou-localization-crit}), Center Distance (Fig.~\ref{fig:cheat-sheet-center-distance}), Mask IoU > 0 (Fig.~\ref{fig:cheat-sheet-iou-localization-crit-0}), and Point inside Mask/ Box/ Approx (Fig.~\ref{fig:cheat-sheet-point-inside}).}

    \begin{tabular}{p{4.5cm}P{1.7cm}P{1.7cm}P{1.5cm}P{3cm}}
    \toprule
    \textbf{Source of potential pitfall} & \textbf{Box/ Approx IoU} & \textbf{Center Distance}  & \textbf{Mask IoU > 0} & \textbf{Point inside Mask/ Box/ Approx} \\\midrule
    
    Importance of structure boundaries  & \textcolor{ODgreen}{\faWarning} (Fig.~\ref{fig:pitfalls-p2-3}a)& \textcolor{ODgreen}{\faWarning}  (Fig.~\ref{fig:pitfalls-p2-3}a)&  \textcolor{ODgreen}{\faWarning} (Fig.~\ref{fig:pitfalls-p2-3}a)& \textcolor{ODgreen}{\faWarning} (Fig.~\ref{fig:pitfalls-p2-3}a)\\
    
    \rowcolor{lightgray}Importance of structure volume  &  & \textcolor{ODgreen}{\faWarning} (Fig.~\ref{fig:outline})& \textcolor{ODgreen}{\faWarning} (Fig.~\ref{fig:outline})& \textcolor{ODgreen}{\faWarning} (Fig.~\ref{fig:outline})\\
    
    Importance of structure center(line)  & \textcolor{ODgreen}{\faWarning} (Fig.\ref{fig:center}, Extended Data Fig.~\ref{fig:pitfalls-p2-2}b)&  & \textcolor{ODgreen}{\faWarning} (Fig.\ref{fig:center}, Extended Data Fig.~\ref{fig:pitfalls-p2-2}b)& \textcolor{ODgreen}{\faWarning} (Fig.\ref{fig:center}, Extended Data Fig.~\ref{fig:pitfalls-p2-2}b)\\
    
    \rowcolor{lightgray}Unequal severity of class confusions  & \textcolor{ODgreen}{\faWarning} (Fig.~\ref{fig:DSC-overunder})& \textcolor{ODgreen}{\faWarning} (Fig.~\ref{fig:DSC-overunder})* & \textcolor{ODgreen}{\faWarning} (Fig.~\ref{fig:DSC-overunder}) &\textcolor{ODgreen}{\faWarning} (Fig.~\ref{fig:DSC-overunder})* \\ 
    
    
    Small structure sizes  & \textcolor{ODgreen}{\faWarning} (Fig.~\ref{fig:boundary-mask-iou}, Extended Data Fig.~\ref{fig:pitfalls-p2-2}a)&  &  &  \\
    
    
    \rowcolor{lightgray}Complex structure shapes  & \textcolor{ODgreen}{\faWarning} (Figs.~\ref{fig:high-variability},~\ref{fig:disconnected})& \textcolor{ODgreen}{\faWarning} (Fig.~\ref{fig:high-variability})& \textcolor{ODgreen}{\faWarning} (Fig.~\ref{fig:high-variability})& \textcolor{ODgreen}{\faWarning} (Fig.~\ref{fig:high-variability})\\
    
    Occurrence of disconnected structures  & \textcolor{ODgreen}{\faWarning} (Fig.~\ref{fig:disconnected})&  &  & Point inside Box: \textcolor{ODgreen}{\faWarning} (Fig.~\ref{fig:disconnected})\\ 
    
    
    \rowcolor{lightgray}Imperfect reference standard& \textcolor{ODgreen}{\faWarning} (Fig.~\ref{fig:pitfalls-p2-3}c)&  & & \\
    
    
    \bottomrule
    \multicolumn{5}{l}{* \textit{Criterion implies point prediction, thus overlap assessment is not applicable.}}\\
    \end{tabular}
\end{subtable}

\end{table}

\newpage
\begin{table}[H]
\caption{\textbf{Overview of sources of pitfalls for \textit{\textcolor{ISpink}{instance segmentation metrics} (Part 1)}} ((a): detection metrics, (b): localization criteria) related to poor metric selection [P2]. A warning sign indicates a potential pitfall for the metric in the corresponding column, in case the property represented by the respective row holds true. Comprehensive illustrations of pitfalls are available in \ref{app:pitfalls}. A comprehensive list of pitfalls is provided separately for each metrics in the metrics cheat sheets (\ref{app:steckbriefe}). Note that we only list sources of pitfalls relevant to the considered metrics. Other sources of pitfalls are neglected for this table.}

\label{tab:pitfalls-overview-is-1}
\hspace*{-1em}
\centering
\tiny

\begin{subtable}{1\textwidth}
\subcaption{\textbf{Detection metrics.} Considered counting metrics: F$_\beta$ Score (Fig.~\ref{fig:cheat-sheet-fbeta}), \acf{PPV} (Fig.~\ref{fig:cheat-sheet-ppv}), \acf{PQ} (Fig.~\ref{fig:cheat-sheet-pq}), Sensitivity (Sens) (Fig.~\ref{fig:cheat-sheet-sensitivity}). Considered multi-threshold metrics: \acf{AP} (Fig.~\ref{fig:cheat-sheet-ap}) and \acf{FROC} (Fig.~\ref{fig:cheat-sheet-froc}).}

    \begin{tabular}{p{4.5cm}P{1.5cm}P{1cm}P{1cm}P{1cm}P{1cm}P{1.5cm}}
    \toprule
    \textbf{Source of potential pitfall} & \textbf{F$_\beta$ Score} & \textbf{PPV} & \textbf{PQ} & \textbf{Sens} & \textbf{AP} & \textbf{FROC Score} \\\midrule
    Unequal severity of class confusions  & \textcolor{ISpink}{\faWarning}* (Fig.~\ref{fig:pitfalls-p2-1}b)& \textcolor{ISpink}{\faWarning} (Fig.~\ref{fig:pitfalls-p2-1}b)& \textcolor{ISpink}{\faWarning} (Fig.~\ref{fig:pitfalls-p2-1}b)& \textcolor{ISpink}{\faWarning} (Fig.~\ref{fig:pitfalls-p2-1}b)& & \\
    
    \rowcolor{lightgray}High class imbalance  & & & & \textcolor{ISpink}{\faWarning} (Fig.~\ref{fig:pitfalls-p2-3}a)&&  \\
    Small test set size  & \textcolor{ISpink}{\faWarning} (Fig.~\ref{fig:auroc-small-sample-sizes})& \textcolor{ISpink}{\faWarning} (Fig.~\ref{fig:auroc-small-sample-sizes})& \textcolor{ISpink}{\faWarning} (Fig.~\ref{fig:auroc-small-sample-sizes})& \textcolor{ISpink}{\faWarning} (Fig.~\ref{fig:auroc-small-sample-sizes})& \textcolor{ISpink}{\faWarning} (Fig.~\ref{fig:auroc-small-sample-sizes})& \textcolor{ISpink}{\faWarning} (Fig.~\ref{fig:auroc-small-sample-sizes})\\
    \rowcolor{lightgray}Lack of predicted class scores  & & & & & \textcolor{ISpink}{\faWarning} (Fig.~\ref{fig:lack-of-scores})& \textcolor{ISpink}{\faWarning} (Fig.~\ref{fig:lack-of-scores})\\ \bottomrule
        \multicolumn{7}{l}{* \textit{The hyperparameter $\beta$ can be used as a penalty for class confusions in the binary case.}}\\
    \multicolumn{7}{l}{  \textit{This property is not applicable to multi-class problems.}} \\
    & &&&&\\
    \end{tabular}
\end{subtable}

\begin{subtable}{1\textwidth}
    \subcaption{\textbf{Localization criteria.} Considered localization criteria: \acf{Boundary IoU} (Fig.~\ref{fig:cheat-sheet-boundary-iou-localization-crit}), \acf{IoR} (Fig.~\ref{fig:cheat-sheet-ior}), Mask IoU (Fig.~\ref{fig:cheat-sheet-iou}).}
    \begin{tabular}{p{6cm}P{2.5cm}P{2cm}P{2cm}}
    \toprule
    \textbf{Source of potential pitfall} & \textbf{Boundary IoU} & \textbf{IoR} & \textbf{Mask IoU}\\\midrule
    
    Importance of structure boundaries  &  & \textcolor{ISpink}{\faWarning} (Fig.~\ref{fig:pitfalls-p2-1}a)& \textcolor{ISpink}{\faWarning} (Fig.~\ref{fig:pitfalls-p2-1}a)\\
    
    \rowcolor{lightgray}Importance of structure volume  & \textcolor{ISpink}{\faWarning} (Fig.~SN 2.4)&& \\
    Importance of structure center(line)  & \textcolor{ISpink}{\faWarning} (Fig.~\ref{fig:center}, Extended Data Fig.~\ref{fig:pitfalls-p2-2}b)& \textcolor{ISpink}{\faWarning} (Fig.~\ref{fig:center}, Extended Data Fig.~\ref{fig:pitfalls-p2-2}b)& \textcolor{ISpink}{\faWarning} (Fig.~\ref{fig:center}, Extended Data Fig.~\ref{fig:pitfalls-p2-2}b) \\
    
    \rowcolor{lightgray}Unequal severity of class confusions  & \textcolor{ISpink}{\faWarning} (Fig.~\ref{fig:DSC-overunder})& \textcolor{ISpink}{\faWarning} (Fig.~\ref{fig:DSC-overunder})& \textcolor{ISpink}{\faWarning} (Fig.~\ref{fig:DSC-overunder})\\ 
    
    Small structure sizes  & & \textcolor{ISpink}{\faWarning} (Fig.~\ref{fig:boundary-mask-iou}, Extended Data Fig.~\ref{fig:pitfalls-p2-2}a)& \textcolor{ISpink}{\faWarning} (Fig.~\ref{fig:boundary-mask-iou} , Extended Data Fig.~\ref{fig:pitfalls-p2-2}a) \\
    
    \rowcolor{lightgray}Complex structure shapes  &  & \textcolor{ISpink}{\faWarning} (Fig.~\ref{fig:complex-shapes})& \textcolor{ISpink}{\faWarning} (Fig.~\ref{fig:complex-shapes})\\
    
    Imperfect reference standard & \textcolor{ISpink}{\faWarning} (Fig.~\ref{fig:low-quality}) & \textcolor{ISpink}{\faWarning} (Fig.~\ref{fig:low-quality}) & \textcolor{ISpink}{\faWarning} (Fig.~\ref{fig:low-quality})\\ \bottomrule
    
    & & &\\
    \end{tabular}
\end{subtable}

\end{table}

\newpage
\begin{table}[H]
\caption{\textbf{Overview of sources of pitfalls for \textit{\textcolor{ISpink}{instance segmentation metrics} (Part 2)}} ((a) per instance segmentation overlap-based metrics, (b) per instance segmentation boundary-based metrics) related to poor metric selection [P2]. A warning sign indicates a potential pitfall for the metric in the corresponding column, in case the property represented by the respective row holds true. Comprehensive illustrations of pitfalls are available in \ref{app:pitfalls}. Note that we only list sources of pitfalls relevant to the considered metrics. Other sources of pitfalls are neglected for this table.}
\label{tab:pitfalls-overview-is-2}
\hspace*{-1em}
\centering
\tiny

\begin{subtable}{1\textwidth}
    \subcaption{\textbf{Per instance segmentation overlap-based metrics.} Considered metrics:\acf{clDice} (Fig.~\ref{fig:cheat-sheet-cldice}), \acf{DSC} (Fig.~\ref{fig:cheat-sheet-dsc}), F$_\beta$ Score (Fig.~\ref{fig:cheat-sheet-fbeta}), \acf{IoU} (Fig.~\ref{fig:cheat-sheet-iou}).}

    \begin{tabular}{p{4.6cm}P{2.2cm}P{3.4cm}P{2.2cm}}
    \toprule
    \textbf{Source of potential pitfall} & \textbf{clDice} & \textbf{DSC/IoU} & \textbf{F$_\beta$ Score} \\ \midrule
    
    Importance of structure boundaries  &  \textcolor{ISpink}{\faWarning} (Fig.~\ref{fig:pitfalls-p2-1}a)&  \textcolor{ISpink}{\faWarning} (Fig.~\ref{fig:pitfalls-p2-1}a)&  \textcolor{ISpink}{\faWarning} (Fig.~\ref{fig:pitfalls-p2-1}a)\\
    
    \rowcolor{lightgray}Importance of structure center(line)  &  &   \textcolor{ISpink}{\faWarning} (Fig.~\ref{fig:center}, Extended Data Fig.~\ref{fig:pitfalls-p2-2}b)&  \textcolor{ISpink}{\faWarning} (Fig.~\ref{fig:center}, Extended Data Fig.~\ref{fig:pitfalls-p2-2}b)\\
    Unequal severity of class confusions  &  \textcolor{ISpink}{\faWarning} (Fig.~\ref{fig:DSC-overunder})&   \textcolor{ISpink}{\faWarning} (Fig.~\ref{fig:DSC-overunder})&  \\
    
   \rowcolor{lightgray} Small structure sizes  &  \textcolor{ISpink}{\faWarning} (Fig.~\ref{fig:boundary-mask-iou} , Extended Data Fig.~\ref{fig:pitfalls-p2-2}a)&   \textcolor{ISpink}{\faWarning} (Fig.~\ref{fig:boundary-mask-iou} , Extended Data Fig.~\ref{fig:pitfalls-p2-2}a)&  \textcolor{ISpink}{\faWarning} (Fig.~\ref{fig:boundary-mask-iou} , Extended Data Fig.~\ref{fig:pitfalls-p2-2}a)\\

    Complex structure shapes &  &   \textcolor{ISpink}{\faWarning} (Fig.~\ref{fig:complex-shapes})&   \textcolor{ISpink}{\faWarning} (Fig.~\ref{fig:complex-shapes})\\

    \rowcolor{lightgray}Imperfect reference standard  & &  \textcolor{ISpink}{\faWarning} (Fig.~\ref{fig:low-quality})&  \textcolor{ISpink}{\faWarning} (Fig.~\ref{fig:low-quality})\\

    \bottomrule
    && \\
    \end{tabular}

\end{subtable}

\begin{subtable}{1\textwidth}
    \subcaption{\textbf{Per instance segmentation boundary-based metrics.} Considered metrics: \acf{ASSD} (Fig.~\ref{fig:cheat-sheet-assd}), \acf{Boundary IoU} (Fig.~\ref{fig:cheat-sheet-boundary-iou}), \acf{HD} (Fig.~\ref{fig:cheat-sheet-hd}), \acf{HD95} (Fig.~\ref{fig:cheat-sheet-hd95}), \acf{MASD} (Fig.~\ref{fig:cheat-sheet-masd}), \acf{NSD} (Fig.~\ref{fig:cheat-sheet-nsd}).}

    \begin{tabular}{p{3.7cm}P{1.2cm}P{1.8cm}P{1.2cm}P{1.2cm}P{1.2cm}P{1.2cm}P{1.21cm}}
    \toprule
    \textbf{Source of potential pitfall} & \textbf{ASSD} & \textbf{Boundary IoU} & \textbf{HD} & \textbf{HD95} & \textbf{MASD} & \textbf{NSD} \\\midrule
    
    Importance of structure volume  & \textcolor{ISpink}{\faWarning} (Fig.~\ref{fig:outline})& \textcolor{ISpink}{\faWarning} (Fig.~\ref{fig:outline})& \textcolor{ISpink}{\faWarning} (Fig.~\ref{fig:outline})& \textcolor{ISpink}{\faWarning} (Fig.~\ref{fig:outline})& \textcolor{ISpink}{\faWarning} (Fig.~\ref{fig:outline})& \textcolor{ISpink}{\faWarning} (Fig.~\ref{fig:outline})\\
    \rowcolor{lightgray}Importance of structure center(line)  & \textcolor{ISpink}{\faWarning} (Fig.~\ref{fig:center}, Extended Data Fig.~\ref{fig:pitfalls-p2-2}b)& \textcolor{ISpink}{\faWarning} (Fig.~\ref{fig:center}, Extended Data Fig.~\ref{fig:pitfalls-p2-2}b)& \textcolor{ISpink}{\faWarning} (Fig.~\ref{fig:center}, Extended Data Fig.~\ref{fig:pitfalls-p2-2}b)& \textcolor{ISpink}{\faWarning} (Fig.~\ref{fig:center}, Extended Data Fig.~\ref{fig:pitfalls-p2-2}b)& \textcolor{ISpink}{\faWarning} (Fig.~\ref{fig:center}, Extended Data Fig.~\ref{fig:pitfalls-p2-2}b)& \textcolor{ISpink}{\faWarning} (Fig.~\ref{fig:center}, Extended Data Fig.~\ref{fig:pitfalls-p2-2}b)\\

    Imperfect reference standard  & \textcolor{ISpink}{\faWarning} (Figs.~\ref{fig:pitfalls-p2-3}c,~\ref{fig:low-quality})& \textcolor{ISpink}{\faWarning} (Figs.~\ref{fig:pitfalls-p2-3}c,~\ref{fig:low-quality})& \textcolor{ISpink}{\faWarning} (Figs.~\ref{fig:pitfalls-p2-3}c,~\ref{fig:low-quality})& \textcolor{ISpink}{\faWarning} (Figs.~5c*,~\ref{fig:low-quality})& \textcolor{ISpink}{\faWarning} (Figs.~\ref{fig:pitfalls-p2-3}c,~\ref{fig:low-quality})&  \\

    \bottomrule
    \multicolumn{7}{l}{* \textit{Can be mitigated by the choice of the percentile.}} 
    \end{tabular}

\end{subtable}

\end{table}

\section*{Code Availability Statement}
\addcontentsline{toc}{chapter}{\protect\numberline{}Code Availability Statement}
We provide reference implementations for all \textit{Metrics Reloaded} metrics within the \ac{MONAI} open-source framework. They are accessible at \url{https://github.com/Project-MONAI/MetricsReloaded}.

\section*{Acknowledgements}
\addcontentsline{toc}{chapter}{\protect\numberline{}Acknowledgements}
This work was initiated by the Helmholtz Association of German Research Centers in the scope of the Helmholtz Imaging Incubator (HI), the MICCAI Special Interest Group for biomedical image analysis challenges, and the benchmarking working group of the MONAI initiative. It has received funding from the European Research Council (ERC) under the European Union’s Horizon 2020 research and innovation programme (grant agreement No. [101002198], NEURAL SPICING) and the Surgical Oncology Program of the National Center for Tumor Diseases (NCT) Heidelberg. It was further supported in part by the Intramural Research Program of the National Institutes of Health Clinical Center as well as by the National Cancer Institute (NCI) and the National Institute of Neurological Disorders and Stroke (NINDS) of the National Institutes of Health (NIH), under award numbers NCI:U01CA242871,NCI:U24CA279629, and NINDS:R01NS042645. The content of this publication is solely the responsibility of the authors and does not represent the official views of the NIH. T.A. acknowledges the Canada Institute for Advanced Research (CIFAR) AI Chairs program, the Natural Sciences and Engineering Research Council of Canada. F.B. was co-funded by the European Union (ERC, TAIPO, 101088594). Views and opinions expressed are however those of the authors only and do not necessarily reflect those of the European Union or the European Research Council. Neither the European Union nor the granting authority can be held responsible for them. M.J.C. acknowledges funding from Wellcome/EPSRC Centre for Medical Engineering (WT203148/Z/16/Z), the Wellcome Trust (WT213038/Z/18/Z), and the InnovateUK funded London AI Centre for Value-Based Healthcare. J.C. is supported by the Federal Ministry of Education and Research (BMBF) under the funding reference 161L0272. V.C. acknowledges funding from NovoNordisk Foundation (NNF21OC0068816) and Independent Research Council Denmark (1134-00017B). B.A.C. was supported by NIH grant P41 GM135019 and grant 2020-225720 from the Chan Zuckerberg Initiative DAF, an advised fund of the Silicon Valley Community Foundation. G.S.C. was supported by Cancer Research UK (programme grant: C49297/A27294). M.M.H. is supported by the Natural Sciences and Engineering Research Council of Canada (RGPIN-2022-05134). A.Ka. is supported by French State Funds managed by the “Agence Nationale de la Recherche (ANR)” - “Investissements d’Avenir” (Investments for the Future), Grant ANR-10-IAHU-02 (IHU Strasbourg). M.K. was funded by the Ministry of Education, Youth and Sports of the Czech Republic (Project LM2018129). Ta.K. was supported in part by 4UH3-CA225021-03, 1U24CA180924-01A1, 3U24CA215109-02, and 1UG3-CA225-021-01 grants from the National Institutes of Health. G.L. receives research funding from the Dutch Research Council, the Dutch Cancer Association, HealthHolland, the European Research Council, the European Union, and the Innovative Medicine Initiative. S.M.R. wishes to acknowledge the Allen Institute for Cell Science founder Paul G. Allen for his vision, encouragement and support. M.R is supported by Innosuisse grant number 31274.1 and Swiss National Science Foundation Grant Number 205320\_212939. C.H.S. is supported by an Alzheimer's Society Junior Fellowship (AS-JF-17-011). R.M.S. is supported by the Intramural Research Program of the NIH Clinical Center. A.T. acknowledges support from Academy of Finland (Profi6 336449 funding program), University of Oulu strategic funding, Finnish Foundation for Cardiovascular Research, Wellbeing Services County of North Ostrobothnia (VTR project K62716), and Terttu foundation. S.A.T. acknowledges the support of Canon Medical and the Royal Academy of Engineering and the Research Chairs and Senior Research Fellowships scheme (grant RCSRF1819\textbackslash 8\textbackslash 25). B.V.C. was supported by Research Foundation Flanders (FWO grant G097322N) and Internal Funds KU Leuven (grant C24M/20/064).\

We would like to thank Peter Bankhead, Gary S. Collins, Robert Haase, Fred Hamprecht, Alan Karthikesalingam, Hannes Kenngott, Peter Mattson, David Moher, Bram Stieltjes, and Manuel Wiesenfarth for the fruitful discussions on this work. \

We would like to thank Sandy Engelhardt, Sven Koehler, M. Alican Noyan, Gorkem Polat, Hassan Rivaz, Julian Schroeter, Anindo Saha, Lalith Sharan, Peter Hirsch, and Matheus Viana for suggesting additional illustrations that can be found in \citep{reinke2021commonarxiv}.

\section*{Competing Interests}
\addcontentsline{toc}{chapter}{\protect\numberline{}Competing Interests}
The authors declare the following competing interests: F.B. is an employee of Siemens AG (Munich, Germany). B.v.G. is a shareholder of Thirona (Nijmegen, NL). B.G. is an employee of HeartFlow Inc (California, USA) and Kheiron Medical Technologies Ltd (London, UK). M.M.H. received an Nvidia GPU Grant. Th. K. is an employee of Lunit (Seoul, South Korea). G.L. is on the advisory board of Canon Healthcare IT (Minnetonka, USA) and is a shareholder of Aiosyn BV (Nijmegen, NL). Na.R. is the founder and CSO of Histofy (New York, USA). Ni.R. is an employee of Nvidia GmbH (Munich, Germany). J.S.-R. reports funding from GSK (Heidelberg, Germany), Pfizer (New York, USA) and Sanofi (Paris, France) and fees from Travere Therapeutics (California, USA), Stadapharm (Bad Vilbel, Germany), Astex Therapeutics (Cambridge, UK), Pfizer (New York, USA), and Grunenthal (Aachen, Germany). R.M.S. receives patent royalties from iCAD (New Hampshire, USA), ScanMed (Nebraska, USA), Philips (Amsterdam, NL), Translation Holdings (Alabama, USA) and PingAn (Shenzhen, China); his lab received research support from PingAn through a Cooperative Research and Development Agreement. S.A.T. receives financial support from Canon Medical Research Europe (Edinburgh, Scotland).

\bibliography{sample-base}
\addcontentsline{toc}{chapter}{\protect\numberline{}References}


\newpage
\section*{Supplementary Methods}
\addcontentsline{toc}{chapter}{\protect\numberline{}Supplementary Methods}
\label{sec:methods}
\subsection*{Literature search}
The literature search of metric pitfalls and limitations was conducted 
on the platform Google Scholar. The checkbox "include patents" was activated and the checkbox "include citations" was deactivated; other default settings were left unchanged. For each metric, a specific search string using the Boolean operators \texttt{OR} and \texttt{AND} was generated as follows: 
\begin{itemize}
    \item (Different notations of the metric name, including synonyms and acronyms, enclosed in quotation marks, respectively, and combined with \texttt{OR})
    \item \texttt{AND} "metric"
    \item \texttt{AND} (different expressions pertaining to the concept of pitfalls, limitations and flaws, enclosed in quotation marks, respectively, and combined with \texttt{OR})
\end{itemize}
For example, the following search string was used for the literature search of \ac{DSC} pitfalls: \texttt{("DSC" OR "Dice Similarity Coefficient" OR "Sørensen–Dice coefficient" OR "F1 score" OR "DCE") AND "metric" AND ("pitfall" OR "limitation" OR "caveat" OR "drawback" OR "shortcoming" OR "weakness" OR "flaw" OR "disadvantage” OR "suffer")}. 

A second literature search dedicated to the pitfalls collected during the Delphi process was conducted on the platforms Google Scholar and Google. This search served the purpose of determining how many of the proposed pitfalls could be found in either existing research literature or online resources such as blogs, assuming that the issue is already roughly known to the person conducting the search. We further determined whether or not a found pitfall was presented in a visual manner. We analyzed the first three results pages (corresponding to thirty results) from each search platform and excluded our own previous work on metric pitfalls from the analysis.

\subsection*{Delphi process}
The collection of pitfalls was achieved 
via a multi-stage Delphi process conducted among an international expert consortium comprised of more than 60 biomedical image analysis experts, as well as community feedback. A Delphi process is a structured group communication process that serves to pool opinions from an expert panel via a series of individual interrogations, usually in the form of questionnaires, interspersed with feedback from the respondents \cite{brown1968delphi}. The technique is widely used for building consensus among experts in medicine, particularly in the development of best practices in areas where evidence may be limited, conflicting, or absent \cite{nasa2021delphi}. Expert selection was initially based on membership in major relevant societies such as the \ac{BIAS} initiative, the \ac{MONAI} Working Group for Evaluation, Reproducibility and Benchmarks, and the \ac{MICCAI} Special Interest Group for Challenges (previously \ac{MICCAI} board working group), as well as a track record of expertise in the areas of metrics, challenges and/or best practices. To reflect as broad a range of application areas and metric pitfalls as possible, the number of consortium members was increased throughout the process to a final number of 62 members. The Delphi process comprised four surveys. Each survey was developed by the coordinating team of the process and sent out to the remaining members of the consortium. Upon completion, the coordinating team then analyzed the results and iteratively refined the list of pitfalls. The main stages of the compilation and consensus building process are detailed in the following:

\begin{enumerate}
    \item \textit{Compilation of pitfall sources:} The primary purpose of the first survey was obtaining agreement on sources of pitfalls. 
    \item \textit{Collection of pitfalls:} The following survey specifically asked for concrete pitfalls in the presence of those problem characteristics. 
    \item \textit{Community feedback:} The proposed list of pitfalls was further complemented by social media-based feedback from the general scientific community.
    \item \textit{Final agreement on pitfalls:} The subsequent survey served to obtain consensus agreement on which pitfalls to include. For each pitfall, it asked whether the pitfall should be included. In addition, the experts were given the opportunity to provide feedback on each pitfall and to suggest further pitfalls. The final collection of pitfalls was illustrated and all metric values were verified by two independent observers.
    \item \textit{Creation of taxonomy:} The collected pitfalls were analyzed and a taxonomy was created. In the final survey, approval of the consortium for the structure and phrasing of the taxonomy and the assignment of specific pitfalls to the taxonomy was obtained.
\end{enumerate}

\subsection*{Expert consortium}
\label{ssec:experts}

The expert consortium consisted of a total of 70 researchers (70\% male, 30\% female) from a total of 65 institutions. The majority of experts (50\%) were professors, followed by postdoctoral researchers (39\%). The median h-index of the consortium was 31.5 (mean: 36; minimum: 6; maximum: 113) and the median academic age was 18 years (mean: 19; minimum: 3; max: 42). Experts were from 19 countries and 5 continents. 60\% of experts had a technical, 6\% a clinical, 3\% a biological, and 23\% a mixed background. Of the 65 institutions, we could identify the number of employees for 89\%. Of those, the majority of institutions had a size between 1,000 and 10,000 employees (57\%), followed by even larger institutions between 10,000 and 100,000 employees (22\%), and smaller institutions below 1,000 employees (20\%). Only a small portion of institutions were above 100,000 employees (2\%).

\newpage
\section*{Supplementary Notes}
\addcontentsline{toc}{chapter}{\protect\numberline{}Supplementary Notes}
\setlength{\parskip}{0.5em}

\setcounter{section}{0}

\renewcommand{\thefigure}{SN~\arabic{section}.\arabic{figure}}
\renewcommand{\thetable}{SN~\arabic{section}.\arabic{table}}

\renewcommand*{\thesection}{SUPPL. NOTE \arabic{section}}
\renewcommand*{\thesubsection}{\arabic{section}.\arabic{subsection}}



\section{Metric Fundamentals}%
\label{app:fundamentals}

The present work focuses on biomedical image analysis problems that can be interpreted as classification tasks at the image, object, or pixel level. The vast majority of metrics for these problem categories are directly or indirectly based on epidemiological principles of \acf{TP}, \acf{FN}, \acf{FP}, \acf{TN}, i.e., the \textit{cardinalities} of the so-called confusion matrix. The \ac{TP}/\ac{FN}/\ac{FP}/\ac{TN} are henceforth referred to as cardinalities. In the case of more than two classes $C$, we also refer to the entries of the $C \times C$ confusion matrix as cardinalities. For simplicity and clarity in notation, we restrict ourselves to the binary case in most examples. Cardinalities can be computed at the image (segment), object, or pixel level. They are typically computed by comparing the prediction of the algorithm to a reference annotation. Modern neural network-based approaches commonly require a threshold to be set in order to convert the algorithm output comprising predicted class scores (also referred to as continuous class scores) to a confusion matrix. For the purpose of metric recommendation, the available metrics can be broadly classified as follows (see also \citep{cao2020mcc}): 

\begin{itemize}
    \item \textbf{Counting metrics} operate directly on the confusion matrix and express the metric value as a function of the cardinalities. In the context of segmentation, they are typically referred to as \textbf{overlap-based} metrics~\cite{taha2015metrics}. We distinguish \textbf{multi-class counting metrics}, which are defined for an arbitrary number of classes and invariant under class order, from \textbf{per-class counting metrics}, which are computed by treating one class as foreground/positive class and all other classes as background. Popular examples for the former include \ac{MCC} or Accuracy, while examples for the latter are Sensitivity, Specificity and \ac{DSC}.
    \item \textbf{Multi-threshold metrics} operate on a dynamic confusion matrix, reflecting the conflicting properties of interest, such as high Sensitivity and high Specificity. Popular examples include the \ac{AUROC} and \ac{AP}.
    \item \textbf{Distance-based metrics} have been designed for semantic and instance segmentation tasks. They operate exclusively on the \acp{TP} and rely on the explicit definition of object boundaries. Popular examples are the \ac{HD} and the \ac{NSD}.
\end{itemize}

Depending on the context (e.g., image-level classification \textit{vs.} semantic segmentation task) and the community (e.g., medical imaging community \textit{vs.} computer vision community), identical metrics are referred to with different terminology. For example, Sensitivity, \ac{TPR} and Recall refer to the same concept. The same holds true for the \ac{DSC} and the F\textsubscript{1} Score. The most relevant metrics for the problem categories in the scope of this paper are introduced in the following.

Most metrics are recommended to be applied per class (except for the multi-class counting metrics), meaning that a potential multi-class problem is converted to multiple binary classification problems, such that each relevant class serves as the positive class once. This results in different confusion matrices depending on which class is used as the positive class. 


\subsection{Image-level Classification}
\label{sec:fundamentals_ilc}

\textbf{Image-level classification} refers to the process of assigning one or multiple labels, or \textit{classes}, to an image. Modern algorithms usually output \textbf{predicted class scores} (or continuous class scores) between 0 and 1 for every image and class, indicating the probability of the image belonging to a specific class. By introducing a threshold (e.g., 0.5), predictions are considered as positive (e.g., cancer = true) if they are above the threshold, or negative if they are below the threshold. Subsequently, predictions are assigned to the cardinalities (e.g., a cancer patient with prediction cancer = true is considered as \ac{TP}) \cite{davis2006relationship}. The most popular classification metrics are counting metrics, operating on a confusion matrix with fixed threshold on the class probabilities, and multi-threshold metrics, as detailed in the following. 

\paragraph{\textbf{Counting metrics}} As stated previously, counting metrics rely on the confusion matrix. We distinguish between per-class and multi-class counting metrics. Popular multi-class counting metrics include:
\begin{description}
    \item[Accuracy] \cite{tharwat2020classification}: Fig.~\ref{fig:cheat-sheet-accuracy}
    \item[\acf{BA}] \cite{tharwat2020classification}: Fig.~\ref{fig:cheat-sheet-ba}
    \item[\acf{EC}] (also referred to as Expected Prediction Error or Expected Loss) \cite{bishop2006pattern, hastie2009elements, ferrer2022analysis}: Fig.~\ref{fig:cheat-sheet-ec} 
    \item[\acf{MCC}] (also referred to as Phi Coefficient) \cite{matthews1975comparison}: Fig.~\ref{fig:cheat-sheet-mcc}
    \item[\acf{WCK}] (also referred to as Weighted Cohen‘s Kappa Coefficient, Weighted Kappa Statistic or Weighted Kappa Score) \cite{cohen1960coefficient}: Fig.~\ref{fig:cheat-sheet-wck}
\end{description}

Popular per-class counting metrics include:
\begin{description}
    \item[F$_\beta$ Score] \cite{van1979information, Chinchor1992}: Fig.~\ref{fig:cheat-sheet-fbeta}
        \item[\acf{NB}] \cite{vickers2006decision}: Fig.~\ref{fig:cheat-sheet-nb}
    \item[\acf{NPV}] \cite{tharwat2020classification}: Fig.~\ref{fig:cheat-sheet-npv}
    \item[\acf{PPV}] (also referred to as Precision) \cite{tharwat2020classification}: Fig.~\ref{fig:cheat-sheet-ppv}
    \item[Sensitivity] (also referred to as Recall, \ac{TPR} or Hit Rate) \cite{tharwat2020classification}: Fig.~\ref{fig:cheat-sheet-sensitivity}
    \item[Specificity] (also referred to as Selectivity or \ac{TNR}) \cite{tharwat2020classification}: Fig.~\ref{fig:cheat-sheet-specificity}
\end{description}

\paragraph{\textbf{Multi-threshold metrics}}
The classical counting metrics presented above rely on fixed thresholds to be set on the predicted class probabilities (if available), resulting in them being based on the cardinalities of the confusion matrix.\textbf{ Multi-threshold metrics} overcome this limitation by calculating metric scores based on multiple thresholds. Popular examples are:

\begin{description}
    \item[\acf{AUROC}] (also referred to as \ac{AUC}, AUC - ROC (Area under the Curve - Receiver Operating Characteristics), C-Index, C-Statistics) \cite{hanley1982meaning}: Fig.~\ref{fig:cheat-sheet-auroc}
    \item[\acf{AP}] \cite{lin2014microsoft, everingham2015pascal}: Fig.~\ref{fig:cheat-sheet-ap}
\end{description}

\paragraph{\textbf{Calibration metrics}} While most research in biomedical image analysis focuses on the discrimination capabilities of classifiers, a complementary property of relevance is the \textit{calibration} of predicted class scores (also known as \textit{confidence scores}). Intuitively speaking, a system is well-calibrated if the predicted class scores (i.e., the output of the model) reflect the true probabilities of the outcome. In practice, this means that calibrated scores match the empirical success rate of associated predictions. For a binary classification task, calibration implies that of all the data samples assigned a predicted score of $0.8$ for the positive class, empirically, $80\%$ belong to this class. Popular examples are:

\begin{description}
    \item[\acf{BS}] \cite{gneiting2007strictly}: Fig.~\ref{fig:cheat-sheet-bs}
    \item[\acf{CWCE}] \cite{kumar2019verified, kull2019beyond}: Fig.~\ref{fig:cheat-sheet-cwce}
    \item[\acf{ECE}] \cite{naeini2015obtaining}: Fig.~\ref{fig:cheat-sheet-ece}
    \item[\acf{ECEKDE}]\cite{popordanoska2022consistent} : Fig.~\ref{fig:cheat-sheet-ecekde}
    \item[\acf{KCE}] \cite{gruber2022better, widmann2019calibration}: Fig.~\ref{fig:cheat-sheet-kce}
    \item[\acf{NLL}] \cite{cybenko1998mathematics}: Fig.~\ref{fig:cheat-sheet-nll}
    \item[\acf{RBS}] \cite{gruber2022better}: Fig.~\ref{fig:cheat-sheet-rbs}
\end{description}

\subsection{Semantic Segmentation}
\label{sec:fundamentals_ss}
\textbf{Semantic segmentation} is commonly defined as the process of partitioning an image into multiple segments/regions. To this end, one or multiple labels are assigned to every pixel such that pixels with the same label share certain characteristics. Semantic segmentation can therefore also be regarded as pixel-level classification. As in image-classification problems, predicted class probabilities are typically calculated for each pixel, deciding on the class affiliation based on a threshold over the class scores \cite{asgari2021deep}. In semantic segmentation problems, the pixel-level classification is typically followed by a post-processing step, in which connected components are defined as objects, and object boundaries are created accordingly. Semantic segmentation metrics can roughly be classified into: (1) counting metrics or overlap-based metrics, for measuring the overlap between the reference annotation and the prediction of the algorithm, (2) distance-based or boundary-based metrics, for measuring the distance between object boundaries, and (3) problem-specific metrics, measuring, for example, object volumes.

\paragraph{\textbf{Counting metrics}} The most frequently used segmentation metrics are \textbf{counting metrics}. In the context of segmentation they are also referred to as \textbf{overlap-based metrics}, as they essentially measure the overlap between a reference mask and the algorithm prediction. Popular examples of overlap-based metrics include:

\begin{description}
    \item[\acf{DSC}] (also referred to as Sørensen–Dice Coefficient, F\textsubscript{1} Score,  Balanced F Score) \cite{dice1945measures}: Fig.~\ref{fig:cheat-sheet-dsc}
    \item[\acf{IoU}] (also referred to as Jaccard Index, Tanimoto Coefficient) \cite{jaccard1912distribution}: Fig.~\ref{fig:cheat-sheet-iou}
    \item[\acf{clDice}] \cite{shit2021cldice}: Fig.~\ref{fig:cheat-sheet-cldice}
\end{description}

\paragraph{\textbf{Distance-based metrics}} Overlap-based metrics are often complemented by \textbf{distance-based metrics} that operate exclusively on the \acp{TP} and compute one or several distances between the reference and the prediction. Besides few exceptions, distance-based metrics are often \textbf{boundary-based metrics} which focus on assessing the accuracy of object boundaries. Popular examples include:

\begin{description}
    \item[\acf{ASSD}] (also referred to as Weighted Bilateral Mean Contour Distance) \cite{yeghiazaryan2015overview}: Fig.~\ref{fig:cheat-sheet-assd}
    \item[\acf{Boundary IoU}] \cite{cheng2021boundary}: Fig.~\ref{fig:cheat-sheet-boundary-iou}
    \item[\acf{HD}] (also referred to as Maximum Symmetric Surface Distance, Hausdorff Metric, Pompeiu–Hausdorff Distance) \cite{huttenlocher1993comparing}: Fig.~\ref{fig:cheat-sheet-hd}
    \item[\acf{HD95}] \cite{huttenlocher1993comparing}: Fig.~\ref{fig:cheat-sheet-hd95}
    \item[\acf{MASD}] (also referred to as Mean Surface Distance) \cite{benevs2015performance}: Fig.~\ref{fig:cheat-sheet-masd}
    \item[\acf{NSD}] (also referred to as Normalized Surface Dice, Surface Distance, Surface Dice, Surface DSC) \cite{nikolov2021clinically}: Fig.~\ref{fig:cheat-sheet-nsd}
\end{description}

\paragraph{\textbf{Problem-specific segmentation metrics}} 
While overlap- and distance-based metrics are the standard metrics used by the general computer vision community, biomedical applications often have special domain-specific requirements. In medical imaging, for example,  the actual volume of an object (e.g., a tumor) may be of particular interest. In this case, \textbf{volume metrics} such as the \textit{Absolute} or \textit{Relative Volume Error} and the \textit{Symmetric Relative Volume Difference} can be computed \cite{nai2021comparison}. 

\subsection{Object Detection}
\label{sec:fundamentals_od}
\textbf{Object detection} refers to the detection of one or multiple objects (or: instances) of a particular class (e.g., lesion) in an image \cite{lin2014microsoft}. The following description assumes single-class problems, but translation to multi-class problems is straightforward, as validation for multiple classes on object level is performed individually per class. Notably, as multiple predictions and reference instances may be present in one image, the predictions need to include localization information, such that reference and predicted objects can be matched. Important design choices with respect to the validation of object detection methods include:
\begin{enumerate}
    \item \textit{How to represent an object?} Representation is typically composed of location information and a class affiliation. The former may for example take the form of a bounding box (i.e., a list of coordinates), a pixel mask, or the object's center point. Additionally, modern algorithms typically assign a confidence value to each object, representing the probability of a prediction corresponding to an actual object of the respective class. Note that a confusion matrix is later computed for a fixed threshold on the predicted class probabilities.\footnote{Please note that we will use the term confidence scores analogously to predicted class probabilities in the context of object detection and instance segmentation.}
    \item \textit{How to decide whether a reference instance was correctly detected?} This step is achieved by applying the \textit{localization criterion}. A localization criterion may, for example, be based on comparing the object centers of the reference and prediction or computing their overlap.
    \item \textit{How to resolve assignment ambiguities?} The above step might lead to ambiguous matchings, such as two predictions being assigned to the same reference object. Several strategies exist for resolving such cases.
\end{enumerate}
\noindent
The following sections provide details on (1) applying the localization criterion, (2) applying the assignment strategy, and (3) computing the actual performance metrics.

\paragraph{\textbf{Localization criterion}}
As one image may contain multiple objects or no object at all, the \textbf{localization criterion} or \textbf{hit criterion} measures the (spatial) similarity between a prediction (represented by a bounding box, pixel mask, center point or similar) and a reference object. It defines whether the prediction \textit{hit/detected} (\ac{TP}) or \textit{missed} (\ac{FP}) the reference. Any reference object not detected by the algorithm is defined as \ac{FN}. Please note that \acp{TN} are not defined for object detection tasks. Popular localization criteria include:

\begin{description}
    \item[Box/Approx \acf{IoU}] \cite{jaccard1912distribution}: Fig.~\ref{fig:cheat-sheet-iou-localization-crit}
    \item[Mask \ac{IoU} > 0] \cite{jaccard1912distribution, wack2012improved}: Fig.~\ref{fig:cheat-sheet-iou-localization-crit-0}
    \item[Center Distance] \cite{gurcan2010pattern}: Fig.~\ref{fig:cheat-sheet-center-distance}
    \item[Point inside Mask/ Box/ Approx]\footnote{\url{https://cada.grand-challenge.org/Assessment/}}: Fig.~\ref{fig:cheat-sheet-point-inside}
\end{description}

\paragraph{\textbf{Assignment strategy}}
The localization criterion alone is not sufficient to extract the final confusion matrix based on a fixed threshold for the predicted class probabilities (confidence scores), as ambiguities can occur. For example, two predictions may have been assigned to the same reference object in the localization step, or vice versa. These ambiguities need to be resolved in a further \textbf{assignment step}. This assignment and thus the resolving of potential assignment ambiguities can be done via different strategies:

\begin{description}
    \item[Greedy (by Score) Matching] \cite{everingham2015pascal}: Fig.~\ref{fig:cheat-sheet-greedy-score}
    \item[Optimal (Hungarian) Matching] \cite{kuhn1955hungarian}: Fig.~\ref{fig:cheat-sheet-hungarian}
    \item[Matching via Overlap > 0.5] \cite{everingham20062005}: Fig.~\ref{fig:cheat-sheet-matching-greater}
    \item[Greedy (by Localization Criterion) Matching] \cite{maier2022metrics}: Fig.~\ref{fig:cheat-sheet-greedy-localization}

\end{description}

\paragraph{\textbf{Metric computation}}
Similar to image-level classification and semantic segmentation algorithms, object detection algorithms are commonly assessed with counting metrics, assuming a fixed confusion matrix. Popular examples include:

\begin{description}
    \item[F$_\beta$ Score] \cite{van1979information, Chinchor1992}: Fig.~\ref{fig:cheat-sheet-fbeta}
    \item[\acf{FPPI}] \cite{van2010comparing,bandos2009area}: Fig.~\ref{fig:cheat-sheet-fppi}
    \item[\acf{PPV}] (also referred to as Precision) \cite{tharwat2020classification}: Fig.~\ref{fig:cheat-sheet-ppv}
    \item[Sensitivity] (also referred to as Recall, \ac{TPR} or Hit Rate) \cite{tharwat2020classification}: Fig.~\ref{fig:cheat-sheet-sensitivity}
\end{description}

Similarly, multi-threshold metrics rely on a range of thresholds. Popular examples are: 
\begin{description}
    \item[\acf{AP}] \cite{lin2014microsoft, everingham2015pascal}: Fig.~\ref{fig:cheat-sheet-ap}
    \item[\acf{FROC} Score] \cite{van2010comparing}: Fig.~\ref{fig:cheat-sheet-froc}
\end{description}

\subsection{Instance Segmentation}
\label{sec:fundamentals_is} 
In contrast to semantic segmentation, \textbf{instance segmentation} problems distinguish different instances of the same class (e.g., different lesions). Similarly to object detection problems, the task is to detect individual instances of the same class, but detection performance is measured by pixel-level correspondences (as in semantic segmentation problems). Optionally, instances can be applied to one of multiple classes. Validation metrics in instance segmentation problems often combine common detection metrics with segmentation metrics applied per instance. For instance, segmentation problems, we consider different localization criteria, namely:

\paragraph{Localization criteria:}
\begin{description}
    \item[\acf{Boundary IoU}] \cite{cheng2021boundary}: Fig.~\ref{fig:cheat-sheet-boundary-iou-localization-crit}
    \item[Mask \ac{IoU}] \cite{jaccard1912distribution}: Fig.~\ref{fig:cheat-sheet-iou-localization-crit}
    \item[\acf{IoR}] \cite{mavska2014benchmark}: Fig~\ref{fig:cheat-sheet-ior}
\end{description}

\paragraph{Additional counting metric:}
If detection and segmentation performance should be assessed simultaneously in a single score, the \textbf{\ac{PQ}} metric can be utilized \cite{kirillov2019panoptic}: Fig.~\ref{fig:cheat-sheet-pq}.

 It should be noted that instance segmentation problems are often phrased as semantic segmentation problems with an additional post-processing step, such as connected component analysis \cite{rosenfeld1966sequential}.

\section{Metric Pitfalls}%
\label{app:pitfalls}
This section presents common limitations of image processing metrics related to [P1] an inadequate choice of problem category (Suppl. Note~\ref{sec:pitf:underlying-task}), [P2] poor metric selection (Suppl. Note~\ref{sec:pitf:poor-selection}) and [P3] poor metric application (Suppl. Note~\ref{sec:pitf:poor-application}) in an illustrated manner. \\

To preserve visual clarity, the most important of the presented metric values may be highlighted with color. Green metric values correspond to a "good" metric value (e.g. a high \textit{Sensitivity} score), whereas red values correspond to a "bad" value (e.g. a low \textit{Sensitivity}). Green check marks indicate desirable behavior of metrics, red crosses indicate undesirable behavior. Please note that a low metric value is not automatically a "bad" score. A metric value should always be put into perspective and compared to inter-rater variability. For simplicity, we still use the terms "good" and "bad/poor" throughout the section. Finally, our illustrations do not provide the concrete class probabilities of the presented classifiers.


\subsection{Pitfalls related to an inadequate choice of the problem category}
\label{sec:pitf:underlying-task}
\begin{figure}[H]
\begin{tcolorbox}[title= Assessing object detection performance at image level yields misleading results, colback=white]
    \centering
    \includegraphics[width=0.8\linewidth]{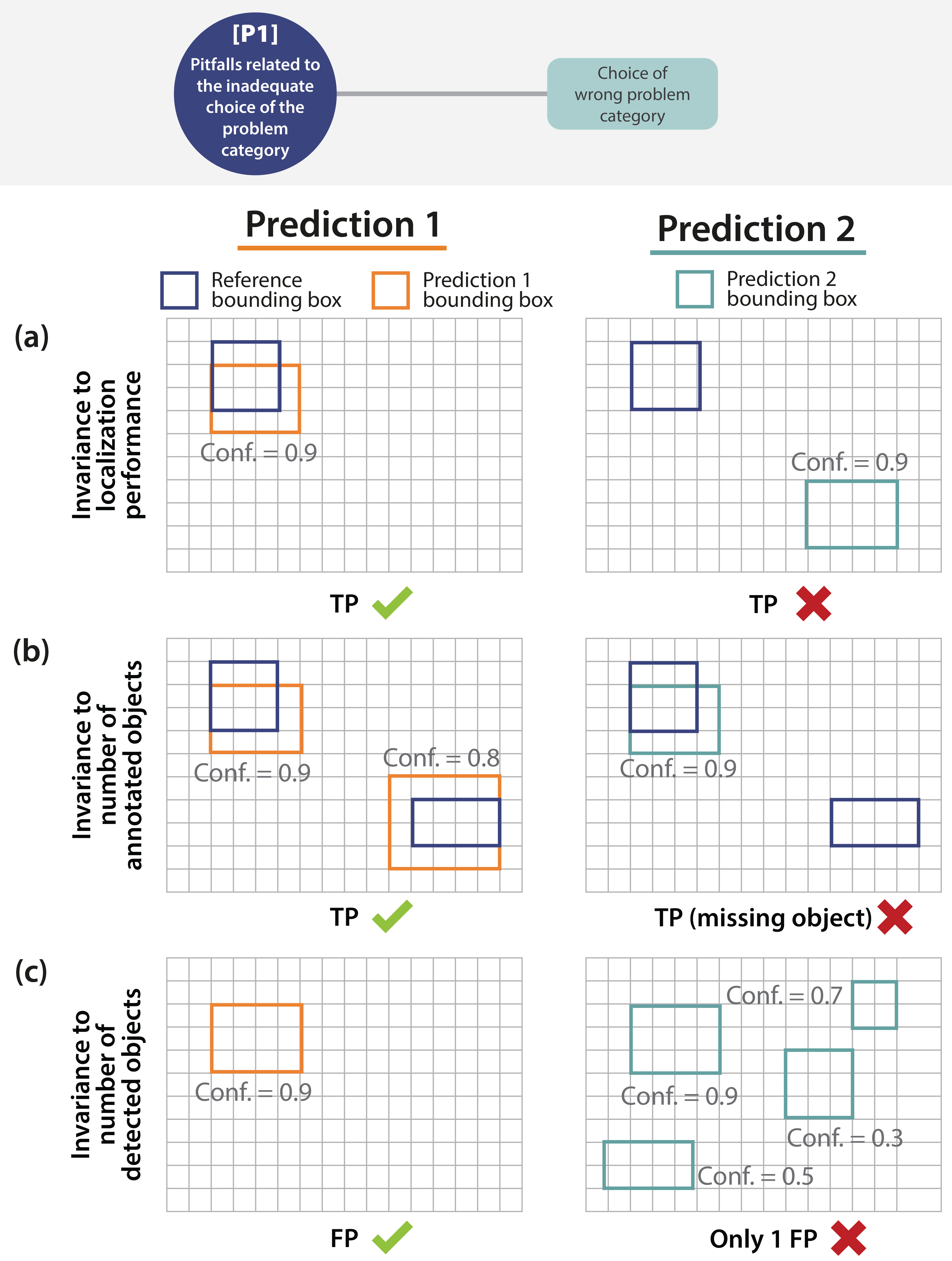}
    \caption{Image-level classification metrics such as the \acf{AUROC} curve can be used to validate object detection models by first aggregating predictions to one image-level score (per class). This validation scheme discards the information on the object matching (localization, number of objects etc.). This leads to several problems:
    \textbf{(a)} The image-level \ac{ROC} curve does not measure the localization performance. Both \textit{Prediction 1} and \textit{2} are considered as \acf{TP} due to their score being very high, although \textit{Prediction 2} does not hit the annotated object. \textbf{(b)} The image-level \ac{ROC} is invariant to the number of annotated objects in an image. The curve does not discriminate between a model detecting all positives (\textit{Prediction 1}) and a model detecting only one of the positives (\textit{Prediction 2}), as long as the maximum score is the same. \textbf{(c)} The image-level \ac{ROC} is invariant to the number of detections in an image. The curve does not discriminate between a model with many False Positives (FP) (\textit{Prediction 2}), and a model with just one \ac{FP} (\textit{Prediction 1}), as long as the maximum score is the same. The class probabilities are represented by confidence scores (Conf.).}
    \label{fig:roc}
\end{tcolorbox}
\end{figure}

\begin{figure}[H]
\begin{tcolorbox}[title= Common metrics may not reflect the domain interest, colback=white]
    \centering
    \includegraphics[width=\linewidth]{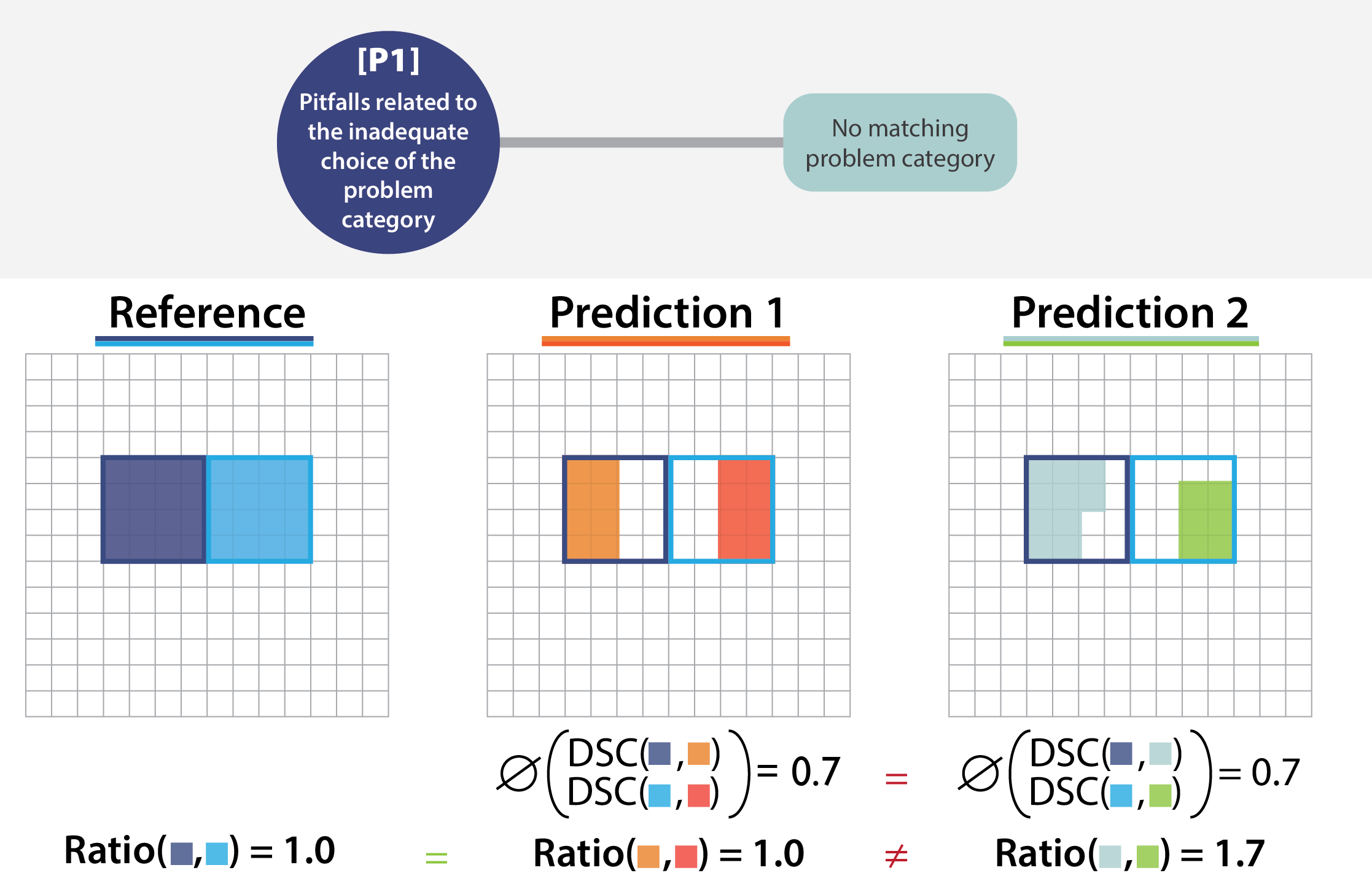}
    \caption{In the absence of a matching problem category for the problem at hand, it may not be possible to find a common metric that ideally captures the domain interest. In this example, accuracy of the ratio between two volumes is the property of interest (e.g., the percentage of blood volume ejected in each cardiac cycle \cite{bamira2018imaging}). Using overlap-based segmentation metrics (here: \acf{DSC}) to measure the volumetric ratio may be misleading. \textit{Predictions 1} and \textit{2} result in similar averaged \ac{DSC} metric values although they result in a different ratio between structure volumes, which is the parameter of interest. $\varnothing$ refers to the average \ac{DSC} values.}
     \label{fig:context-ratio}
\end{tcolorbox}
\end{figure}

\hfill\\
Performance metrics are typically expected to reflect a domain-specific  (e.g., clinical) validation goal. Previous research, however, suggests that this is often not the case \cite{saha2021anatomical}. Before choosing validation metrics, the correct problem category needs to be defined. In the following, we  present pitfalls related to metrics not being applied to the appropriate problem category. These can either be associated with a wrong choice of the problem category (here: Figs.~\ref{fig:pitfalls-p1} and~\ref{fig:roc}; more examples are provided in \citep{reinke2021commonarxiv}) or the lack of a matching problem category (Fig.~\ref{fig:context-ratio}).

\newpage

\newpage
\subsection{Pitfalls related to poor metric selection}
\label{sec:pitf:poor-selection}
Validation metrics typically assess a specific property of interest. Thus, a metric designed for a particular purpose often cannot be used to appropriately validate another property. This is due to both the limitations as well as the mathematical properties of individual metrics, both of which are often neglected. In this section, we present pitfalls related to poor metric selection.

\subsubsection{Pitfalls related to disregard of the domain interest}
\label{sec:pitf:poor-selection-domain}
\begin{figure}[H]
\begin{tcolorbox}[title= Volume-based metrics alone are inadequate for assessing performance, colback=white]
    \centering
    \includegraphics[width=0.9\linewidth]{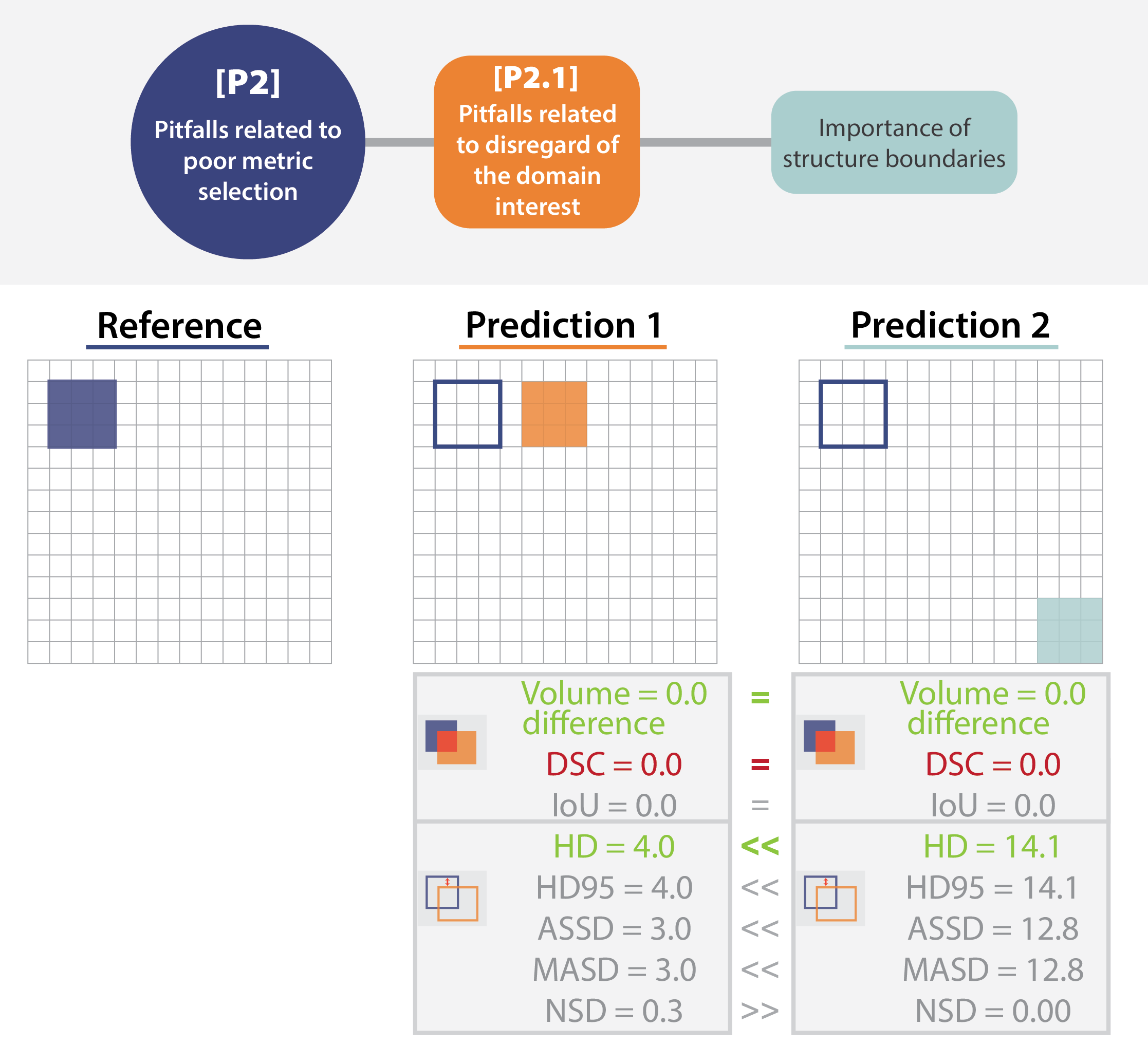}
    \caption{Effect of only focusing on object volume. Both \textit{Predictions 1} and \textit{2} result in the correct volume difference of 0, but do not overlap the reference (\acf{DSC} and \acf{IoU} of 0). Only the boundary-based measures (\acf{HD}, \acf{HD95}, \acf{ASSD}, \acf{MASD}, and \acf{NSD}) recognize the mislocalization. This pitfall is also relevant for localization criteria such as Box/Approx/Mask \ac{IoU}, Center Distance, Mask \ac{IoU} > 0, Point inside Mask/Box/Approx, and \acf{IoR} .}
     \label{fig:volume}
\end{tcolorbox}
\end{figure}

\begin{figure}[H]
\begin{tcolorbox}[title= Boundary-based metrics disregard holes in the segmentation, colback=white]
    \centering
    \includegraphics[width=1\linewidth]{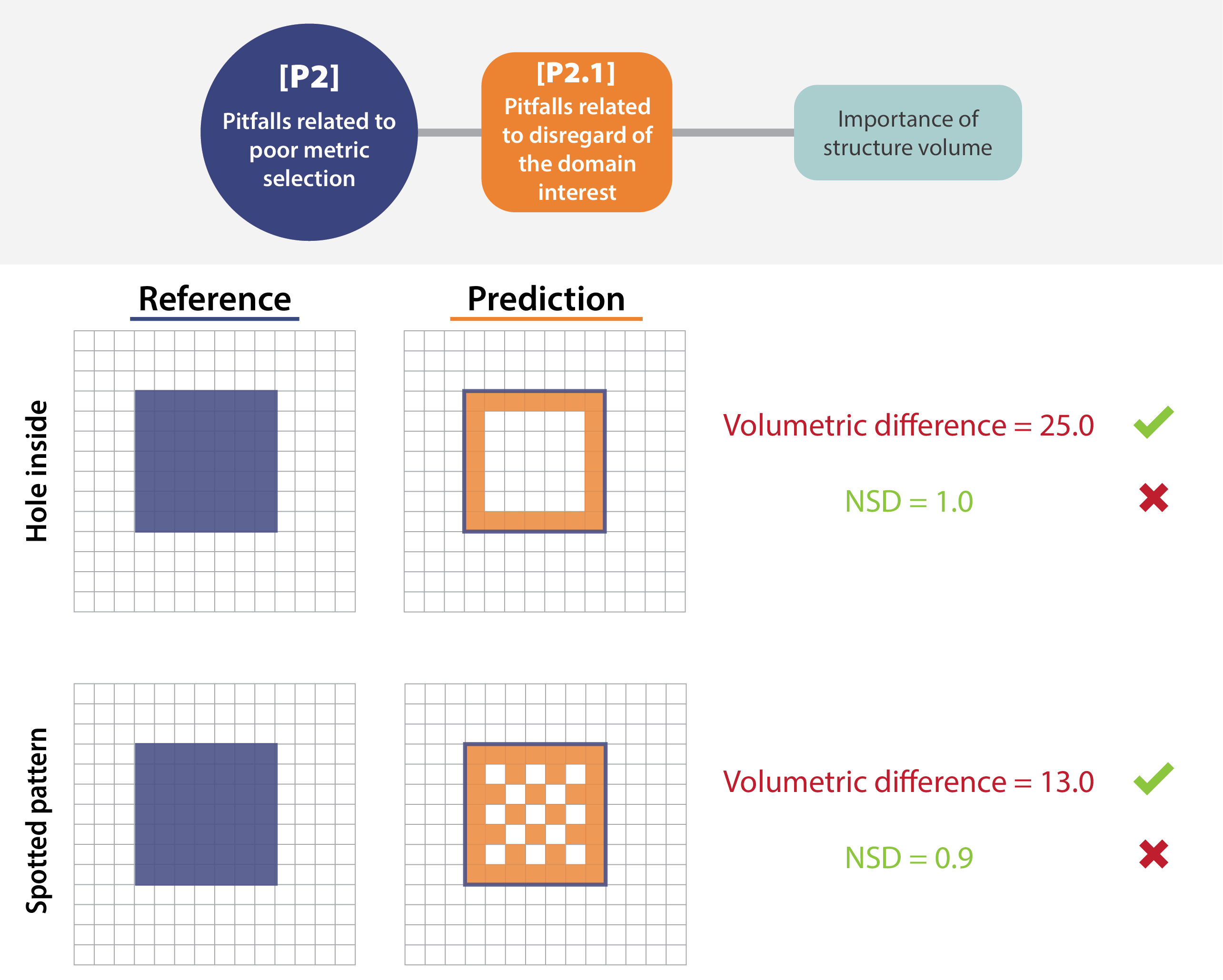}
    \caption{Boundary-based metrics commonly ignore the overlap between structures and are thus insensitive to holes in structures. In the examples, the Prediction respectively features a hole or spotted pattern within the object. Boundary-based metrics (here: \acf{NSD}) do not recognize this problem, yielding (near) perfect metric scores of 1.0 and 0.9, whereas the volumetric difference reflects the fact that the inner area is inadequately predicted. \ac{NSD} was calculated for $\tau=2$. This pitfall is also relevant for other boundary-based metrics such as \acf{ASSD}, \acf{Boundary IoU}, \acf{HD}, \acf{HD95}, and \acf{MASD}, as well as localization criteria such as Center Distance, Mask \ac{IoU} > 0, Point inside Mask/Box/Appeox, \ac{Boundary IoU}, \acf{IoR}, and Mask \ac{IoU}.}
     \label{fig:outline}
\end{tcolorbox}
\end{figure}

\begin{figure}[H]
\begin{tcolorbox}[title= Overlap-based metrics are unaware of object centers, colback=white]
    \centering
    \includegraphics[width=0.8\linewidth]{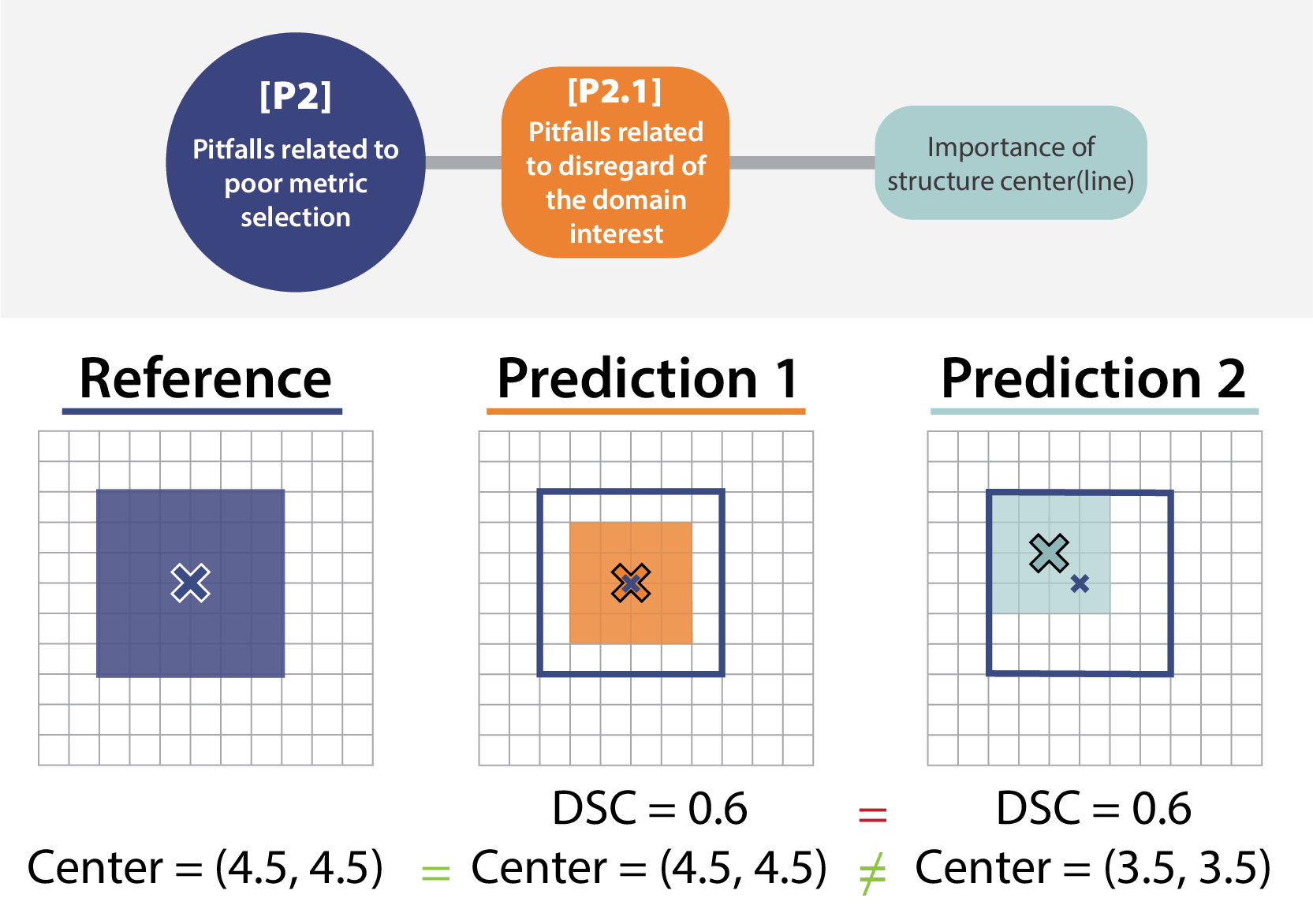}
    \caption{The most common counting-based metrics are poor proxies for the center point alignment. Here, \textit{Predictions 1} and \textit{2} yield the same \acf{DSC} value although \textit{Prediction 1} approximates the location of the object much better. This pitfall is also relevant for other boundary- and overlap-based metrics such as \acf{ASSD}, Boundary \acf{IoU}, \acf{HD}, \acf{HD95}, \ac{IoU}, pixel-level F$_\beta$ Score, and \acf{MASD}, and localization criteria such as Box/Approx/Mask \ac{IoU}, Mask \ac{IoU} > 0, Point inside Mask/Box/Approx, Boundary \ac{IoU}, and \acf{IoR}.}
     \label{fig:center}
\end{tcolorbox}
\end{figure}

\begin{figure}[H]
\begin{tcolorbox}[title= Common calibration metrics falsely imply perfect calibration, colback=white]
    \centering
    \includegraphics[width=0.9\linewidth]{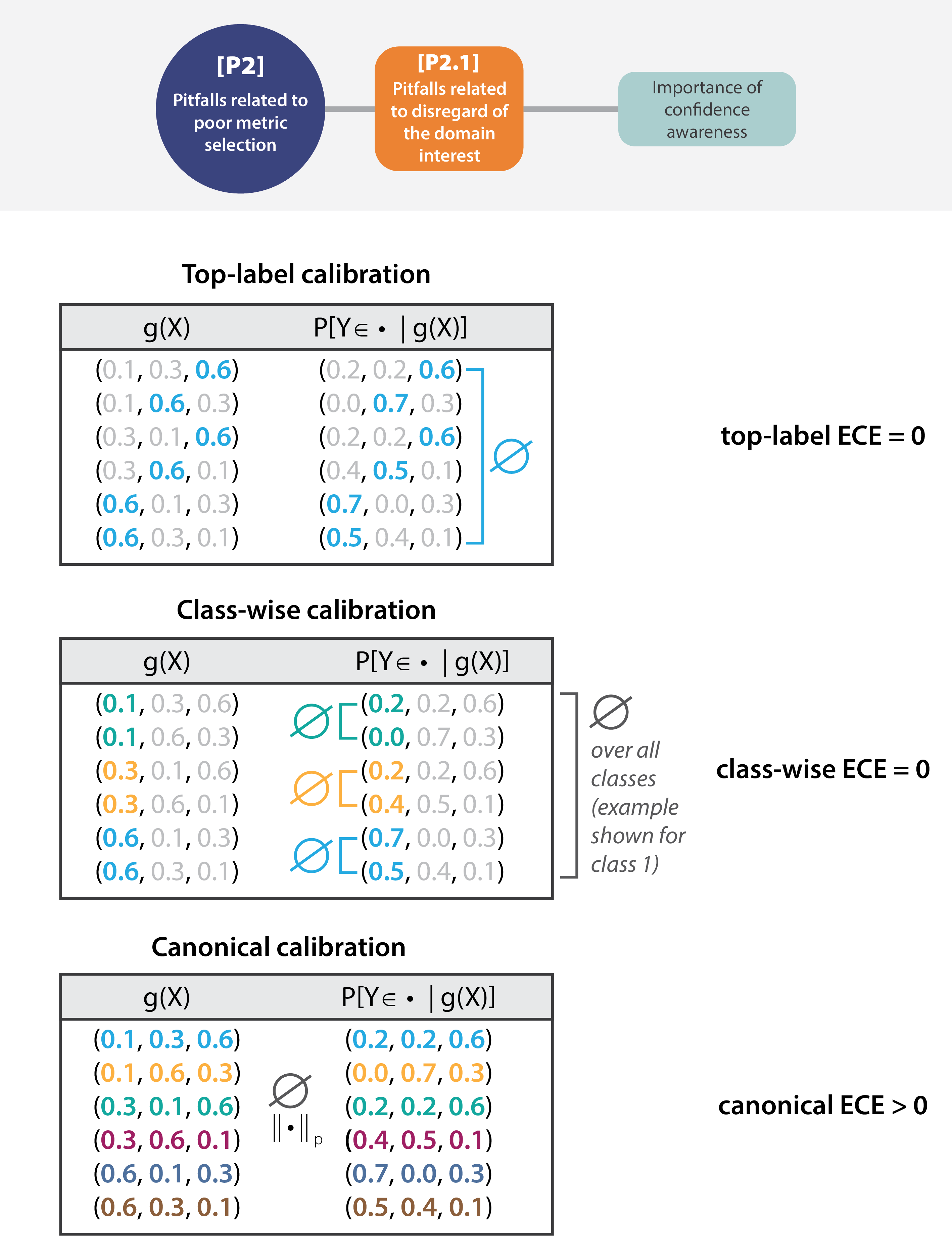}
    \caption{Effect of different definitions of calibration on the \acf{ECE} when focusing on confidence or predicted class scores (confidence awareness). For top-label calibration, only the maximum values of the predicted class scores $g(X)$ are considered, while all other values are neglected, resulting in a perfect calibration for this example. Similarly, for class-wise calibration, the predicted class scores are compared class-wise per value, also yielding a perfect score. Only canonical calibration considers all components of the predicted class score vectors, showing that the model is not perfectly calibrated \citep{gruber2022better,vaicenavicius2019evaluating}. A more detailed insight in different definitions of calibration is given in \citep{maier2022metrics}. It should be noted that discrimination metrics generally do not assess calibration performance, i.e., perfect discrimination does not imply good calibration performance.}
    \label{fig:calibration}
\end{tcolorbox}
\end{figure}

\begin{figure}[H]
\begin{tcolorbox}[title= Comparison of metric scores across data sets may be misleading, colback=white]
    \centering
    \includegraphics[width=0.8\linewidth]{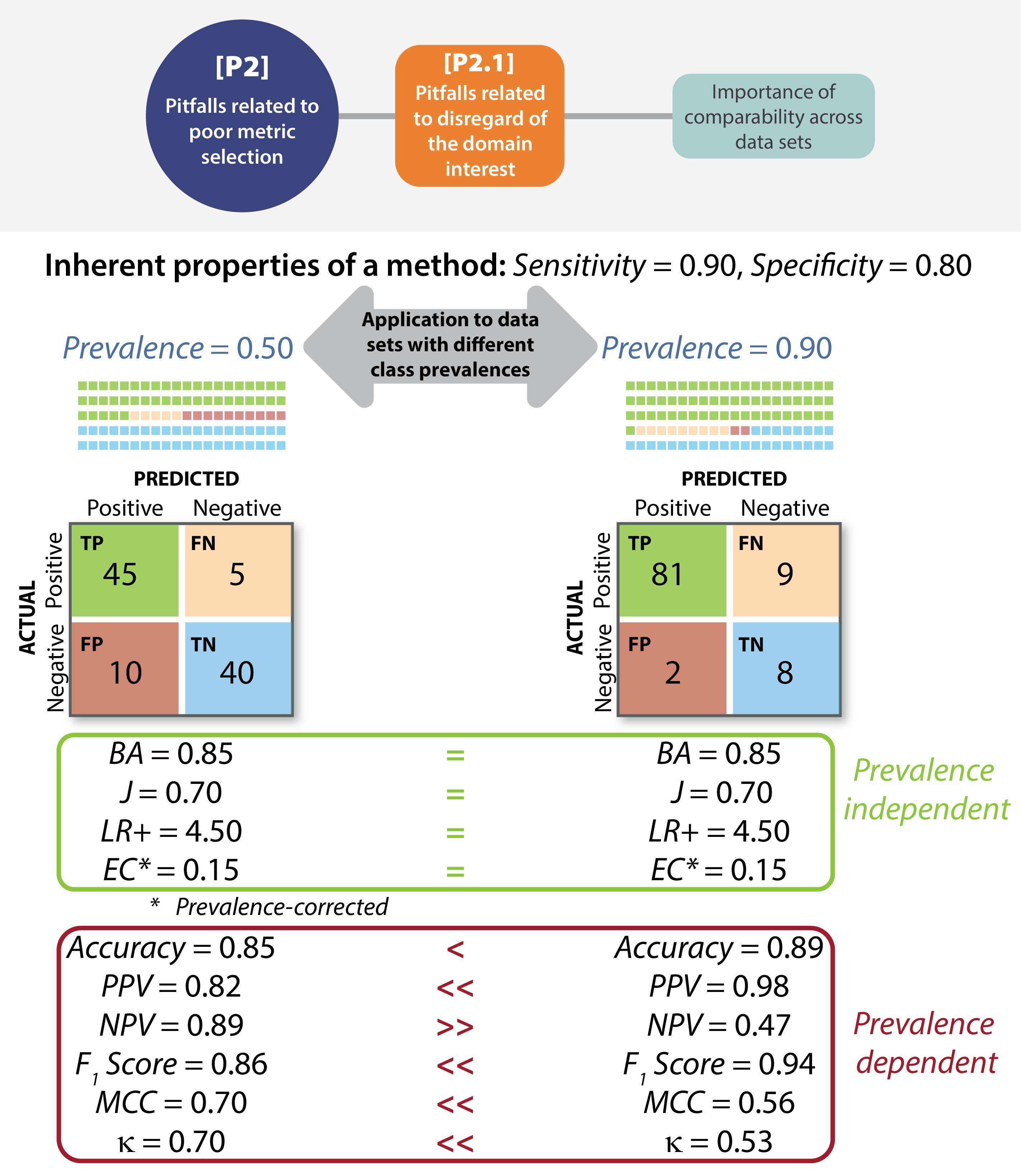}
    \caption{Effect of prevalence dependency. An algorithm with specific inherent properties (here: Sensitivity of 0.9 and Specificity of 0.8) may perform completely differently on different data sets if the prevalences differ (here: 50\% (left) and 90\% (right)) and prevalence-dependent metrics are used for validation (here: Accuracy, \acf{PPV}, \acf{NPV}, F$_1$ Score, \acf{MCC}, Cohen's Kappa $\kappa$). In contrast, prevalence-independent metrics (here: \acf{BA}, Youden's Index J, \acf{LR+}, and \acf{EC}) can be used to compare validation results across different data sets. Used abbreviations: \acf{TP}, \acf{FN}, \acf{FP} and \acf{TN}. This pitfall is also relevant for other counting metrics such as \acf{NB}.}
    \label{fig:prevalence-dependency}
\end{tcolorbox}
\end{figure}

\begin{figure}[H]
\begin{tcolorbox}[title= Overlap-based metrics prefer oversegmentation over undersegmentation, colback=white]
    \centering
    \includegraphics[width=1\linewidth]{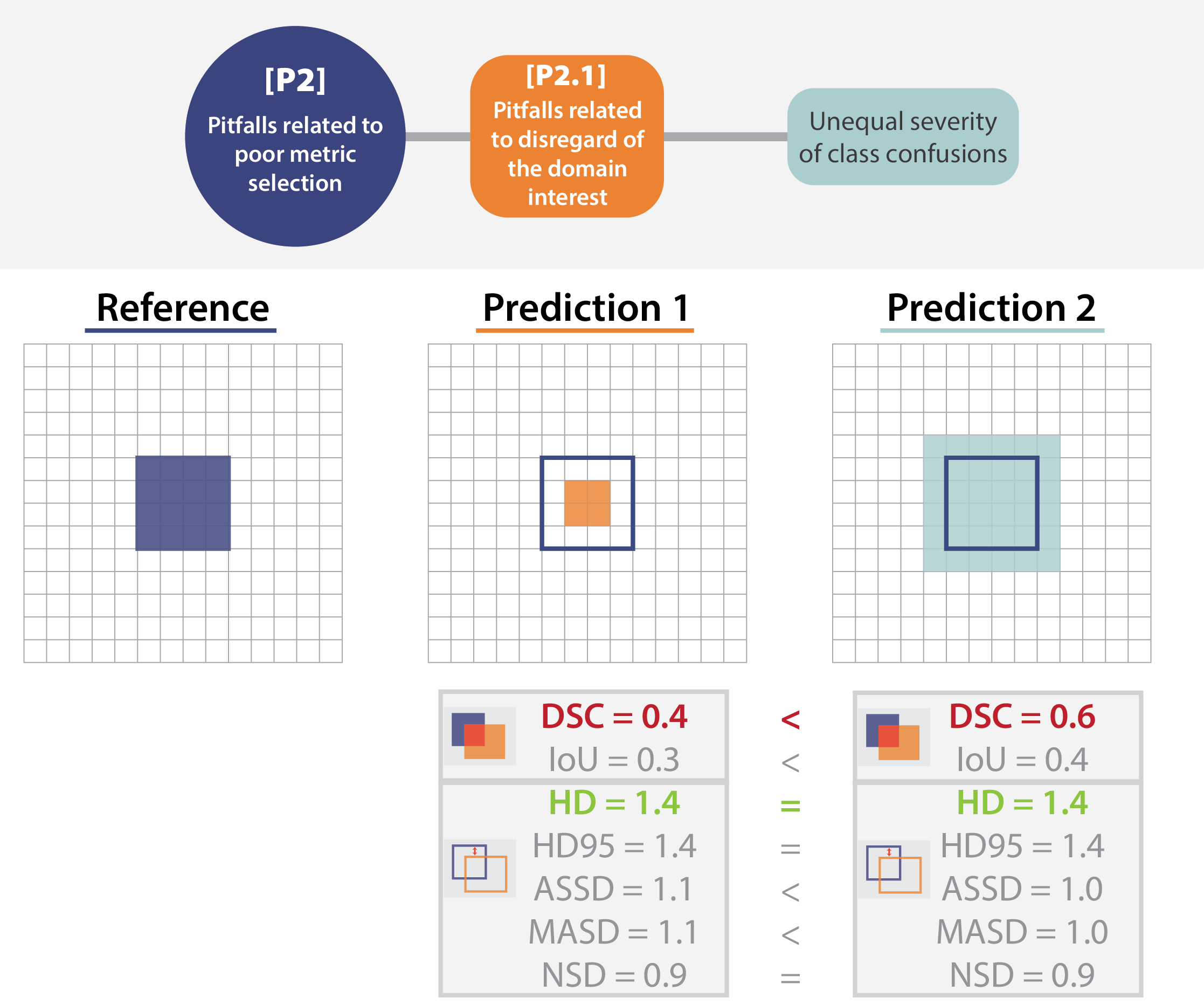}
    \caption{Effect of undersegmentation \textit{vs.} oversegmentation. The outlines of the predictions of two algorithms (\textit{Prediction 1/2}) differ in only a single layer of pixels (\textit{Prediction 1}: undersegmentation -- smaller structure compared to reference, \textit{Prediction 2}: oversegmentation -- larger structure compared to reference). This has no (or only a minor) effect on the \acf{HD}/(95\%), the \acf{NSD}, \ac{MASD}, and the \acf{ASSD}, but yields a substantially different \acf{DSC} or \acf{IoU} score \cite{taha2015metrics, yeghiazaryan2018family}. If penalizing of either over- or undersegmentation is desired (unequal severity of class confusions), other metrics such as the F$_\beta$ Score provide specific penalties for either depending on the chosen hyperparameter $\beta$. This pitfall is also relevant for other overlap-based metrics such as \acf{clDice} and localization criteria such as Box/Approx/Mask \ac{IoU}, Boundary \ac{IoU}, and \acf{IoR}.}
     \label{fig:DSC-overunder}
\end{tcolorbox}
\end{figure}

\begin{figure}[H]
\begin{tcolorbox}[title= Common metrics disregard cost-benefit analysis, colback=white]
    \centering
    \includegraphics[width=1\linewidth]{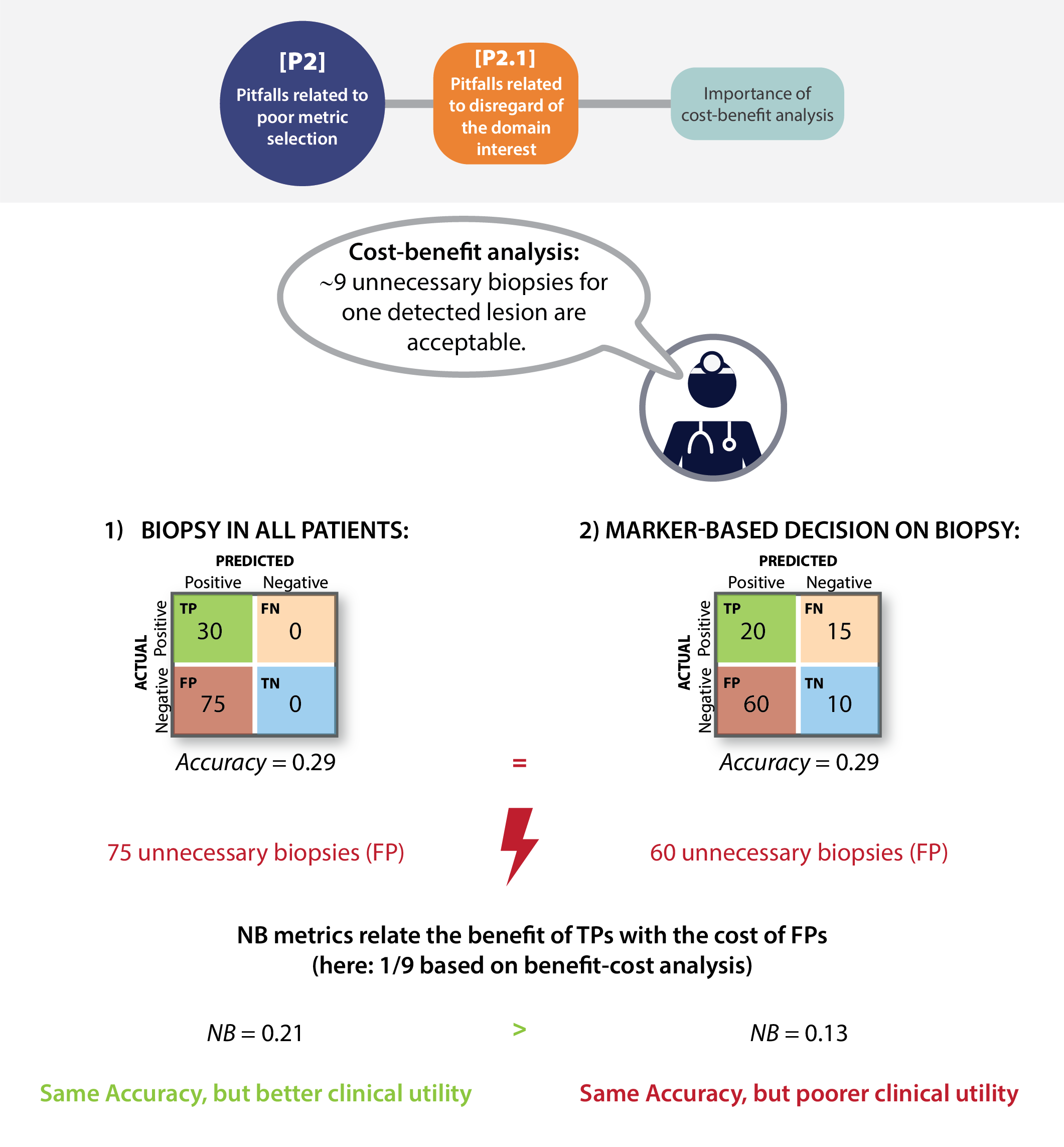}
    \caption{Effect of neglecting a cost-benefit analysis. In a cost-benefit analysis, clinicians are able to define a risk-specific exchange rate that is used in the computation of the \acf{NB} metric. Common metrics such as Accuracy do not consider this analysis and would favor the marker-based decision on biopsy, while \ac{NB} indicates that biopsies of all patients actually yield a better clinical outcome \citep{vickers2016net}. This pitfall is also relevant for other counting metrics such as \acf{BA}, \acf{LR+}, \acf{MCC}, \acf{NPV}, \acf{PPV}, Sensitivity, and Specificity. For binary problems, the hyperparameter $\beta$ of the F$_\beta$ Score can be used as a dynamic penalty for class confusions.}
     \label{fig:cost-benefit}
\end{tcolorbox}
\end{figure}

Several requirements for metric selection arise from the domain interest, which may clash with particular metric limitations. In the following, we present pitfalls related to disregard of the domain interest, stemming from the following sources:

\begin{itemize}
    \item Importance of structure boundaries (Figs.~\ref{fig:pitfalls-p2-1}a and~\ref{fig:volume})
    \item Importance of structure volume (Fig.~\ref{fig:outline})
    \item Importance of structure center(line) (Fig.~\ref{fig:center})
    \item Importance of confidence awareness (Fig.~\ref{fig:calibration})
    \item Importance of comparability across data sets (Figs.~\ref{fig:prevalence-dependency})
    \item Unequal severity of class confusions (Figs.~\ref{fig:pitfalls-p2-1}b and~\ref{fig:DSC-overunder})
    \item Importance of cost-benefit analysis (Fig.~\ref{fig:cost-benefit})
\end{itemize}

\newpage

\newpage
\subsubsection{Pitfalls related to disregard of the properties of the target structure}
\label{sec:pitf:poor-selection-target}
\begin{figure}[H]
\begin{tcolorbox}[title= Common localization criteria and overlap-based metrics are sensitive to structure sizes, colback=white]
    \centering
    \includegraphics[width=0.9\linewidth]{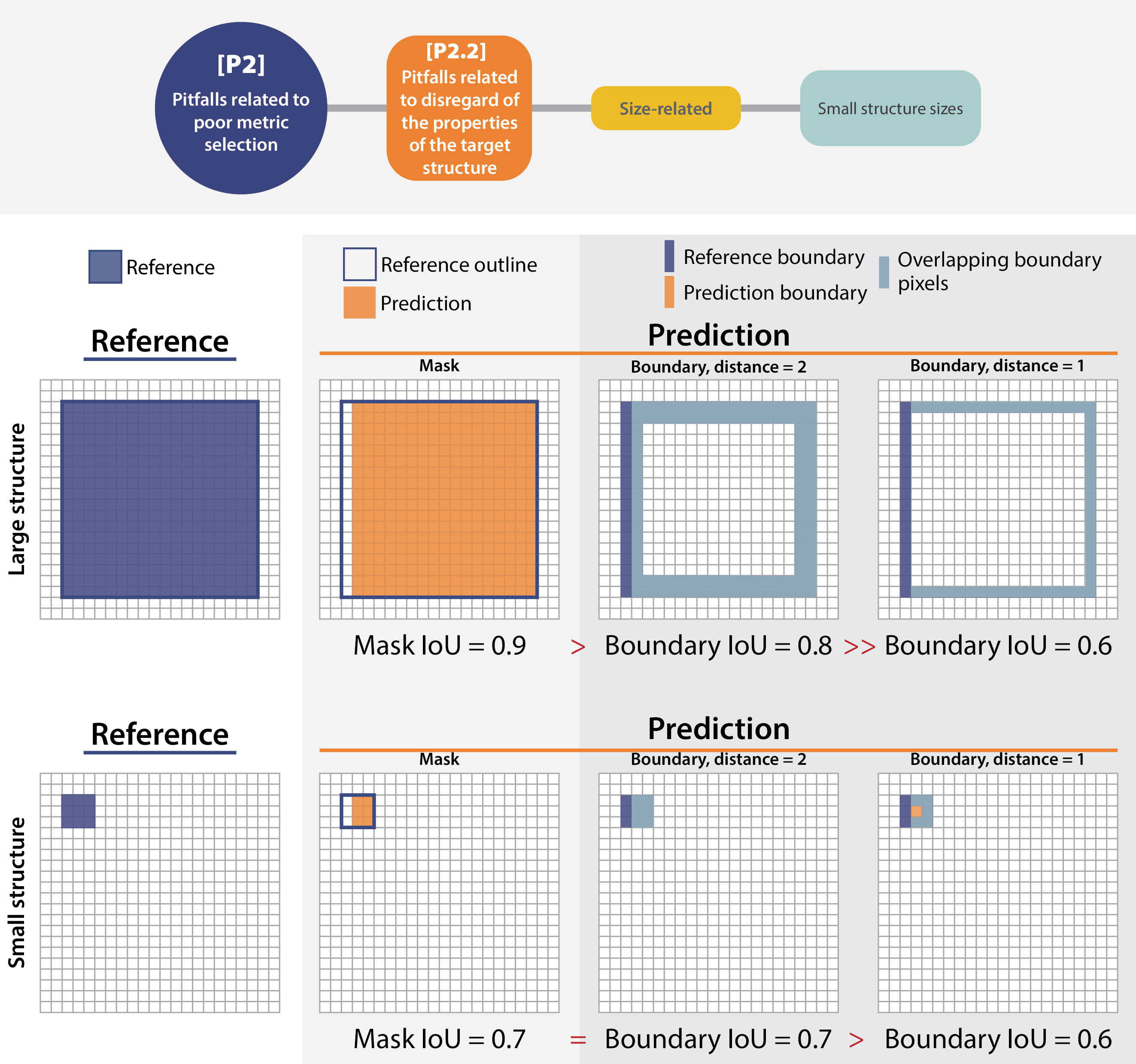}
    \caption{Comparison of Mask and Boundary Intersection over Union (IoU) localization criteria in the case of particular importance of structure boundaries. Overlapping pixels from the reference and prediction are shown in light blue. The Mask IoU (second column) is less sensitive to boundary errors for large objects. The Boundary IoU (third and fourth column) especially considers contours, (1) yields smaller metric scores, thus penalizing errors in the boundaries, and (2) is more invariant to structure sizes, leading to very similar values for large and small structures (fourth column) \cite{cheng2021boundary}. This pitfall is also relevant for other overlap-based metrics such as \acf{clDice}, \acf{DSC}, and pixel-level F$_\beta$ Score, as well as localization criteria such as Box/Approx \ac{IoU} and \acf{IoR}.}
     \label{fig:boundary-mask-iou} 
\end{tcolorbox}
\end{figure}

\begin{figure}[H]
\begin{tcolorbox}[title= Effect of high variability of structure sizes, colback=white]
    \centering
    \includegraphics[width=\linewidth]{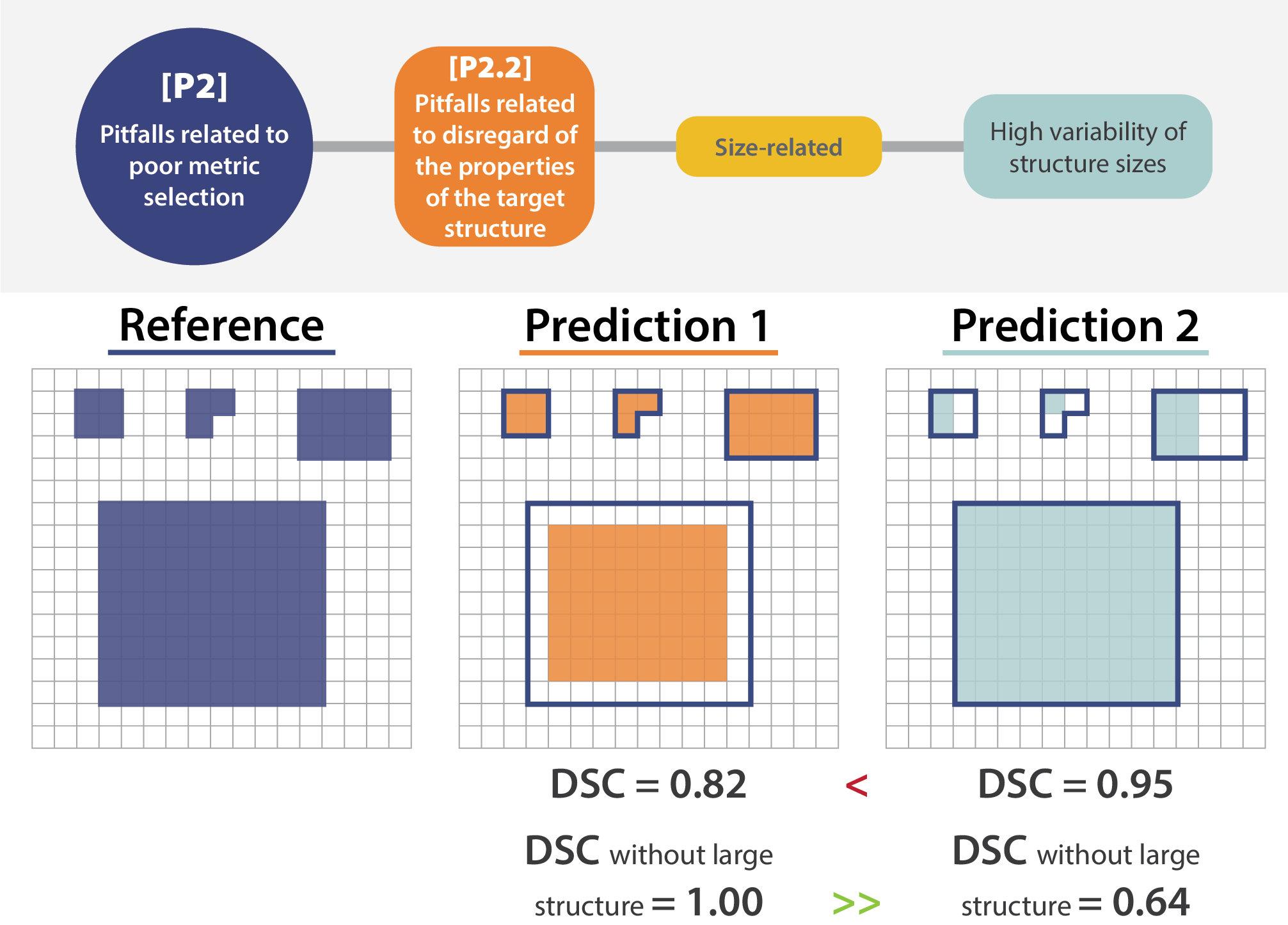}
    \caption{Large structures completely dominate overlap-based metrics in semantic segmentation problems. While \textit{Prediction 1} perfectly segments all three small structures, the metric score (here: \acf{DSC}) is much worse compared to the score of \textit{Prediction 2}, with only one perfect prediction for the large structure. This is highlighted by only computing the metric without the large structure. This pitfall is also relevant for other overlap-based metrics such as \acf{clDice}, \acf{DSC}, and pixel-level F$_\beta$ Score, as well as localization criteria such as Mask/Box/Approx \acf{IoU} and \acf{IoR}.}
    \label{fig:high-variability}
\end{tcolorbox}
\end{figure}

For problems that require capturing local properties (object detection, semantic or instance segmentation), the properties of the target structures to be localized and/or segmented may have severe implications for metric  choice. Pitfalls can be further subdivided into \textit{size-related} and \textit{shape- and topology-related} pitfalls. In the following, we present pitfalls stemming from the following sources:

\textbf{Size-related pitfalls:}
\begin{itemize}
    \item Small structure sizes (Extended Data Fig.~\ref{fig:pitfalls-p2-2}a and Fig.~\ref{fig:boundary-mask-iou})
    \item High variability of structure sizes (Fig.~\ref{fig:high-variability})
\end{itemize}

\begin{figure}[H]
\begin{tcolorbox}[title=Common metrics are unaware of object shapes, colback=white]
    \centering
    \includegraphics[width=1\linewidth]{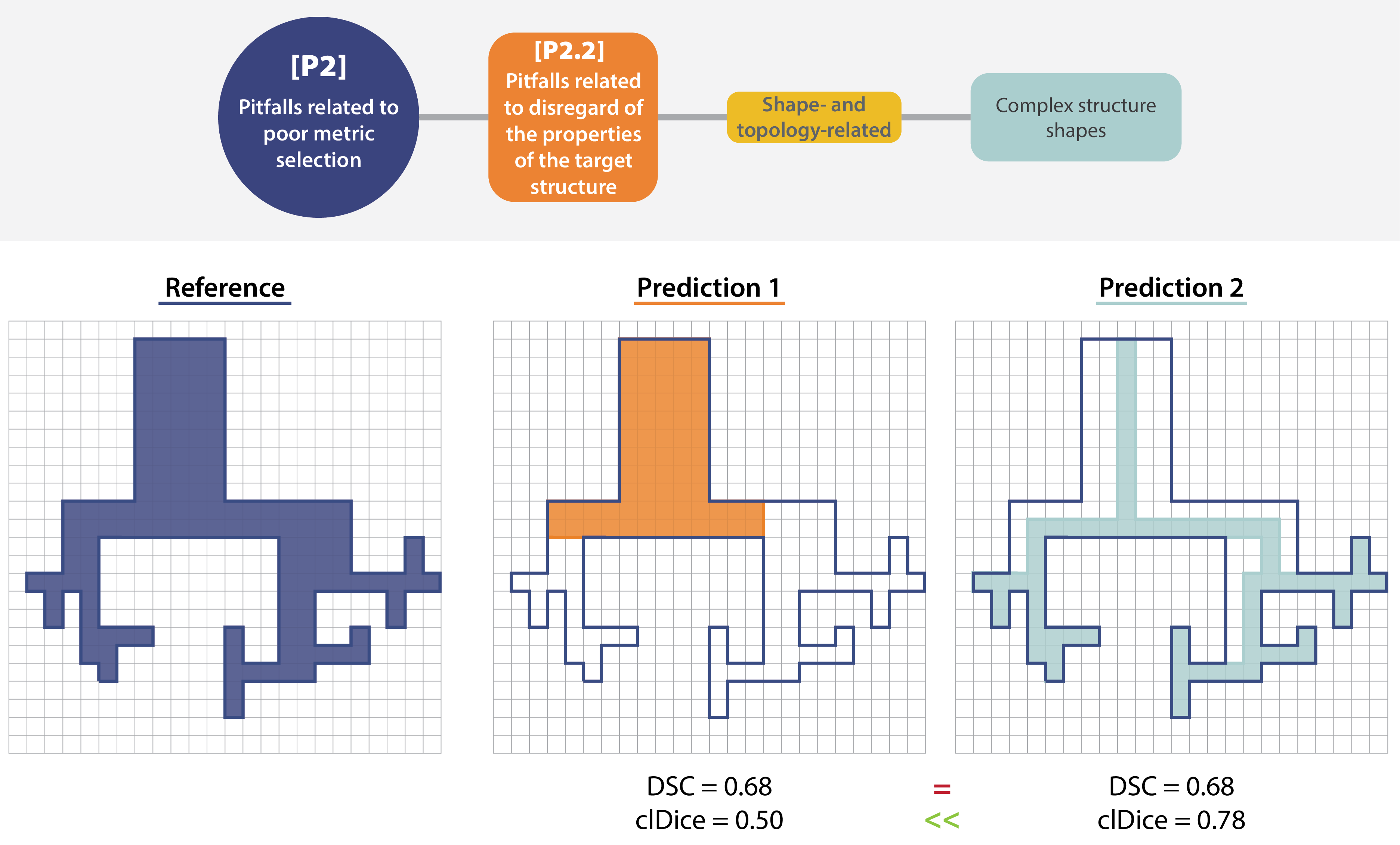}
    \caption{Effect of complex shapes. Common overlap-based metrics such as the \acf{DSC} are unaware of complex structure shapes and treat \textit{Predictions 1} and \textit{2} equally. The \acf{clDice} uncovers that \textit{Prediction 1} misses the fine-granular branches of the reference and favors \textit{Prediction 2}, which focuses on the object's center line and better captures its fine branches. This pitfall is also relevant for other overlap-based metrics such as \acf{IoU} and pixel-level F$_\beta$ Score, and localization criteria such as Box/Approx/Mask \ac{IoU}, Center Distance, Mask \ac{IoU} > 0, Point inside Mask/Box/Approx, and \acf{IoR}.}
     \label{fig:complex-shapes}
\end{tcolorbox}
\end{figure}

\begin{figure}[H]
\begin{tcolorbox}[title= Common metrics do not account for hierarchical label structure, colback=white]
    \centering
    \includegraphics[width=1\linewidth]{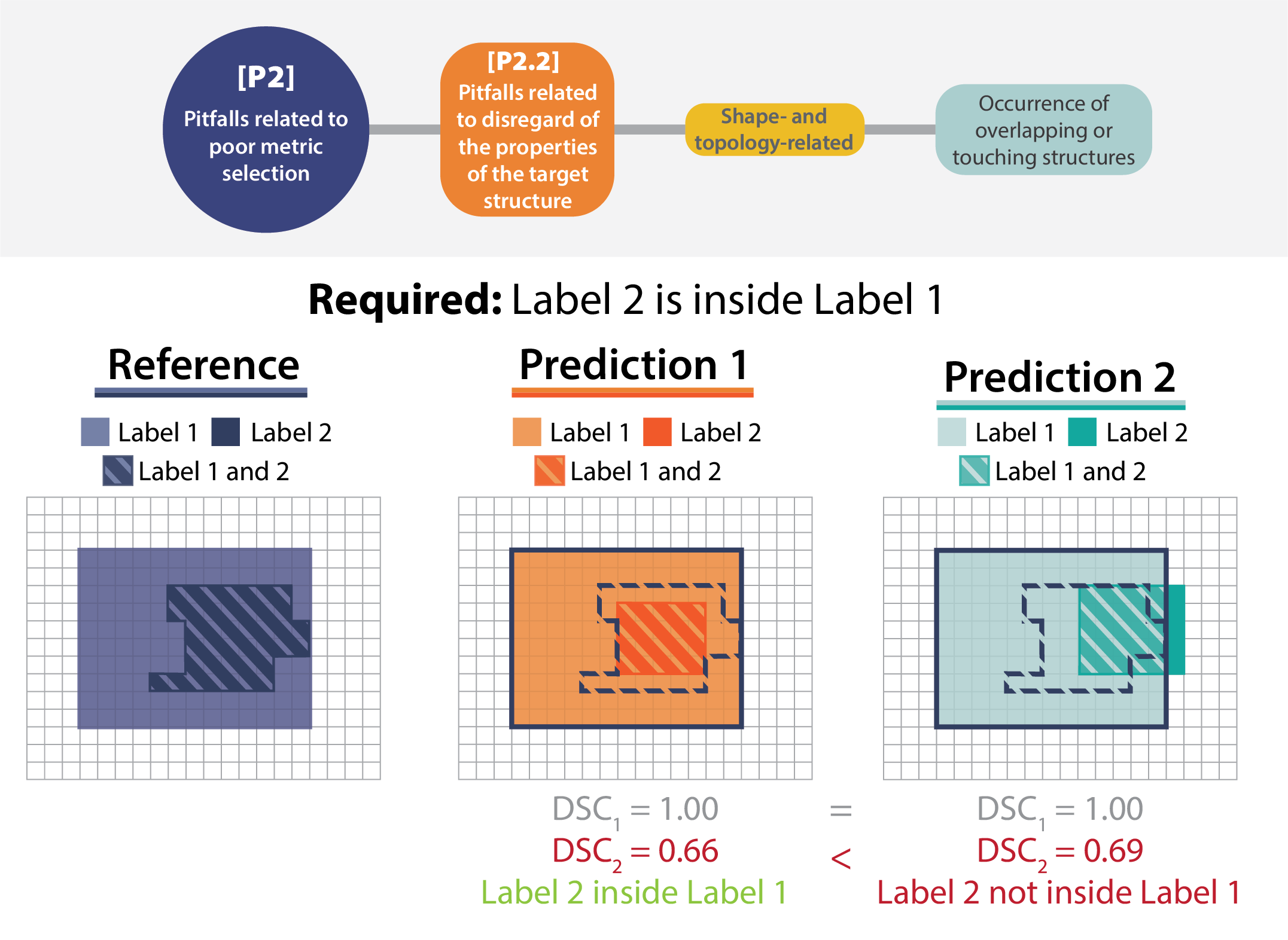}
    \caption{Effect of nested multi-label structures. The requirement of \textit{Label 2} being inside \textit{Label 1} is violated by \textit{Prediction 2}. Nevertheless, \textit{Prediction 2} has a higher \acf{DSC} score compared to \textit{Prediction 1}, which adheres to the requirement. This pitfall is also relevant for other boundary- and overlap-based metrics such as \acf{ASSD}, Boundary \acf{IoU}, \acf{clDice}, \acf{HD}, \acf{HD95}, \ac{IoU}, pixel-level F$_\beta$ Score, \acf{MASD}, and \acf{NSD}.}
    \label{fig:multi-labels}
\end{tcolorbox}
\end{figure}

\begin{figure}[H]
\begin{tcolorbox}[title= Bounding boxes are inadequate for representing complex shapes and disconnected structures, colback=white]
    \centering
    \includegraphics[width=1\linewidth]{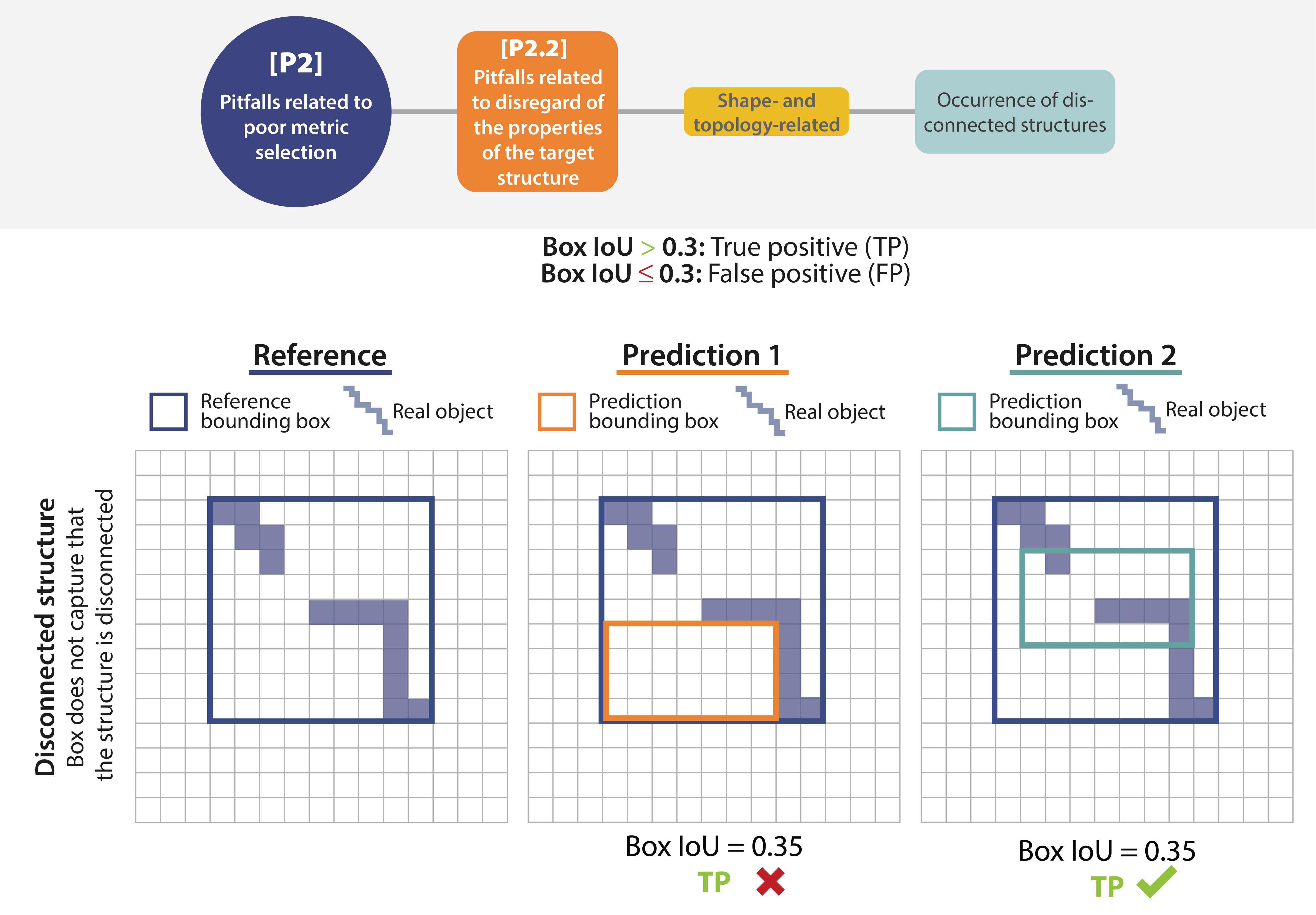}
    \caption{Bounding boxes are not well-suited for representing disconnected shapes, in particular multi-component structures. \textit{Predictions 1} and \textit{2} both yield a \acf{TP} detection, as the Box \acf{IoU} is larger than the threshold 0.3. However, \textit{Prediction 1} does not hit the real object at all.}
    \label{fig:disconnected}
\end{tcolorbox}
\end{figure}

\textbf{Shape- and topology-related pitfalls}
\begin{itemize}
    \item Complex structure shapes (Extended Data Fig.~\ref{fig:pitfalls-p2-2}b and Fig.~\ref{fig:complex-shapes})
    \item Occurrence of overlapping or touching structures (Fig.~\ref{fig:multi-labels})
    \item Occurrence of disconnected structures (Fig.~\ref{fig:disconnected})
\end{itemize}

\newpage

\newpage
\subsubsection{Pitfalls related to disregard of the properties of the data set and algorithm output}
\label{sec:pitf:poor-selection-data-output}
\begin{figure}[H]
\begin{tcolorbox}[title= Common metrics yield implausible results in the presence of class imbalance, colback=white]
    \centering
    \includegraphics[width=0.8\linewidth]{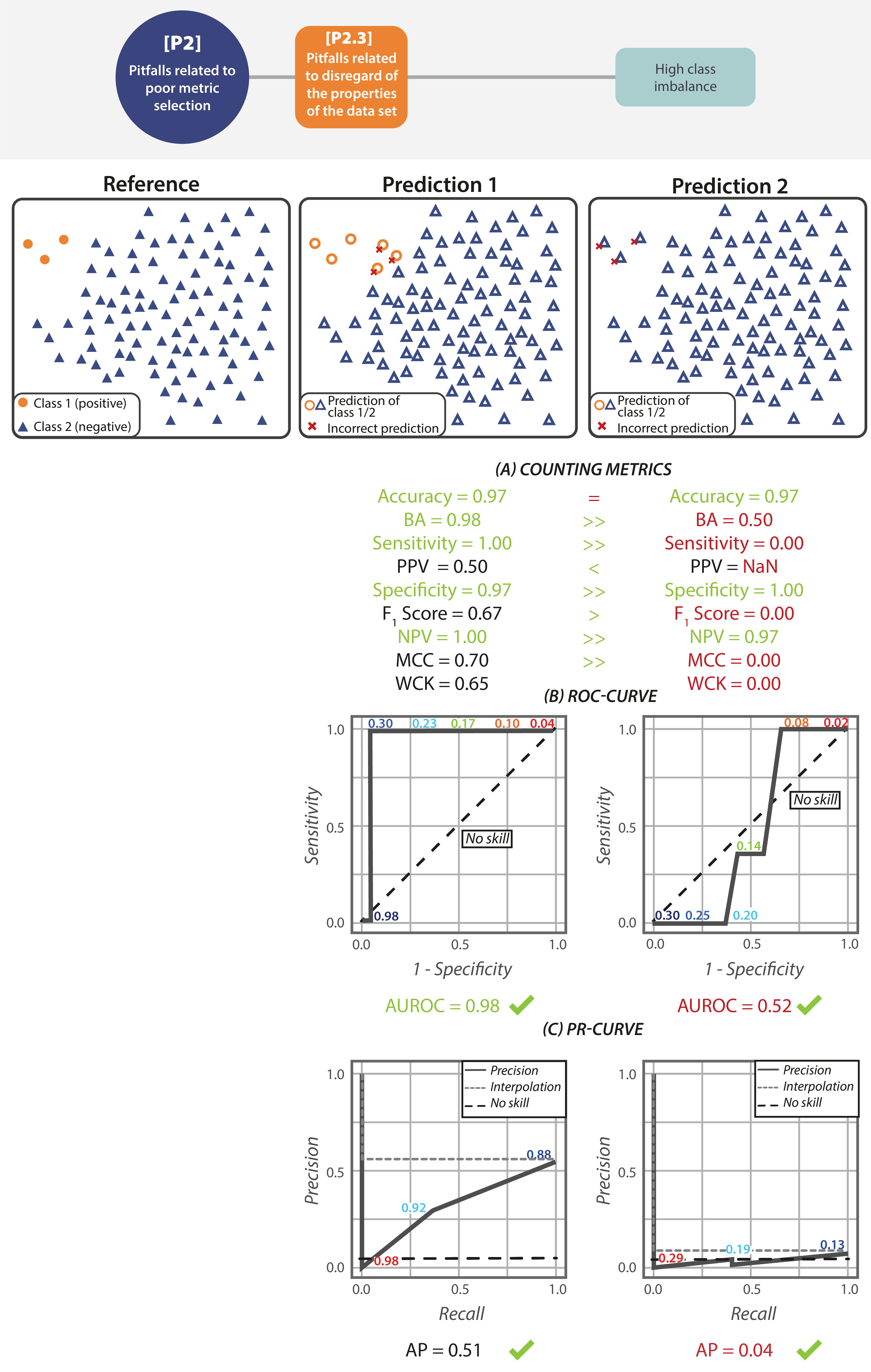}
    \caption{Effect of class imbalance. Not every metric is designed to reflect class imbalance \citep{chicco2020advantages}. In the case of underrepresented classes, an unsuitable metric, such as Accuracy, yields a high value even if the classifier performs very poorly for one of the classes (here: \textit{Prediction 2}). Multi-threshold metrics, such as the \acf{AUROC} and the \acf{AP}, reveal the weakness, indicating that \textit{Prediction 2} is not better than random guessing. For comparison, a no-skill classifier (random guessing) is shown as a black dashed line. For the \acf{PR} curves, the interpolation applied to compute the \ac{AP} metric is shown as a dashed grey line. Thresholds used for curve generation are provided as small numbers above the curve. Further abbreviations: \acf{PPV}, \acf{NPV}, \acf{MCC}, \acf{WCK}. This pitfall is also relevant for other counting metrics such as \acf{NB}.}
    \label{fig:class-imbalance}
\end{tcolorbox}
\end{figure}

\begin{figure}[H]
\begin{tcolorbox}[title= Common multi-threshold metrics are not well-suited for small sample sizes, colback=white]
    \centering
    \includegraphics[width=0.8\linewidth]{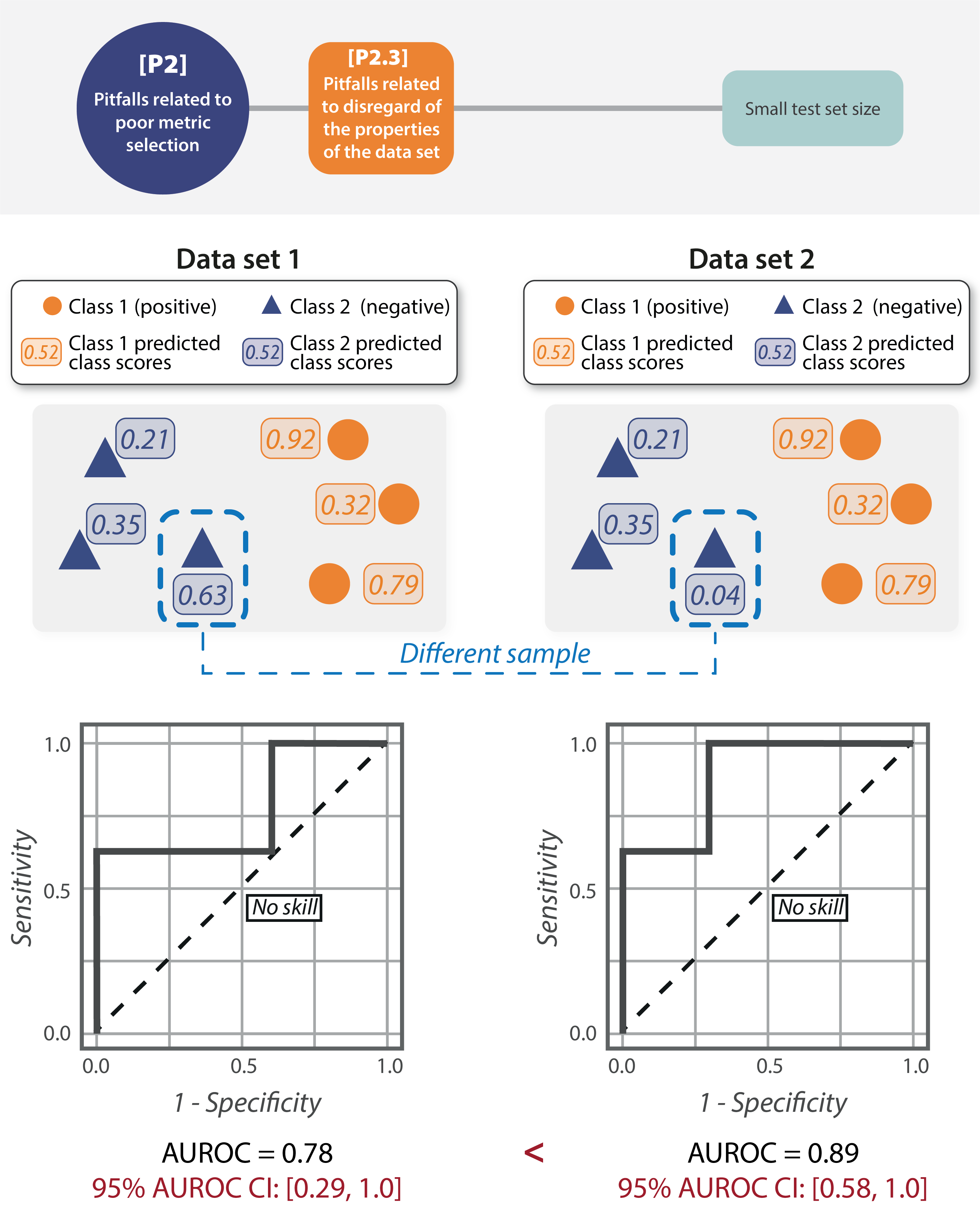}
    \caption{Effect of calculating the \acf{AUROC} for very small sample sizes. The \ac{AUROC} is very unstable for small sample sizes. \textit{Data sets} 1 and 2 only contain six samples each, for which only one predicted score differs between sets. Drawing the \acf{ROC} curve and calculating the \ac{AUROC} leads to a large difference in scores between both data sets. The 95\% \acf{CI} reveals that there is a large range of possible \ac{AUROC} values. \ac{CI}s were calculated based on \cite{delong1988comparing}. This pitfall is also relevant for other counting metrics such as Accuracy, \acf{AP}, \acf{BA}, \acf{EC}, F$_\beta$ Score, \acf{FROC} Score, \acf{LR+}, \acf{MCC}, \acf{NB}, \acf{NPV}, \acf{PPV}, Sensitivity, Specificity, and \acf{WCK}.}
    \label{fig:auroc-small-sample-sizes}
\end{tcolorbox}
\end{figure}

\begin{figure}[H]
\begin{tcolorbox}[title= Common metrics do not account for inter-rater variability, colback=white]
    \centering
    \includegraphics[width=1\linewidth]{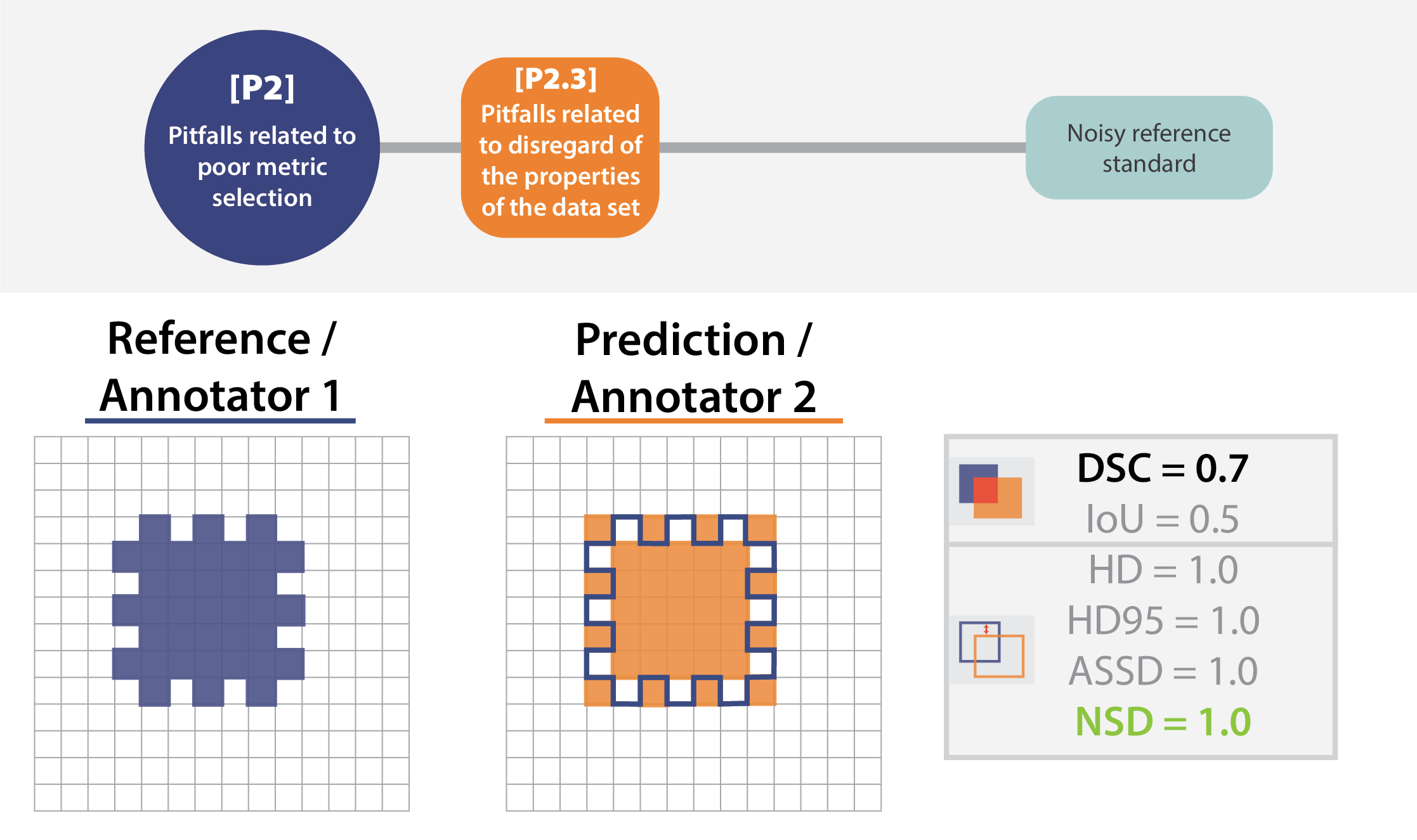}
    \caption{Effect of inter-rater variability between two annotators. Assessing the performance of \textit{Annotator 2} while using an reference annotation created by \textit{Annotator 1} leads to a low \acf{DSC} score because inter-rater variability is not taken into account by common overlap-based metrics. In contrast, the \acf{NSD}, applied with a threshold of $\tau = 1$, captures this variability. It should be noted, however, that this effect occurs primarily in small structures as overlap-based metrics tend to be robust to variations in the object boundaries in large structures. Further abbreviations: \acf{IoU}, \acf{HD}, \acf{HD95}, \acf{ASSD}, \acf{MASD}. This pitfall is also relevant for other boundary- and overlap-based metrics \ac{Boundary IoU}, \acf{clDice}, pixel-level F$_\beta$ Score and \acf{MASD} and localization criteria such as Mask \ac{IoU} > 0, Point inside Mask, \ac{Boundary IoU}, \ac{IoR}, and Mask \ac{IoU}.}
    \label{fig:low-quality}
\end{tcolorbox}
\end{figure}

Properties of the data set such as class imbalances or high inter-rater variability may directly affect metric values. Pitfalls can be further subdivided into \textit{class-related} and \textit{reference-related} pitfalls. For reference-based metrics, the algorithm output will be compared against the reference annotation to compute a metric score. Thus, the content and format of the prediction is of high relevance for metric choice. In the following, we present pitfalls stemming from the following sources:

\textbf{[P2.3] Disregard of the properties of the data set}
\begin{itemize}
    \item High class imbalance (Figs.~\ref{fig:pitfalls-p2-3}a and~\ref{fig:class-imbalance}) 
    \item Small test set size (Figs.~\ref{fig:pitfalls-p2-3}b and~\ref{fig:auroc-small-sample-sizes}) 
    \item Imperfect reference standard (Figs.~\ref{fig:pitfalls-p2-3}c and~\ref{fig:low-quality})
\end{itemize}

\begin{figure}[H]
\begin{tcolorbox}[title= Empty reference or prediction leads to invalid scores, colback=white]
    \centering
    \includegraphics[width=0.8\linewidth]{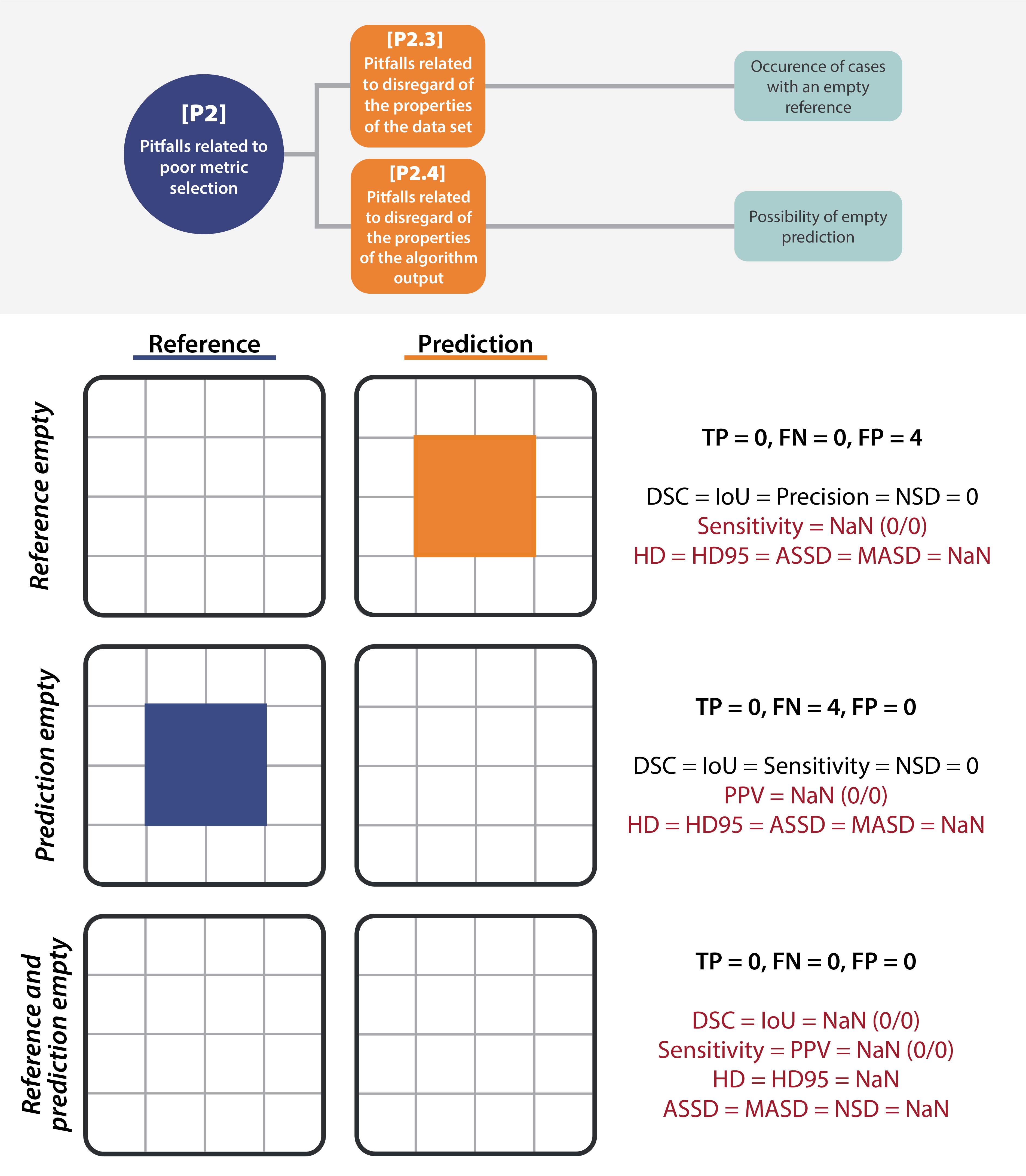}
    \caption{Effect of empty references or predictions when applying common metrics per image (here for semantic segmentation). Empty images lead to division by zero for many common metrics as the numbers of the \acp{TP}, \acp{FP}, \acp{FN} turn zero. Used abbreviations: \acf{ASSD}, \acf{DSC}, \acf{HD}, \acf{HD95}, \acf{IoU}, \acf{MASD}, \acf{NaN}, \acf{NSD}. This pitfall is also relevant for other boundary-based, overlap-based and counting metrics such as \ac{Boundary IoU}, \acf{clDice}, F$_\beta$ Score, \acf{NPV}, \acf{PPV}, Sensitivity, and Specificity.}
    \label{fig:empty}
\end{tcolorbox}
\end{figure}

\begin{figure}[H]
\begin{tcolorbox}[title= Common segmentation metrics are not well-suited for overlapping structures, colback=white]
    \centering
    \includegraphics[width=1\linewidth]{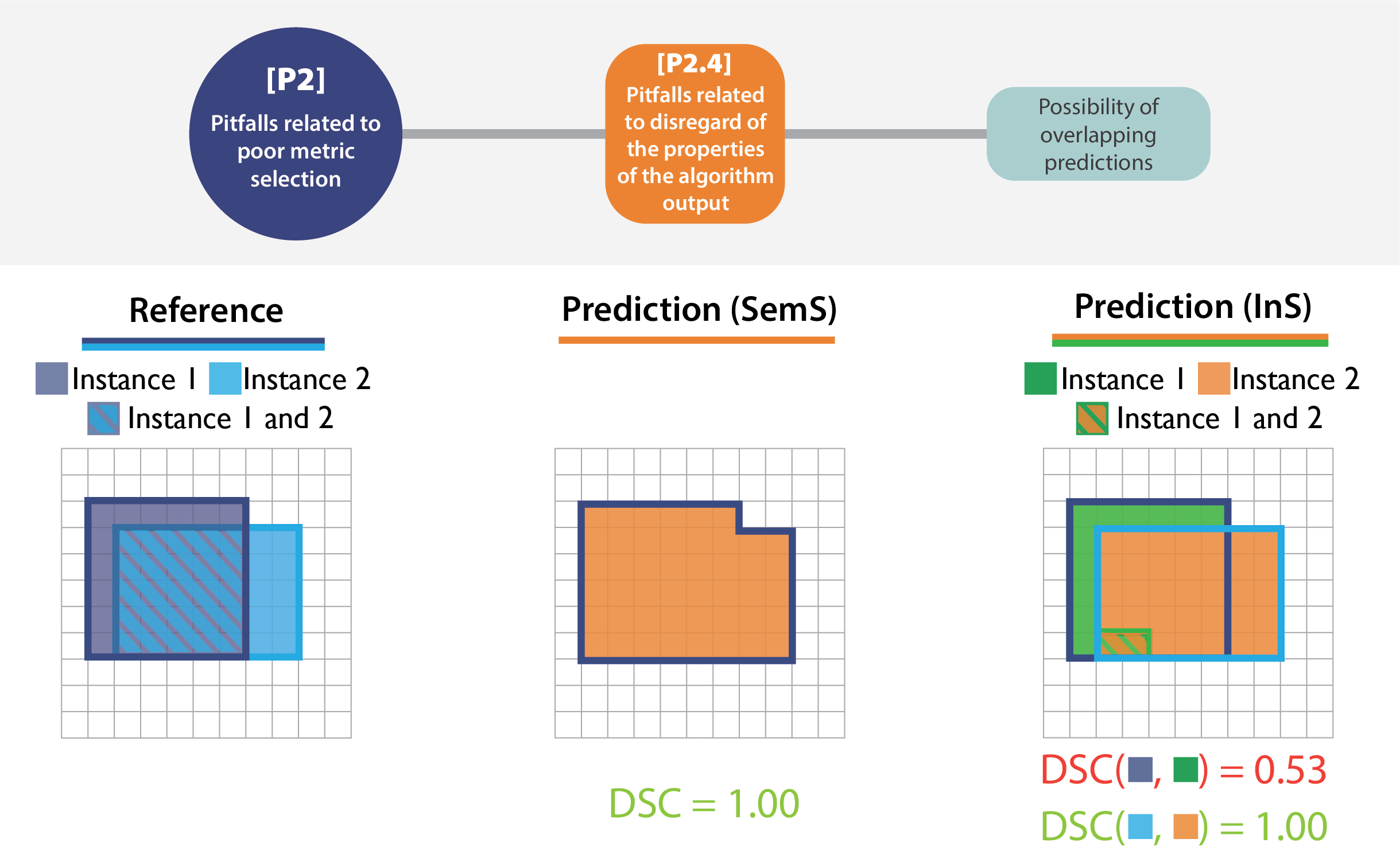}
    \caption{Effect of overlapping predictions in segmentation problems. In semantic segmentation problem (SemS; right), overlapping predictions are merged into a single object, yielding a perfect metric score. Phrasing the problem as an instance segmentation problem reveals that the dark blue instance is not well-approximated at all. This issue is not revealed by common metrics if only semantic segmentation is performed (here: \acf{DSC}). This pitfall is also relevant for other boundary- and overlap-based metrics such as \acf{ASSD}, Boundary \acf{IoU}, \acf{clDice}, pixel-level F$_\beta$ Score, \acf{HD}, \acf{HD95}, \ac{IoU}, \acf{MASD}, and \acf{NSD}.}
    \label{fig:overlapping-pred}
\end{tcolorbox}
\end{figure}

\begin{figure}[H]
\begin{tcolorbox}[title= Selection of multi-threshold metrics in the absence of predicted class scores, colback=white]
    \centering
    \includegraphics[width=1\linewidth]{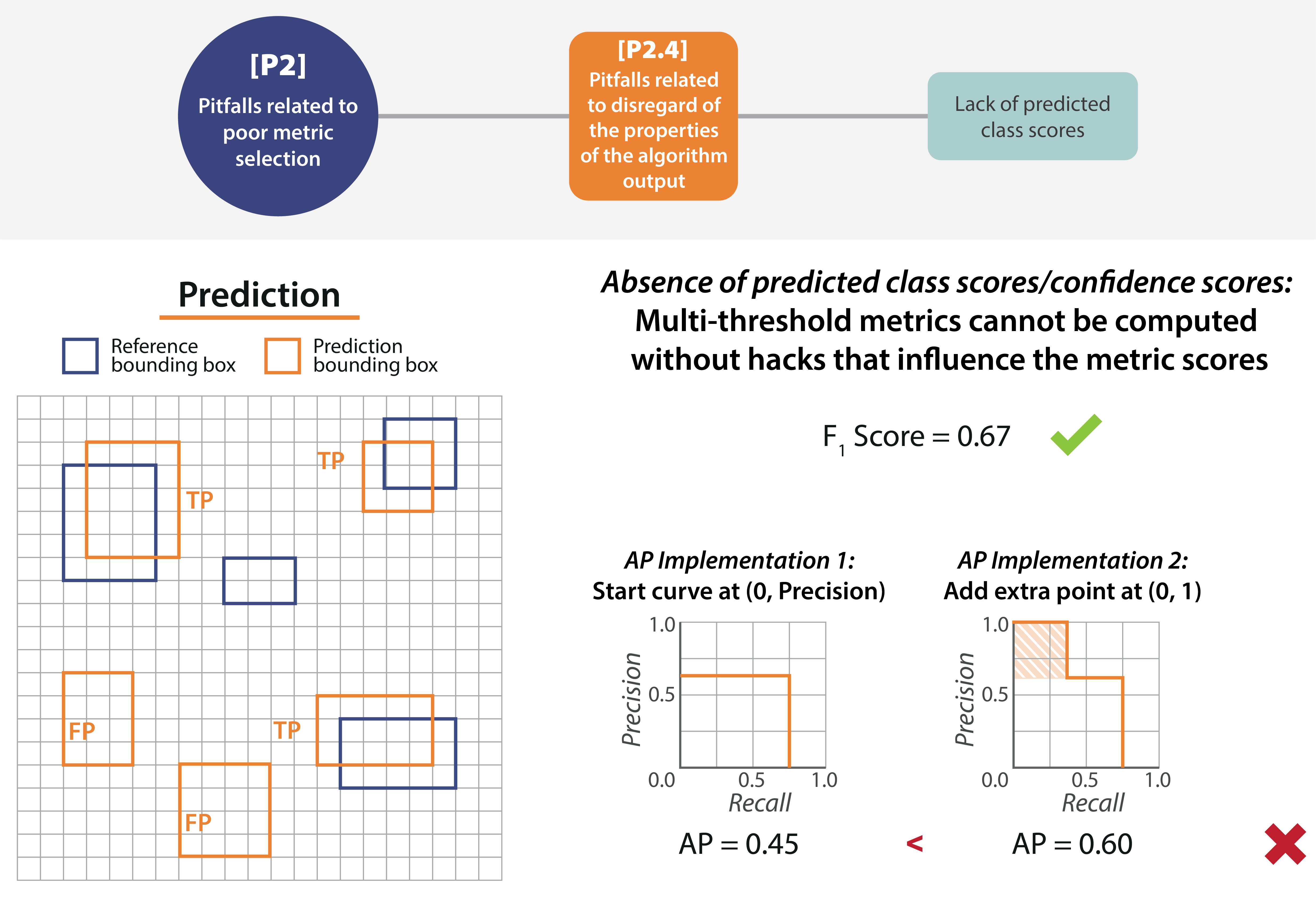}
    \caption{Multi-threshold metrics should only be computed if predicted class scores are available, although an increasing body of work  computes multi-threshold metrics such as \ac{AP} in the absence of class scores (e.g., \citep{bai2017deep, de2017semantic, gao2019ssap, hirsch2020patchperpix, kulikov2020instance}). Otherwise, the strategy chosen for compensating the lack of class scores (here reflected by \textit{Implementations 1} and \textit{2}) leads to metric scores that are less well interpretable than those of established counting metrics working on a fixed confusion matrix (here: F$_\text{1}$ Score). This pitfall is also relevant for other multi-threshold metrics such as \acf{AUROC} and \acf{FROC} Score.}
    \label{fig:lack-of-scores}
\end{tcolorbox}
\end{figure}

\textbf{[P2.4] Disregard of the properties of the algorithm output}
\begin{itemize}
    \item Possibility of empty prediction (Extended Data Fig.~\ref{fig:pitfalls-p2-4}b and Fig.~\ref{fig:empty})
    \item Possibility of overlapping predictions (Extended Data Fig.~\ref{fig:pitfalls-p2-4}a and Fig.~\ref{fig:overlapping-pred})
    \item  Lack of predicted class scores (Fig.~\ref{fig:lack-of-scores})
\end{itemize}

\newpage 


\newpage
\subsection{Pitfalls related to poor metric application}
\label{sec:pitf:poor-application}
\hfill\\A data set typically contains several hundreds or thousands of images. When analyzing, aggregating and combining metric values, a number of factors need to be taken into account. 

\subsubsection{Pitfalls related to inadequate metric implementation}
\label{sec:pitf:poor-application-impl}
\begin{figure}[H]
\begin{tcolorbox}[title= Lack of standardization leads to variation in metric scores, colback=white]
    \centering
    \includegraphics[width=1\linewidth]{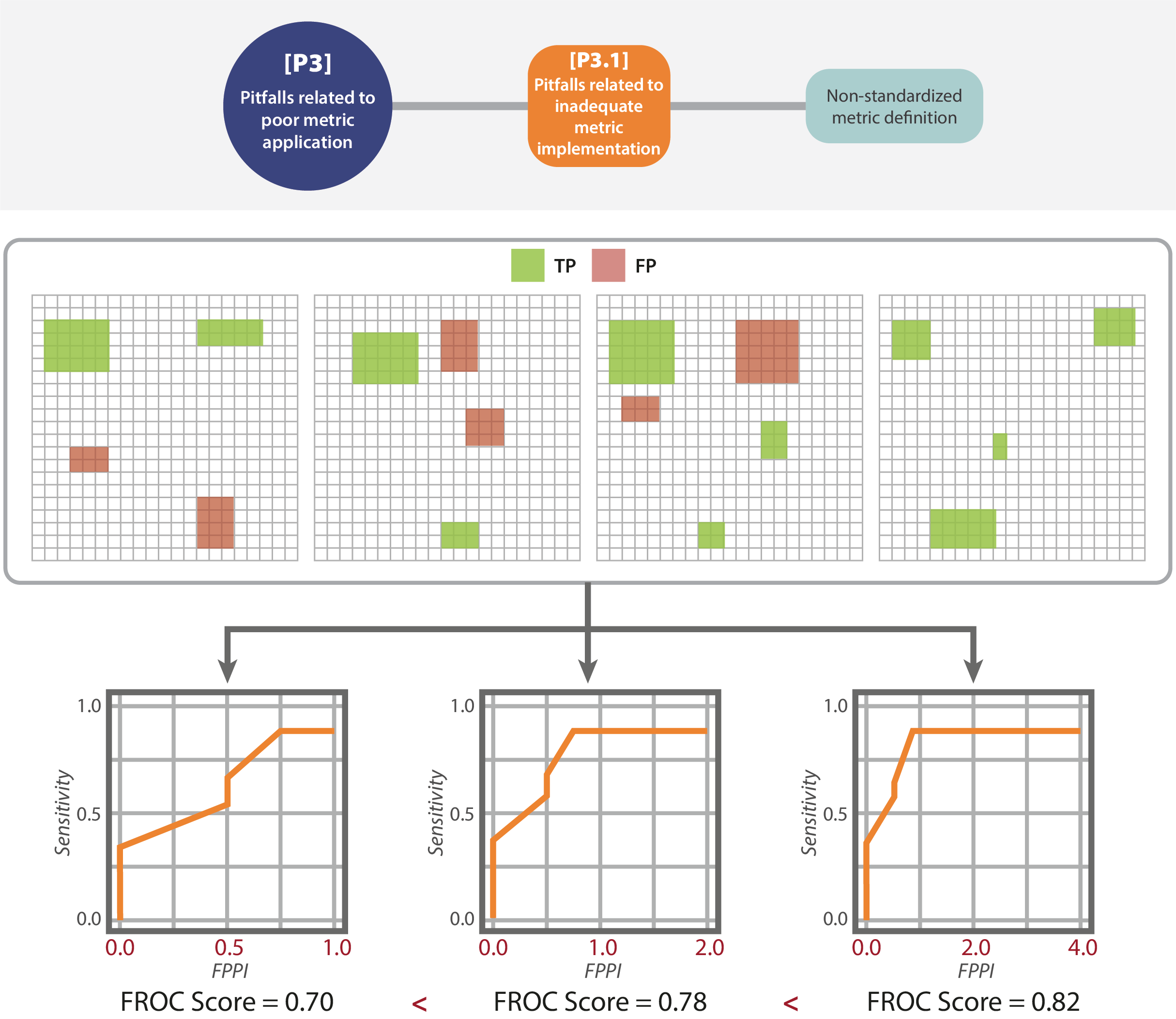}
    \caption{Effect of defining different ranges for the \acf{FPPI} (which are unbounded to the top) used to draw the \acf{FROC} curve for the same prediction (top). The resulting \ac{FROC} Scores differ for different boundaries of the x-axis used for the \ac{FPPI} ([0, 1], [0, 2] and [0, 4]). Publications make use of different ranges for the x-axis, complicating comparison between works.}
    \label{fig:FROC-no-standard}
\end{tcolorbox}
\end{figure}

\begin{figure}[H]
\begin{tcolorbox}[title= Common metrics suffer from discretization issues, colback=white]
    \centering
    \includegraphics[width=0.8\linewidth]{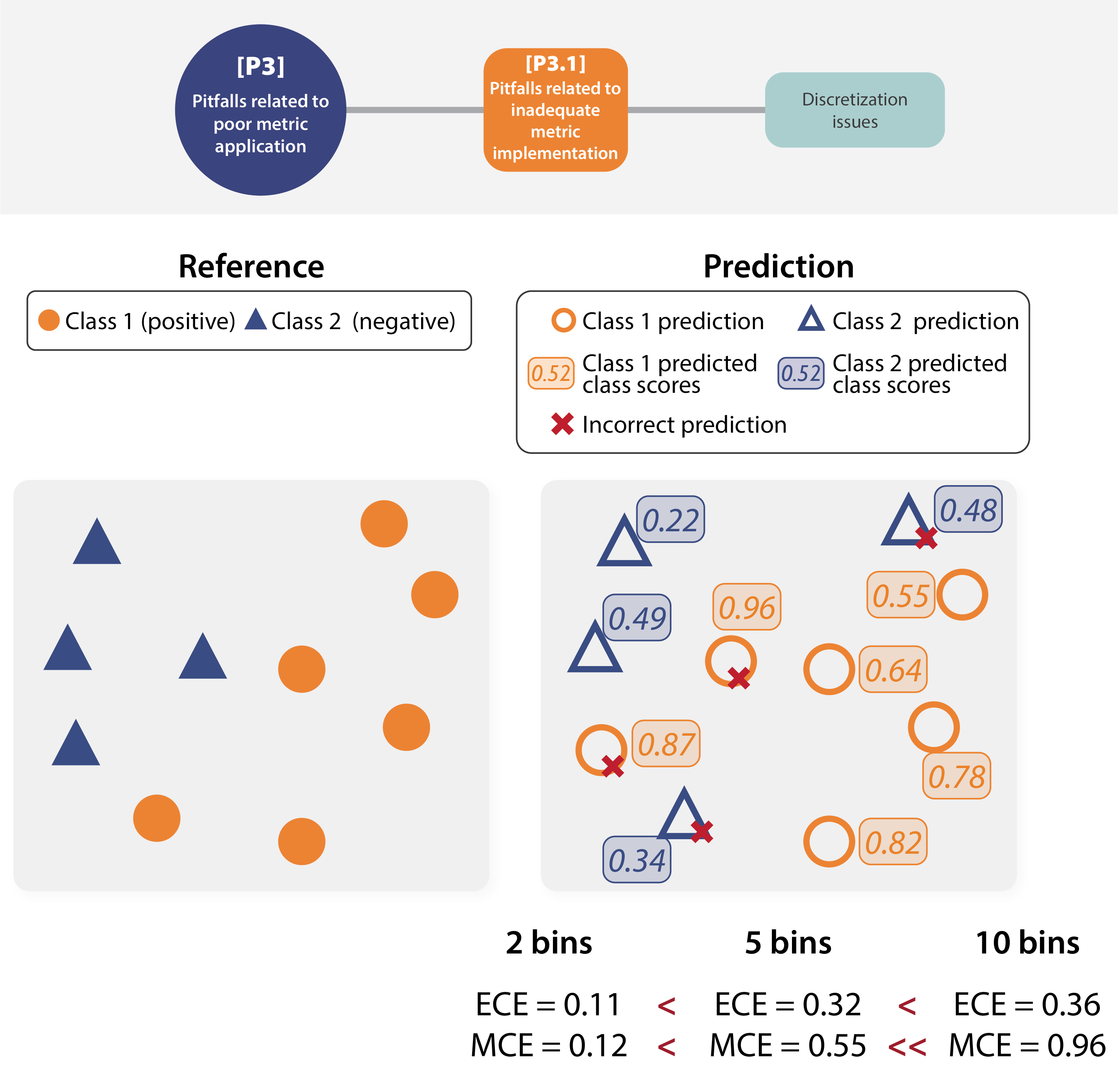}
    \caption{Effect of choosing different bins for calculating the \acf{ECE} and \acf{MCE}. Three different strategies are chosen for the binning of the interval [0, 1] of the predicted class scores of the \textit{Prediction}. The resulting metric scores are substantially affected by the number of bins \citep{guoCalibrationModernNeural2017}.}
    \label{fig:ece-mce-bins}
\end{tcolorbox}
\end{figure}

\begin{figure}[H]
\begin{tcolorbox}[title= Choice of hyperparameters may have largely impact metric scores, colback=white]
    \centering
    \includegraphics[width=1\linewidth]{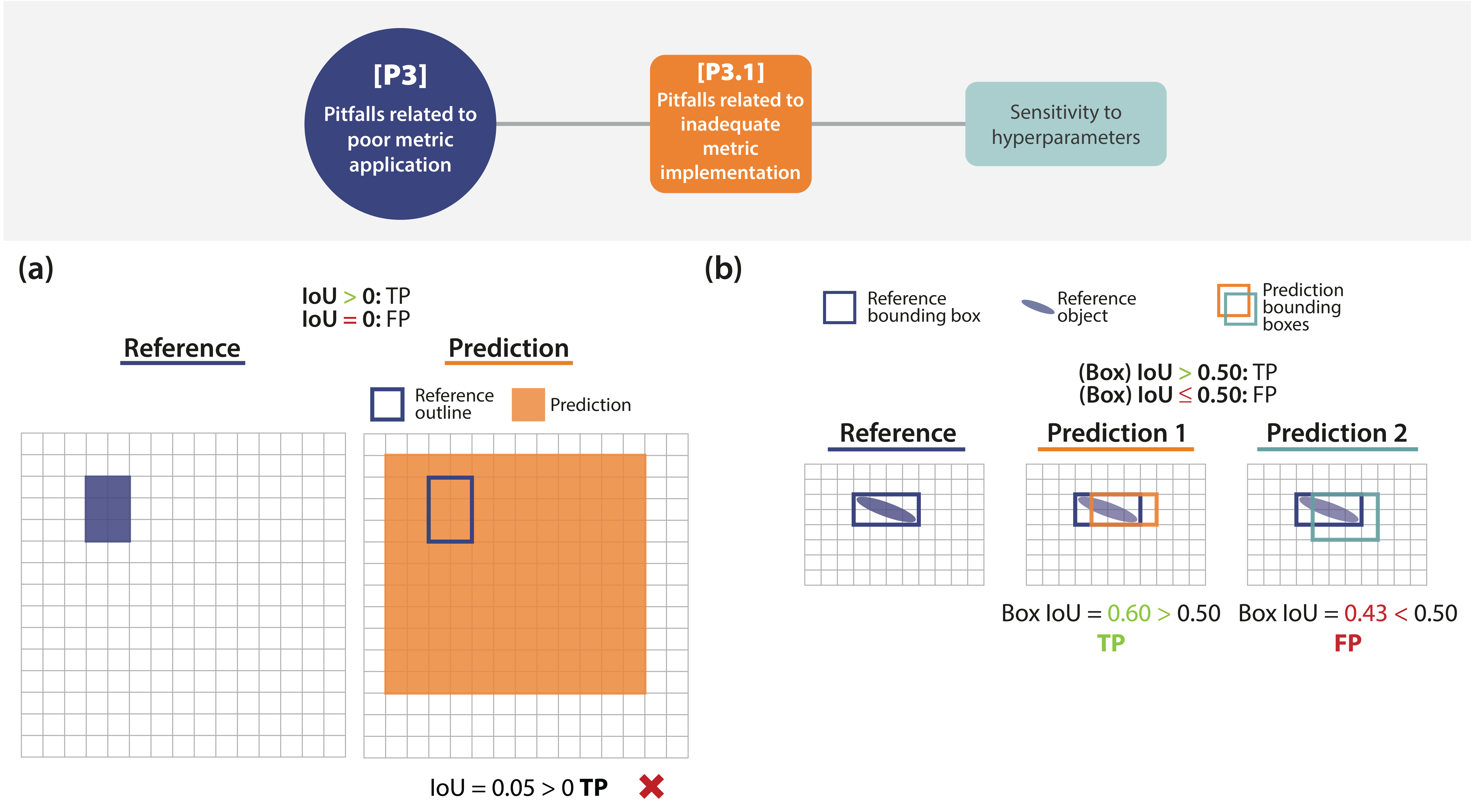}
    \caption{Effect of the \acf{IoU} threshold on the localization (here Box IoU). \textbf{(a)} When defining a \acf{TP} by a very loose \ac{IoU} > 0, the resulting localizations may be deceived by very large predictions. \textbf{(b)} On the other hand, a strict \ac{IoU} criterion may be problematic when the bounding box does not approximate the target structure shapes well. Although \textit{Predictions 1} and \textit{2} are very similar (differing in one pixel in one dimension), only \textit{Prediction 1} is a \ac{TP} because the number of bounding box pixels increases quadratically with the size of diagonal narrow structures. Further abbreviation: \acf{FP}.}
    \label{fig:sensitivity-hyperparam}
\end{tcolorbox}
\end{figure}

\begin{figure}[H]
\begin{tcolorbox}[title= Per-class tuning of the decision threshold yields misleading results, colback=white]
    \centering
    \includegraphics[width=0.75\linewidth]{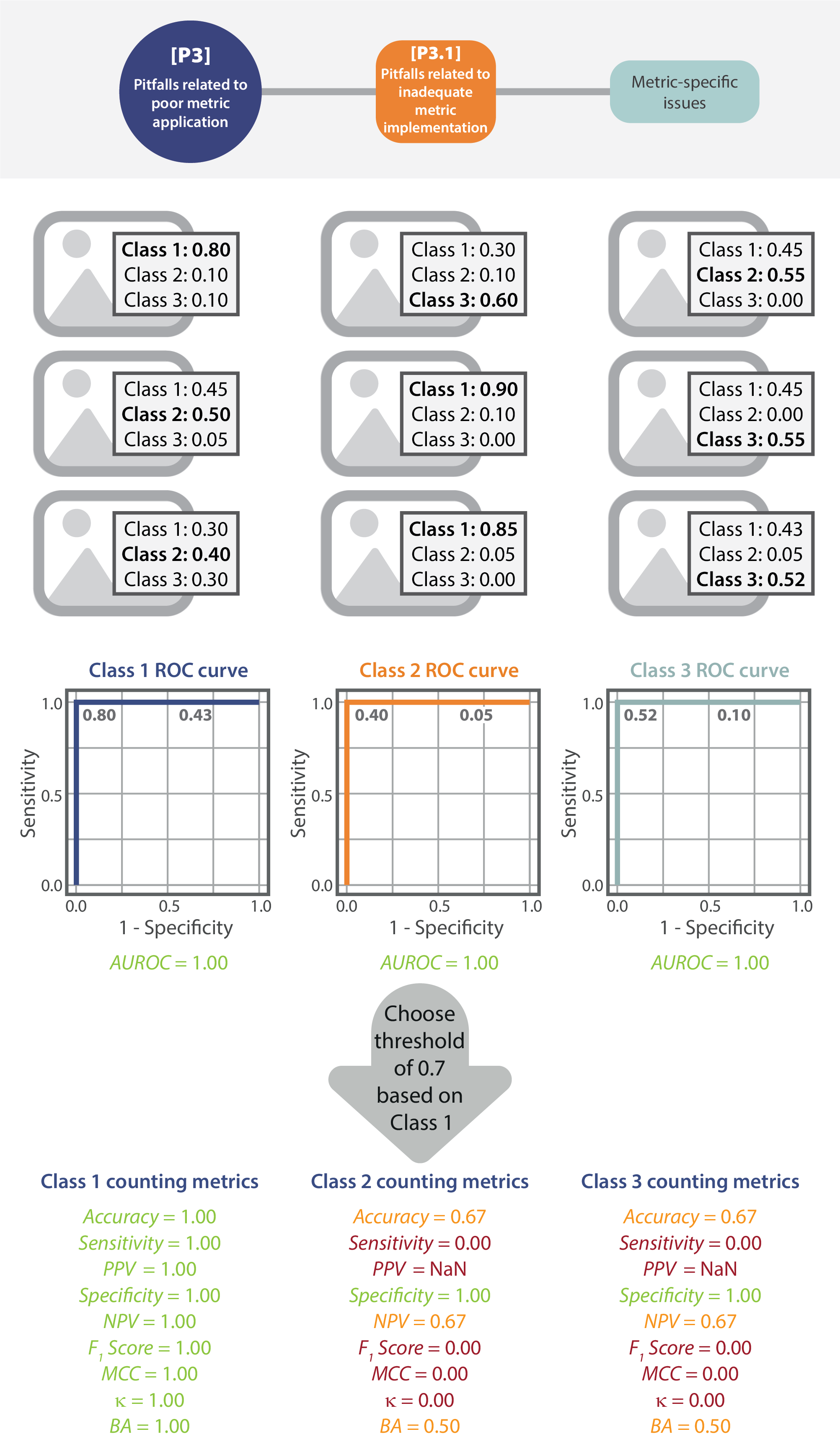}
    \caption{Effect of the determination of a global threshold for all classes based on a single class. In a data set of three classes and nine images, the \acf{AUROC} score is 1.0 for every class. In practice, however, a global decision threshold needs to be set in multi-class problems, which typically renders substantially worse results. Here, the optimal threshold for \textit{Class 1} yields poor results for \textit{Classes 2} and \textit{3} (see e.g., \citep{krause2018grader, de2018clinically}). Used abbreviations: \acf{PPV}, \acf{NPV}, \acf{MCC}, Cohen's Kappa $\kappa$, and \acf{BA}.}
    \label{fig:auroc-classes-threshold}
\end{tcolorbox}
\end{figure}

The implementation of metrics is, unfortunately, not standardized. While some metrics are straightforward to implement, others require more advanced techniques and offer a variety of implementation possibilities. Sources of metric implementation pitfalls include:

\begin{itemize}
    \item Non-standardized definitions (Figs.~\ref{fig:pitfalls-p3}a and~\ref{fig:FROC-no-standard})
    \item Discretization issues (Fig.~\ref{fig:ece-mce-bins})
    \item Sensitivity to hyperparameters (Fig.~\ref{fig:sensitivity-hyperparam})
    \item Metric-specific issues (Fig.~\ref{fig:auroc-classes-threshold})
\end{itemize}
\newpage

\subsubsection{Pitfalls related to inadequate metric aggregation}
\label{sec:pitf:poor-application-aggregation}
\begin{figure}[H]
\begin{tcolorbox}[title= Standard aggregation schemes disregard hierarchical class structures, colback=white]
    \centering
    \includegraphics[width=1\linewidth]{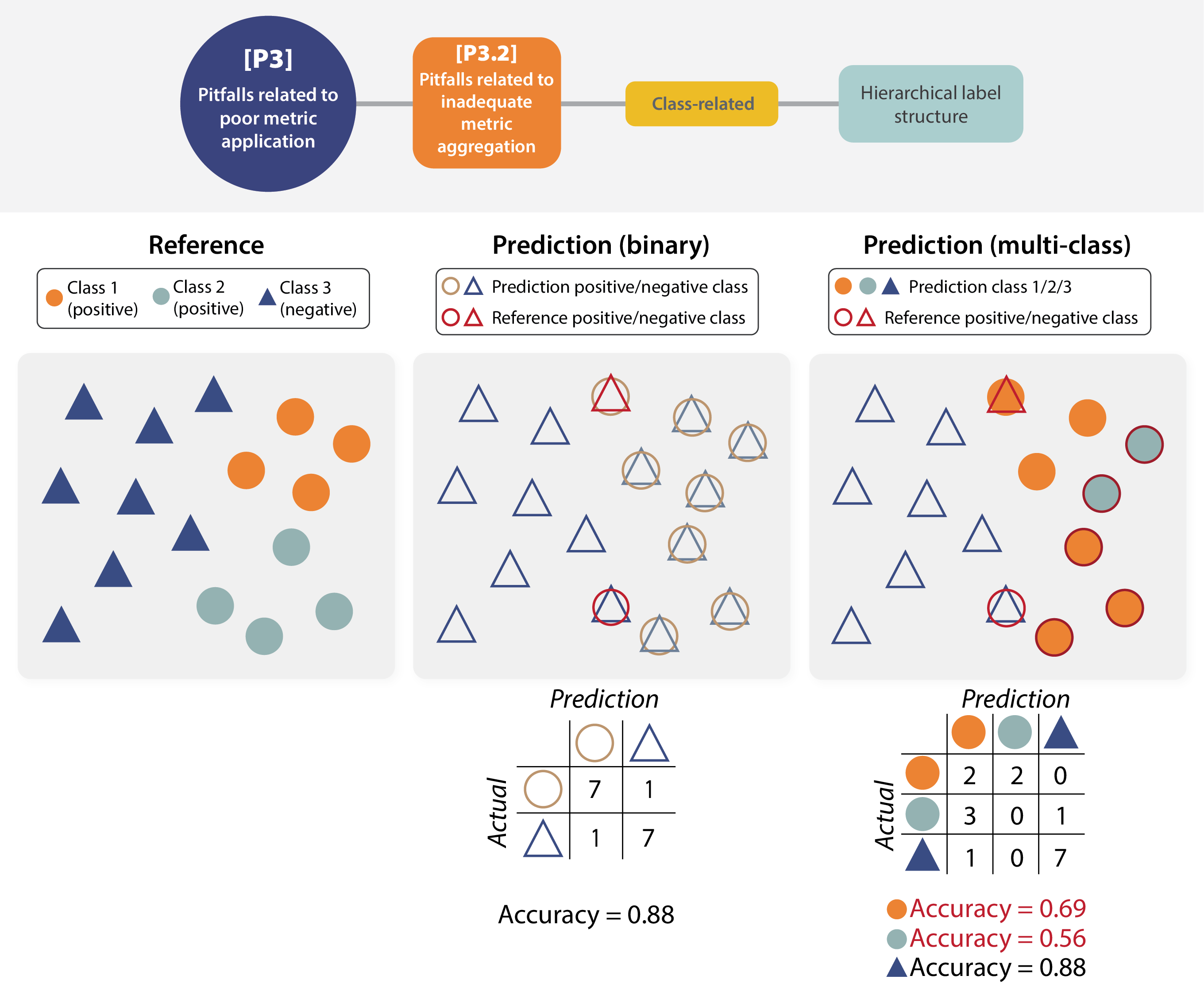}
    \caption{Classes in categorical classification may be hierarchically structured, for example in the form of multiple positive classes and one negative class. The phrasing of the problem as binary \textit{vs.} multi-class hugely affects the validation result. Binary classification (middle), differentiating triangles from circles, yields a good Accuracy, while per-class validation yields a poor score because the two circle classes cannot be distinguished well. Incorrect predictions are overlaid by a red shape of the correct reference class.}
    \label{fig:multi-class}
\end{tcolorbox}
\end{figure}

\begin{figure}[H]
\begin{tcolorbox}[title= Lack of per-class validation conceals important information, colback=white]
    \centering
    \includegraphics[width=1\linewidth]{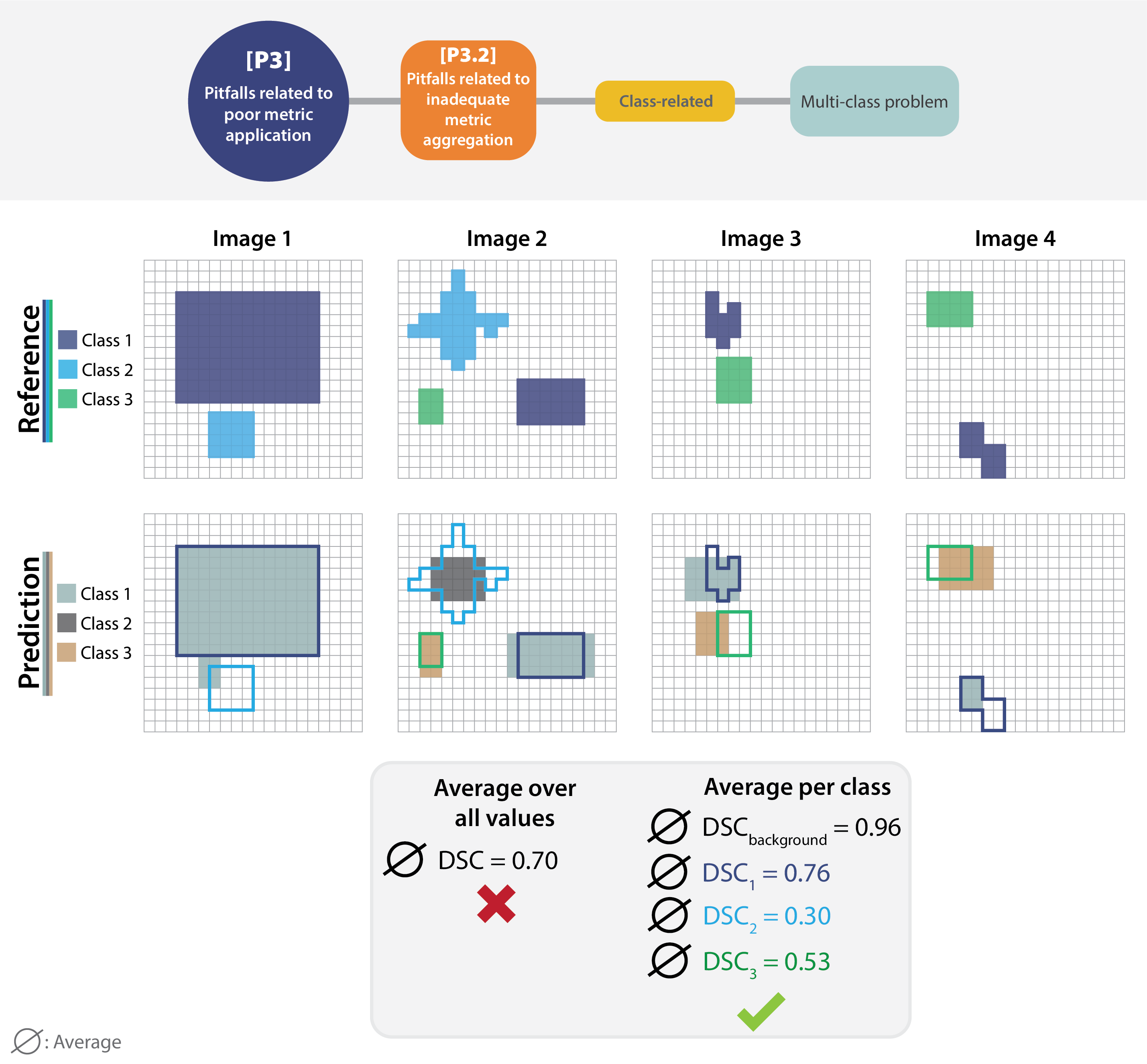}
    \caption{Effect of ignoring the presence of multiple classes when aggregating metric values (here: using the mean). The overall average of all \acf{DSC} scores for the four images is 0.7. Averaging per class reveals a very low performance for \textit{Classes 2} and \textit{3}. $\varnothing$ refers to the average \ac{DSC} values.}
    \label{fig:aggr-per-class}
\end{tcolorbox}
\end{figure}

When aggregating metric values over multiple cases (data points), the method of metric aggregation should be clearly defined and reported including details for example on the aggregation operator (e.g., mean or median) and missing value handling. In addition, special care should be taken when aggregating across classes or different hierarchy levels. Pitfalls can be further subdivided into \textit{class-related} and \textit{data set-related} pitfalls. In the following, we present pitfalls stemming from the following sources:

\textbf{Class-related pitfalls}
\begin{itemize}
    \item Hierarchical label structure (Fig.~\ref{fig:multi-class})
    \item Multi-class problem (Fig.~\ref{fig:aggr-per-class})
\end{itemize}

\begin{figure}[H]
\begin{tcolorbox}[title= Inter-class dependencies are concealed in standard aggregation schemes, colback=white]
    \centering
    \includegraphics[width=0.6\linewidth]{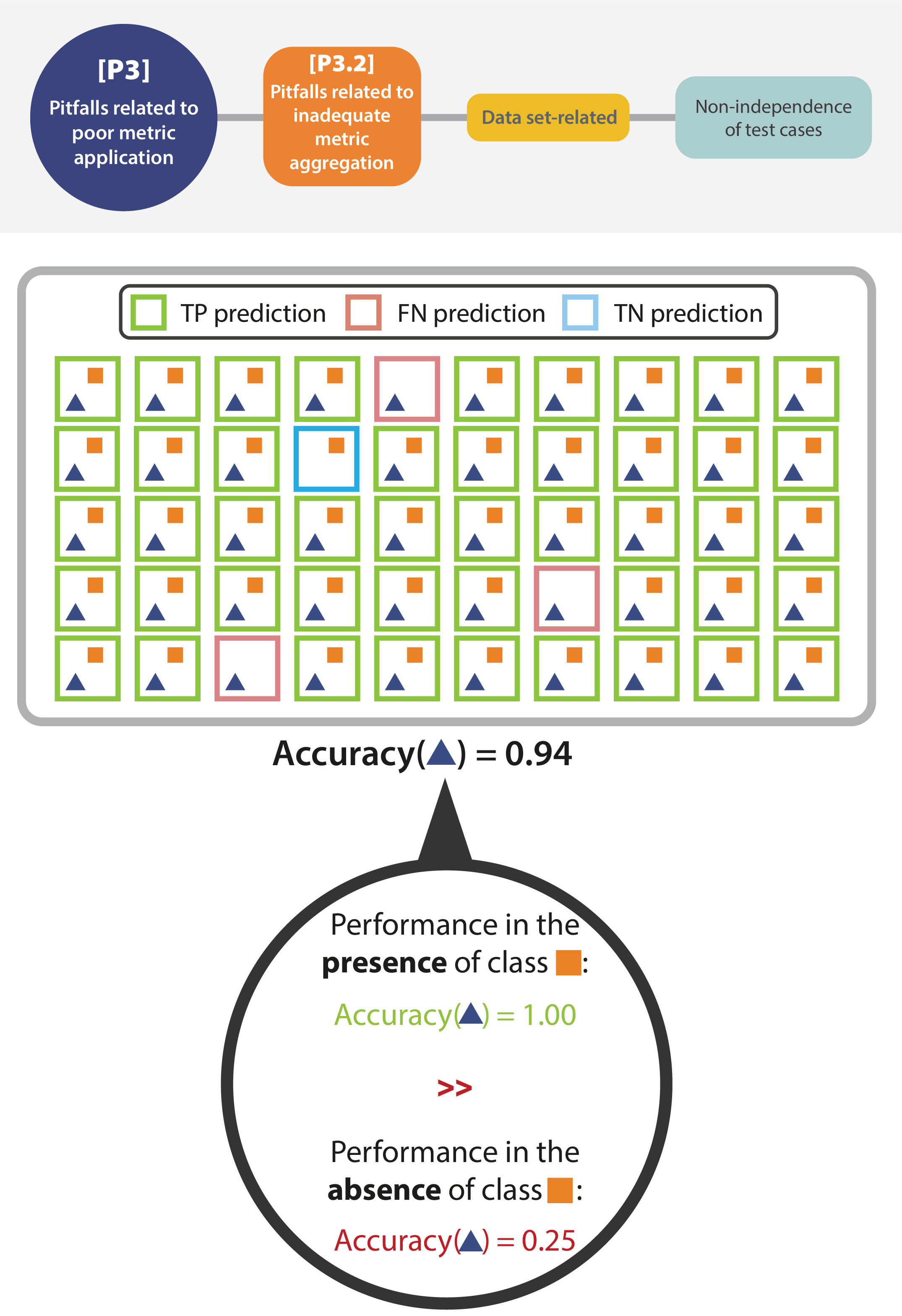}
    \caption{Effect of interdependencies between classes. A prediction may show a near-perfect Accuracy score of 0.94 for the dark blue triangle as it frequently appears in conjunction with the orange square. By calculating the Accuracy in the \textit{presence} and \textit{absence} of the orange square class, it can be seen that the algorithm only works well in the presence of the orange square class.}
    \label{fig:multi-class-interdependencies}
\end{tcolorbox}
\end{figure}

\begin{figure}[H]
\begin{tcolorbox}[title= Lack of stratification conceal biases, colback=white]
    \centering
    \includegraphics[width=0.6\linewidth]{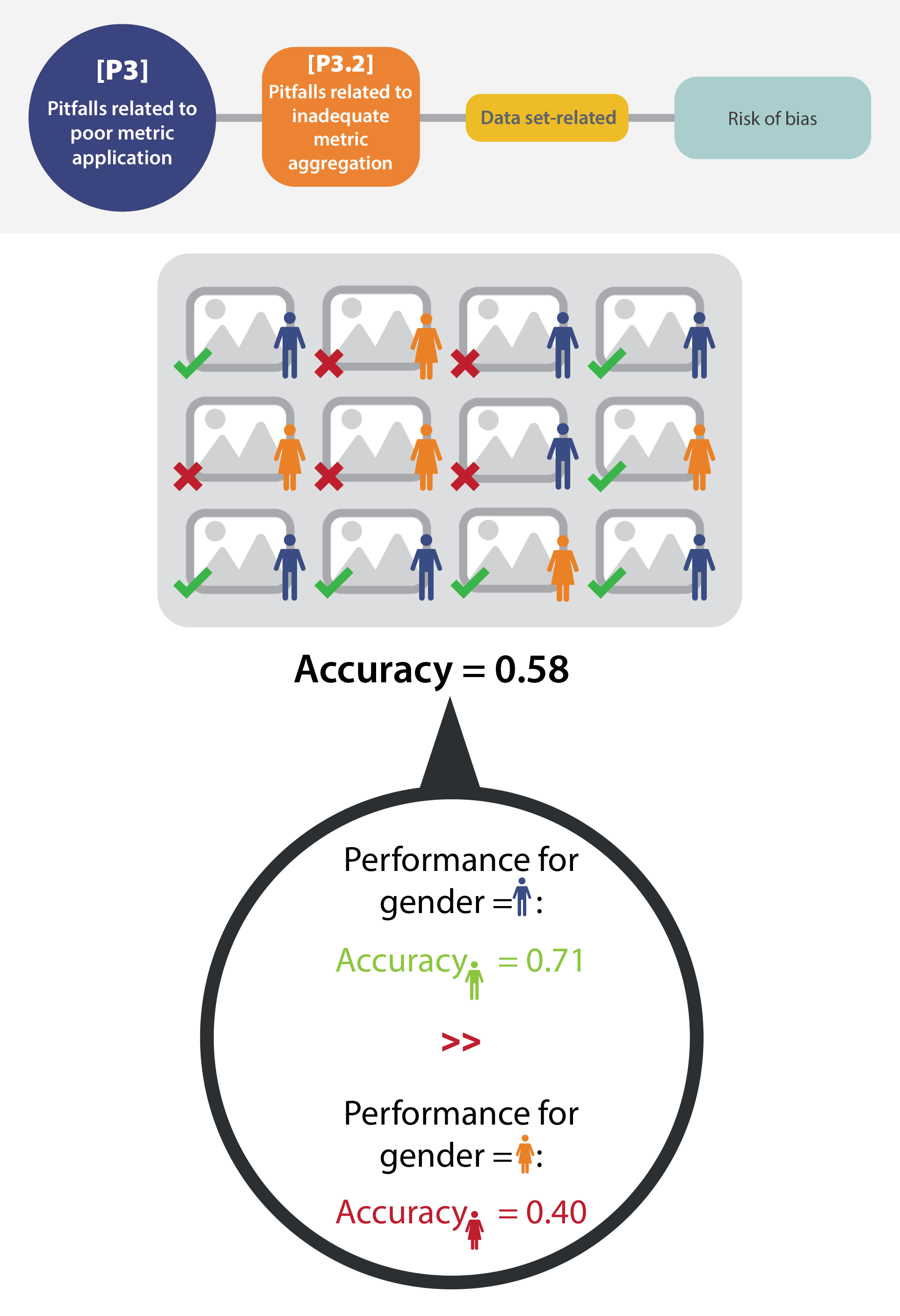}
    \caption{Effect of disregarding relevant meta-information (here: gender). When ignoring the available meta-information of the patient's gender per image, any metric (here: \textit{Accuracy}) fails to reveal that the algorithm performs much better for men compared to women. In this example, correct predictions are marked by a green check mark, incorrect predictions by a red cross.}
    \label{fig:stratification-gender}
\end{tcolorbox}
\end{figure}

\begin{figure}[H]
\begin{tcolorbox}[title= Lack of missing data handling strategy yields misleading results, colback=white]
    \centering
    \includegraphics[width=0.9\linewidth]{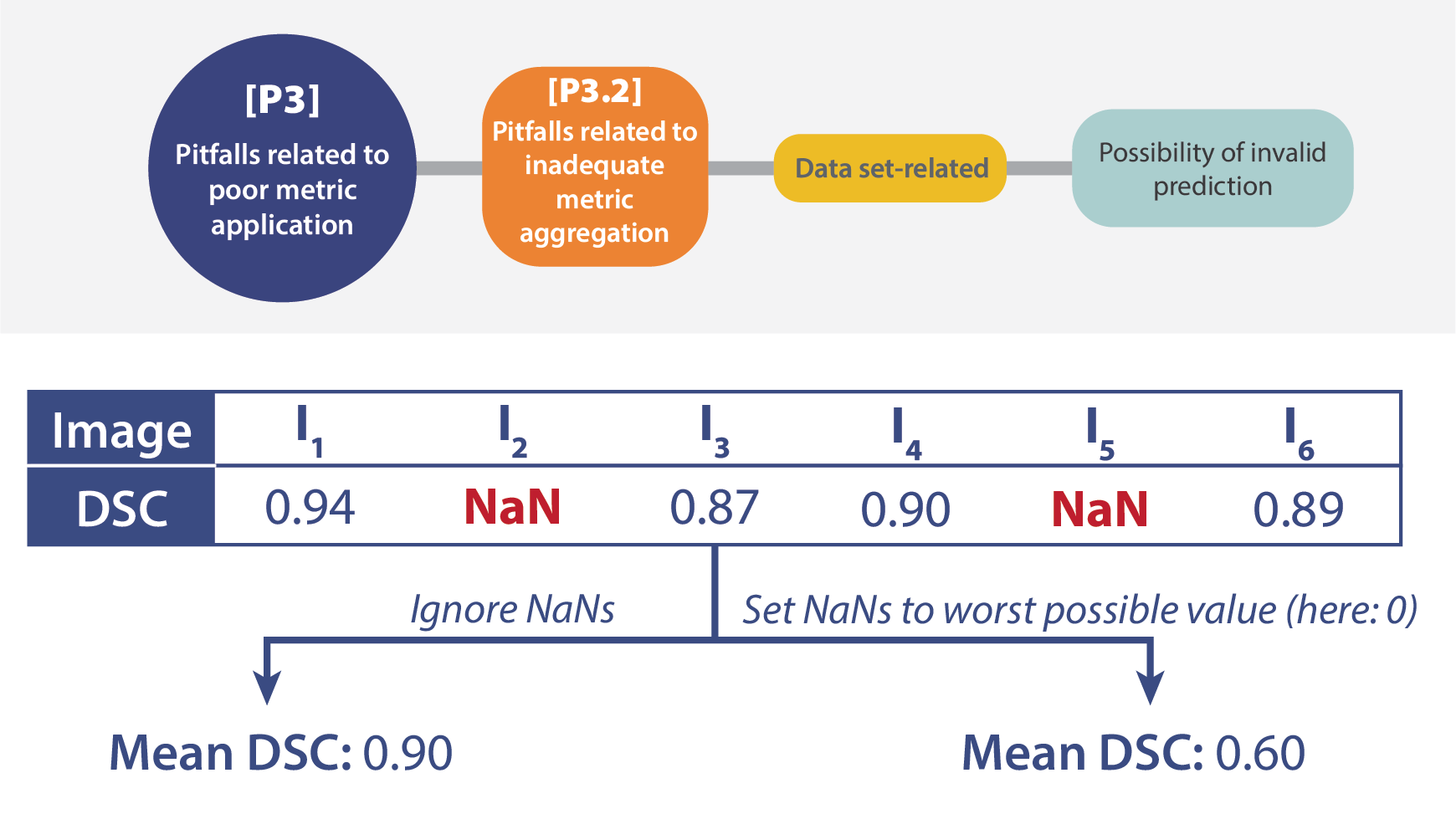}
    \caption{Effect of invalid predictions (missing values) when aggregating metric values. In this example, ignoring missing values leads to a substantially higher \acf{DSC}) compared to setting missing values to the worst possible value (here: 0).}
    \label{fig:missings}
\end{tcolorbox}
\end{figure}

\textbf{Data set-related pitfalls}
\begin{itemize}
    \item Non-independence of test cases (Figs.~\ref{fig:pitfalls-p3}b and~\ref{fig:multi-class-interdependencies})
    \item Risk of bias (Fig.~\ref{fig:stratification-gender})
    \item Possibility of invalid prediction (Fig.~\ref{fig:missings})
\end{itemize}

\newpage

\newpage
\subsubsection{Pitfalls related to inadequate ranking scheme}
\label{sec:pitf:poor-application-ranking}
\begin{figure}[H]
\begin{tcolorbox}[title= Related metrics may yield identical rankings, colback=white]
    \centering
    \includegraphics[width=1\linewidth]{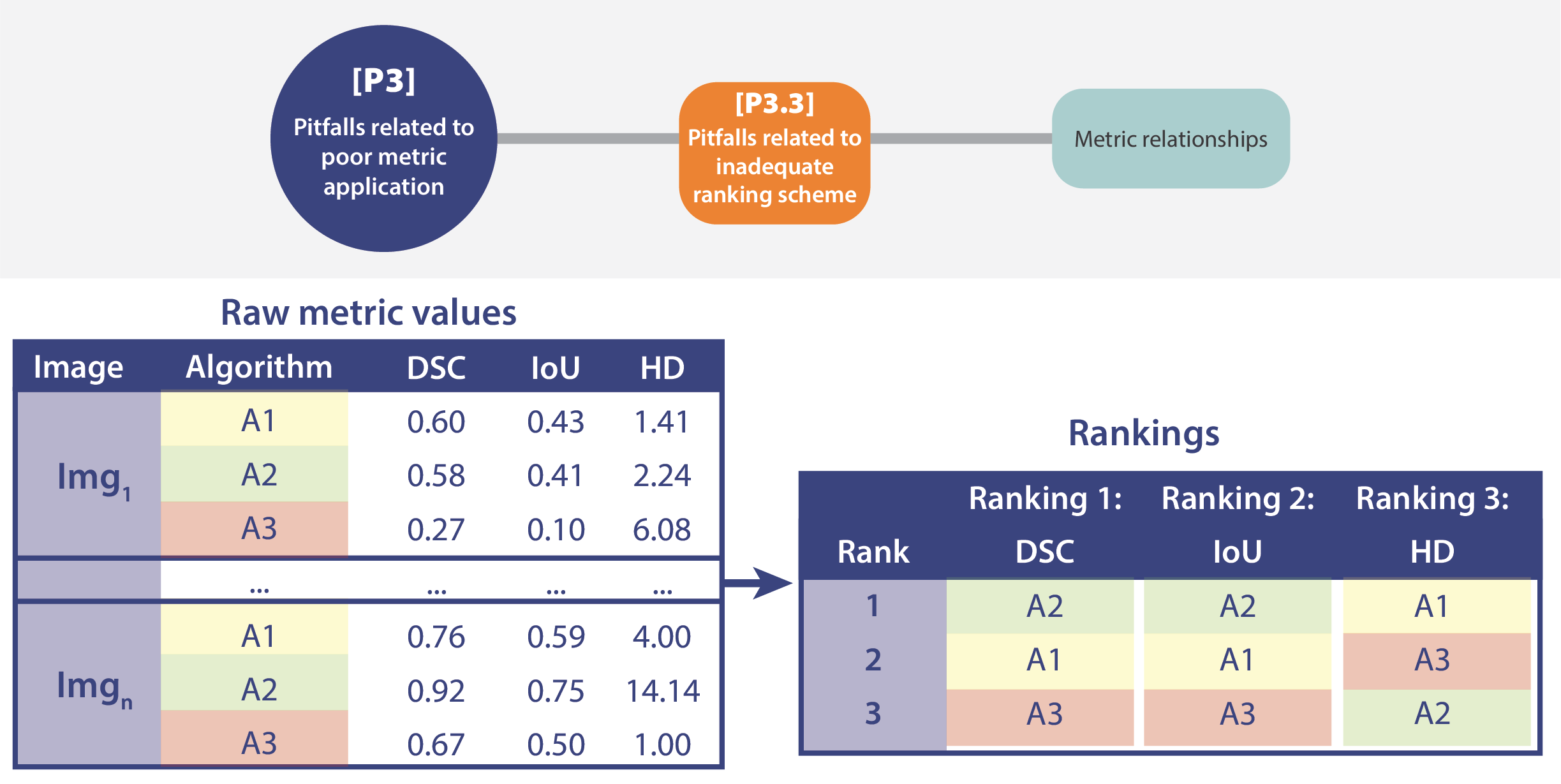}
    \caption{Effect of using mathematically closely related metrics. The \acf{DSC} and \acf{IoU} typically lead to the same ranking, whereas metrics from different families (here: \acf{HD}) may lead to substantially different rankings \citep{taha2014formal, taha2015metrics}. Combining metrics that are related will not provide additional information for a ranking, and having multiple metrics measuring the same properties may overrule rankings of other properties (here: \ac{HD}).}
    \label{fig:combination}
\end{tcolorbox}
\end{figure}

\begin{figure}[H]
\begin{tcolorbox}[title= Ranking tables do not reflect ranking uncertainty, colback=white]
    \centering
    \includegraphics[width=0.8\linewidth]{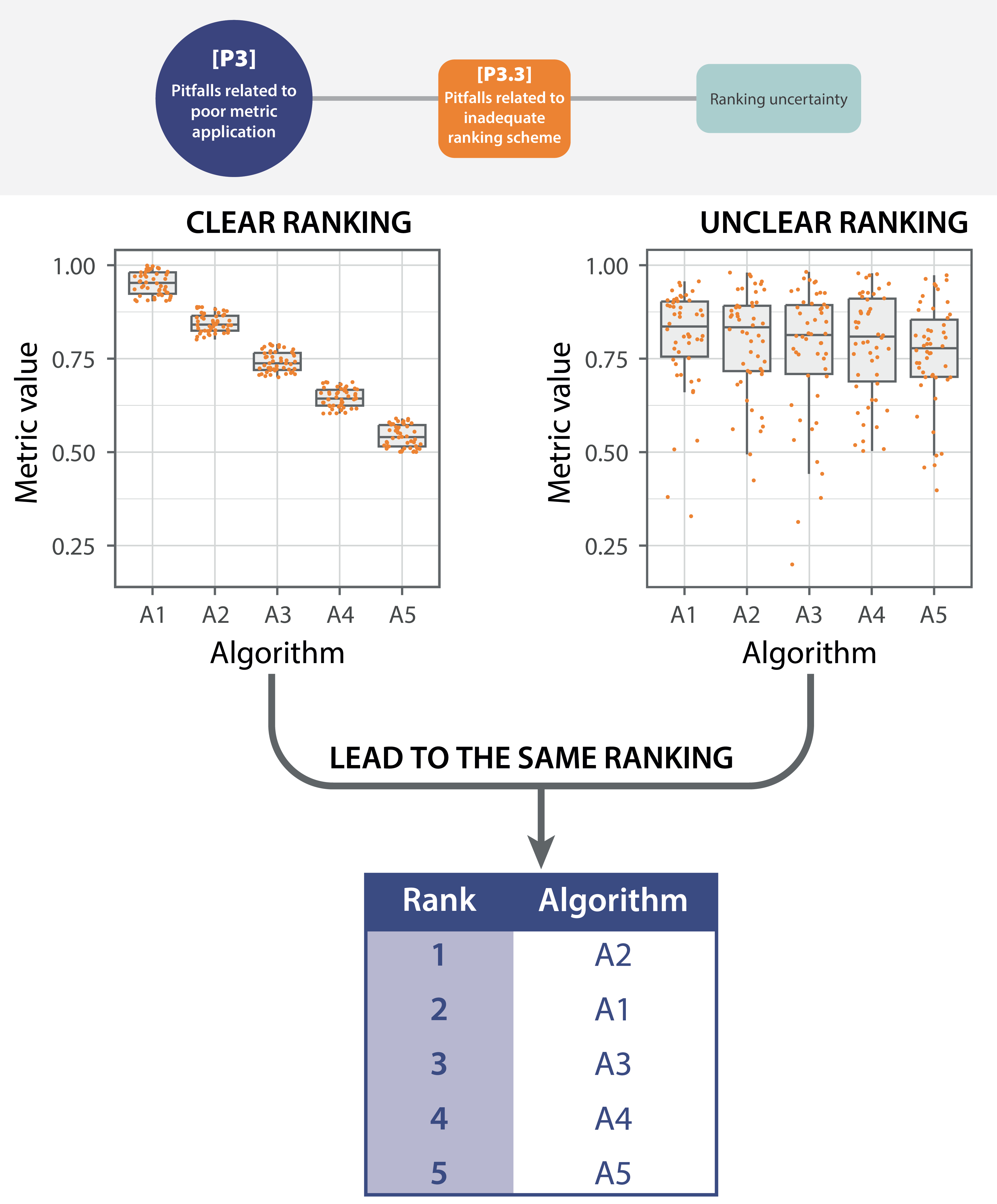}
    \caption{Effect of ranking uncertainty. The results of two benchmarking experiments with five algorithms $A1$-$A5$ differ substantially, as shown by the boxplots of the metric values for every algorithm. While the left situation introduces a clear ranking visible from the boxplots, the right use case is not clear as performance is very similar across algorithms. However, both situations lead to the same ranking \citep{maier2018rankings,wiesenfarth2021methods}. Thus, solely providing ranking tables conceals information on ranking uncertainty.}
    \label{fig:ranking-uncertainty}
\end{tcolorbox}
\end{figure}

Rankings are often created to compare algorithm performances. In this context, we present pitfalls stemming from the following sources:

\begin{itemize}
    \item Metric relationships (Fig.~\ref{fig:combination})
    \item Ranking uncertainty (Fig.~\ref{fig:ranking-uncertainty})
\end{itemize}




\newpage
\newpage
\subsubsection{Pitfalls related to inadequate metric reporting}
\label{sec:pitf:poor-application-reporting}
\begin{figure}[H]
\begin{tcolorbox}[title= The non-determinism of neural networks effects metric results, colback=white]
    \centering
    \includegraphics[width=0.7\linewidth]{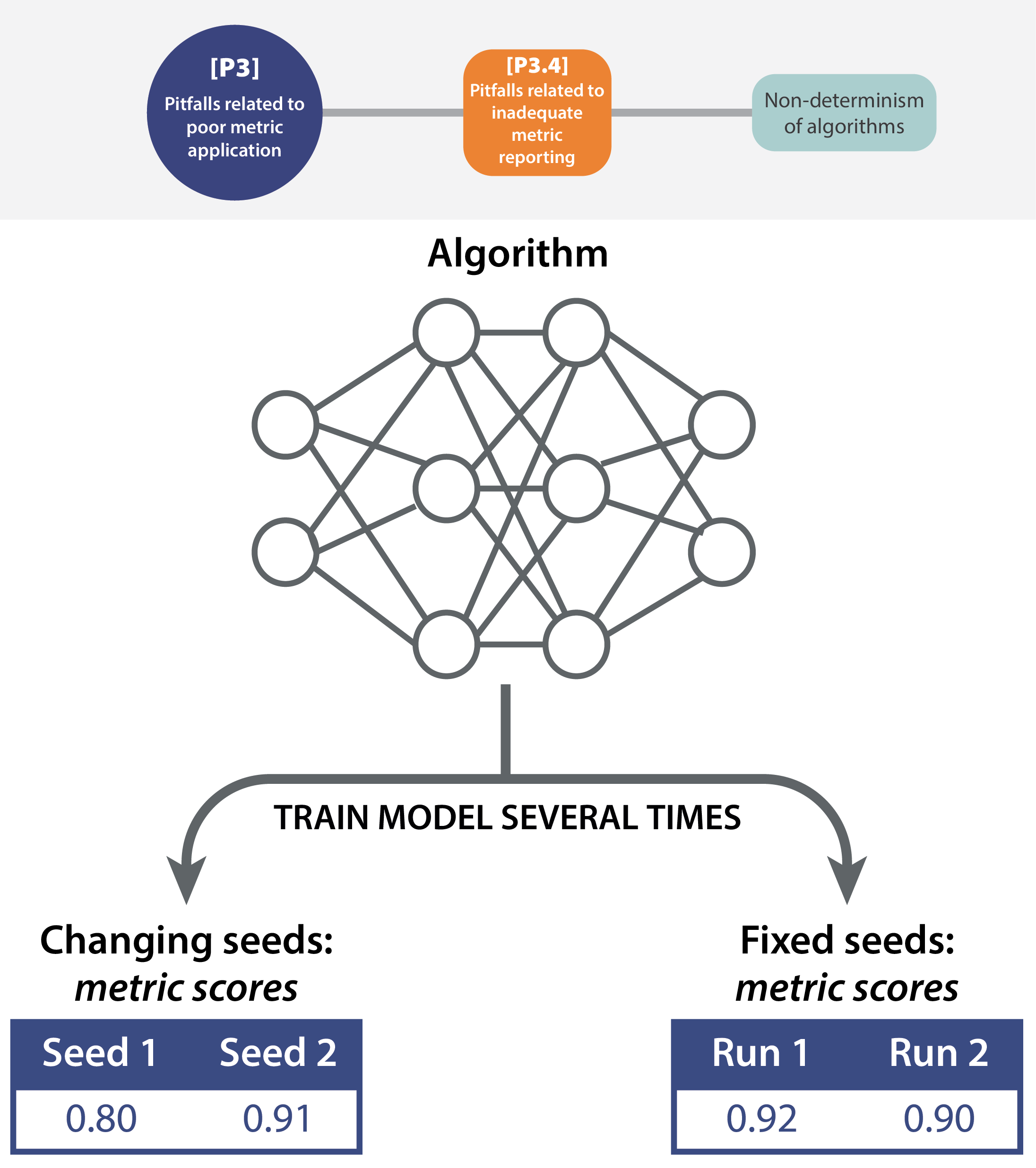}
    \caption[Effect of non-determinism of \acf{AI} algorithms. An algorithm trained under identical conditions may yield different results when changing seeds (left), but also with fixed seeds (right). The latter may, for example, be caused by parallel processes, the order of threads, auto-selection of primitive operations, and other factors. Fixing seeds does not guarantee reproducibility even for the same hardware/software configuration as many software libraries have a degree of randomness on their operations.]{Effect of non-determinism of \acf{AI} algorithms. An algorithm trained under identical conditions may yield different results when changing seeds (left), but also with fixed seeds (right). The latter may, for example, be caused by parallel processes, order of threads, auto-selection of primitive operations, and other factors \cite{pham2020problems}\protect\footnotemark. Fixing seeds does not guarantee reproducibility even for the same hardware/software configuration as many software libraries have a degree of randomness on their operations.}
    \label{fig:non-determinism}
\end{tcolorbox}
\end{figure}

\begin{figure}[H]
\begin{tcolorbox}[title= Common visualization schemes conceal relevant information, colback=white]
    \centering
    \includegraphics[width=0.9\linewidth]{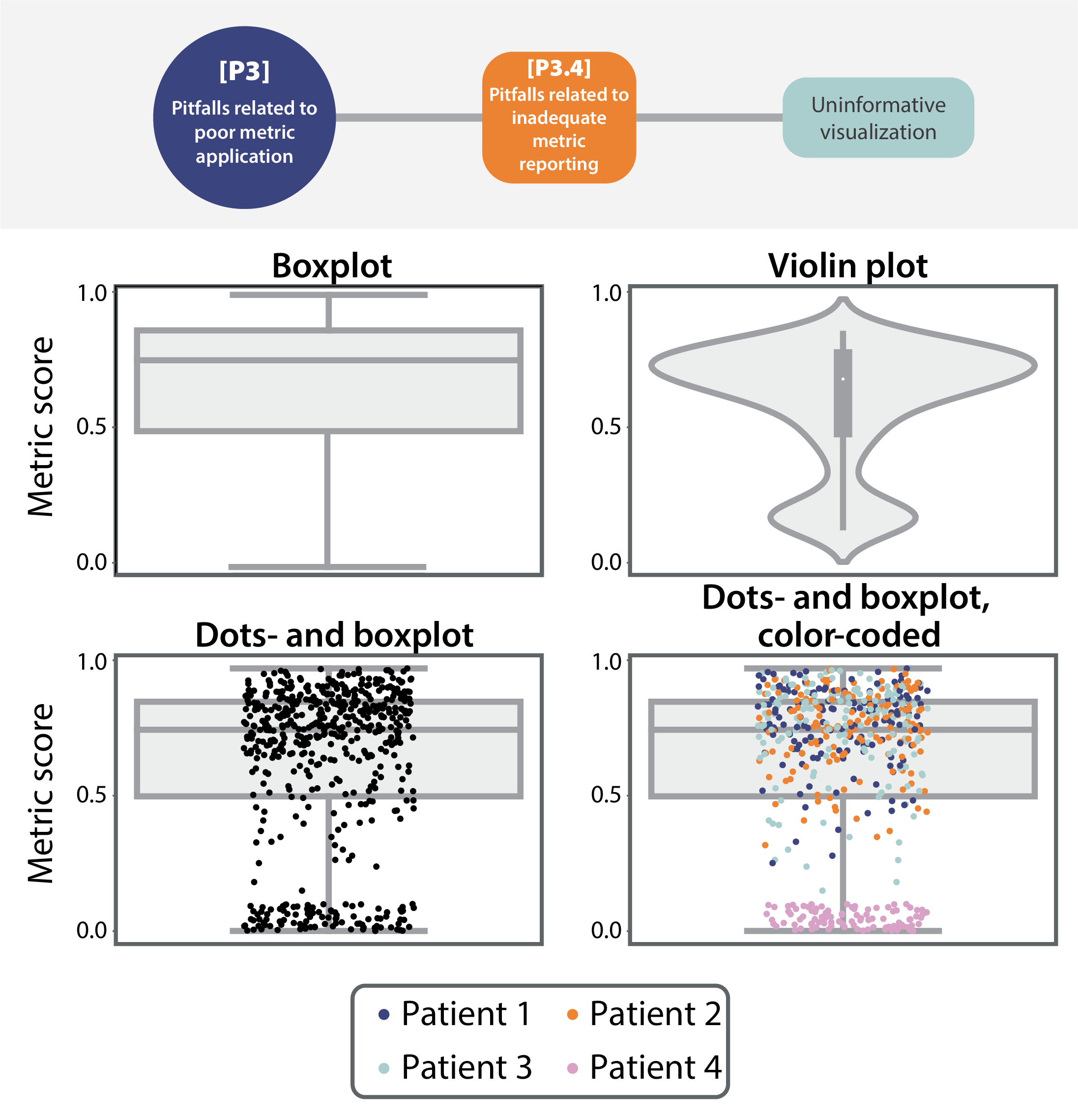}
    \caption{Effect of different visualization types. A single boxplot (top left) does not provide sufficient information about the raw metric value distribution (here: \acf{DSC}). Using a violin plot (top right) or adding the raw metric values as jittered dots on top (bottom left) adds important information. In the case of non-independent validation data, color/shape-coding helps reveal data clusters (bottom right).}
    \label{fig:raw-metric-values-boxplot}
\end{tcolorbox}
\end{figure}

A thorough reporting of metric values and aggregates is important both in terms of transparency and interpretability. However, several pitfalls are to be avoided in this regard. Sources of metric reporting pitfalls include:

\begin{itemize}
    \item Non-determinism of algorithms (Fig.~\ref{fig:non-determinism})
    \item Uninformative visualization (Figs.~\ref{fig:pitfalls-p3}c and~\ref{fig:raw-metric-values-boxplot})
\end{itemize}

\footnotetext{See for example: \url{https://pytorch.org/docs/stable/notes/randomness.html}}


\newpage
\subsubsection{Pitfalls related to inadequate interpretation of metric values}
\label{sec:pitf:poor-application-interpretation}
\begin{figure}[H]
\begin{tcolorbox}[title= Image resolution affects metric scores, colback=white]
    \centering
    \includegraphics[width=1\linewidth]{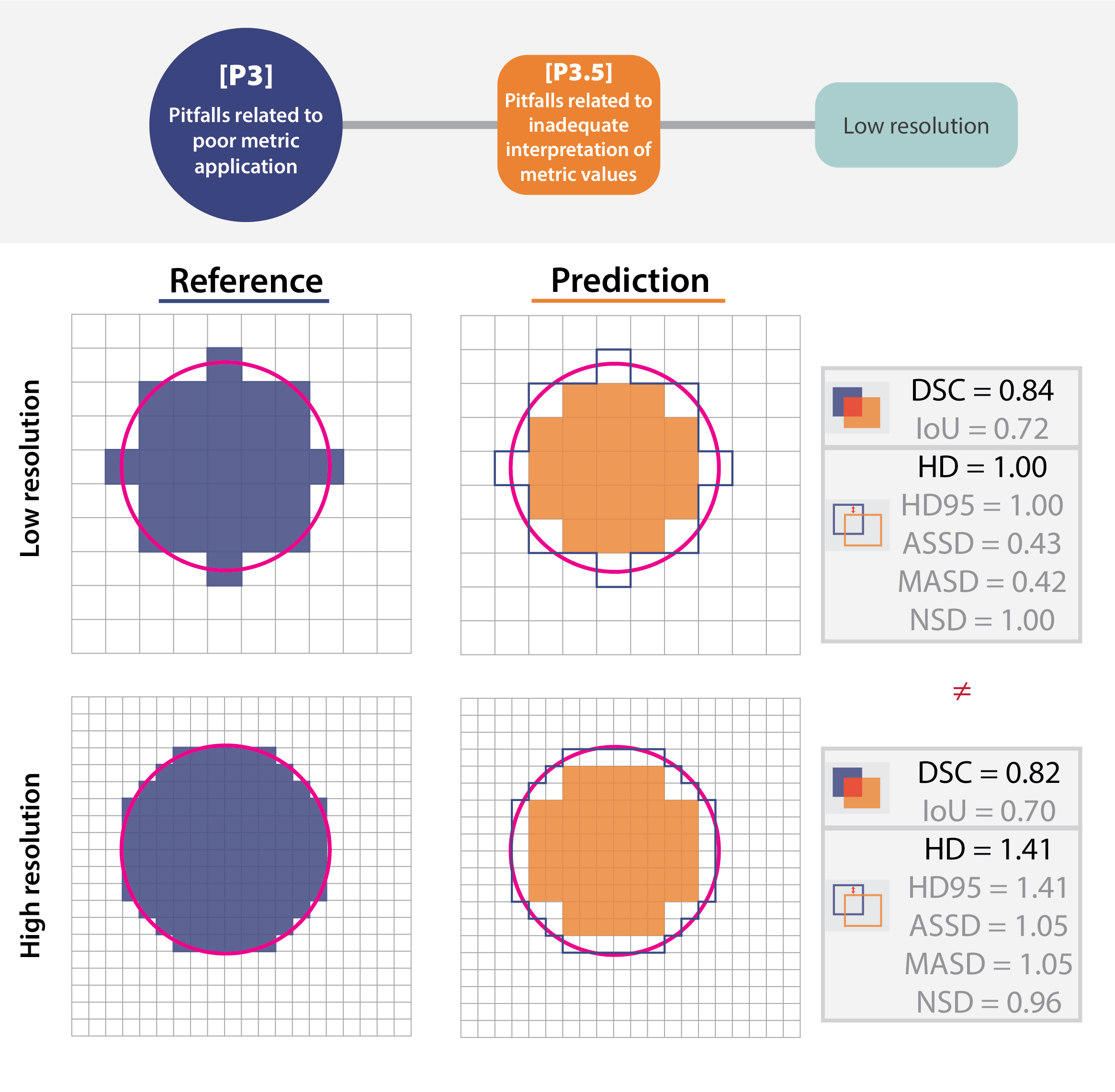}
    \caption{Effect of different grid sizes. Differences in the grid size (resolution) of an image highly influence the image and the reference annotation (dark blue shape (reference) \textit{vs.} pink outline (desired circle shape)), with a prediction of the exact same shape leading to different metric scores. Abbreviations: \acf{DSC}, \acf{IoU}, \acf{HD}, \acf{HD95}, \acf{ASSD}, \acf{MASD}, \acf{NSD}.}
     \label{fig:DSC-grid-size}
\end{tcolorbox}
\end{figure}

\begin{figure}[H]
\begin{tcolorbox}[title= Lower bounds of metrics may not be achievable in practice, colback=white]
    \centering
    \includegraphics[width=0.7\linewidth]{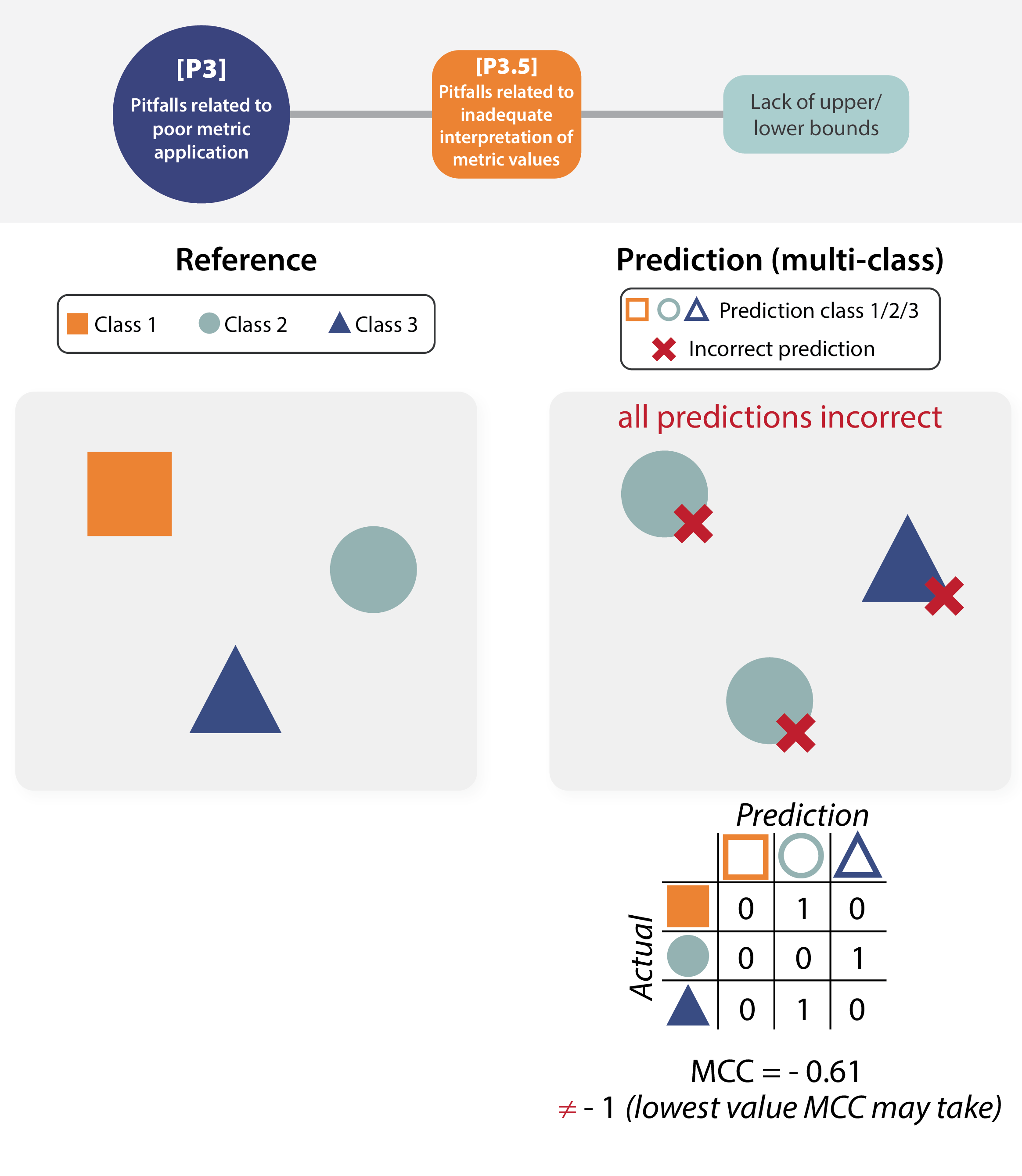}
    \caption{Effect of theoretical bounds that may not be achievable in practice. In this multi-class example, all samples were predicted incorrectly. However, the theoretical lowest value for the \acf{MCC} metric (-1) cannot be achieved in this situation, rendering interpretation difficult.}
    \label{fig:lack-bounds}
\end{tcolorbox}
\end{figure}

\begin{figure}[H]
\begin{tcolorbox}[title= Metric score differences leading to different rankings may be irrelevant, colback=white]
    \centering
    \includegraphics[width=1\linewidth]{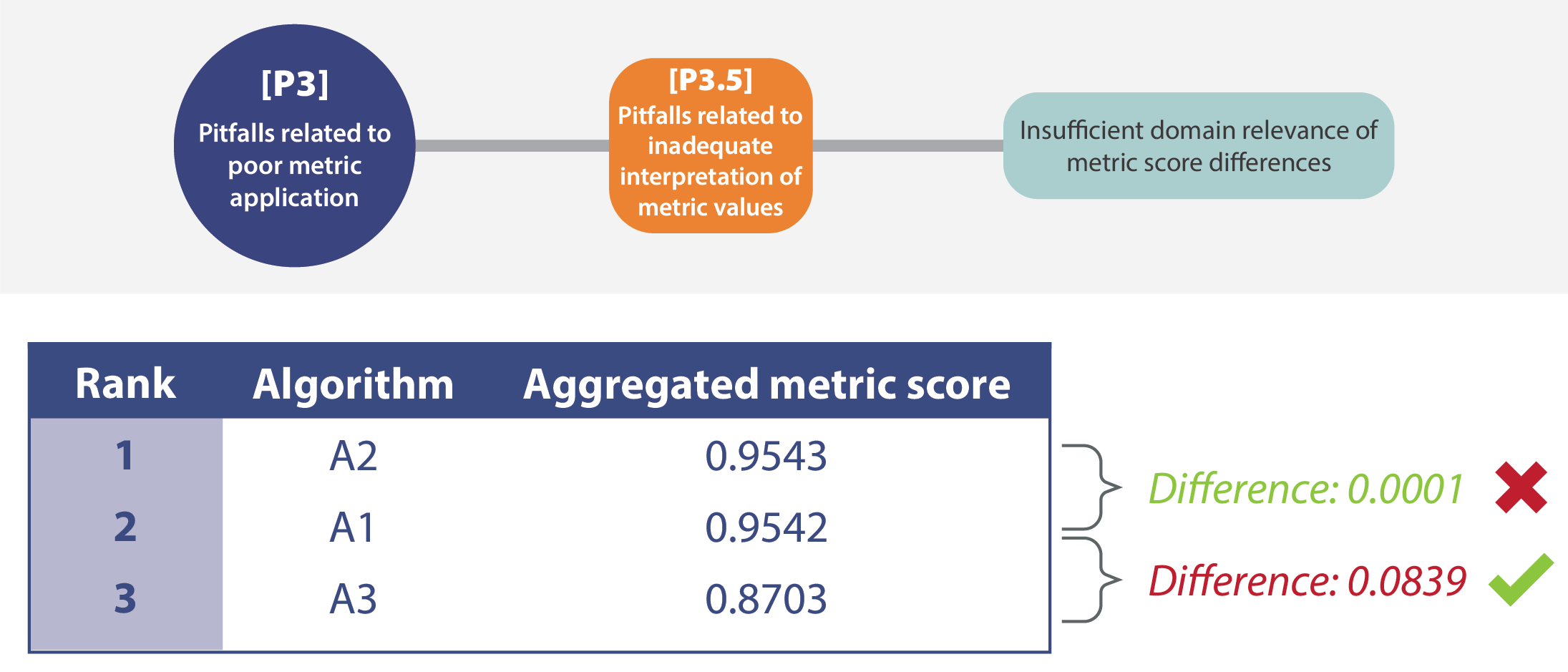}
    \caption{Effect of irrelevant metric score differences in rankings. The difference of the metric score aggregates of algorithms \textit{A1} and \textit{A2} is extremely low and not of biomedical relevance. However, the numerical difference would assign them different ranks.}
    \label{fig:ranking-relevance}
\end{tcolorbox}
\end{figure}

Interpreting metric scores and aggregates is an important step in algorithm performance analysis. However, several pitfalls can arise from interpretation. In the following, we  present pitfalls related to:

\begin{itemize}
    \item Low resolution (Fig.~\ref{fig:DSC-grid-size})
    \item Lack of upper/lower bounds (Fig.~\ref{fig:lack-bounds})
    \item Insufficient domain relevance of metric score differences (Fig.~\ref{fig:ranking-relevance})
\end{itemize}
\newpage

\newpage
\section{Metric Profiles}
\label{app:steckbriefe}
\begin{figure}[H]
    \centering
    \includegraphics[width=0.9\textwidth]{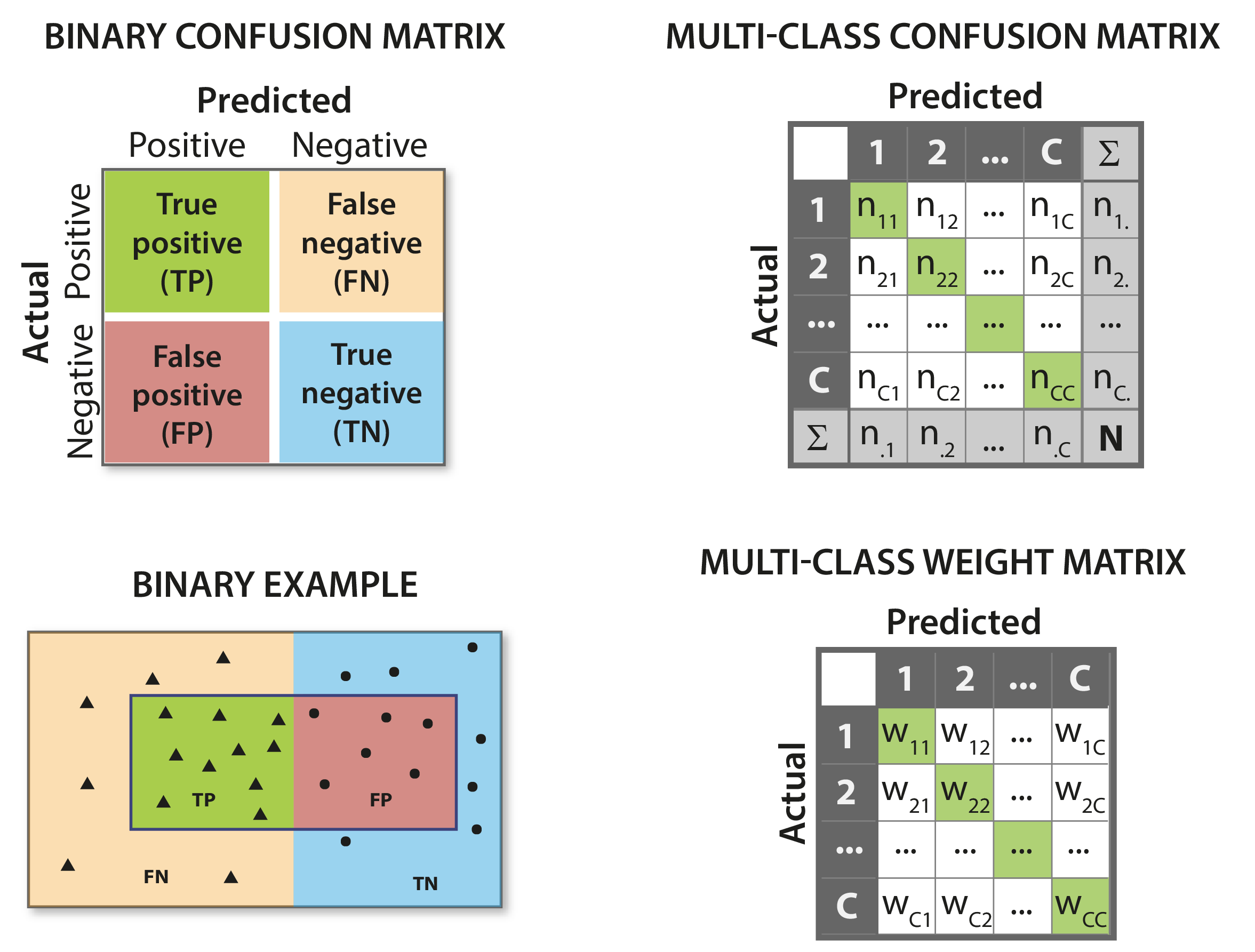}
    \caption{Schematic example of the confusion matrix for two and for $C$ classes. For the latter case, we also present a weight or cost matrix with weights $w_{ij} > 0$ without loss of generality. For the binary confusion matrix, we show an example illustrating the cardinalities for a prediction of triangles and circles.}
    \label{fig:confusion-matrix}
\end{figure}

\begin{figure}[H]
    \centering
    \includegraphics[width=\textwidth]{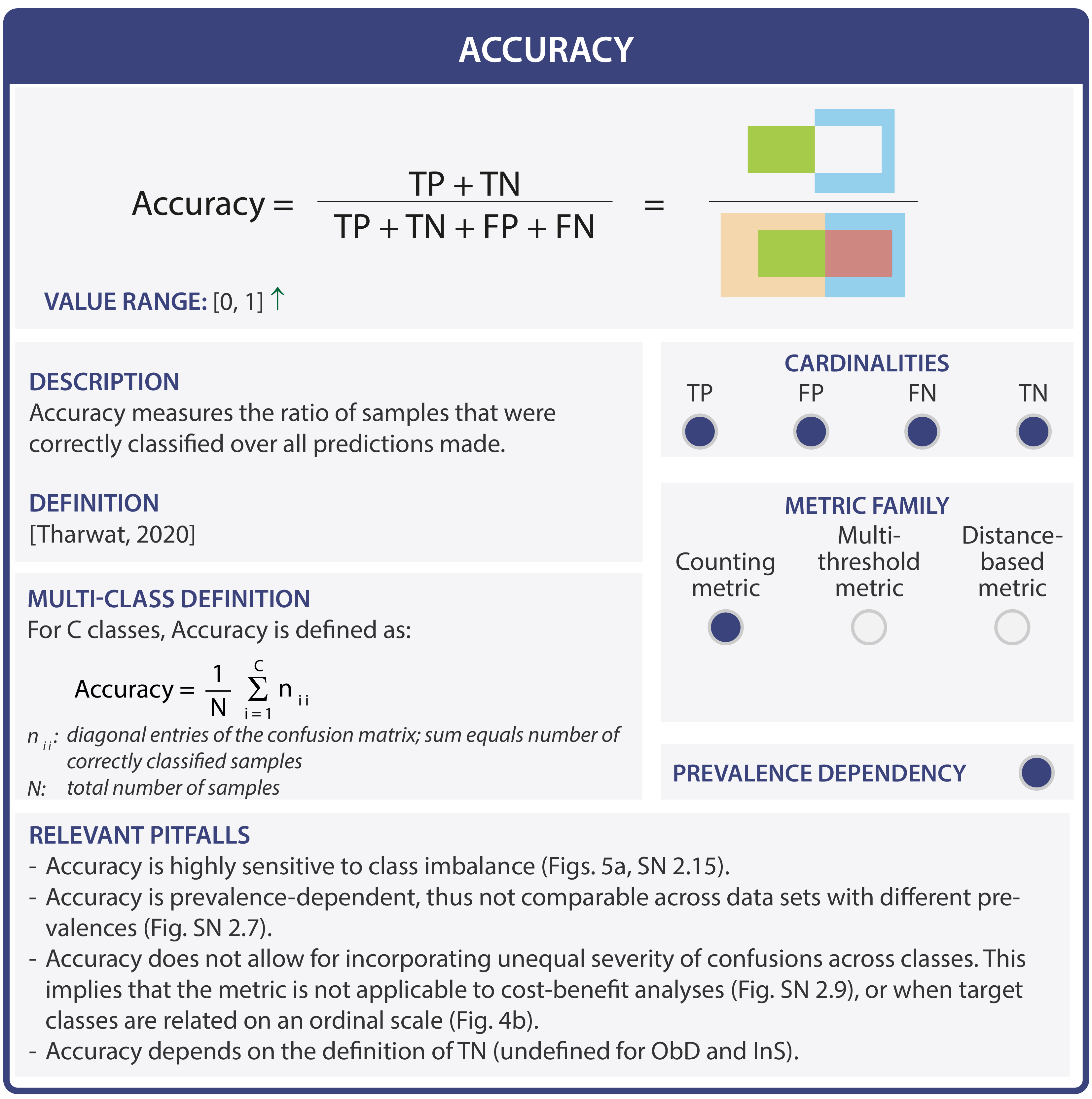}
    \caption{Metric profile of Accuracy. The upward arrow in the value range indicates that higher values are better than lower values. Abbreviations: \acf{FN}, \acf{FP}, \acf{InS}, \acf{ObD}, \acf{TN}, \acf{TP}. Reference: Tharwat, 2020: \cite{tharwat2020classification}. Mentioned figures: Figs.~4b, 5a, \ref{fig:prevalence-dependency}, \ref{fig:cost-benefit}, \ref{fig:class-imbalance}.}
    \label{fig:cheat-sheet-accuracy}
\end{figure}

\begin{figure}[H]
    \centering
    \includegraphics[width=\textwidth]{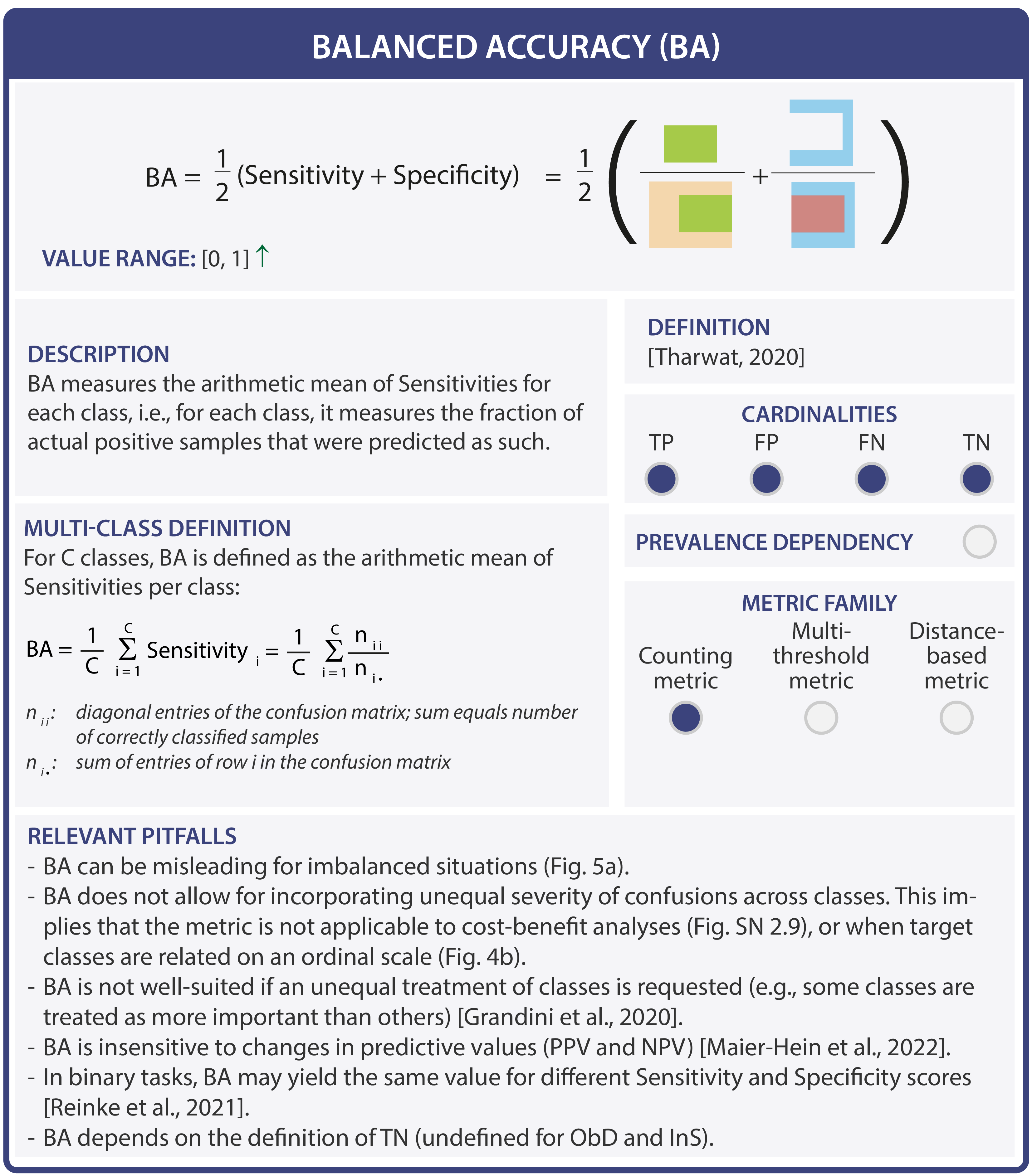}
    \caption{Metric profile of \acf{BA}. The upward arrow in the value range indicates that higher values are better than lower values. Abbreviations: \acf{FN}, \acf{FP}, \acf{InS}, \acf{ObD}, \acf{PPV}, \acf{TN}, \acf{TP}. References: Grandini et al., 2020: \cite{grandini2020metrics}, Maier-Hein et al., 2022: \cite{maier2022metrics}, Reinke et al., 2021: \cite{reinke2021commonarxiv}, Tharwat, 2020: \cite{tharwat2020classification}. Mentioned figures: Figs.~4b, 5a, \ref{fig:cost-benefit}.}
    \label{fig:cheat-sheet-ba}
\end{figure}

\begin{figure}[H]
    \centering
    \includegraphics[width=\textwidth]{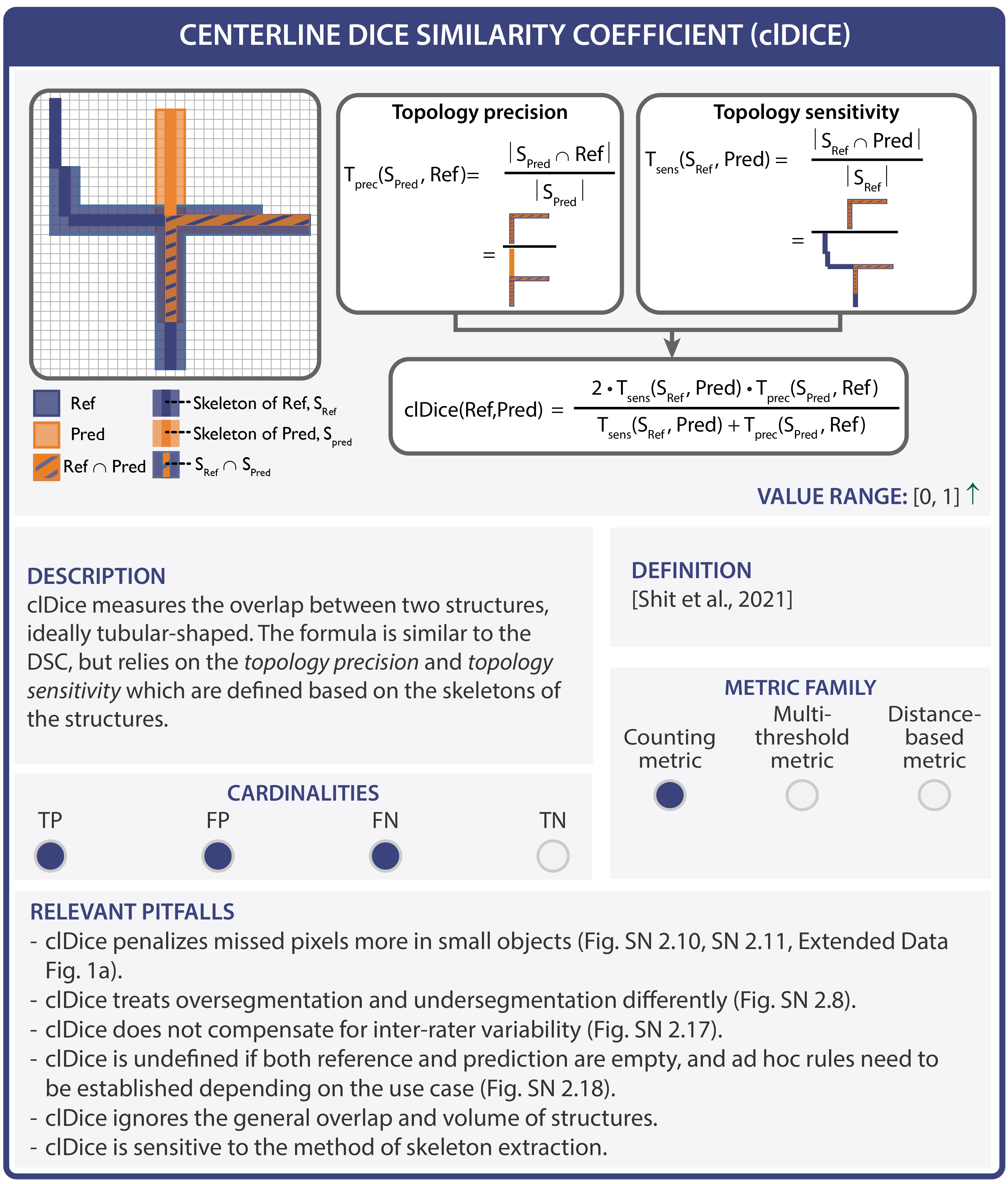}
    \caption{Metric profile of \acf{clDice}. The upward arrow in the value range indicates that higher values are better than lower values. Abbreviations: \acf{FN}, \acf{FP}, \acf{TN}, \acf{TP}. Reference: Shit et al., 2021: \cite{shit2021cldice}. Mentioned figures: Extended Data Fig.~1a, Figs.~\ref{fig:DSC-overunder}, \ref{fig:boundary-mask-iou}, \ref{fig:high-variability}, \ref{fig:low-quality}, \ref{fig:empty}.}
    \label{fig:cheat-sheet-cldice}
\end{figure}

\begin{figure}[H]
    \centering
    \includegraphics[width=\textwidth]{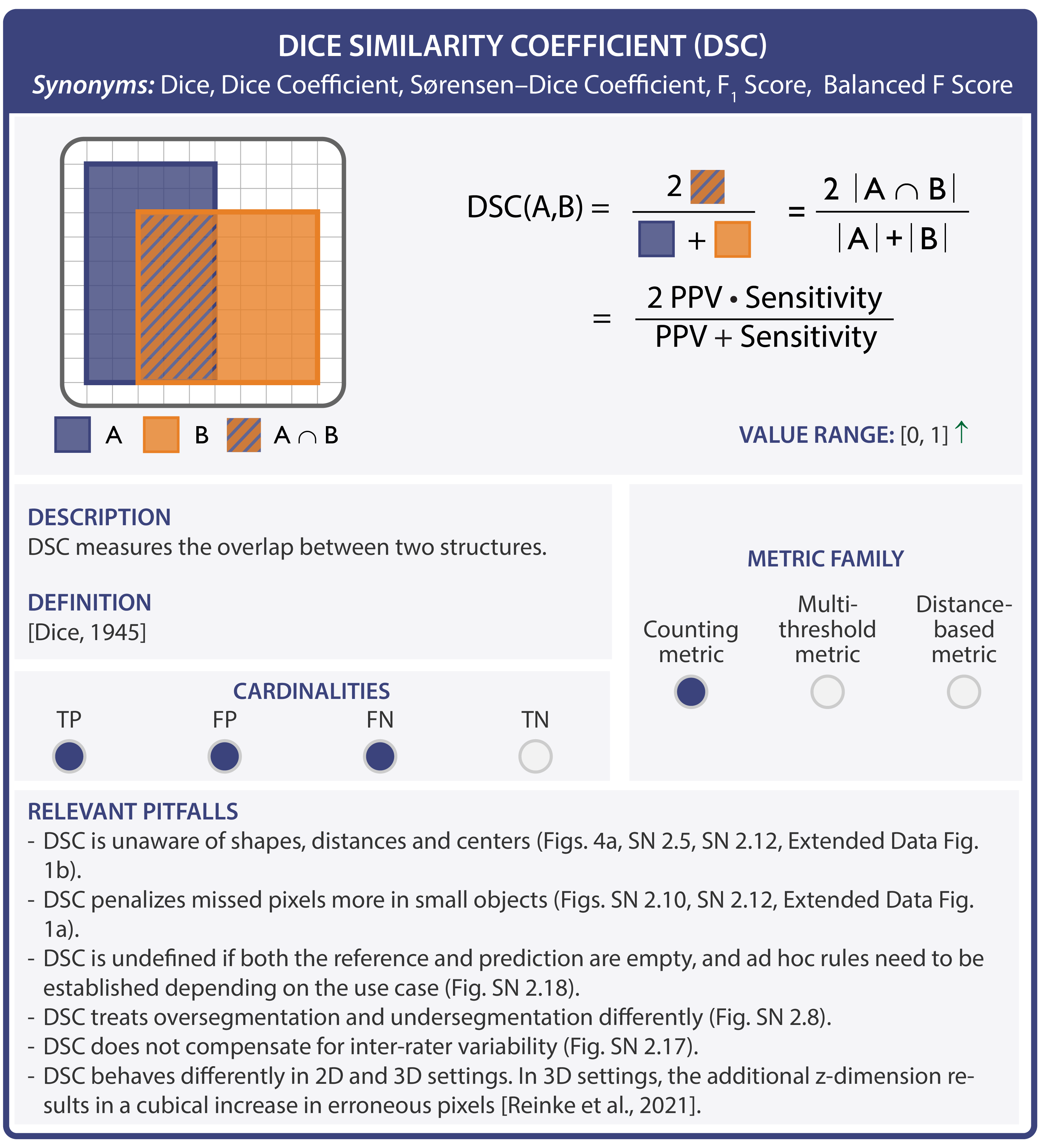}
    \caption{Metric profile of \acf{DSC}. The upward arrow in the value range indicates that higher values are better than lower values. Abbreviations: \acf{FN}, \acf{FP}, \acf{PPV}, \acf{TN}, \acf{TP}. References: Dice, 1945: \cite{dice1945measures}, Reinke et al., 2021: \cite{reinke2021commonarxiv}. Mentioned figures: Figs.~4a, \ref{fig:center}, \ref{fig:DSC-overunder}, \ref{fig:boundary-mask-iou}, \ref{fig:high-variability}, \ref{fig:complex-shapes}, \ref{fig:low-quality}, \ref{fig:empty}, Extended Data Fig.~1a-b.}
    \label{fig:cheat-sheet-dsc}
\end{figure}

\begin{figure}[H]
    \centering
    \includegraphics[width=1\textwidth]{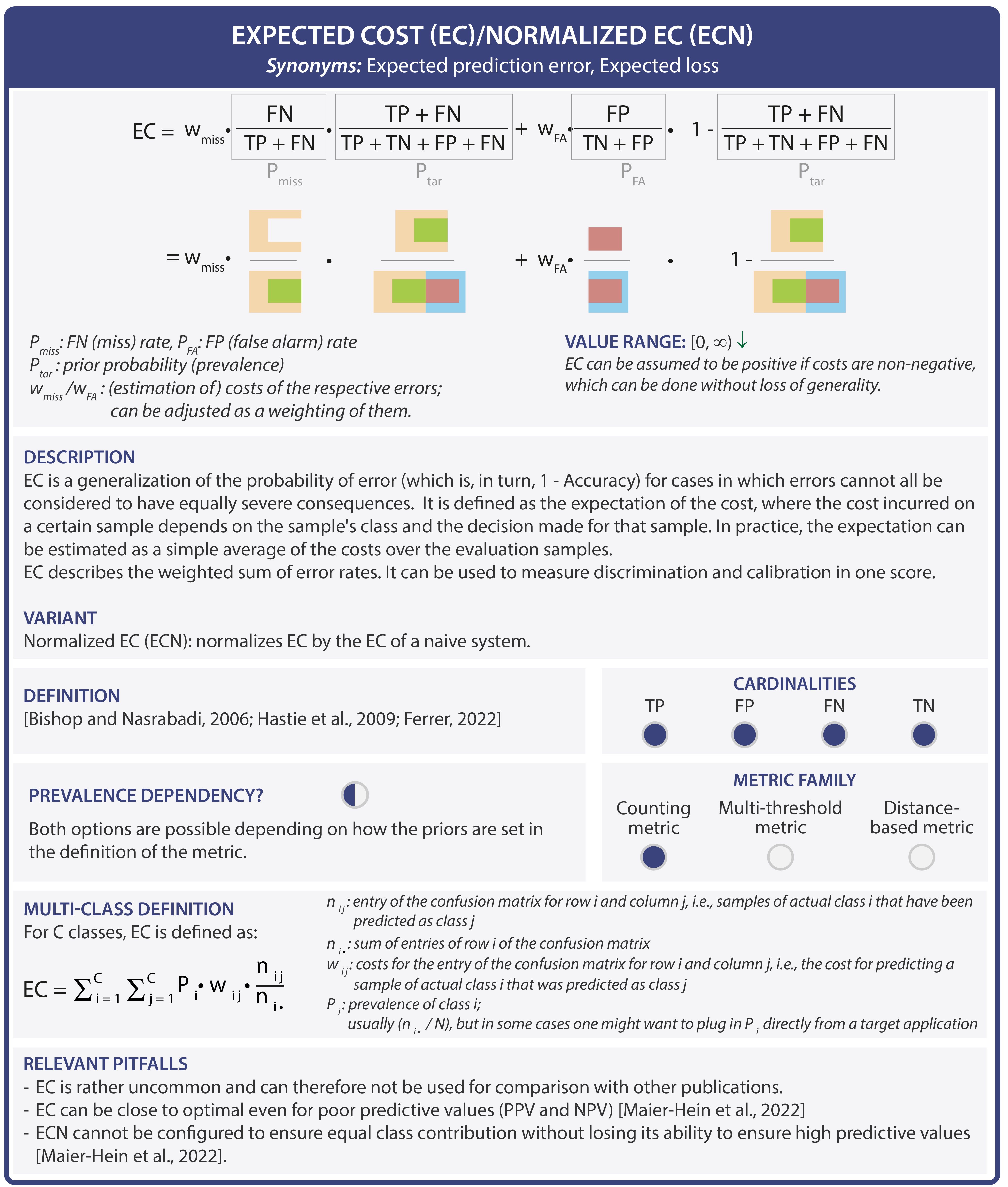}
    \caption{Metric profile of \acf{EC}. The downward arrow in the value range indicates that lower values are better than higher values. Abbreviations: \acf{FN}, \acf{FP}, \acf{TN}, \acf{TP}. References: Bishop and Nasrabadi, 2006: \cite{bishop2006pattern}, Ferrer 2022: \cite{ferrer2022analysis}, Hastie et al., 2009:  \cite{hastie2009elements}, Maier-Hein et al., 2022: \cite{maier2022metrics}.}
    \label{fig:cheat-sheet-ec}
\end{figure}

\begin{figure}[H]
    \centering
    \includegraphics[width=\textwidth]{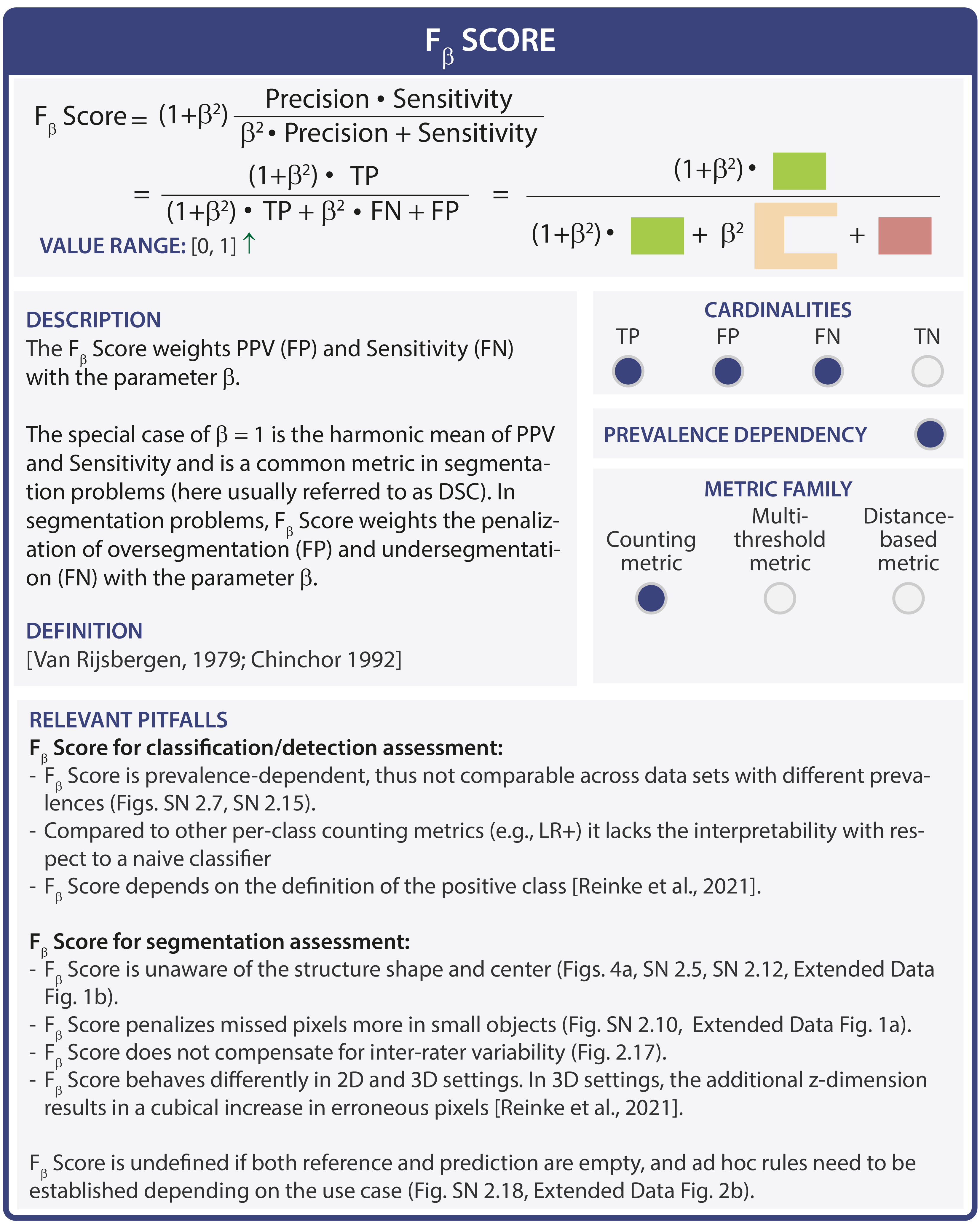}
    \caption{Metric profile of F$_\beta$ Score.\cite{Chinchor1992, van1979information}. The upward arrow in the value range indicates that higher values are better than lower values. Abbreviations: \acf{DSC}, \acf{FN}, \acf{FP}, \acf{PPV}, \acf{TN}, \acf{TP}.References: Chinchor 1992: \cite{Chinchor1992}, Reinke et al., 2021: \cite{reinke2021commonarxiv}, Van Rijsbergen, 1979: \cite{van1979information}. Mentioned figures: Figs.~4a, \ref{fig:center}, \ref{fig:prevalence-dependency}, \ref{fig:boundary-mask-iou}, \ref{fig:complex-shapes}, \ref{fig:class-imbalance}, \ref{fig:low-quality}, \ref{fig:empty}, Extended Data Figs.~1a-b and 2b.}
    \label{fig:cheat-sheet-fbeta}
\end{figure}

\begin{figure}[H]
    \centering
    \includegraphics[width=\textwidth]{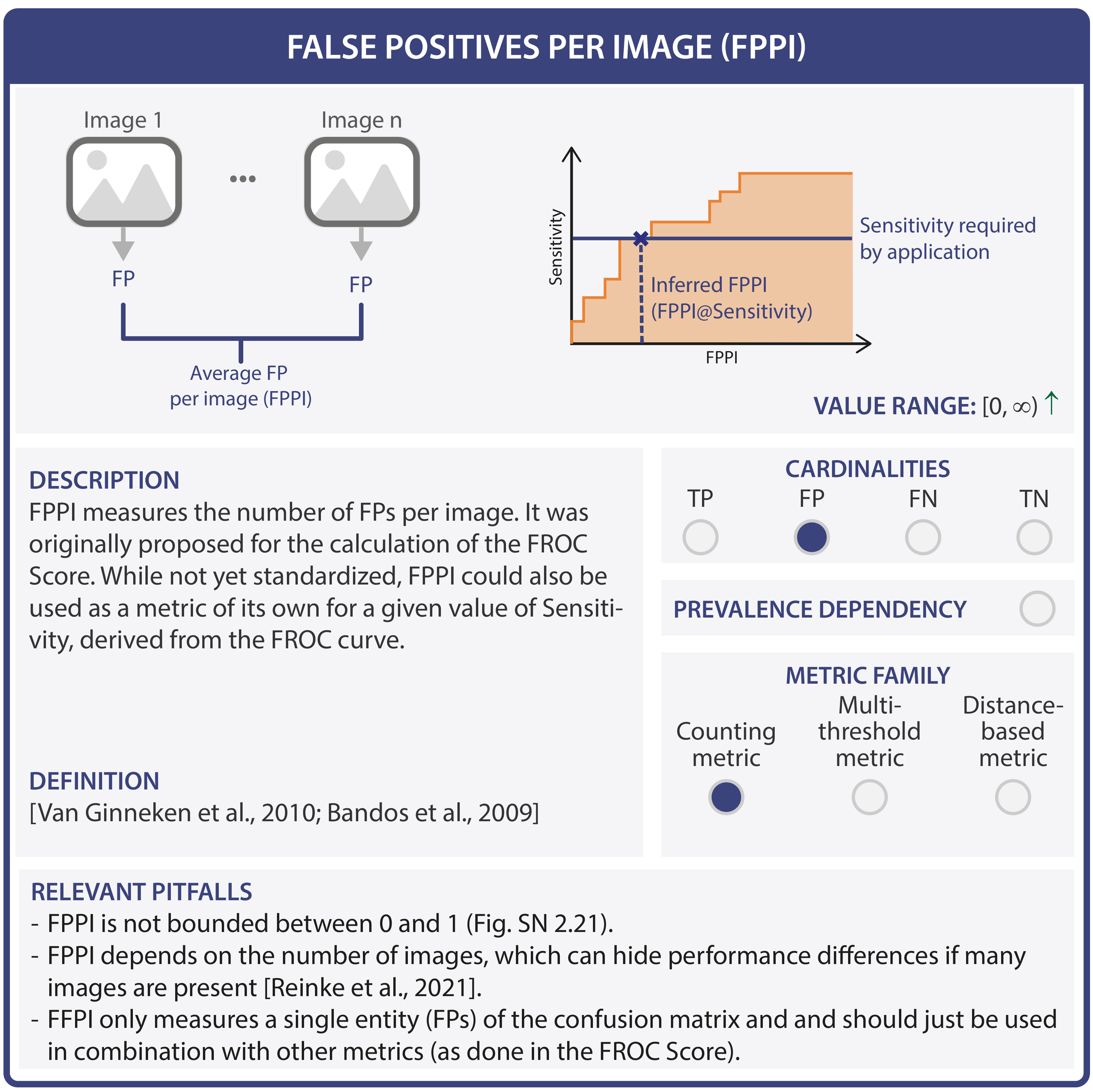}
    \caption{Metric profile of \acf{FPPI}. The upward arrow in the value range indicates that higher values are better than lower values. Abbreviations: \acf{FN}, \acf{FP}, \acf{FROC}, \acf{TN}, \acf{TP}. References: Bandos et al., 2009: \cite{bandos2009area}, Reinke et al., 2021: \cite{reinke2021commonarxiv}, Van Ginneken et al., 2010: \cite{van2010comparing}. Mentioned figure: Fig.~\ref{fig:FROC-no-standard}.}
    \label{fig:cheat-sheet-fppi}
\end{figure}

\begin{figure}[H]
    \centering
    \includegraphics[width=\textwidth]{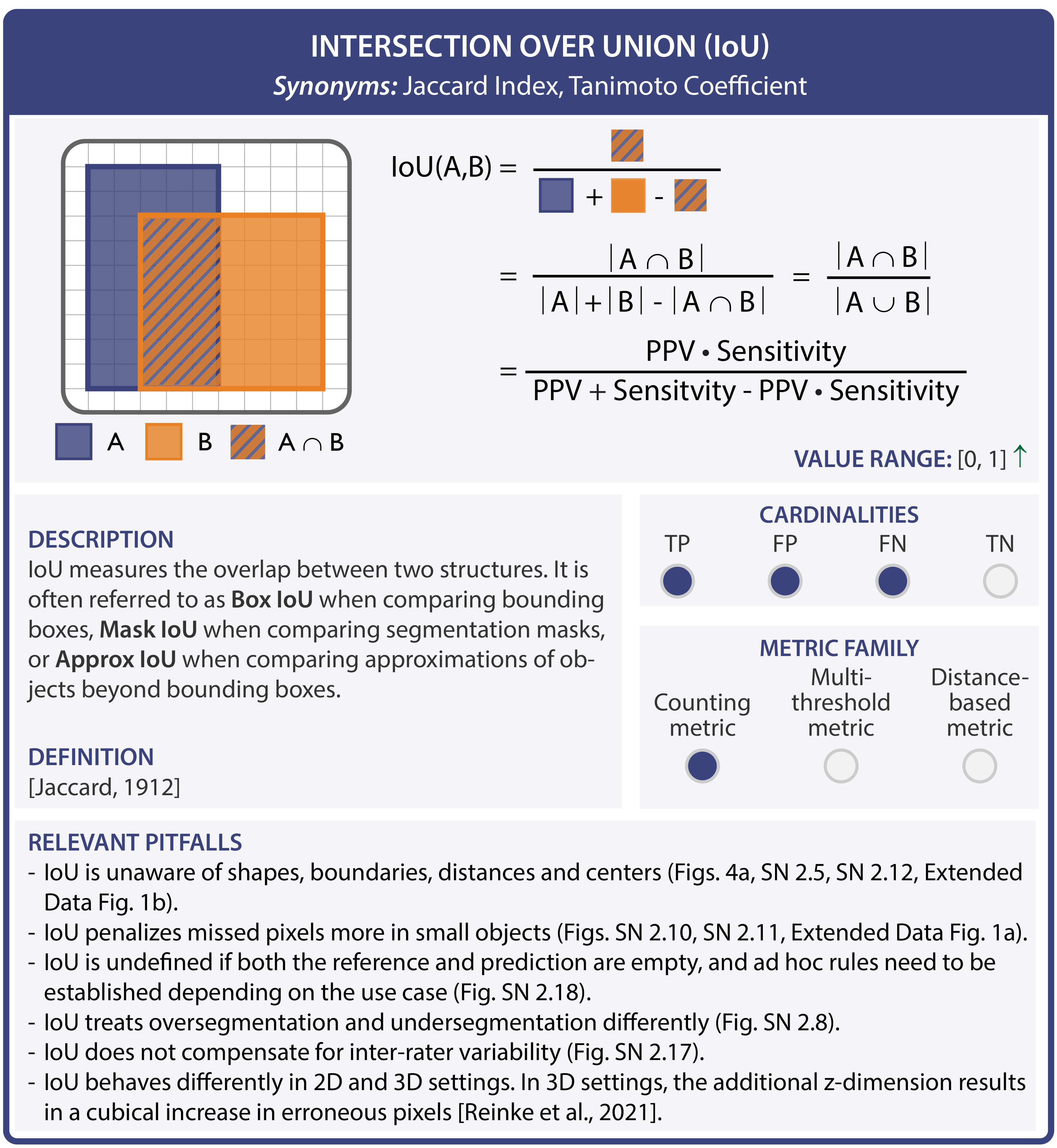}
    \caption{Metric profile of \acf{IoU}. The upward arrow in the value range indicates that higher values are better than lower values. Abbreviations: \acf{FN}, \acf{FP}, \acf{PPV}, \acf{TN}, \acf{TP}. References: Jaccard, 1912: \cite{jaccard1912distribution}, Reinke et al., 2021: \cite{reinke2021commonarxiv}. Mentioned figures: Figs.~4a, \ref{fig:center}, \ref{fig:DSC-overunder}, \ref{fig:boundary-mask-iou}, \ref{fig:high-variability}, \ref{fig:complex-shapes}, \ref{fig:low-quality}, \ref{fig:empty}, Extended Data Fig.~1a-b.}
    \label{fig:cheat-sheet-iou}
\end{figure}

\begin{figure}[H]
    \centering
    \includegraphics[width=\textwidth]{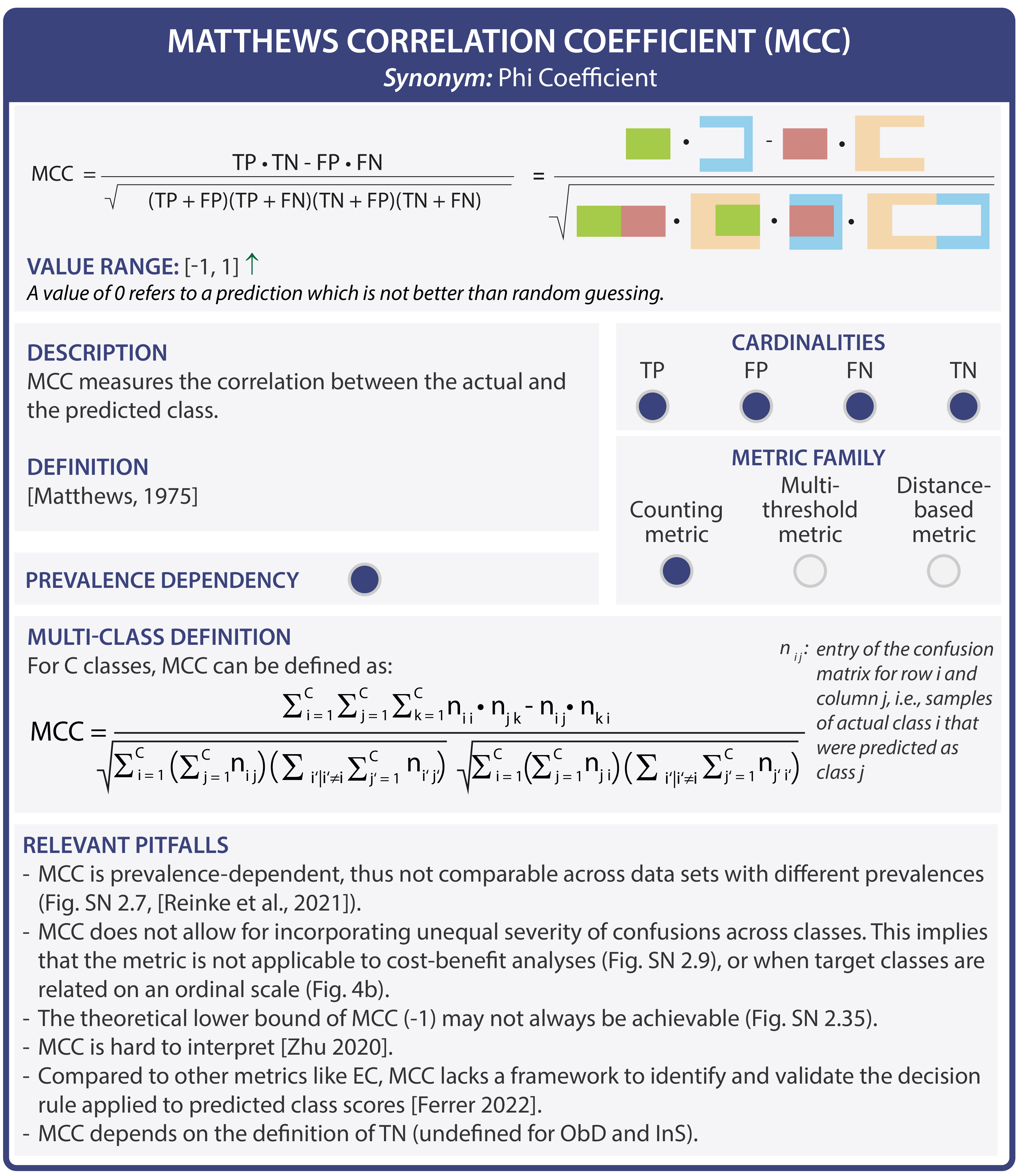}
    \caption{Metric profile of \acf{MCC}. The upward arrow in the value range indicates that higher values are better than lower values. Abbreviations: \acf{EC}, \acf{FN}, \acf{FP}, \acf{InS}, \acf{ObD}, \acf{TN}, \acf{TP}. References: Ferrer, 2022: \cite{ferrer2022analysis}, Matthews, 1975: \cite{matthews1975comparison}, Reinke et al., 2021: \cite{reinke2021commonarxiv}, Zhu, 2020: \cite{zhu2020performance}. Mentioned figures: Figs.~4b, \ref{fig:prevalence-dependency}, \ref{fig:cost-benefit}, \ref{fig:lack-bounds}.}
    \label{fig:cheat-sheet-mcc}
\end{figure}

\begin{figure}[H]
    \centering
    \includegraphics[width=\textwidth]{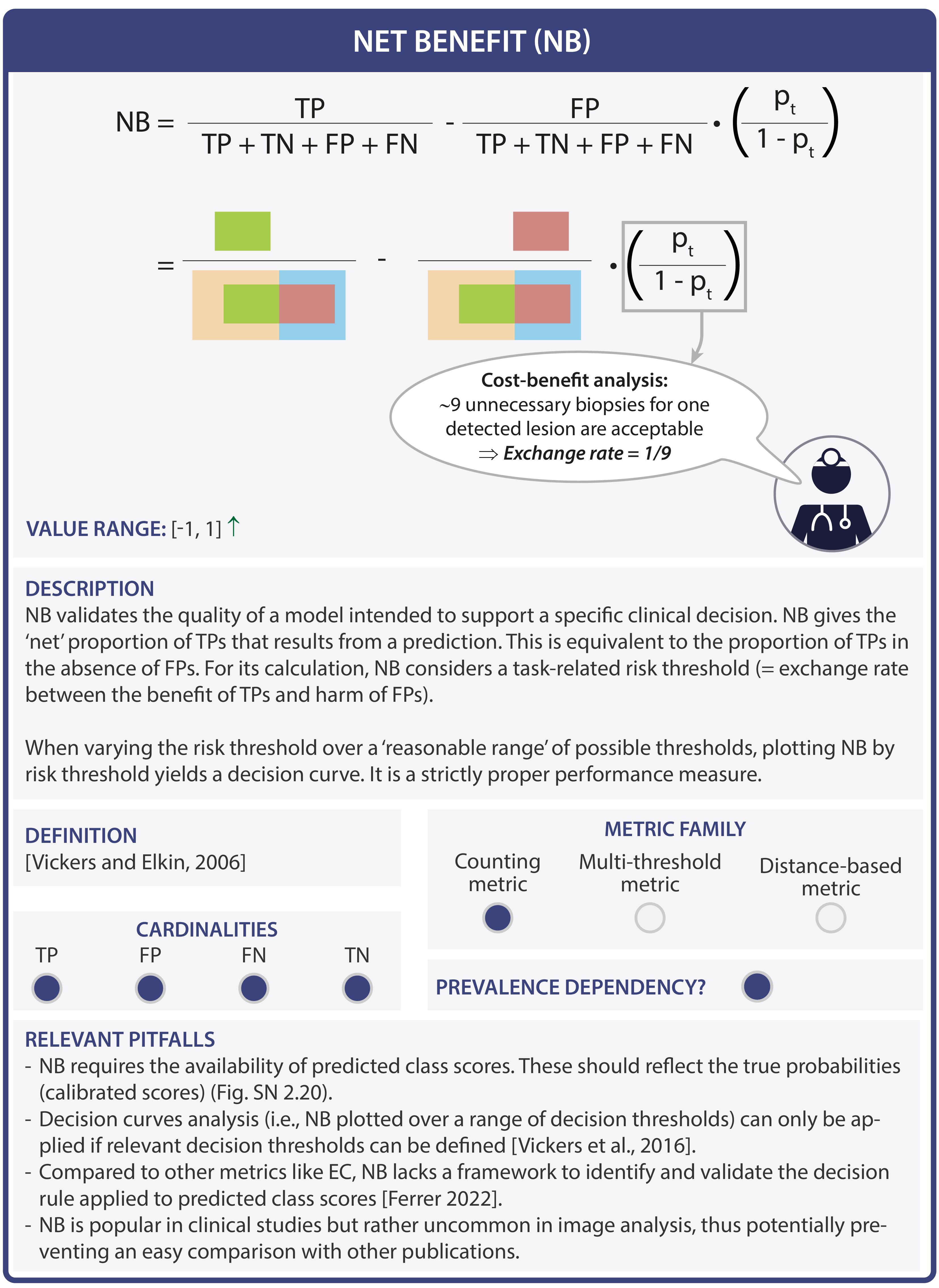}
    \caption{Metric profile of \acf{NB}. The upward arrow in the value range indicates that higher values are better than lower values. Abbreviations: \acf{EC}, \acf{FN}, \acf{FP}, \acf{TN}, \acf{TP}. References: Ferrer, 2022: \cite{ferrer2022analysis}, Vickers and Elkin, 2006: \cite{vickers2006decision}, Vickers et al., 2016: \cite{vickers2016net}. Mentioned figure: Fig.~\ref{fig:lack-of-scores}.}
    \label{fig:cheat-sheet-nb}
\end{figure}

\begin{figure}[H]
    \centering
    \includegraphics[width=\textwidth]{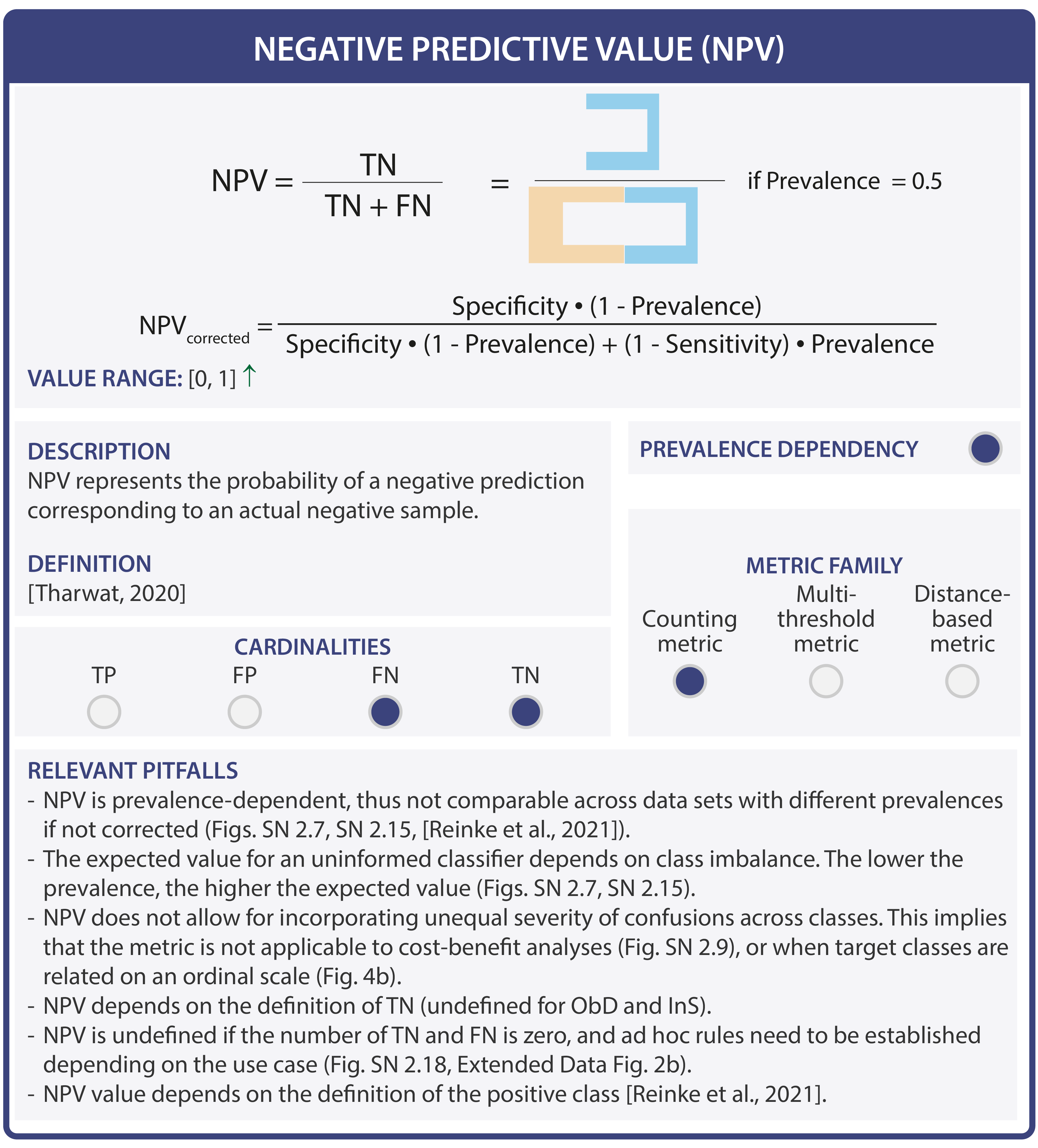}
    \caption{Metric profile of \acf{NPV}. The upward arrow in the value range indicates that higher values are better than lower values. Abbreviations: \acf{FN}, \acf{FP}, \acf{InS}, \acf{ObD}, \acf{TN}, \acf{TP}. References: Reinke et al., 2021: \cite{reinke2021commonarxiv}, Tharwat, 2020: \cite{tharwat2020classification}. Mentioned figures: Figs.~4b, \ref{fig:prevalence-dependency}, \ref{fig:cost-benefit}, \ref{fig:class-imbalance}, \ref{fig:empty}, Extended Data Fig.~2b.}
    \label{fig:cheat-sheet-npv}
\end{figure}

\begin{figure}[H]
    \centering
    \includegraphics[width=\textwidth]{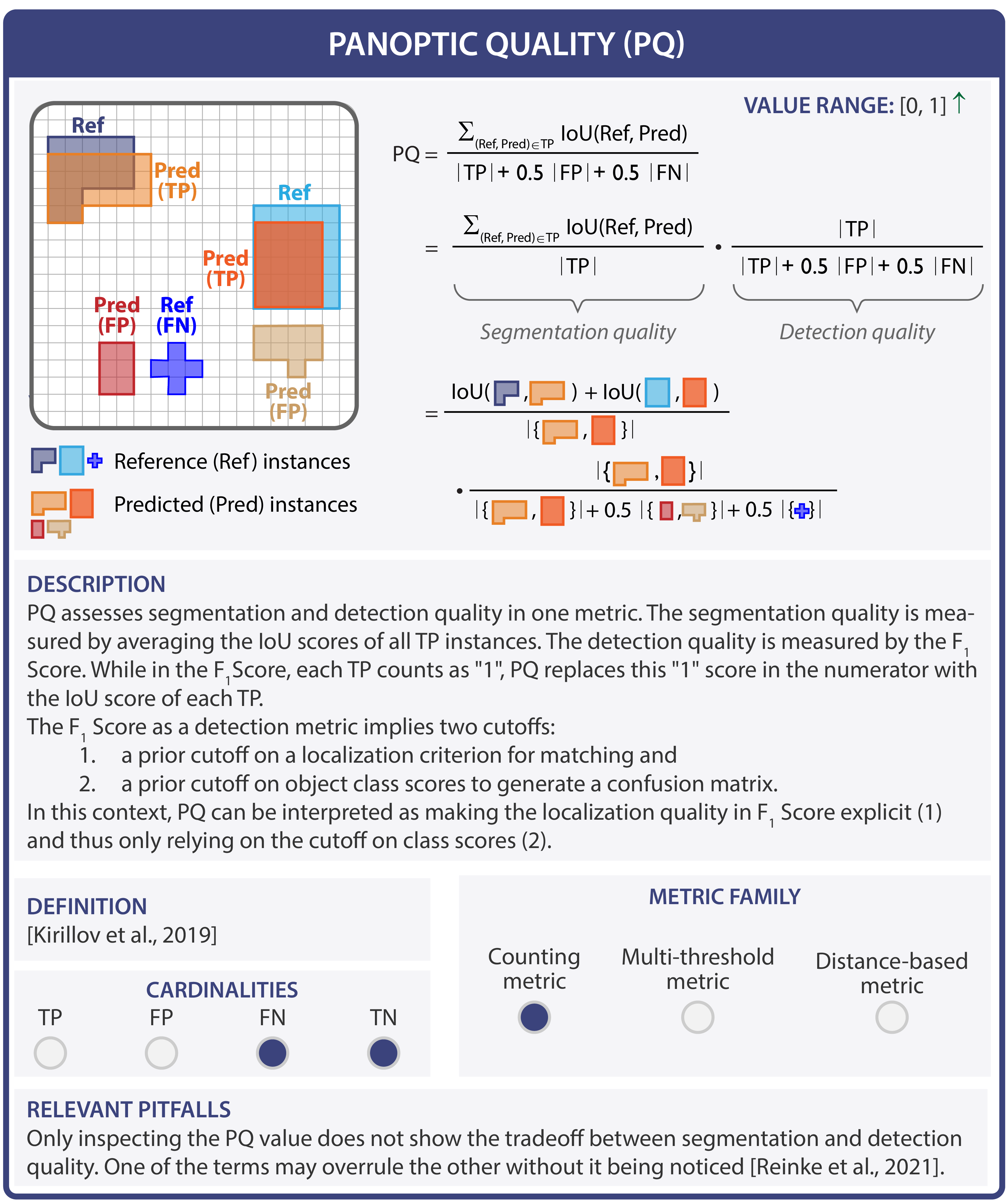}
    \caption{Metric profile of \acf{PQ}. The upward arrow in the value range indicates that higher values are better than lower values. Abbreviations: \acf{AP}, \acf{FN}, \acf{FP}, \acf{FROC}, \acf{IoU}, \acf{TN}, \acf{TP}. References: Kirillov et al., 2019: \cite{kirillov2019panoptic}, Reinke et al., 2021: \cite{reinke2021commonarxiv}.}
    \label{fig:cheat-sheet-pq}
\end{figure}

\acresetall
This section presents profiles for the metrics deemed particularly relevant by the \textit{Metrics Reloaded} consortium~\cite{maier2022metrics}. For each metric, the respective description, formula, and value range (upward arrow: higher values better than lower values; downward arrow: lower values are better than higher values) are provided, along with further important characteristics, such as the used cardinalities of a confusion matrix, or potential prevalence dependency. Finally, relevant pitfalls are highlighted. Many of the presented metrics rely on the confusion matrix, which is illustrated in Fig.~\ref{fig:confusion-matrix}.

\newpage
\subsection{Discrimination metrics} 
\subsubsection{Counting metrics}
\hfill+
\FloatBarrier

\newpage

\begin{figure}[H]
    \centering
    \includegraphics[width=\textwidth]{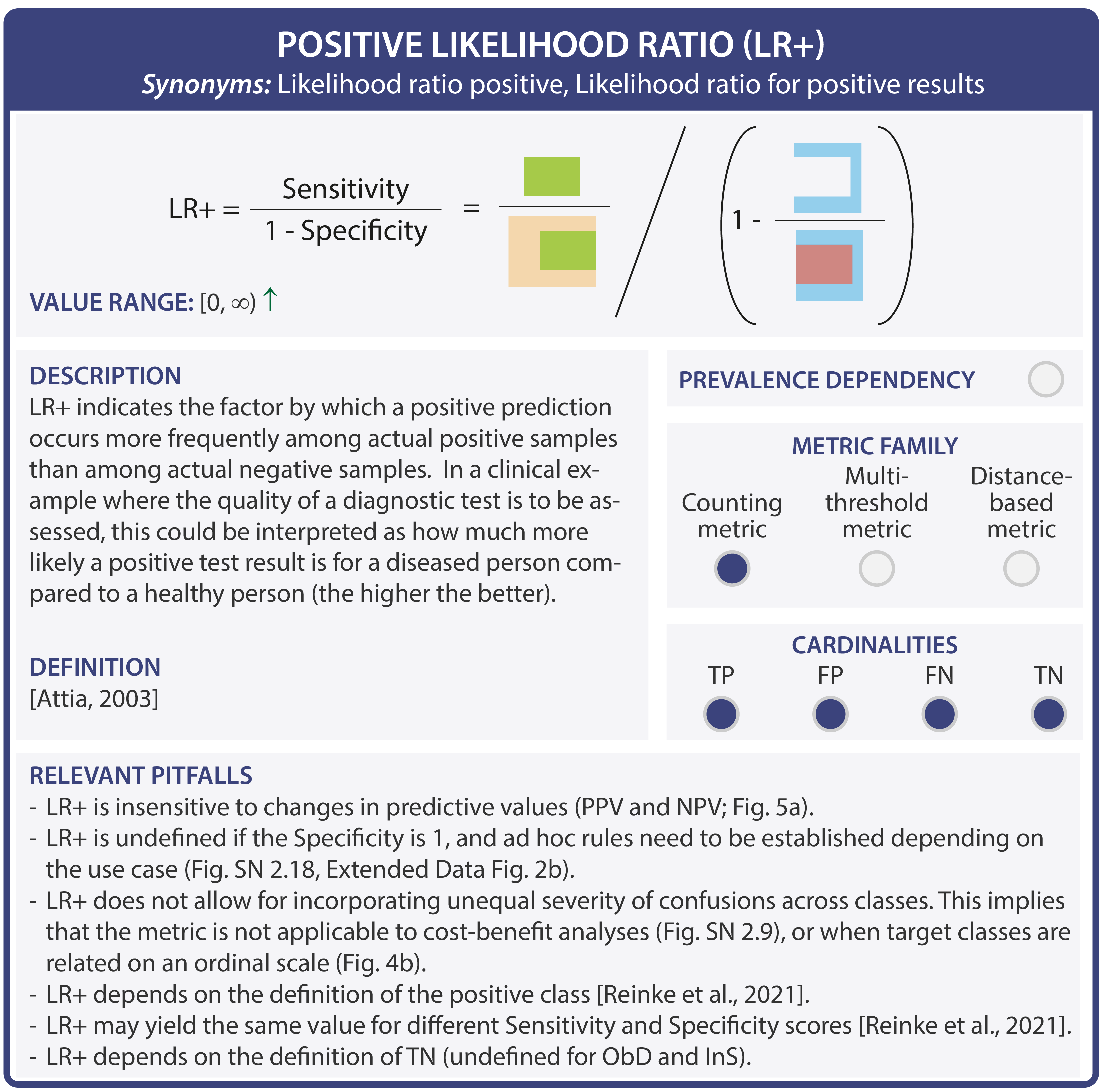}
    \caption{Metric profile of \acf{LR+}. The upward arrow in the value range indicates that higher values are better than lower values. Abbreviations: \acf{FN}, \acf{FP}, \acf{InS}, \acf{ObD}, \acf{PPV}, \acf{TN}, \acf{TP}. References: Attia, 2003: \cite{wales2003moving}, Reinke et al., 2021: \cite{reinke2021commonarxiv}. Mentioned figures: Figs.~4b, 5a, \ref{fig:cost-benefit}, \ref{fig:empty}, Extended Data Fig.~2b.}
    \label{fig:cheat-sheet-lr+}
\end{figure}

\begin{figure}[H]
    \centering
    \includegraphics[width=\textwidth]{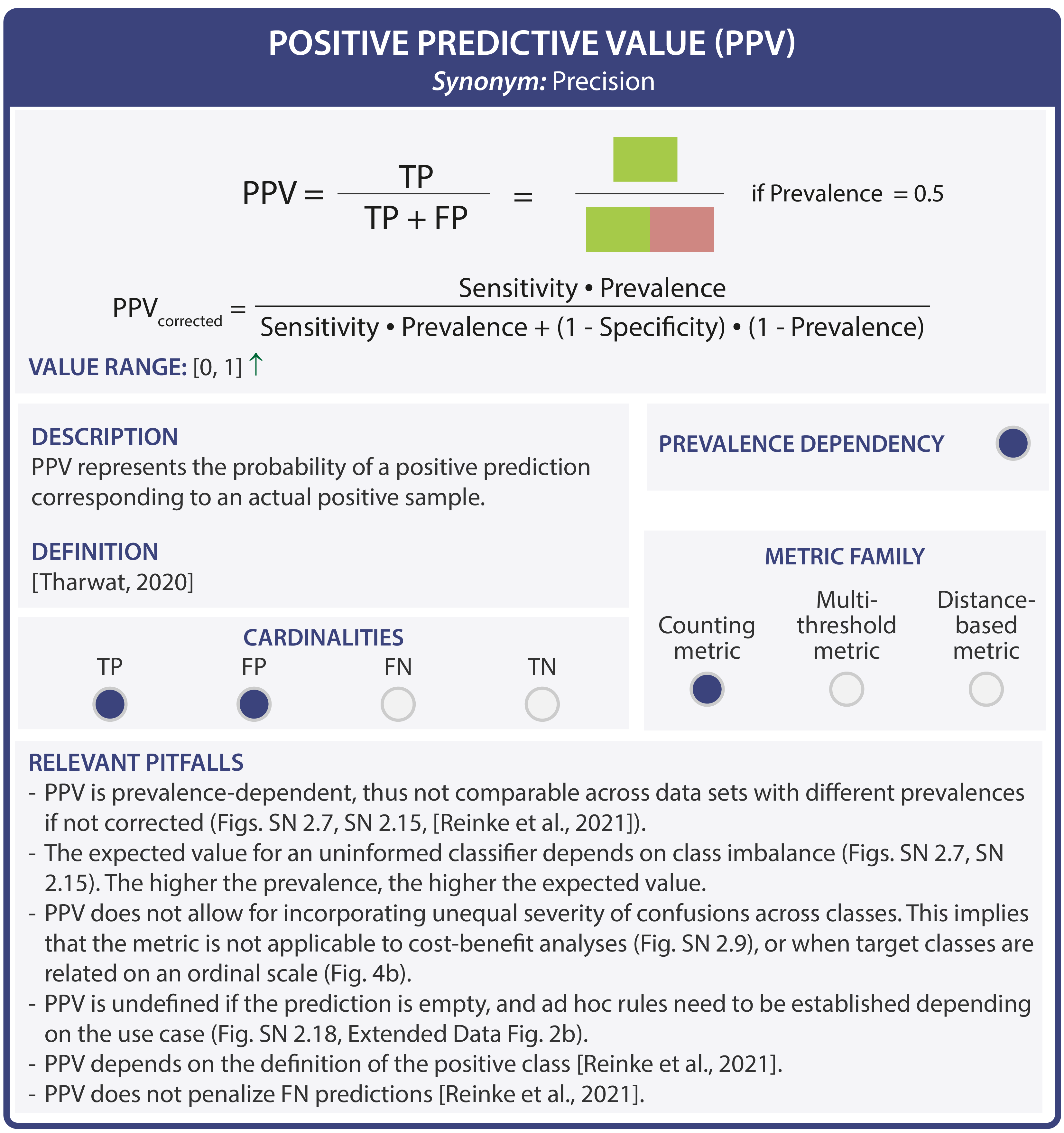}
    \caption{Metric profile of the \acf{PPV}. The upward arrow in the value range indicates that higher values are better than lower values. Abbreviations used in the figure: \acf{FN}, \acf{FP}, \acf{InS}, \acf{ObD}, \acf{TN}, \acf{TP}. References used in the figure: Reinke et al., 2021: \cite{reinke2021commonarxiv}, Tharwat, 2020:  \cite{tharwat2020classification}. Mentioned figures: Figs.~4b, \ref{fig:prevalence-dependency}, \ref{fig:cost-benefit}, \ref{fig:class-imbalance}, \ref{fig:empty}, Extended Data Fig.~2b.}
    \label{fig:cheat-sheet-ppv}
\end{figure}

\begin{figure}[H]
    \centering
    \includegraphics[width=\textwidth]{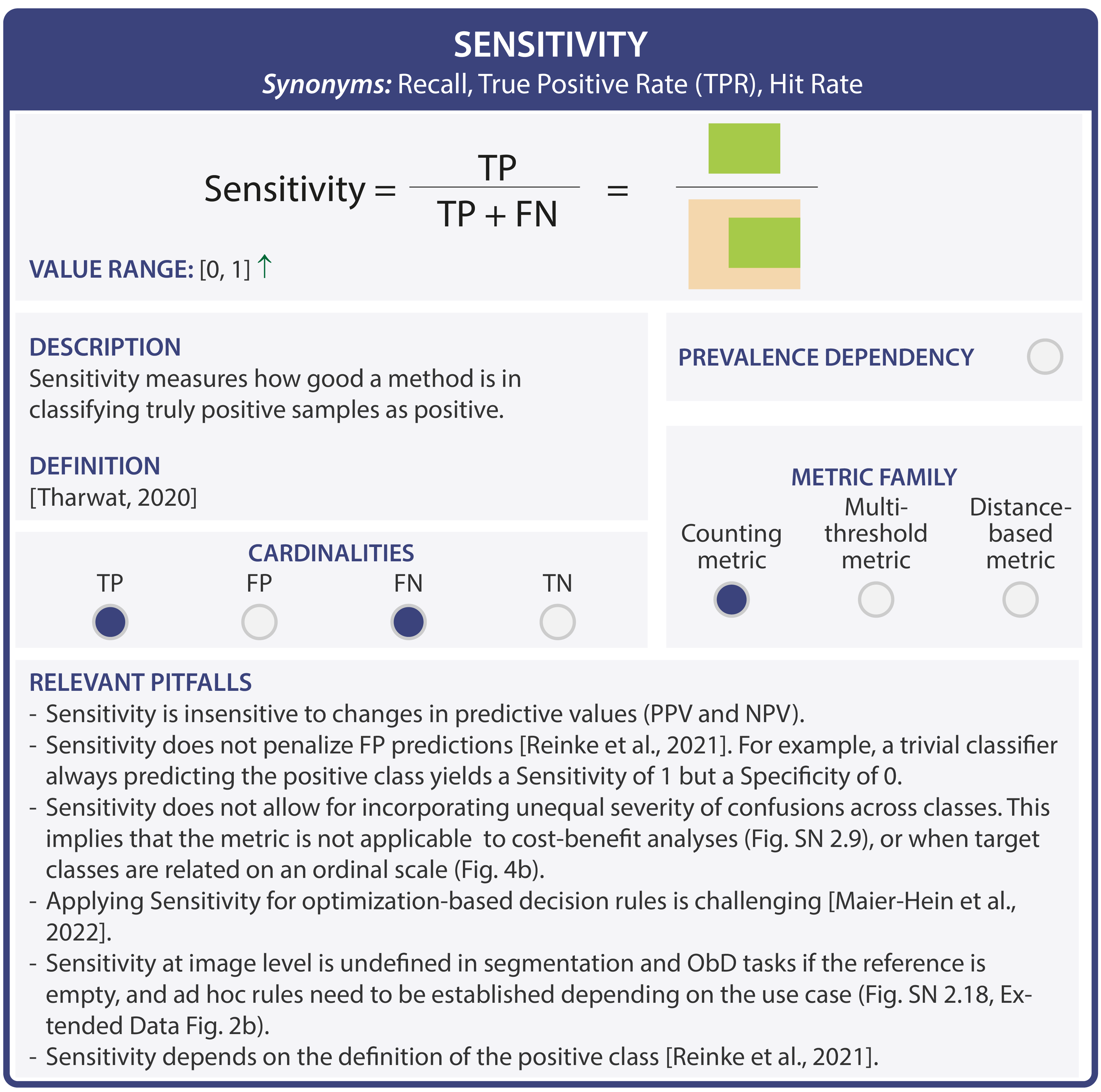}
    \caption{Metric profile of Sensitivity. The upward arrow in the value range indicates that higher values are better than lower values. Abbreviations: \acf{FN}, \acf{FP}, \acf{ObD}, \acf{PPV}, \acf{TN}, \acf{TP}. References: Maier-Hein et al., 2022: \cite{maier2022metrics}, Reinke et al., 2021: \cite{reinke2021commonarxiv}, Tharwat, 2020:  \cite{tharwat2020classification}. Mentioned figures: Figs.~4b, \ref{fig:cost-benefit}, \ref{fig:empty}, Extended Data Fig.~2b.}
    \label{fig:cheat-sheet-sensitivity}
\end{figure}

\begin{figure}[H]
    \centering
    \includegraphics[width=1\textwidth]{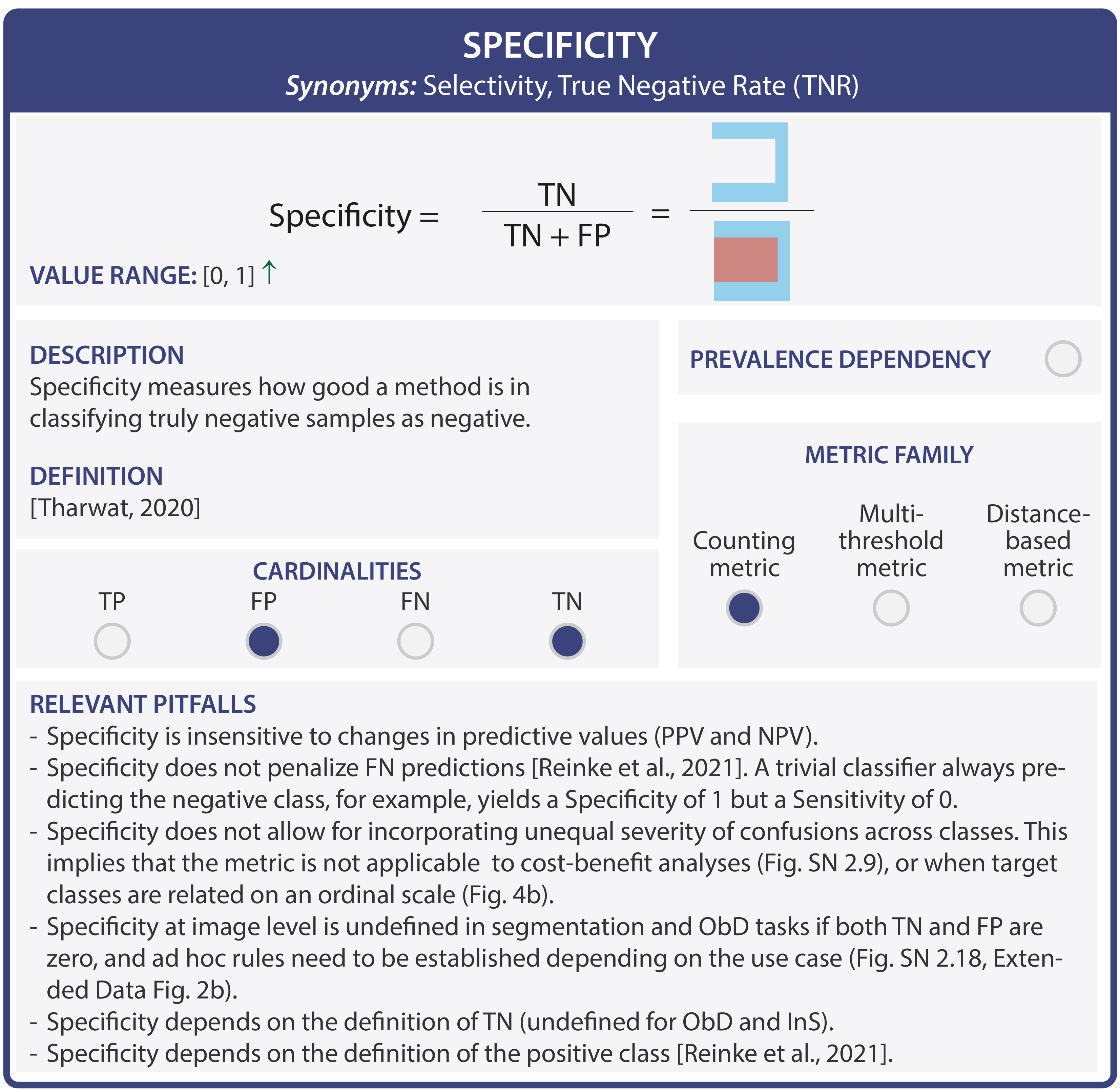}
    \caption{Metric profile of Specificity. The upward arrow in the value range indicates that higher values are better than lower values. Abbreviations: \acf{FN}, \acf{FP}, \acf{TN}, \acf{TP}. References: Reinke et al., 2021: \cite{reinke2021commonarxiv}, Tharwat, 2020: \cite{tharwat2020classification}. Mentioned figures: Figs.~4b \ref{fig:cost-benefit}, \ref{fig:empty}, Extended Data Fig.~2b.}
    \label{fig:cheat-sheet-specificity}
\end{figure}

\begin{figure}[H]
    \centering
    \includegraphics[width=0.9\textwidth]{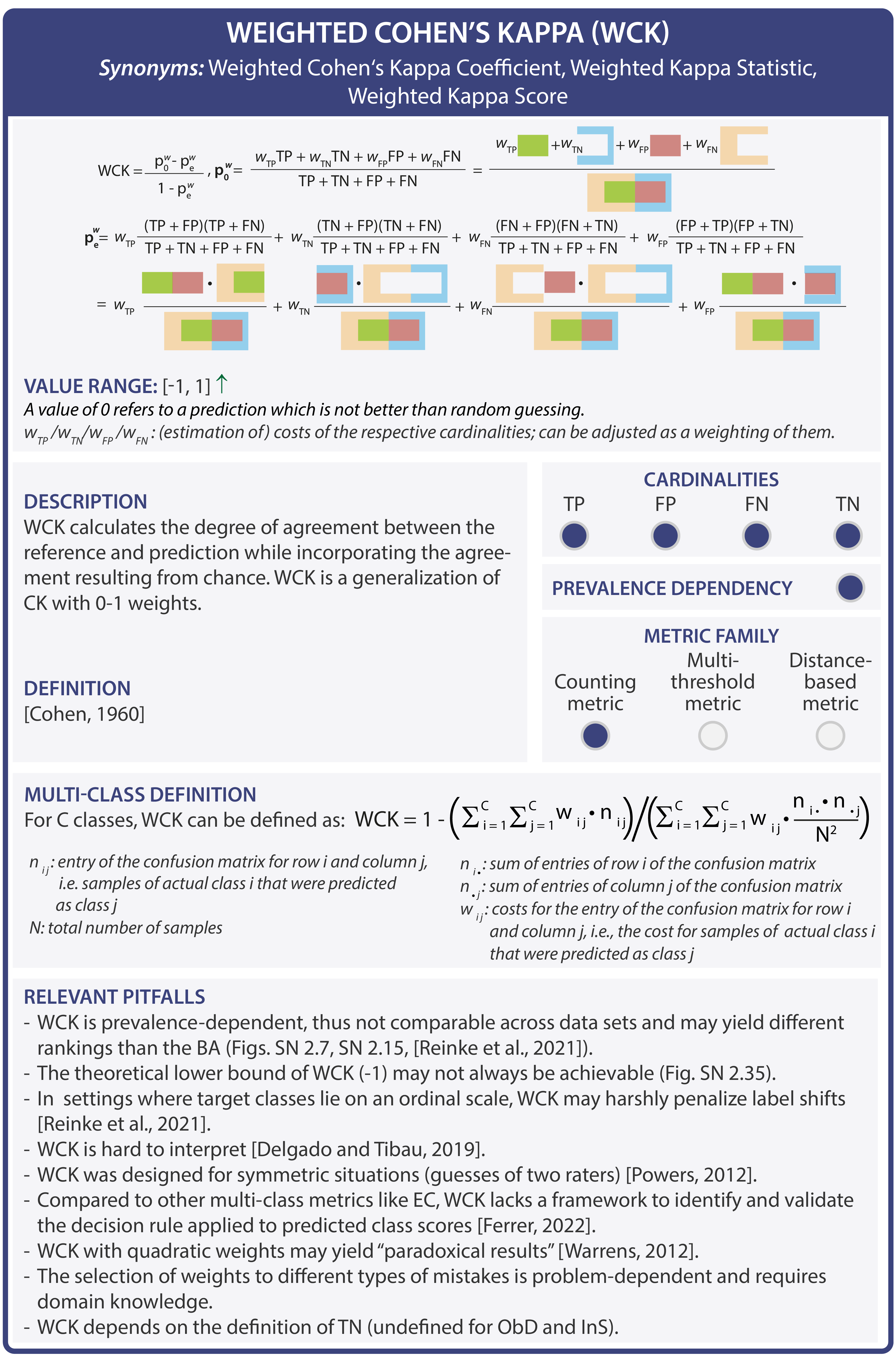}
    \caption{Metric profile of \acf{WCK}. The upward arrow in the value range indicates that higher values are better than lower values. Abbreviations: \acf{BA}, \acf{CK}, \acf{EC}, \acf{FN}, \acf{FP}, \acf{InS}, \acf{ObD}, \acf{TN}, \acf{TP}. References: Cohen, 1960: \cite{cohen1960coefficient}, Delgado and Tibau, 2019: \cite{delgado2019cohen}, Ferrer, 2022: \cite{ferrer2022analysis}, Powers, 2012: \cite{powers2012problem}, Reinke et al., 2021: \cite{reinke2021commonarxiv}, Warrens, 2012: \cite{warrens2012some}. Mentioned figures: Figs.~\ref{fig:prevalence-dependency}, \ref{fig:class-imbalance}, \ref{fig:lack-bounds}.}
    \label{fig:cheat-sheet-wck}
\end{figure}

\newpage
\subsubsection{Multi-threshold metrics}
\hfill
\begin{figure}[H]
    \centering
    \includegraphics[width=\textwidth]{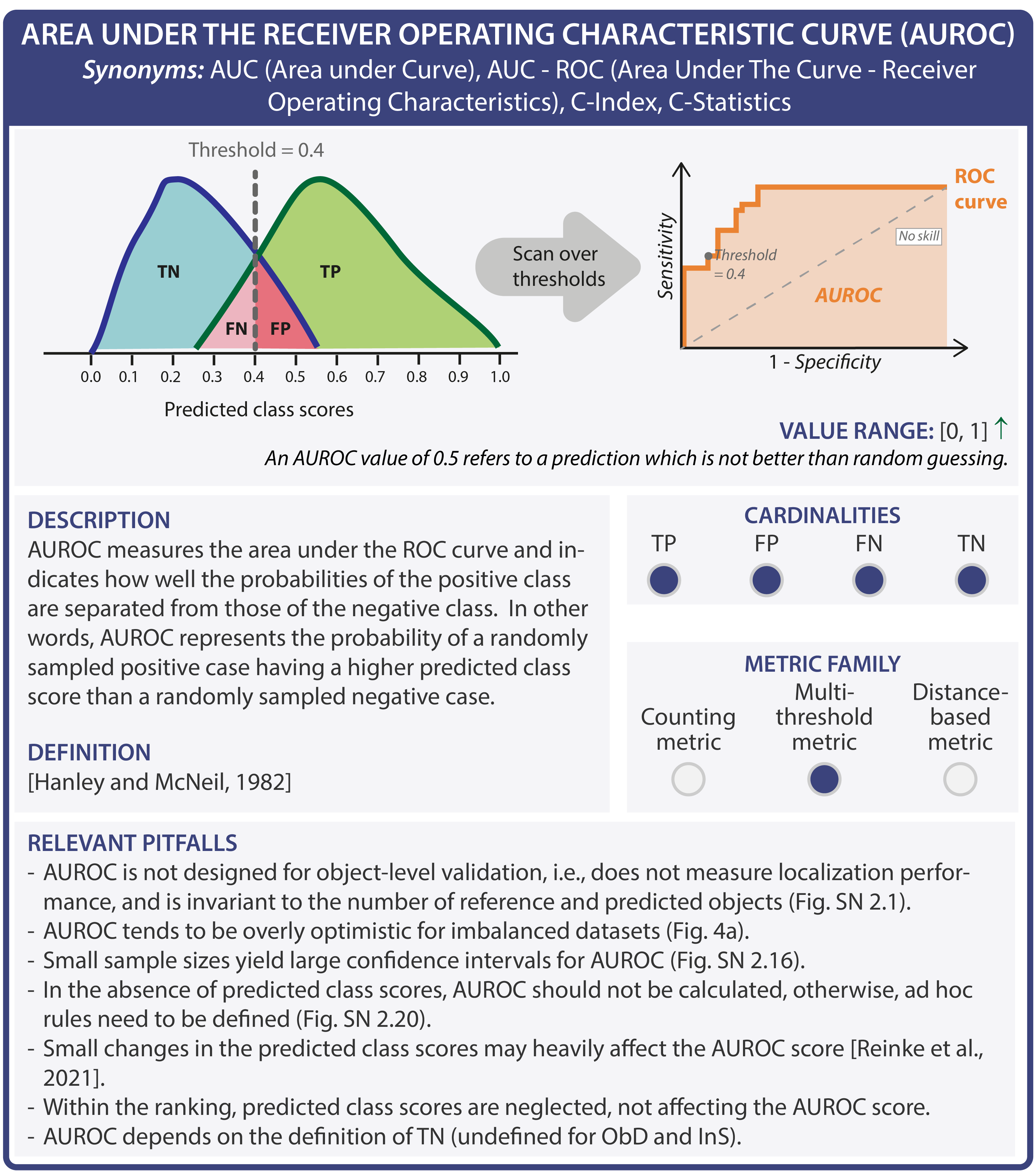}
    \caption{Metric profile of \acf{AUROC}. The upward arrow in the value range indicates that higher values are better than lower values. Abbreviations: \acf{FN}, \acf{FP}, \acf{InS}, \acf{ObD}, \acf{ROC}, \acf{TN}, \acf{TP}. References: Hanley and McNeil, 1982: \cite{hanley1982meaning}, Reinke et al., 2021: \cite{reinke2021commonarxiv}. Mentioned figures: Figs.~5a, \ref{fig:roc}, \ref{fig:auroc-small-sample-sizes}, \ref{fig:lack-of-scores}.}
    \label{fig:cheat-sheet-auroc}
\end{figure}
\FloatBarrier
\newpage
\begin{figure}[H]
    \centering
    \includegraphics[width=0.9\textwidth]{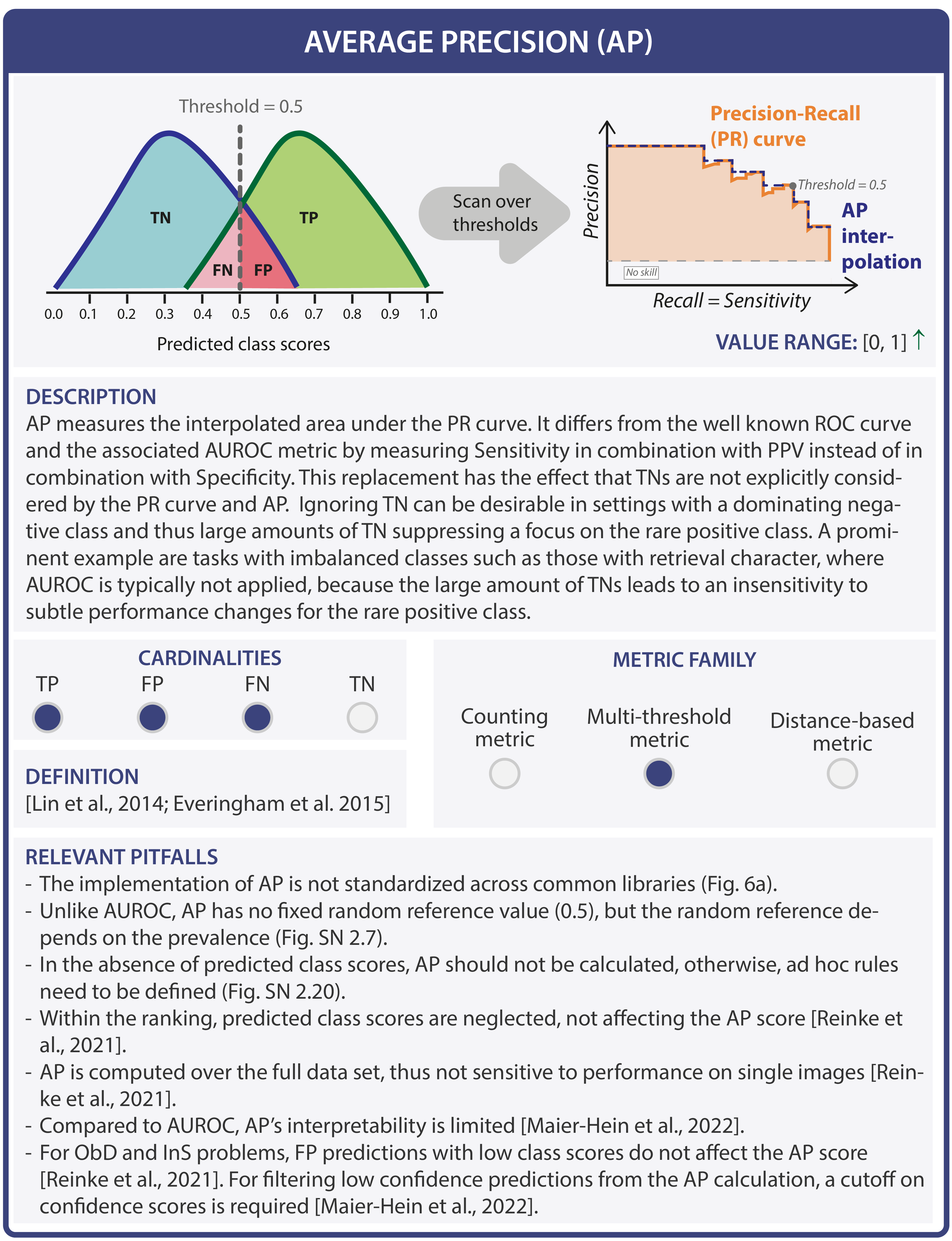}
    \caption{Metric profile of \acf{AP}. The upward arrow in the value range indicates that higher values are better than lower values. Abbreviations: \acf{AUROC}, \acf{FN}, \acf{FP}, \acf{InS}, \acf{ObD}, \acf{PR}, \acf{TN}, \acf{TP}. References: Everingham et al., 2015: \cite{everingham2010pascal}, Lin et al., 2014: \cite{lin2014microsoft}, Maier-Hein et al., 2022: \cite{maier2022metrics}, Reinke et al., 2021: \cite{reinke2021commonarxiv}.
    Mentioned figures: Figs.~6a, \ref{fig:prevalence-dependency}, \ref{fig:lack-of-scores}.}
    \label{fig:cheat-sheet-ap}
\end{figure}

\begin{figure}[H]
    \centering
    \includegraphics[width=\textwidth]{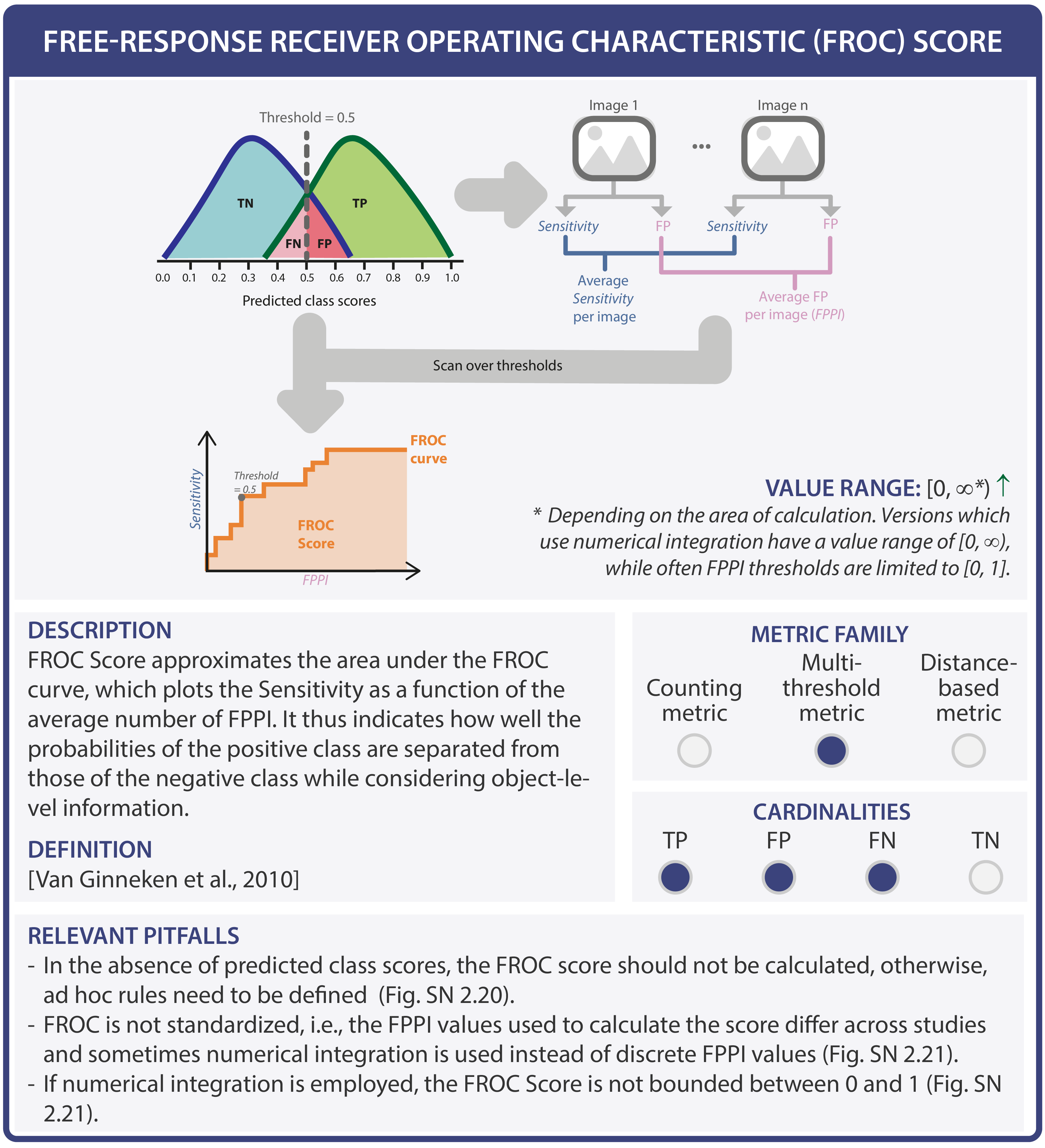}
    \caption{Metric profile of \acf{FROC}. The upward arrow in the value range indicates that higher values are better than lower values. Abbreviations: \acf{FN}, \acf{FP}, \acf{FPPI}, \acf{TN}, \acf{TP}. References: Van Ginneken et al., 2010: \cite{van2010comparing}. Mentioned figures: Figs.~\ref{fig:lack-of-scores}, \ref{fig:FROC-no-standard}.}
    \label{fig:cheat-sheet-froc}
\end{figure}

\newpage
\FloatBarrier
\subsubsection{Distance-based metrics}
\hfill
\FloatBarrier
\begin{figure}[H]
    \centering
    \includegraphics[width=\textwidth]{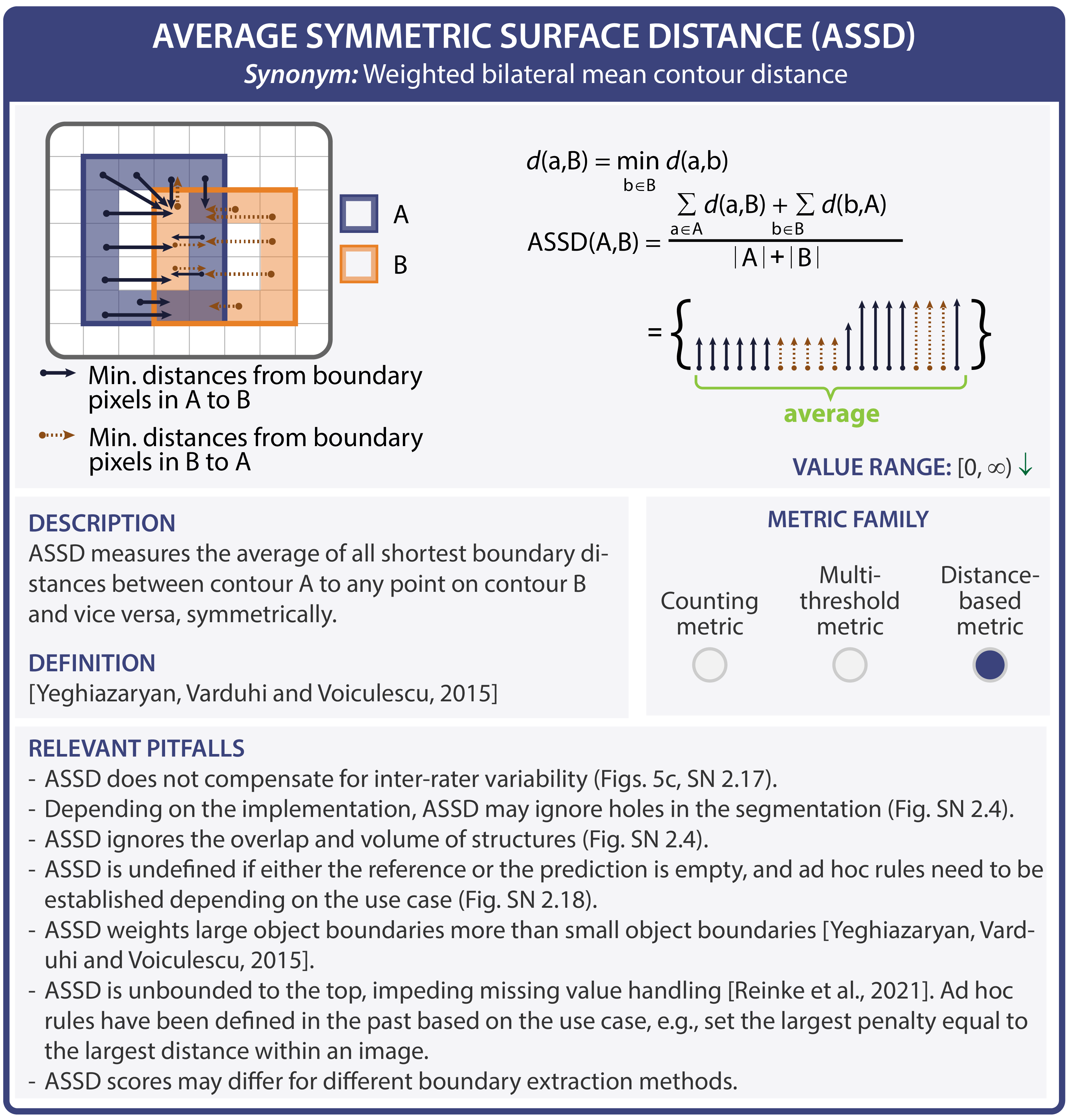}
    \caption{Metric profile of \acf{ASSD}. The downward arrow in the value range indicates that lower values are better than higher values. Abbreviation: \acf{SemS}. References: Reinke et al., 2021: \cite{reinke2021commonarxiv}, Yeghiazaryan, Varduhi and Voiculescu, 2015: \cite{yeghiazaryan2015overview}. Mentioned figures: Figs.~5c, \ref{fig:outline}, \ref{fig:low-quality}, \ref{fig:empty}.}
    \label{fig:cheat-sheet-assd}
\end{figure}
\FloatBarrier
\newpage
\begin{figure}[H]
    \centering
    \includegraphics[width=\textwidth]{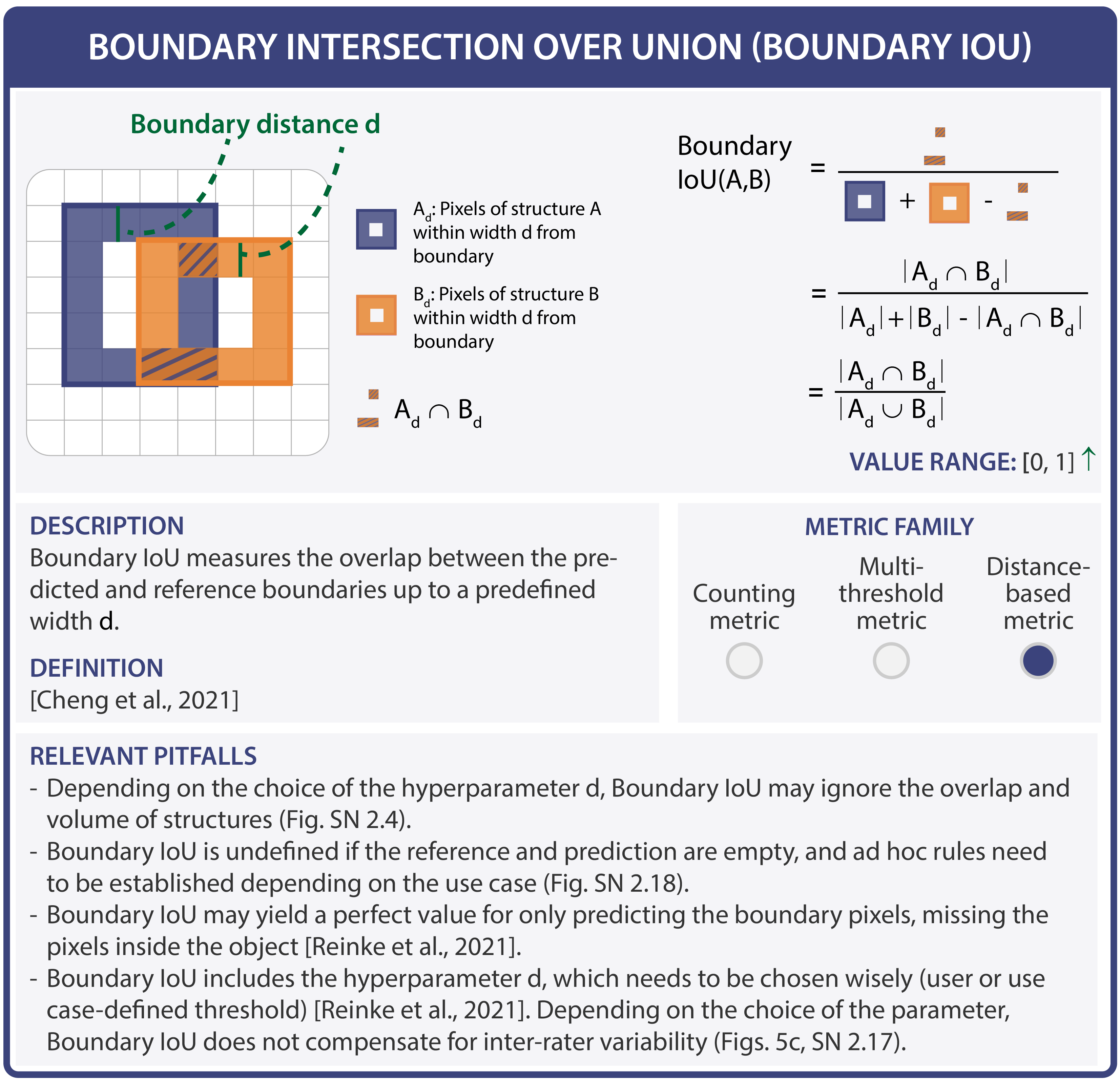}
    \caption{Metric profile of the Boundary \acf{IoU}. The upward arrow in the value range indicates that higher values are better than lower values. References: Cheng et al., 2021: \cite{cheng2021boundary}, Reinke et al., 2021: \cite{reinke2021commonarxiv}. Mentioned figures: Figs.~5c, \ref{fig:outline}, \ref{fig:low-quality}, \ref{fig:empty}.}
    \label{fig:cheat-sheet-boundary-iou}
\end{figure}

\begin{figure}[H]
    \centering
    \includegraphics[width=\textwidth]{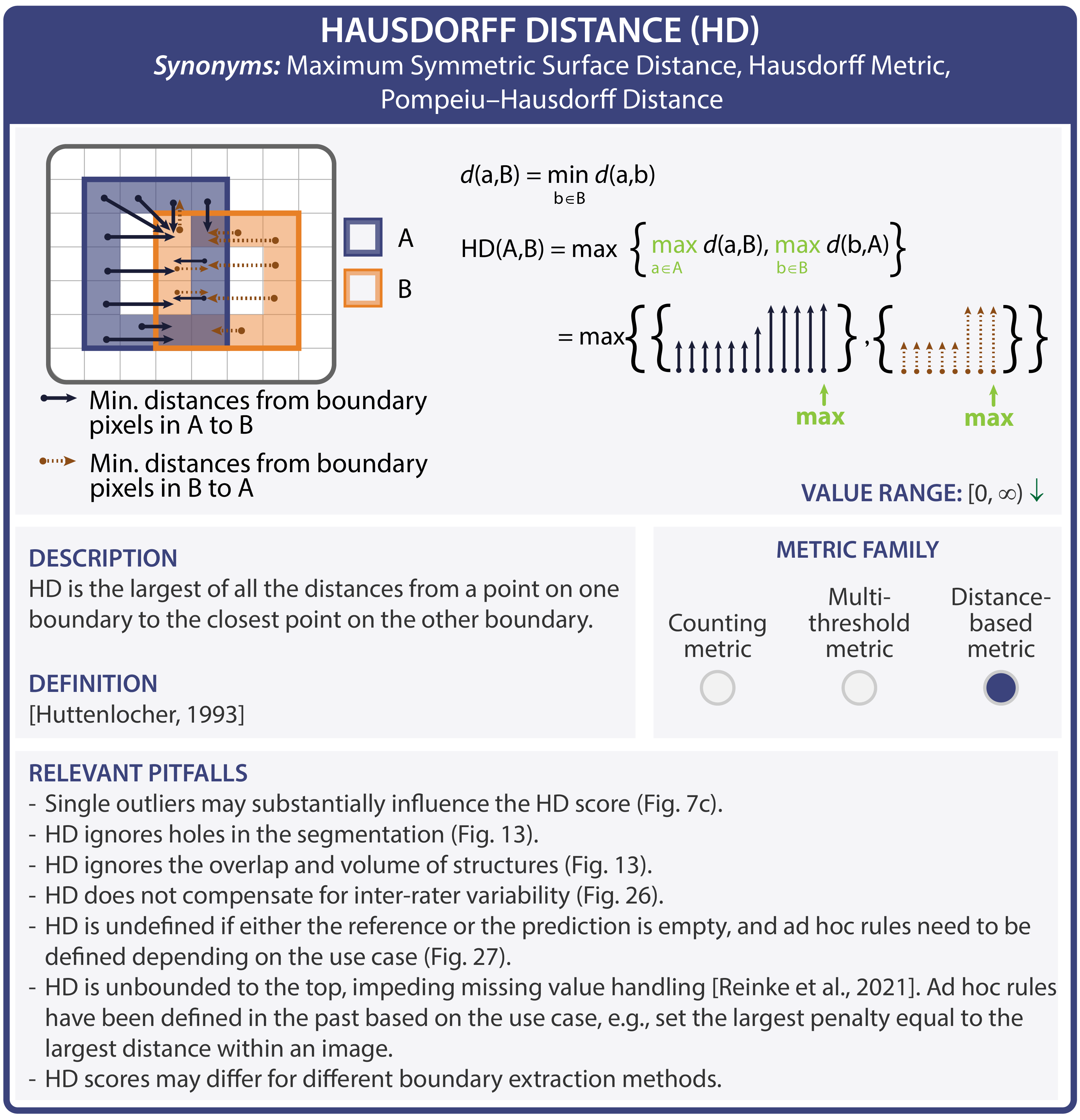}
    \caption{Metric profile of \acf{HD}. The downward arrow in the value range indicates that lower values are better than higher values. Abbreviation: \acf{SemS}. References : Huttenlocher, 1993: \cite{huttenlocher1993comparing}, Reinke et al., 2021: \cite{reinke2021commonarxiv}. Mentioned figures: Figs.~5c, \ref{fig:outline}, \ref{fig:low-quality}, \ref{fig:empty}.}
    \label{fig:cheat-sheet-hd}
\end{figure}

\begin{figure}[H]
    \centering
    \includegraphics[width=\textwidth]{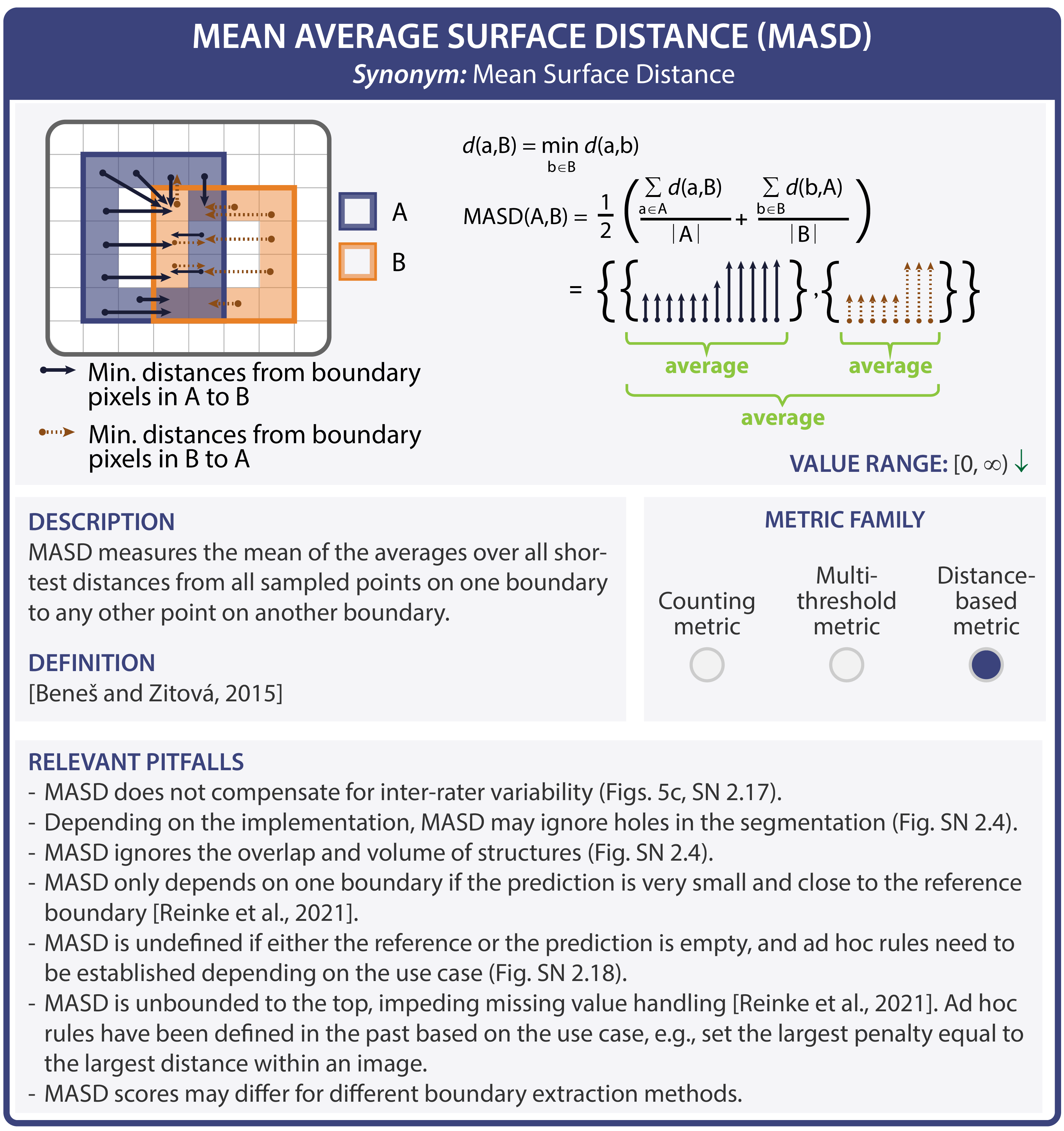}
    \caption{Metric profile of \acf{MASD}. The downward arrow in the value range indicates that lower values are better than higher values. Abbreviation: \acf{SemS}. References: Beneš and Zitová, 2015: \cite{benevs2015performance}, Reinke et al., 2021: \cite{reinke2021commonarxiv}. Mentioned figures: Figs.~5c, \ref{fig:outline}, \ref{fig:low-quality}, \ref{fig:empty}.}
    \label{fig:cheat-sheet-masd}
\end{figure}

\begin{figure}[H]
    \centering
    \includegraphics[width=\textwidth]{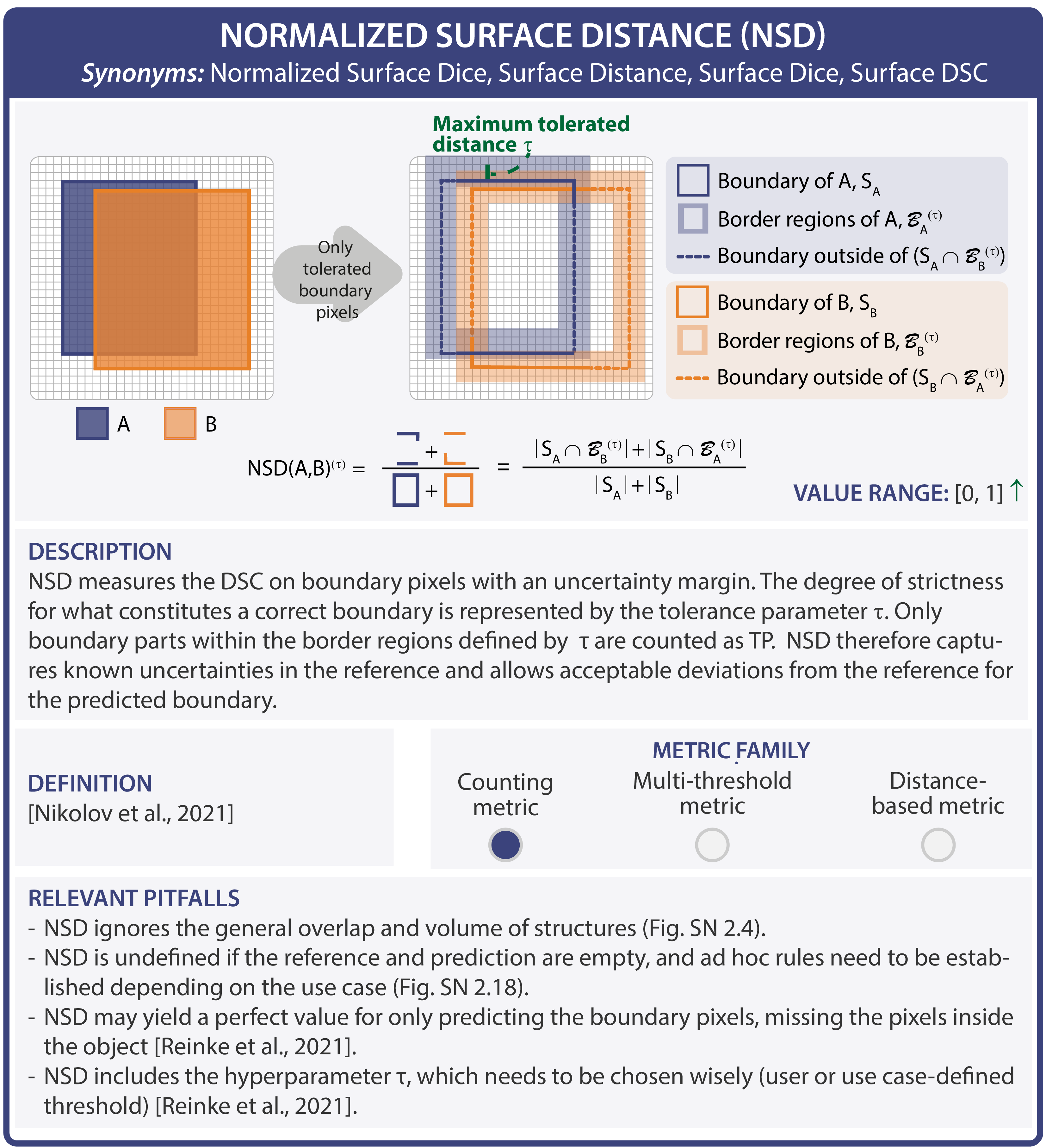}
    \caption{Metric profile of \acf{NSD}. The upward arrow in the value range indicates that higher values are better than lower values. Abbreviation: \acf{DSC}. References: Nikolov et al., 2021: \cite{nikolov2021clinically}, Reinke et al., 2021: \cite{reinke2021commonarxiv}. Mentioned figures: Figs.~\ref{fig:outline}, \ref{fig:empty}.}
    \label{fig:cheat-sheet-nsd}
\end{figure}

\begin{figure}[H]
    \centering
    \includegraphics[width=\textwidth]{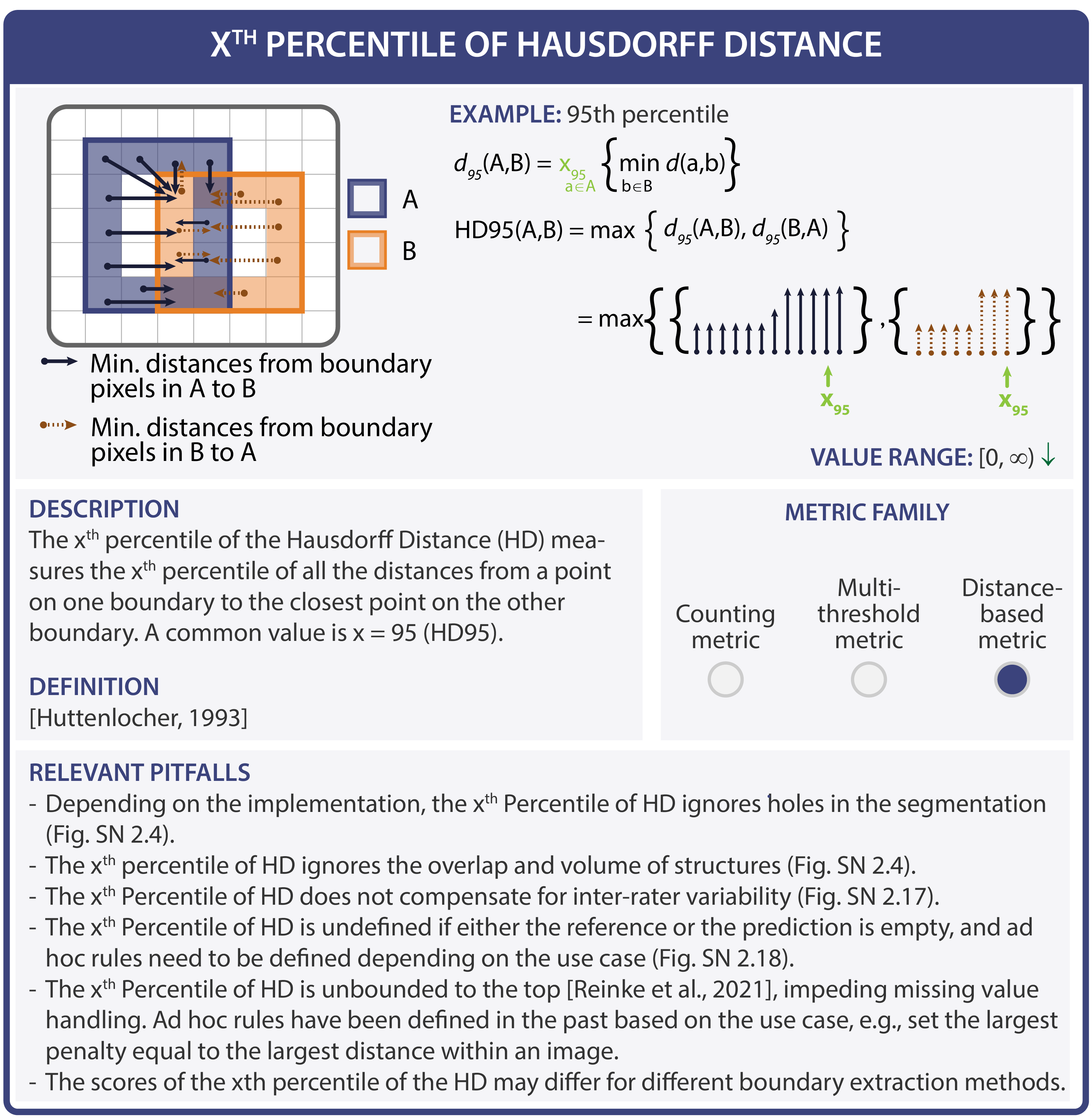}
    \caption{Metric profile of X\textsuperscript{th} Percentile of \acf{HD}. The downward arrow in the value range indicates that lower values are better than higher values. Abbreviations: \acf{HD}, \acf{SemS}. References: Huttenlocher, 1993: \cite{huttenlocher1993comparing}, Reinke et al., 2021: \cite{reinke2021commonarxiv}. Mentioned figures: Figs.~\ref{fig:outline}, \ref{fig:low-quality}, \ref{fig:empty}.}
    \label{fig:cheat-sheet-hd95}
\end{figure}

\newpage
\subsection{Calibration metrics}
\label{app:steckbriefe:calibration}
\begin{figure}[H]
    \centering
    \includegraphics[width=\textwidth]{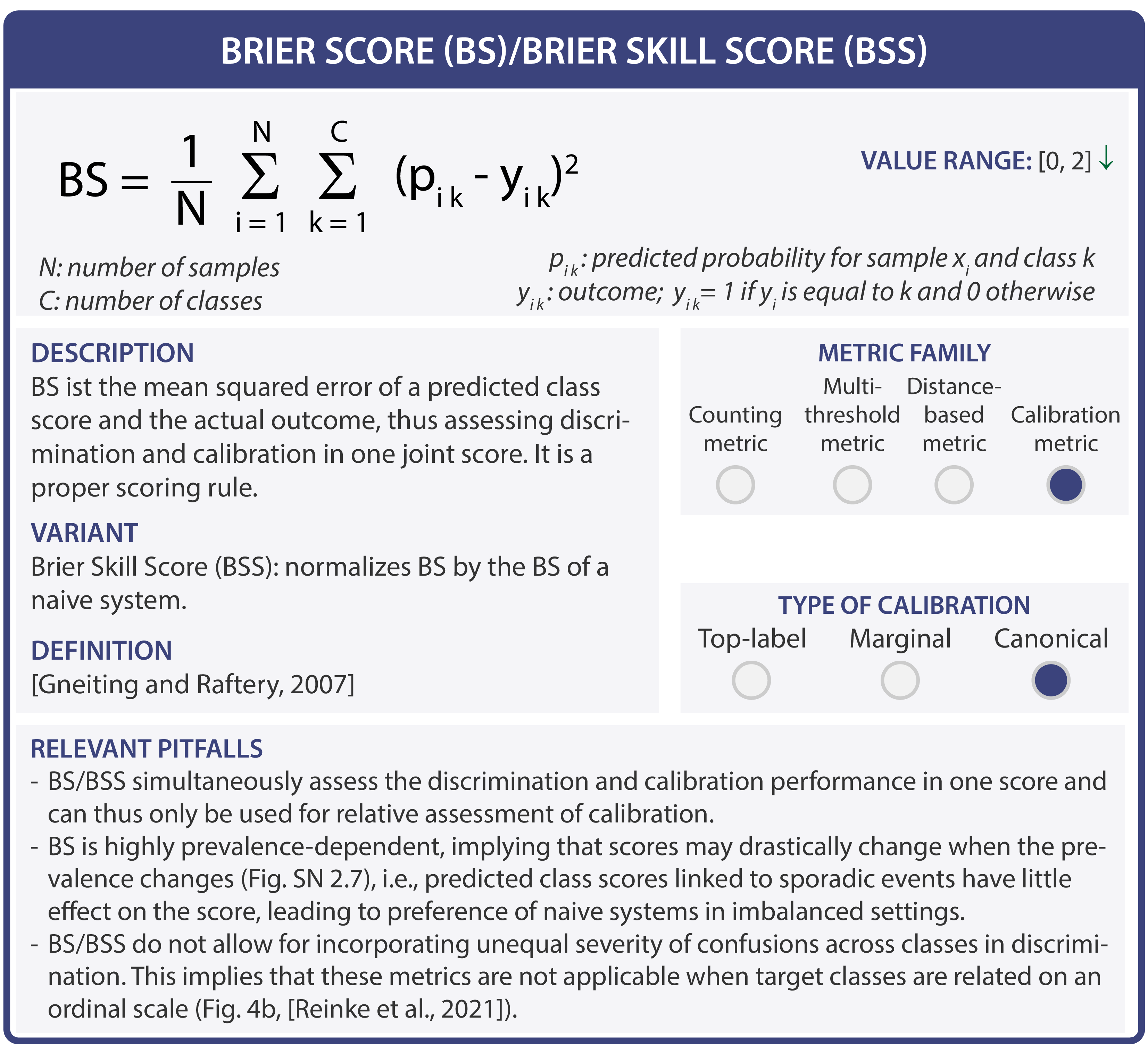}
    \caption{Metric profile of \acf{BS}. The downward arrow in the value range indicates that lower values are better than higher values. Abbreviation: \acf{BSS}. References: Gneiting and Raftery, 2007: \cite{gneiting2007strictly}, Reinke et al., 2021: \cite{reinke2021commonarxiv}. Mentioned figure: Fig.~\ref{fig:prevalence-dependency}.}
    \label{fig:cheat-sheet-bs}
\end{figure}
\FloatBarrier
\newpage
\begin{figure}[H]
    \centering
    \includegraphics[width=\textwidth]{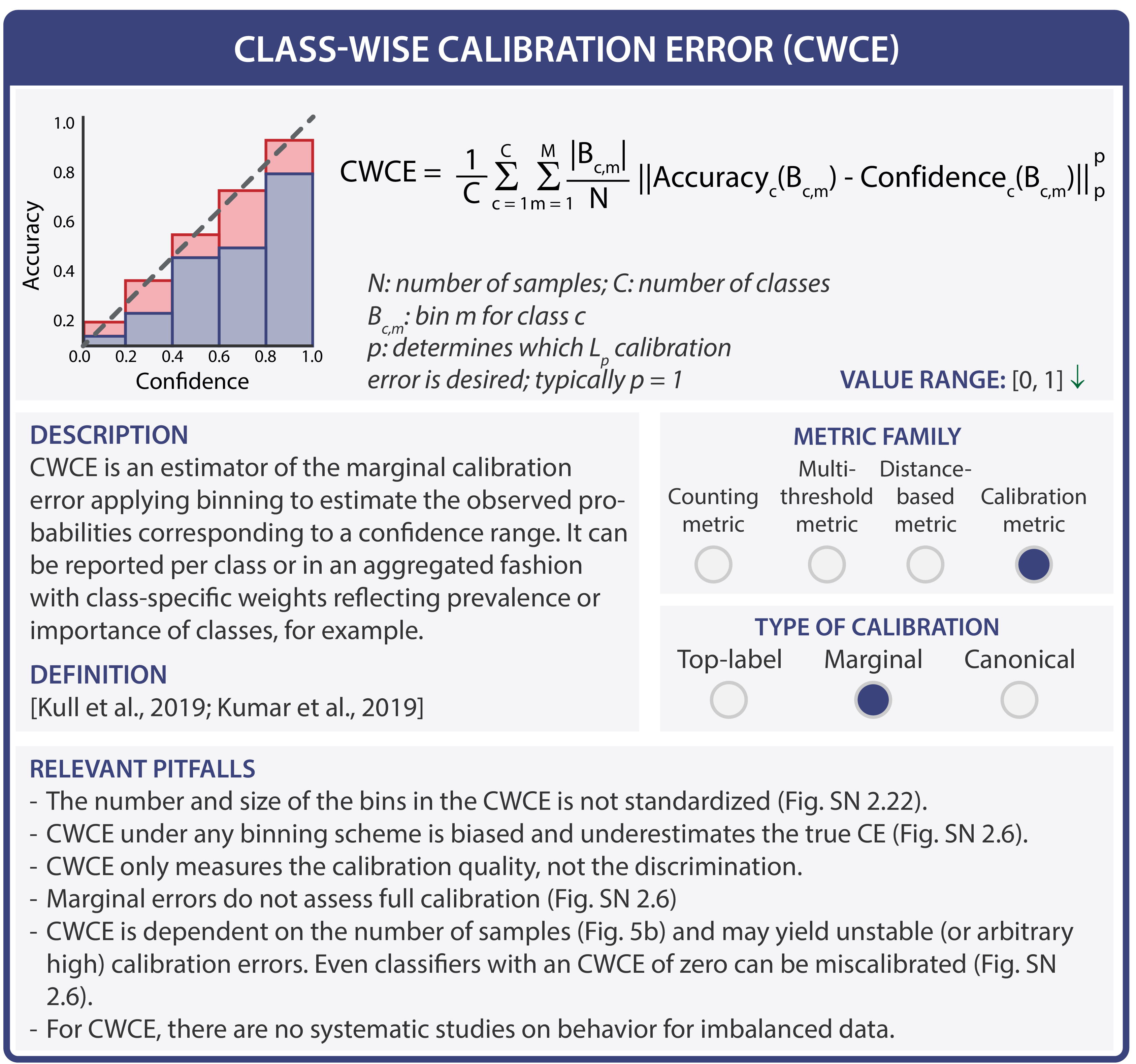}
    \caption{Metric profile of  \acf{CWCE}. The downward arrow in the value range indicates that lower values are better than higher values. References: Kumar et al., 2019: \cite{kumar2019verified}, Kull et al., 2019: \cite{kull2019beyond}. Mentioned figures: Figs.~5b, \ref{fig:calibration}, \ref{fig:ece-mce-bins}.}
    \label{fig:cheat-sheet-cwce}
\end{figure}

\begin{figure}[H]
    \centering
    \includegraphics[width=\textwidth]{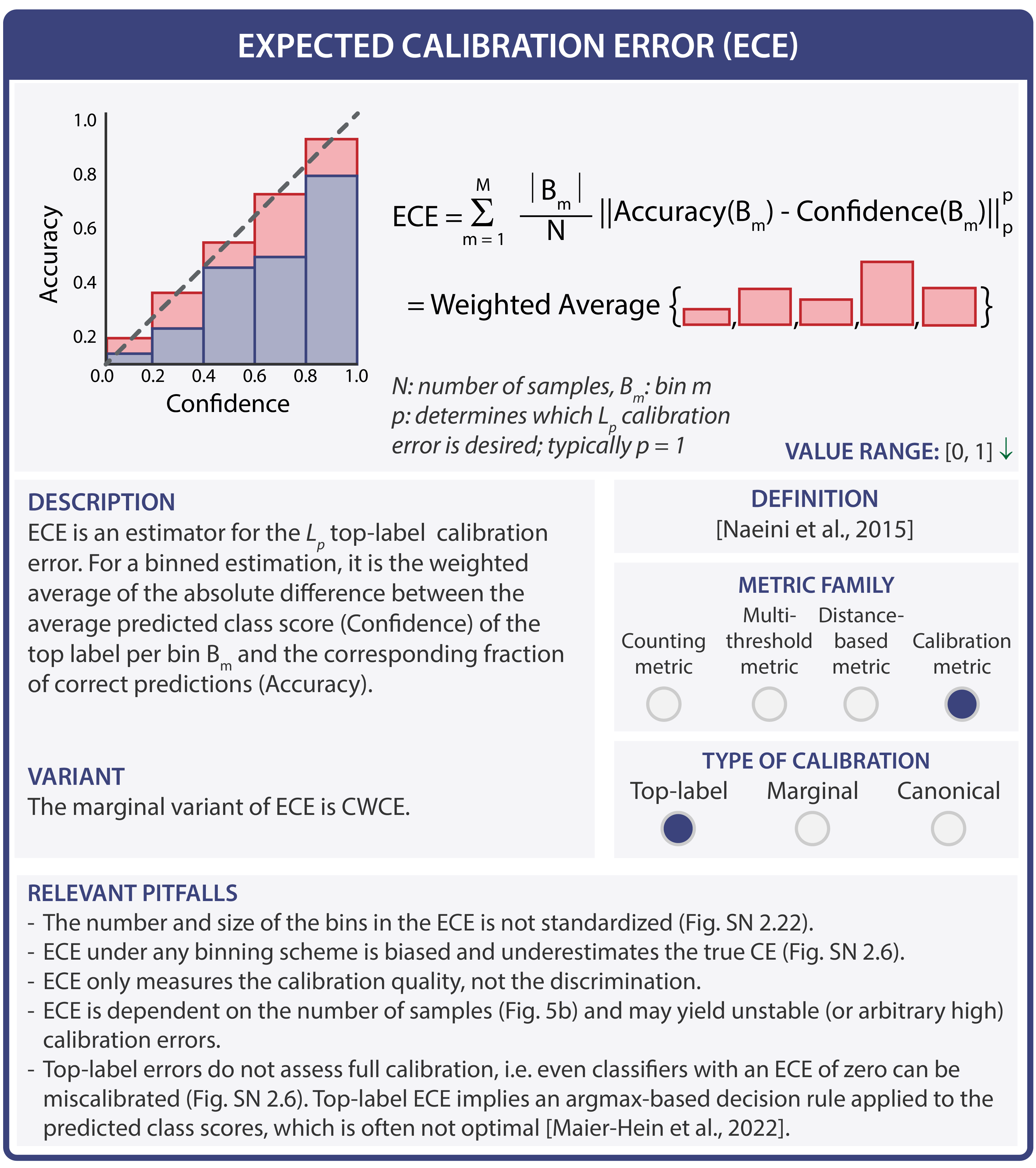}
    \caption{Metric profile of \acf{ECE}. The downward arrow in the value range indicates that lower values are better than higher values. References: Maier-Hein et al., 2022: \cite{maier2022metrics}, Naeini et al., 2015: \cite{naeini2015obtaining}, Reinke et al., 2021: \cite{reinke2021commonarxiv}. Mentioned figures: Figs.~5b, \ref{fig:calibration}, \ref{fig:ece-mce-bins}.}
    \label{fig:cheat-sheet-ece}
\end{figure}

\begin{figure}[H]
    \centering
    \includegraphics[width=\textwidth]{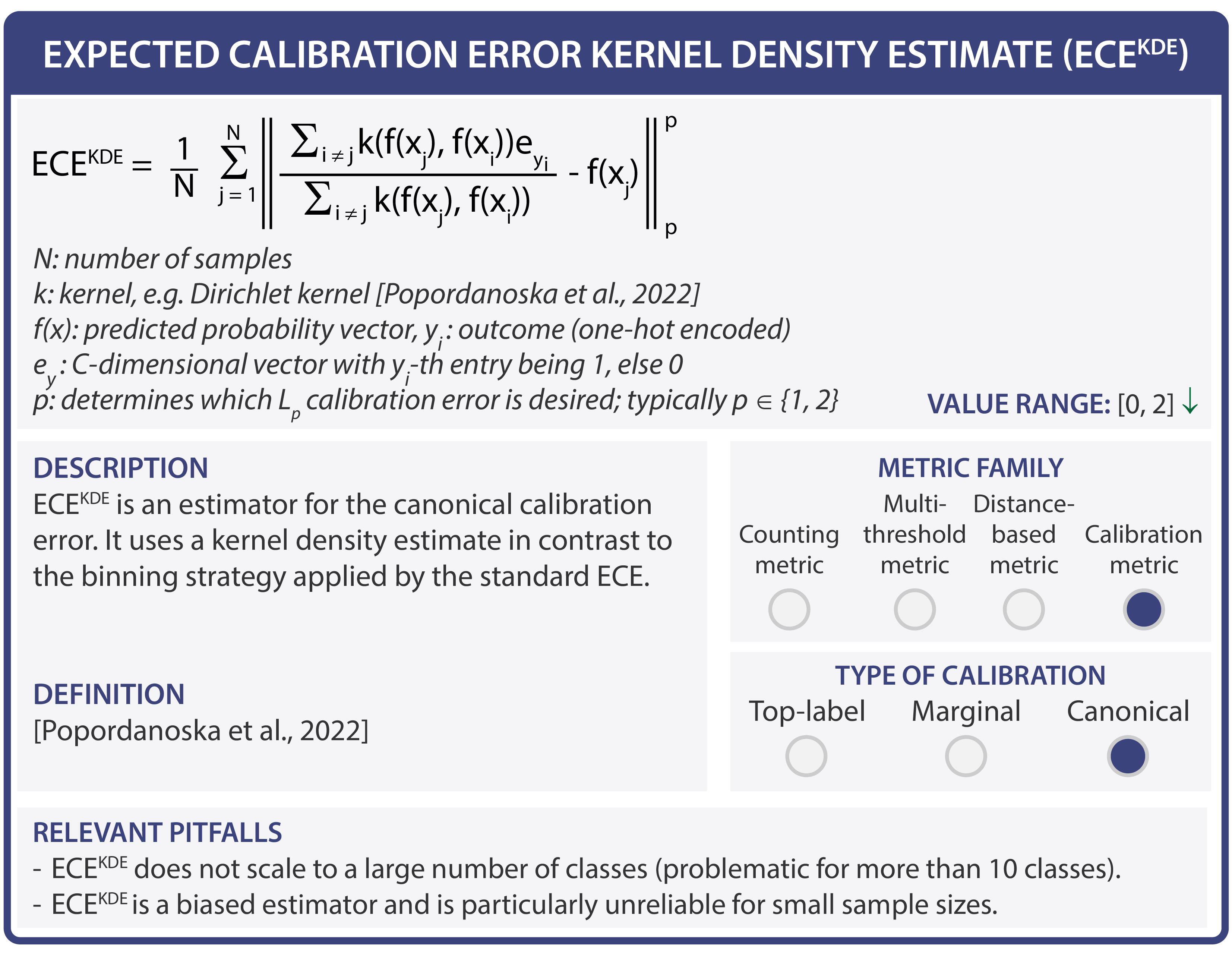}
    \caption{Metric profile of  \acf{ECEKDE}. The downward arrow in the value range indicates that lower values are better than higher values. Abbreviation: \acf{ECE}. Reference used in the figure: Popordanoska et al., 2022: \cite{popordanoska2022consistent}.}
    \label{fig:cheat-sheet-ecekde}
\end{figure}

\begin{figure}[H]
    \centering
    \includegraphics[width=\textwidth]{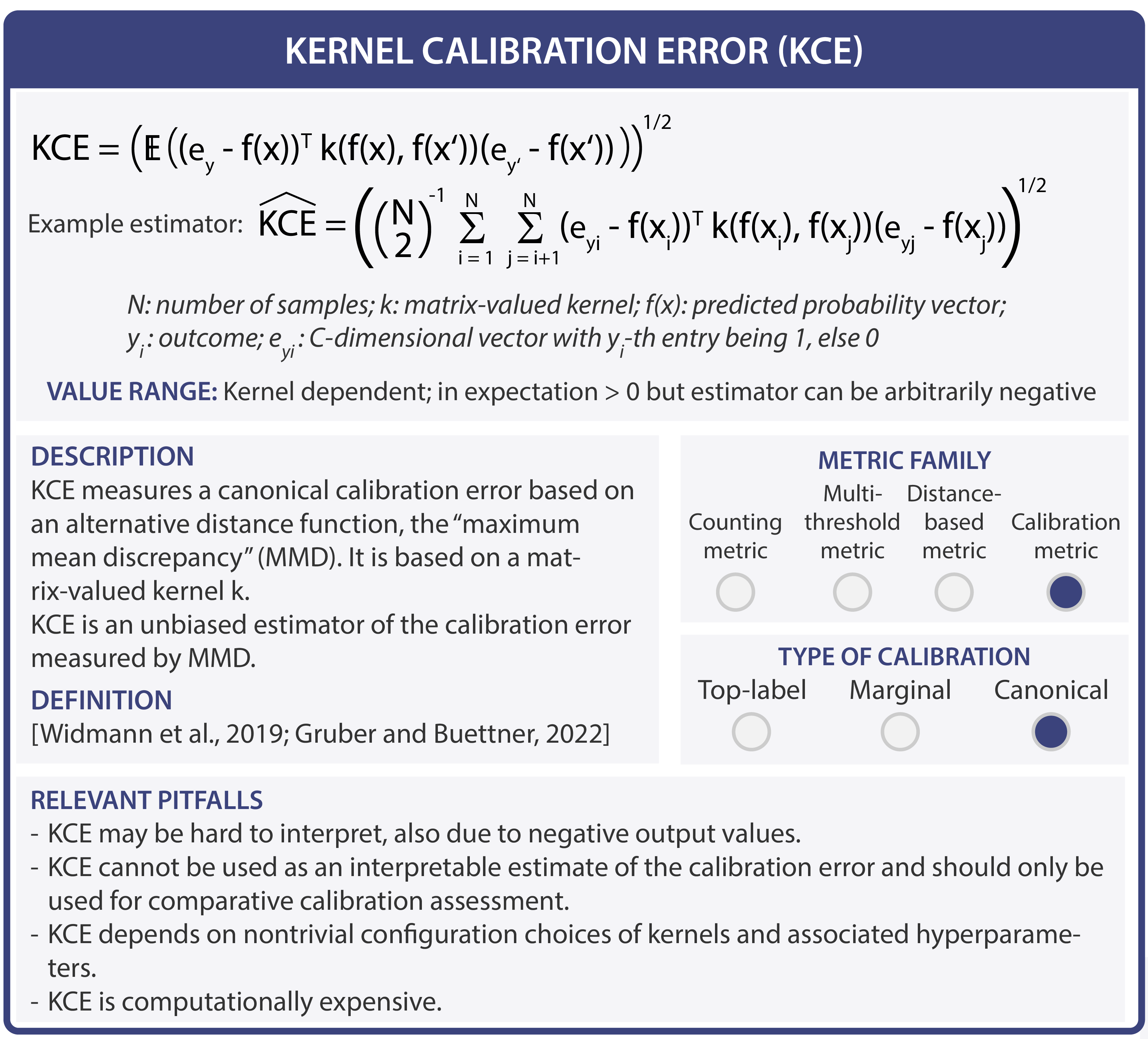}
    \caption{Metric profile of \acf{KCE}. References: Gruber and Buettner, 2022: \cite{gruber2022better}, Widmann et al., 2019: \cite{widmann2019calibration}.}
    \label{fig:cheat-sheet-kce}
\end{figure}

\begin{figure}[H]
    \centering
    \includegraphics[width=\textwidth]{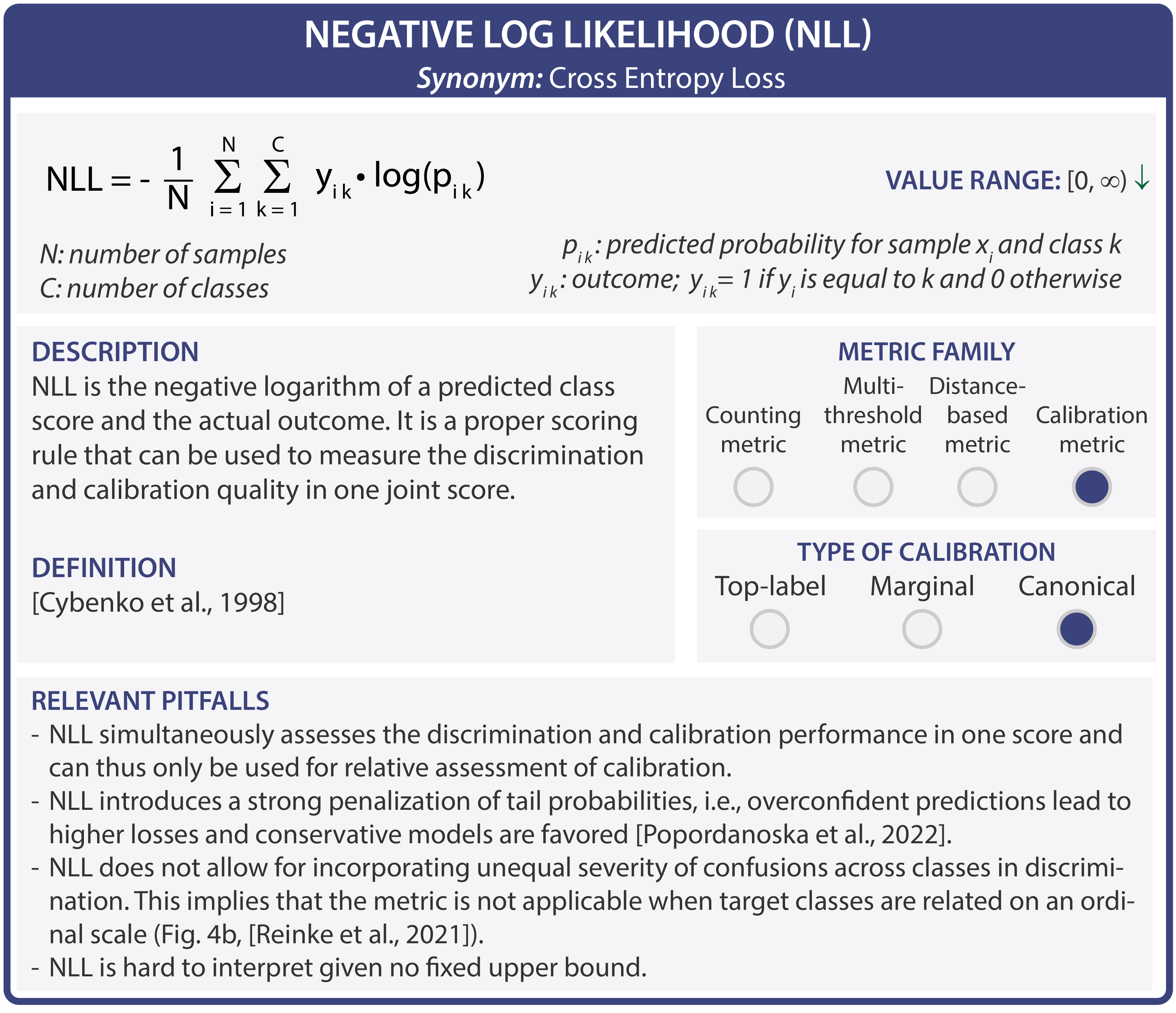}
    \caption{Metric profile of \acf{NLL}. The downward arrow in the value range indicates that lower values are better than higher values. References: Cybenko et al., 1998: \cite{cybenko1998mathematics}, Popordanoska et al., 2022: \cite{popordanoska2022consistent}, Reinke et al., 2021: \cite{reinke2021commonarxiv}. Mentioned figure: Fig.~5b.}
    \label{fig:cheat-sheet-nll}
\end{figure}

\begin{figure}[H]
    \centering
    \includegraphics[width=\textwidth]{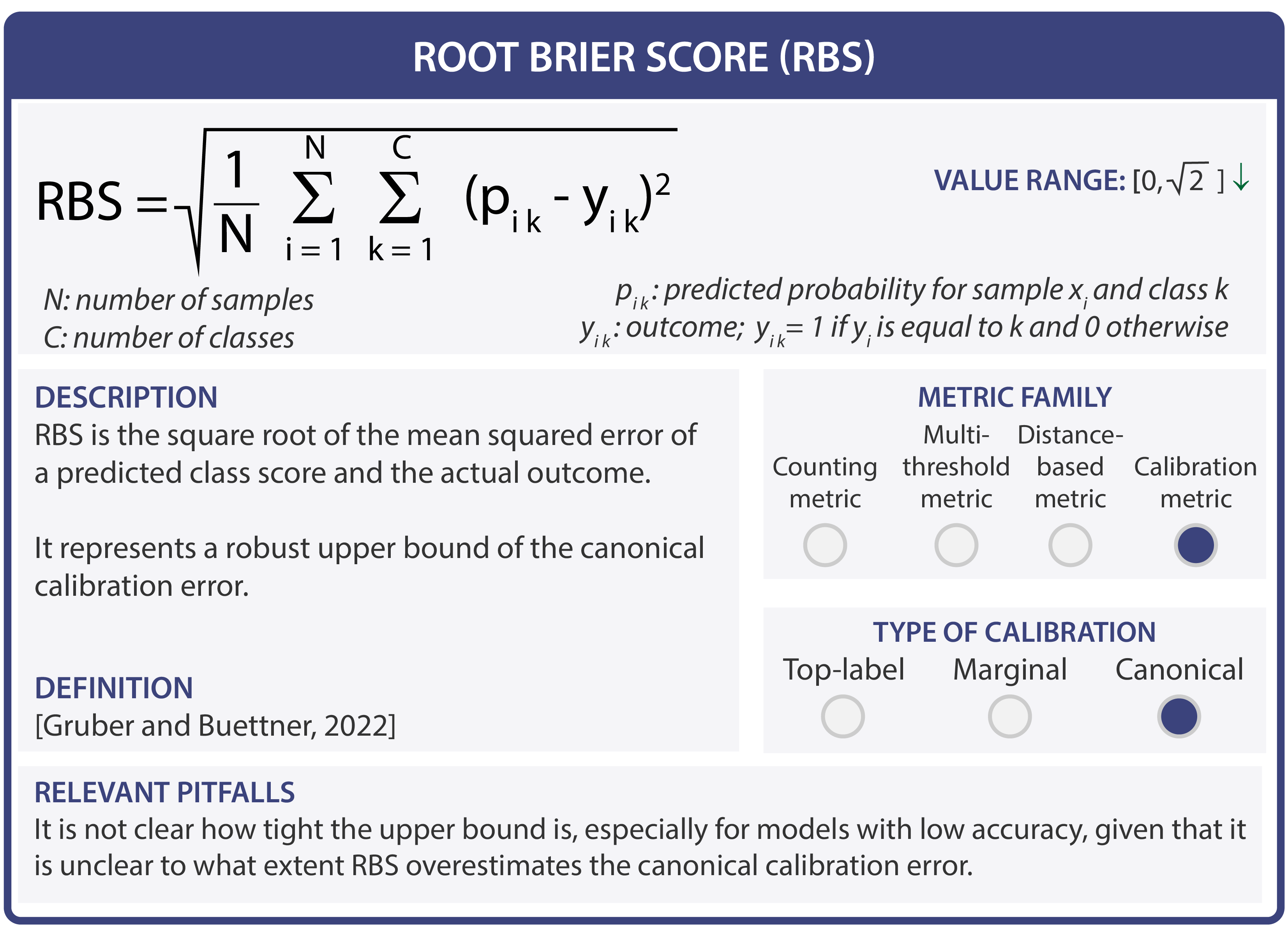}
    \caption{Metric profile of  \acf{RBS}. The downward arrow in the value range indicates that lower values are better than higher values. Reference: Gruber and Buettner, 2022: \cite{gruber2022better}.}
    \label{fig:cheat-sheet-rbs}
\end{figure}

\newpage
\subsection{Localization criteria}
\begin{figure}[H]
    \centering
    \includegraphics[width=\textwidth]{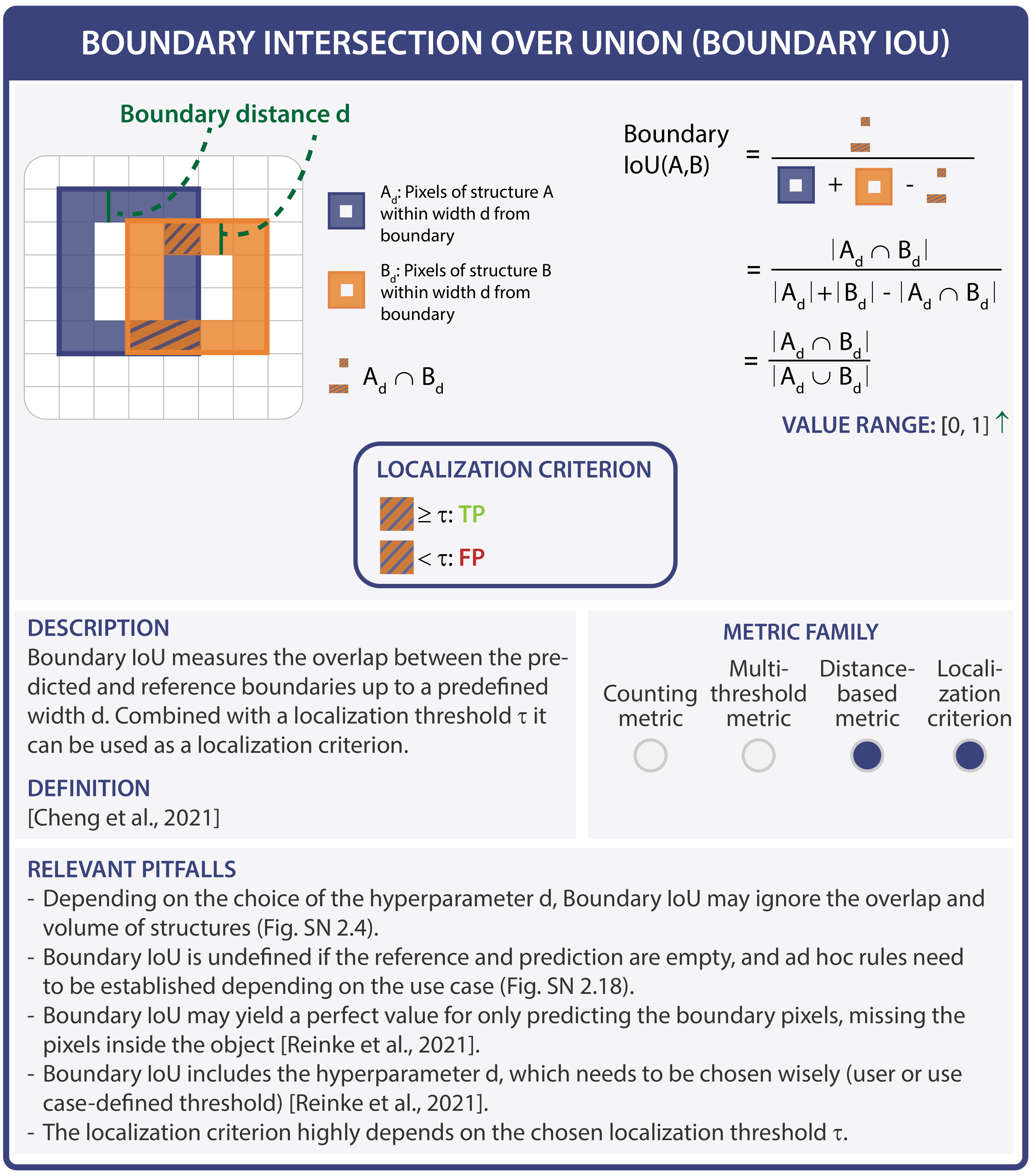}
    \caption{Metric profile of the Boundary \acf{IoU} localization criterion. The upward arrow in the value range indicates that higher values of Boundary \ac{IoU} are better than lower values. References: Cheng et al., 2021: \cite{cheng2021boundary}, Reinke et al., 2021: \cite{reinke2021commonarxiv}. Mentioned figures: Figs.~\ref{fig:outline}, \ref{fig:empty}.}
    \label{fig:cheat-sheet-boundary-iou-localization-crit}
\end{figure}
\FloatBarrier

\newpage
\begin{figure}[H]
    \centering
    \includegraphics[width=\textwidth]{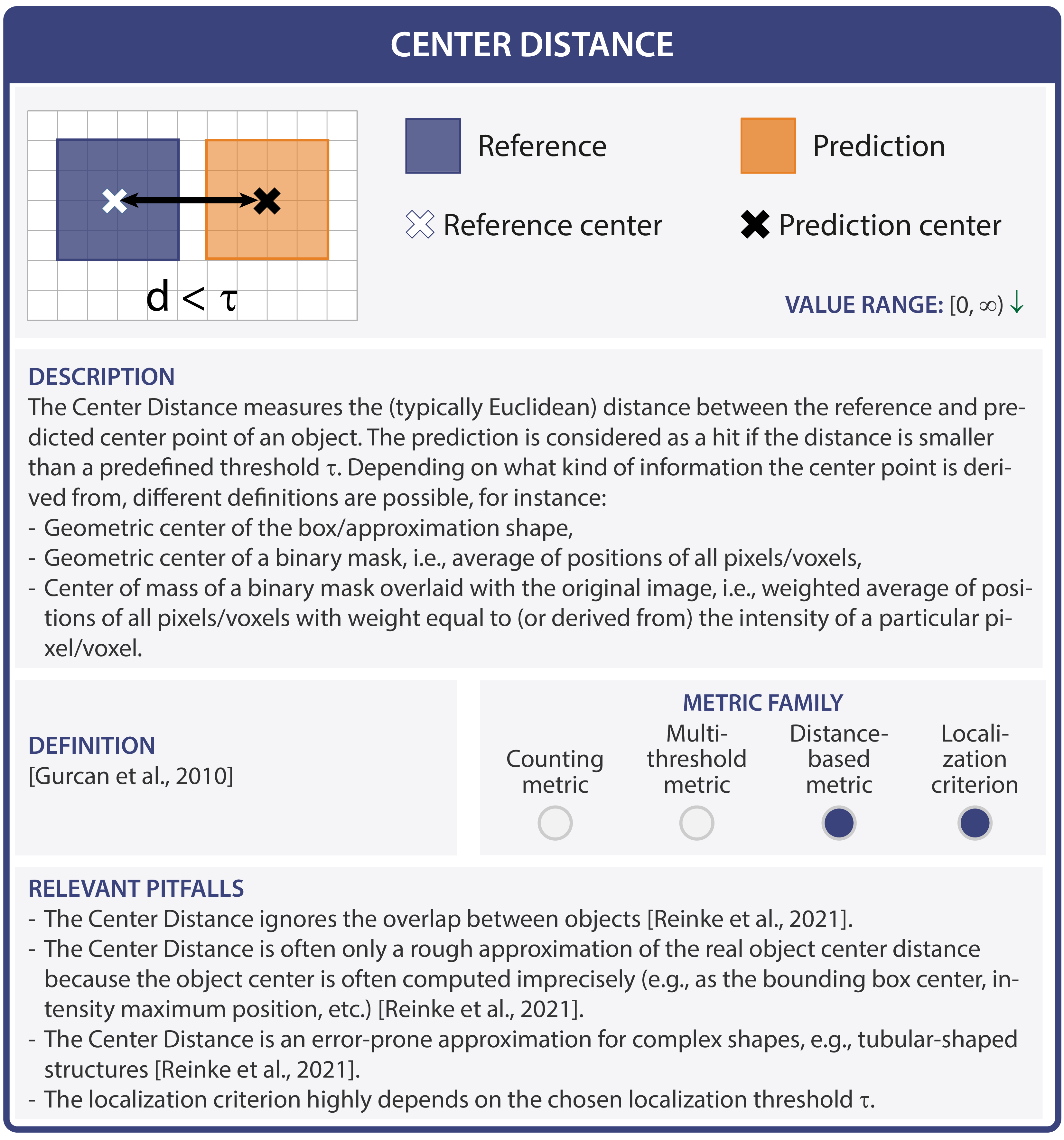}
    \caption{Metric profile of the Center Distance localization criterion. The downward arrow in the value range indicates that lower values of the Center Distance are better than higher values. References: Gurcan et al., 2010: \cite{gurcan2010pattern}, Reinke et al., 2021: \cite{reinke2021commonarxiv}.}
    \label{fig:cheat-sheet-center-distance}
\end{figure}

\begin{figure}[H]
    \centering
    \includegraphics[width=\textwidth]{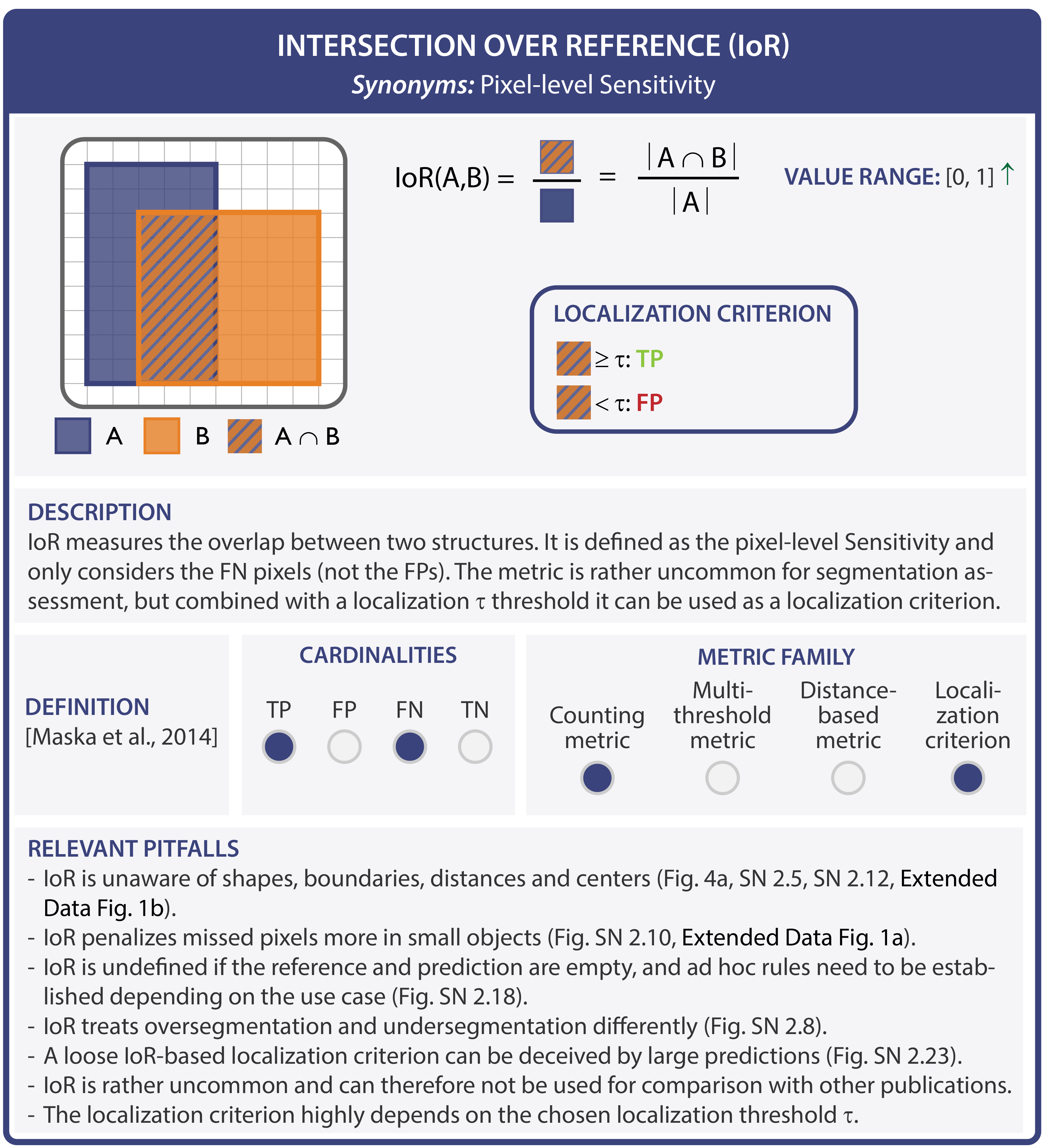}
    \caption{Metric profile of the \acf{IoR} localization criterion. The upward arrow in the value range indicates that higher values of \ac{IoR} are better than lower values. Abbreviations: \acf{FN}, \acf{FP}, \acf{TN}, \acf{TP}. References: Maška et al., 2014: \cite{mavska2014benchmark}, Reinke et al., 2021: \cite{reinke2021commonarxiv}. Mentioned figures: Figs.~4a, \ref{fig:center}, \ref{fig:DSC-overunder}, \ref{fig:boundary-mask-iou}, \ref{fig:high-variability}, \ref{fig:complex-shapes}, \ref{fig:empty}, \ref{fig:sensitivity-hyperparam}, Extended Data Fig.~1b.}
    \label{fig:cheat-sheet-ior}
\end{figure}

\begin{figure}[H]
    \centering
    \includegraphics[width=\textwidth]{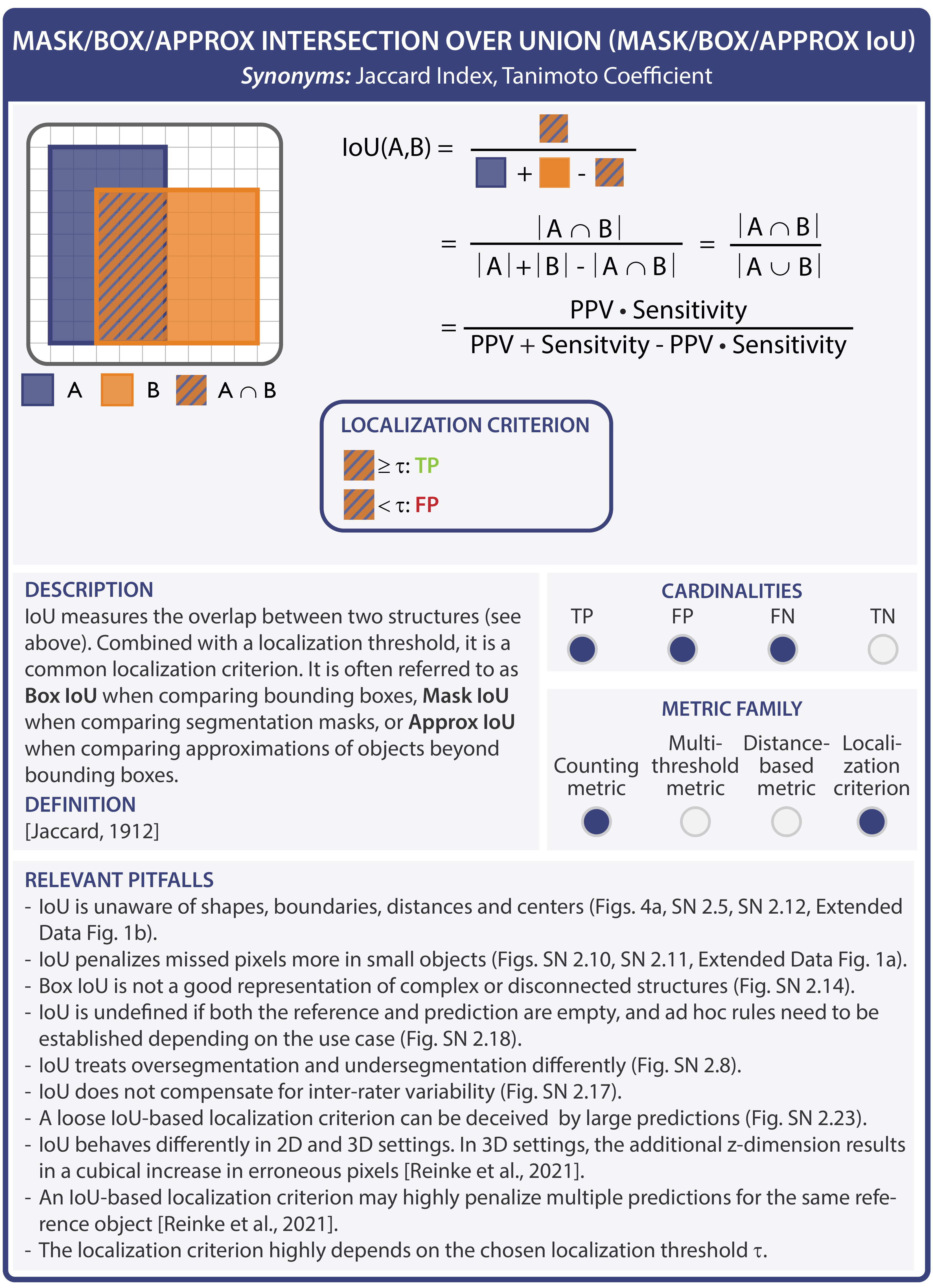}
    \caption{Metric profile of the Mask/Box/Approx \acf{IoU} localization criterion. Abbreviations: \acf{FN}, \acf{FP}, \acf{TN}, \acf{TP}. References: Jaccard, 1912: \cite{jaccard1912distribution}, Reinke et al., 2021: \cite{reinke2021commonarxiv}. Mentioned figures: Figs.~4a, \ref{fig:center}, \ref{fig:DSC-overunder}, \ref{fig:boundary-mask-iou}, \ref{fig:high-variability},\ref{fig:complex-shapes}, \ref{fig:disconnected}, \ref{fig:low-quality}, \ref{fig:empty}, \ref{fig:sensitivity-hyperparam}, Extended Data Fig.~1a-b.}
    \label{fig:cheat-sheet-iou-localization-crit}
\end{figure}

\begin{figure}[H]
    \centering
    \includegraphics[width=\textwidth]{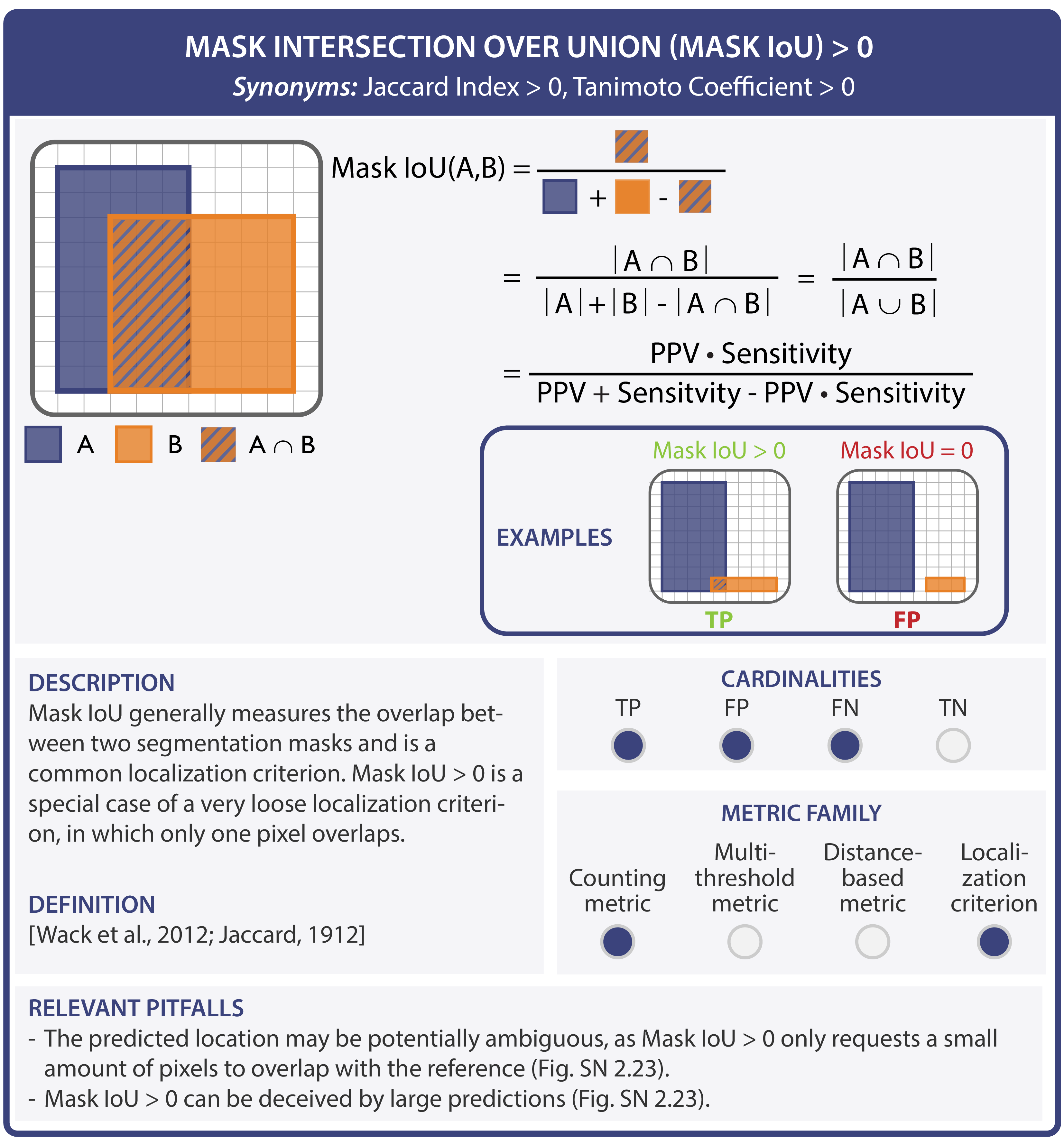}
    \caption{Metric profile of the Mask \acf{IoU} > 0 localization criterion. Abbreviations: \acf{FN}, \acf{FP}, \acf{TN}, \acf{TP}. References: Jaccard, 1912: \cite{jaccard1912distribution}, Wack et al., 2012: \cite{wack2012improved}. Mentioned figure: Fig.~\ref{fig:sensitivity-hyperparam}.}
    \label{fig:cheat-sheet-iou-localization-crit-0}
\end{figure}

\begin{figure}[H]
    \centering
    \includegraphics[width=\textwidth]{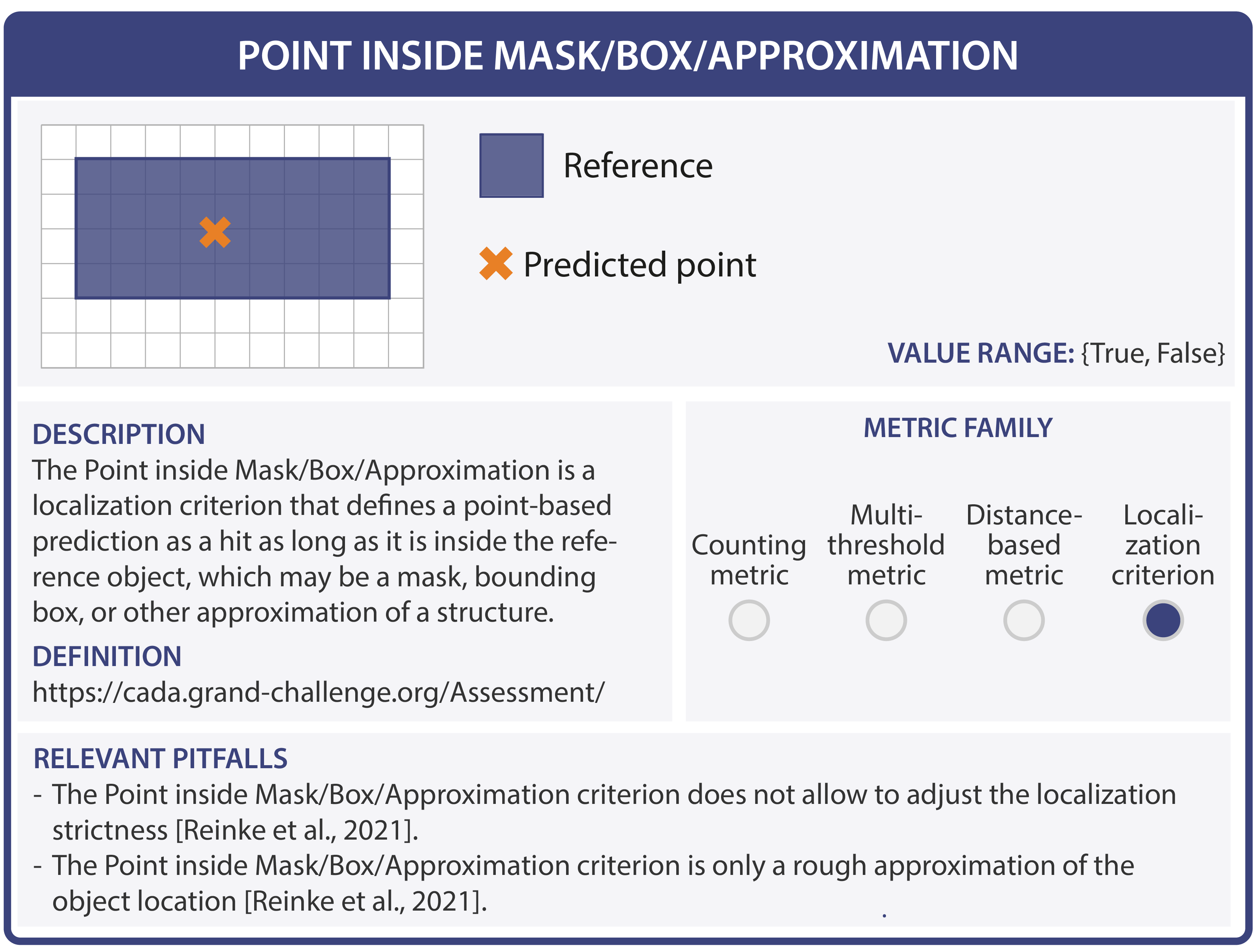}
    \caption{Metric profile of  Point inside Mask/Box/Approximation. References: \url{https://cada.grand-challenge.org/Assessment/}, Reinke et al., 2021: \cite{reinke2021commonarxiv}.}
    \label{fig:cheat-sheet-point-inside}
\end{figure}

\newpage
\subsection{Assignment strategies}

\begin{figure}[H]
    \centering
    \includegraphics[width=\textwidth]{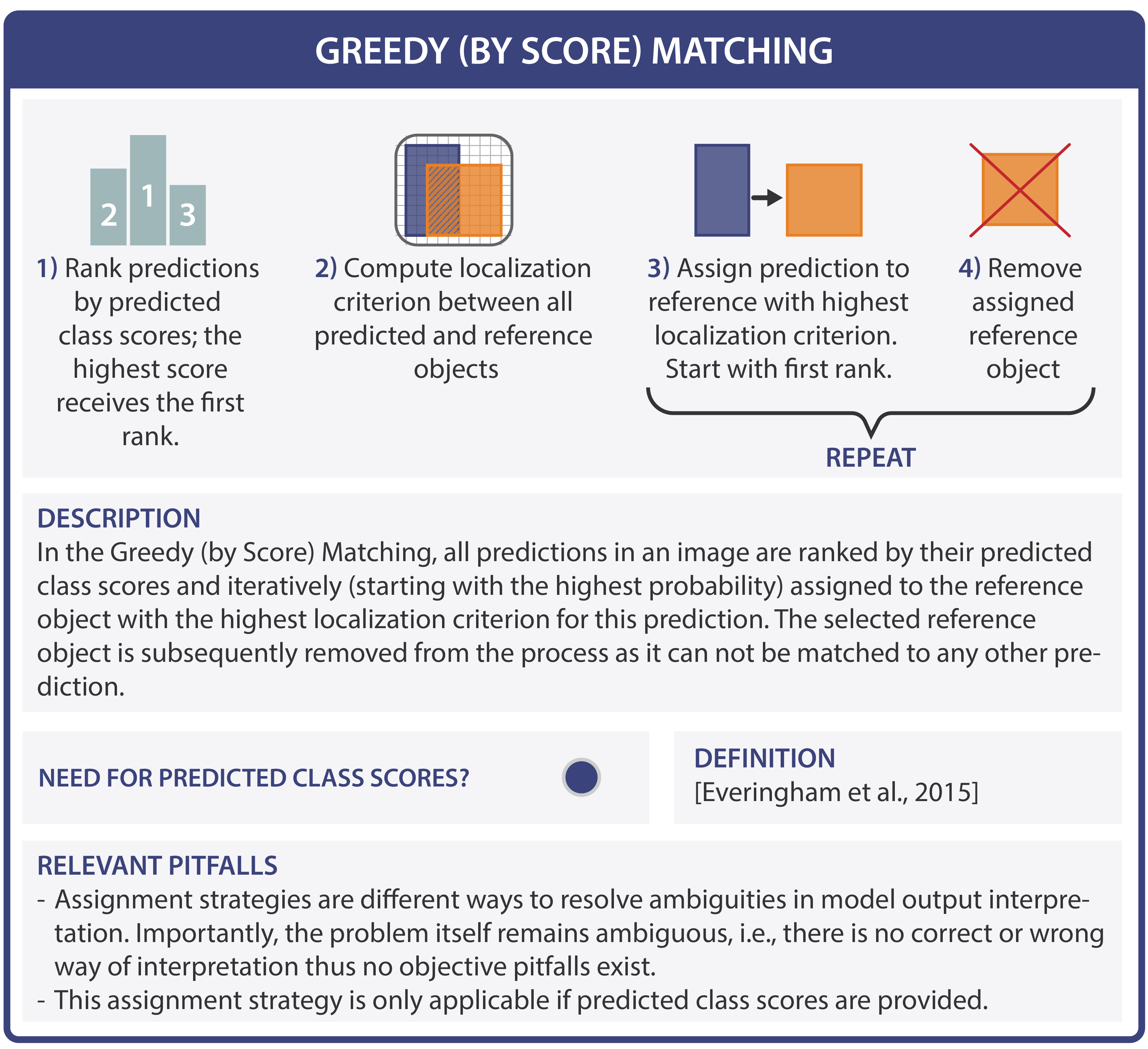}
    \caption{Cheat Sheet for the Greedy (by Score) Matching. Reference used in the figure: Everingham et al., 2015: \cite{everingham2015pascal}.}
    \label{fig:cheat-sheet-greedy-score}
\end{figure}
\FloatBarrier
\newpage
\begin{figure}[H]
    \centering
    \includegraphics[width=\textwidth]{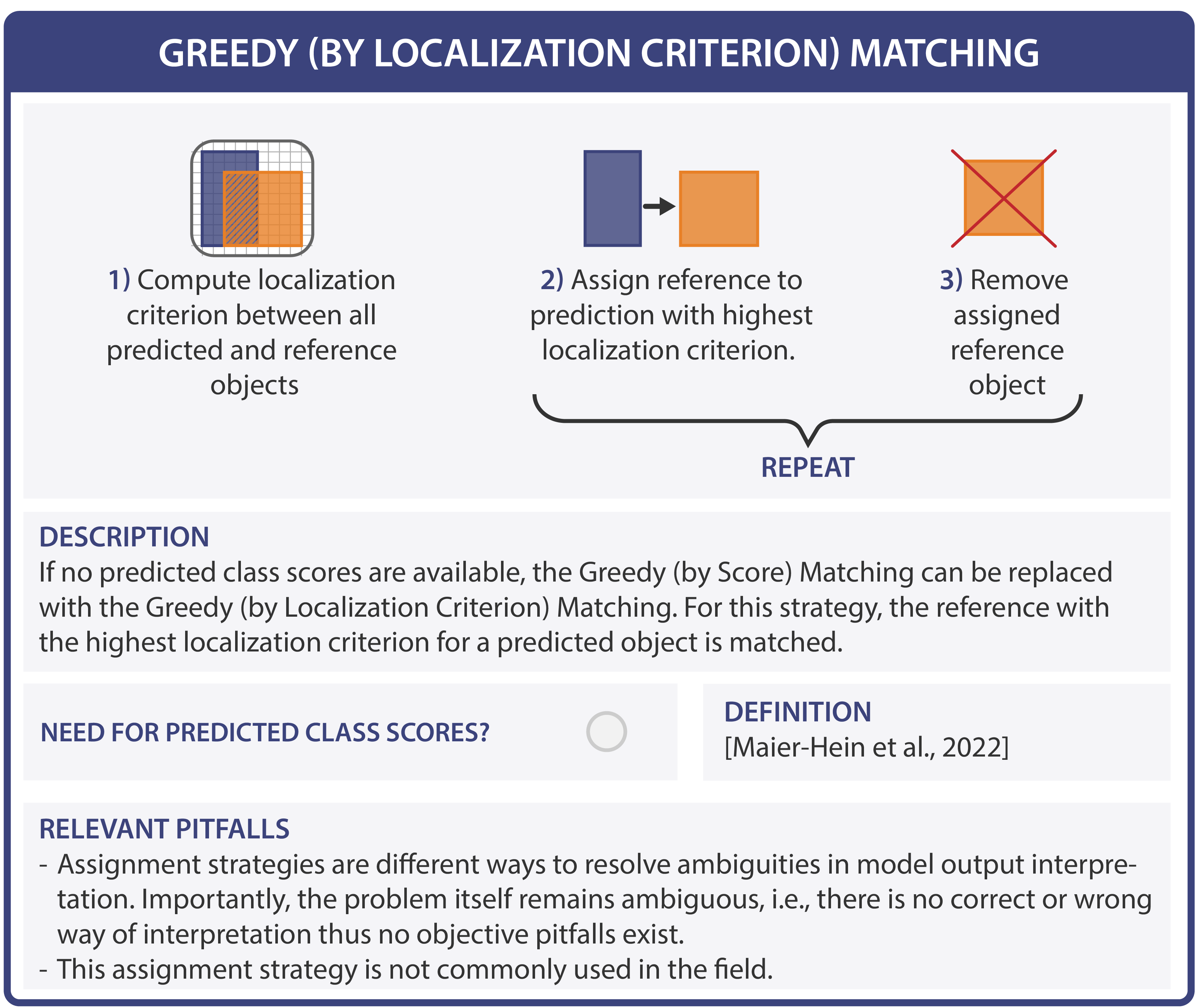}
    \caption{Cheat Sheet for the Greedy (by Localization Criterion) Matching. Reference used in the figure: Maier-Hein et al., 2022: \cite{maier2022metrics}.}
    \label{fig:cheat-sheet-greedy-localization}
\end{figure}

\begin{figure}[H]
    \centering
    \includegraphics[width=\textwidth]{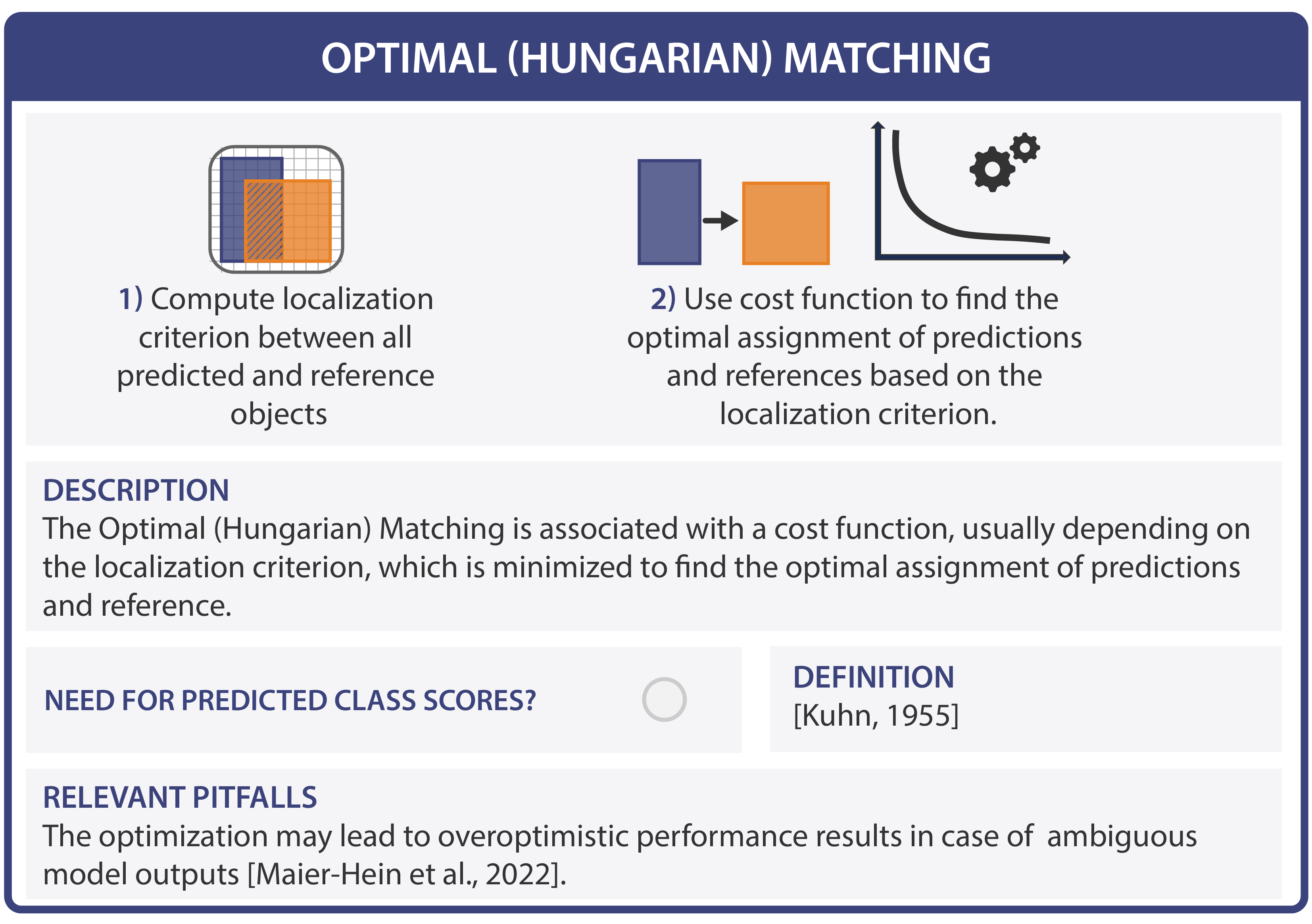}
    \caption{Cheat Sheet for the Optimal (Hungarian) Matching. References used in the figure: Kuhn et al., 1955: \cite{kuhn1955hungarian}, Maier-Hein et al., 2022: \cite{maier2022metrics}.}
    \label{fig:cheat-sheet-hungarian}
\end{figure}

\begin{figure}[H]
    \centering
    \includegraphics[width=\textwidth]{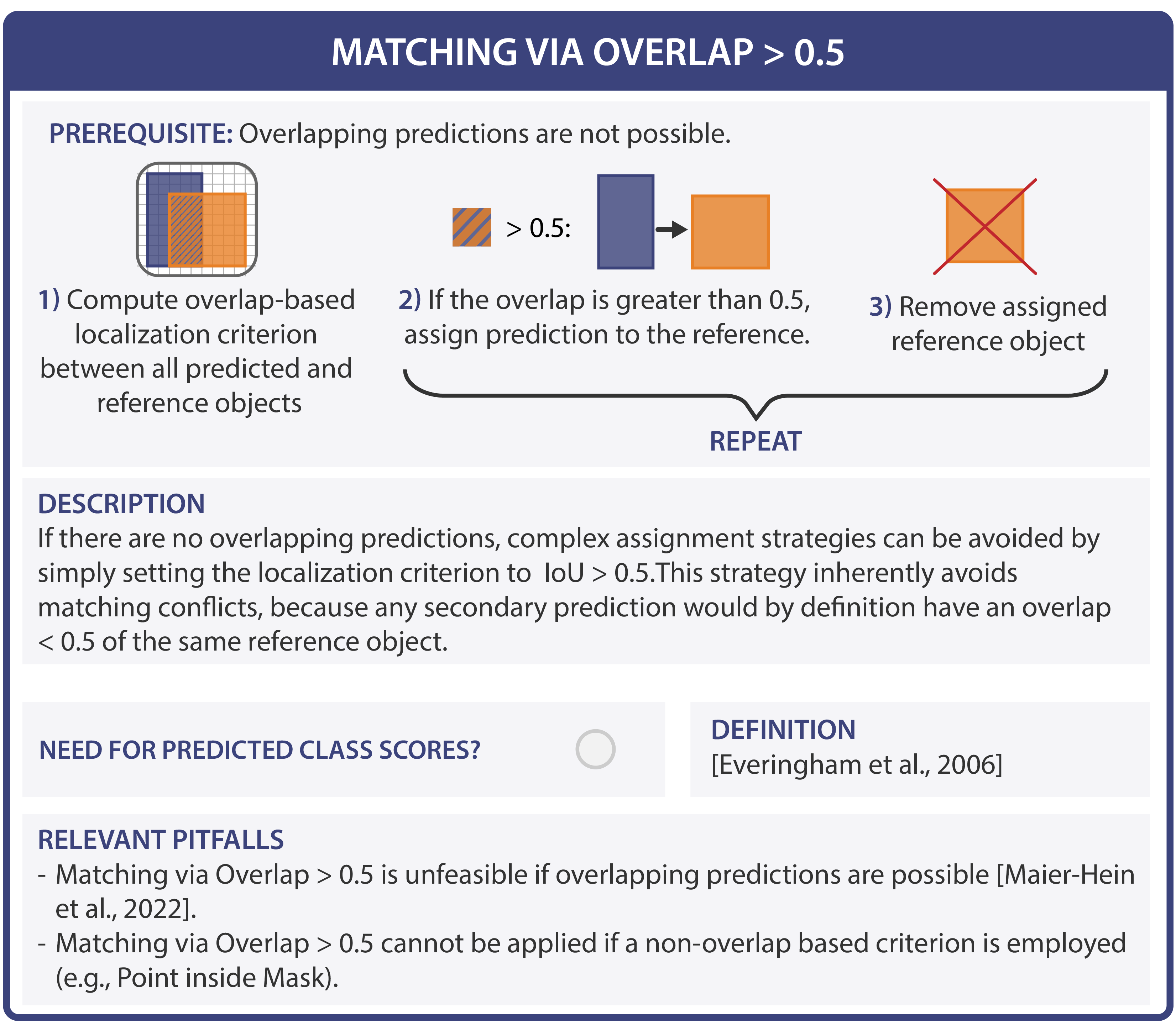}
    \caption{Cheat Sheet for the Matching via Overlap > 0.5. References used in the figure: Everingham et al., 2006: \cite{everingham20062005}, Maier-Hein et al., 2022: \cite{maier2022metrics}.}
    \label{fig:cheat-sheet-matching-greater}
\end{figure}

\newpage
\section*{Acronyms}
\label{app:acronyms}
\begin{acronym}[]
\acro{Acc}{Accuracy}
\acro{AI}{artificial intelligence}
\acro{AP}{Average Precision}
\acro{ASSD}{Average Symmetric Surface Distance}
\acro{AUC}{Area under the Curve}
\acro{AUROC}{Area under the Receiver Operating Characteristic Curve}
\acro{BA}{Balanced Accuracy}
\acro{BIAS}{Biomedical Image Analysis ChallengeS}
\acro{Boundary IoU}{Boundary Intersection over Union}
\acro{Box IoU}{Box Intersection over Union}
\acro{BM}{Bookmaker Informedness}
\acro{BPMN}{Business Process Model and Notation}
\acro{BS}{Brier Score}
\acro{BSS}{Brier Skill Score}
\acro{CI}{Confidence Interval}
\acro{clDice}{centerline Dice Similarity Coefficient}
\acro{COCO}{Common Objects in Context}
\acro{CK}{Cohen's Kappa}
\acro{CWCE}{Class-Wise Calibration Error}
\acro{CT}{computed tomography}
\acro{DSC}{Dice Similarity Coefficient}
\acro{EC}{Expected Cost}
\acro{ECE}{Expected Calibration Error}
\acro{ECEKDE}[ECE\textsuperscript{KDE}]{Expected Calibration Error Kernel Density Estimate}
\acro{ER}{Error Rate}
\acro{FN}{False Negative}
\acrodefplural{FN}{False Negatives}
\acro{FP}{False Positive}
\acrodefplural{FP}{False Positives}
\acro{FPPI}{False Positives per Image}
\acro{FPR}{False Positive Rate}
\acro{FROC}{Free-Response Receiver Operating Characteristic}
\acro{HD}{Hausdorff Distance}
\acro{HD95}{Hausdorff Distance 95th Percentile}
\acro{ImLC}{Image-level Classification}
\acro{InS}{Instance Segmentation}
\acro{IoU}{Intersection over Union}
\acro{IoR}{Intersection over Reference}
\acro{J}{Youden's Index}
\acro{LR+}{Positive Likelihood Ratio}
\acro{KCE}{Kernel Calibration Error}
\acro{LS}{Logarithmic Score}
\acro{mAP}{mean Average Precision}
\acro{MASD}{Mean Average Surface Distance}
\acro{MCC}{Matthews Correlation Coefficient}
\acro{MCE}{Maximum Calibration Error}
\acro{MCP}{Maximum Class Probability}
\acro{MICCAI}{Medical Image Computing and Computer Assisted Interventions}
\acro{ML}{machine learning}
\acro{MONAI}{Medical Open Network
for Artificial Intelligence}
\acro{MRI}{Magnetic Resonance Imaging}
\acro{NaN}{Not a Number}
\acro{NB}{Net Benefit}
\acro{NPV}{Negative Predictive Value}
\acro{NLL}{Negative Log Likelihood}
\acro{NSD}{Normalized Surface Distance}
\acro{PPV}{Positive Predictive Value}
\acro{ObD}{Object Detection}
\acro{PQ}{Panoptic Quality}
\acro{PHI}{Protected Health Information}
\acro{PR}{Precision-Recall}
\acro{RBS}{Root Brier Score}
\acro{RI}{Rand Index}
\acro{ROC}{Receiver Operating Characteristic}
\acro{SemS}{Semantic Segmentation}
\acro{TN}{True Negative}
\acro{TNR}{True Negative Rate}
\acro{TP}{True Positive}
\acro{TPR}{True Positive Rate}
\acro{VoI}{Variation of Information}
\acro{WCK}{Weighted Cohen's Kappa}
\acro{WSI}{Whole Slide Imaging}
\acro{X$^{th}$ Percentile HD}{X$^{th}$ Percentile Hausdorff Distance}

\end{acronym}

\end{document}